\documentclass[opre,nonblindrev]{informs3} 
\usepackage{newtxtext,newtxmath}
\OneAndAHalfSpacedXI 

\def\footnoterule{\kern-1pt \hrule width 12pc \kern 2.6pt }
\usepackage{natbib}
 \bibpunct[, ]{(}{)}{,}{a}{}{,}%
 \def\bibfont{\small}%
\TheoremsNumberedThrough     
\ECRepeatTheorems

\EquationsNumberedThrough    

\usepackage[dvipsnames]{xcolor}

\usepackage{silence}
\usepackage{amsmath,amssymb,url,braket,bbm}
\usepackage{stfloats}
\usepackage{lipsum}
\usepackage{upgreek} 
\usepackage{algorithm}
\usepackage{algorithmic}
\usepackage{hyperref}
\usepackage{mathrsfs}
\usepackage{overpic}
\usepackage{tikz}
\usetikzlibrary{decorations, patterns.meta}
\allowdisplaybreaks[4]
\newcommand\scalemath[2]{\scalebox{#1}{\mbox{\ensuremath{\displaystyle #2}}}}
\usepackage{adjustbox}
\usepackage{caption}

\usepackage{comment}
\usepackage{subfigure}
\usepackage{mathtools}
\usepackage{etoolbox}
\AtBeginEnvironment{dcases}{\selectfont}
\usepackage{enumitem}
\usepackage{graphicx}

\usepackage{booktabs}

\usepackage{xcolor}
\usepackage{soul}

\DeclareMathOperator{\E}{\mathbb{E}}

\usepackage{bm}

\newcommand{\Z}{{Z}}
\newcommand{\tbl}[1]{#1}
\newcommand{\IN}{I^{\mathrm N}}
\newcommand{\JN}{J^{\mathrm N}}
\newcommand{\JNSV}{J^{\mathrm {NSV}}}
\newcommand{\IHNE}{I^{\mathrm{HNE}}}
\newcommand{\fMNL}{{f^{\small \mathrm {MNL}}}}
\begin{document}


\RUNAUTHOR{Yang and Feng}

\RUNTITLE{Learning to Select and Rank from Choice-Based Feedback: A Simple Nested Approach}

\TITLE{Learning to Select and Rank from Choice-Based Feedback: A Simple Nested Approach}

\ARTICLEAUTHORS{%
	\AUTHOR{Junwen Yang$^{1}$ \qquad Yifan Feng$^{1, 2}$\\*[1ex]
		{\footnotesize$^1$ Institute of Operations Research and Analytics, National University of Singapore}
		\\*[0.1ex]
		{\footnotesize$^2$ NUS Business School, National University of Singapore}
}}

\ABSTRACT{We study a ranking and selection problem of learning from choice-based feedback with dynamic assortments. In this problem, a company sequentially displays a set of items to a population of customers and collects their choices as feedback. The only information available about the underlying choice model is that the choice probabilities are consistent with some unknown true strict ranking over the items. The objective is to identify, with the fewest samples, the most preferred item or the full ranking over the items at a high confidence level. We propose novel and simple algorithms for both learning goals through a nested approach. For best-item identification, we introduce \textsc{Nested
		Elimination} (\textsc{NE}), and for full-ranking identification, we introduce \textsc{Nested
		Partition} (\textsc{NP}).
	Both algorithms are fast to run and admit instance-specific, non-asymptotic sample-complexity guarantees.\tbl{By comparing these guarantees with information-theoretic lower bounds, we establish that both algorithms are asymptotically worst-case optimal.} Our analysis is based on an analytical framework that characterizes the system dynamics through analyzing a sequence of multi-dimensional random walks. \tbl{We further extend the problem to incorporate capacity-constrained
		displays.} Numerical experiments on synthetic and real data corroborate our
	theory.
}%

\HISTORY{A preliminary version of the paper appeared in the \emph{Proceedings of the 40th International Conference on Machine Learning (ICML 2023)}; see \citet{yang23b}. This version: August 22, 2026.}

\maketitle

%


\section{Introduction}
\label{section_intro}

Understanding customer preferences is fundamental to decision-making across various domains, including marketing, e-commerce, and recommendation systems. Nowadays, advances in the Internet and computing technologies have significantly enhanced the sophistication of preference learning systems, enabling them to operate in real-time, adapt dynamically, provide personalized results, and scale efficiently. These advancements have unlocked novel applications. For instance, a business model innovation in e-commerce is \textit{crowdvoting}, where companies systematically gather consumer feedback on new product prototypes to determine which products to bring to market (see \citealp{King2013,marinesi2013information,araman2022diffusion} for related reports and studies). More broadly, digital surveys have become increasingly prevalent, which allow businesses to better understand consumer preferences. These developments underscore the importance of designing efficient preference learning systems. For example, in applications such as crowdvoting, well-designed feedback mechanisms can help businesses avoid delays in new product introduction while minimizing the risks of commercializing poorly received products. In the case of digital surveys, efficient data collection is crucial, as participant compensation can make sample inefficiency financially burdensome (see \citealp{liu2023value} for a related study).

Motivated by these preference learning applications, we investigate a class of ranking-and-selection problems under a specific feedback structure, which we refer to as \textit{choice-based feedback}. To illustrate, consider a company seeking to understand customer preferences among a set of items (e.g., product prototypes for commercialization). The company may pursue one of two objectives: identifying the best item or ranking the entire set of items. To achieve these goals, the company can present subsets of items to customers, asking them to select their favorite within each set. The company can dynamically adjust these display sets based on previous feedback. The central challenge lies in designing these display sets to make the learning process \textit{efficient} -- minimizing the cost of feedback collection while ensuring high accuracy in the final outcomes.

Choice-based feedback offers both opportunities and challenges. On the one hand, choices and comparisons provide a natural and intuitive form of feedback. Its advantages over alternative formats such as ratings or scores are discussed across various disciplines such as opinion research (\citealp{krosnick1988test}), psychology (\citealp{goffin2011all}) and computer science (\citealp{shah2014better}). On the other hand, the combinatorial nature of display sets (also known as assortments) introduces significant complexity, especially when combined with dynamic learning aspects. Systematic studies of this problem remain relatively nascent.

In this regard, our paper contributes to an emerging literature that brings machine learning and operations research tools to this type of problem (see \citealp{JMLR:v19:17-607}, \citealp{chen2018nearly}, \citealp{feng2022robust}, \citealp{araman2022diffusion}, \citealp{feng2023mallows} for related works). Among these, the work by \citet{feng2022robust} is most closely related to ours. They introduced a relatively general framework for modeling customer preferences through discrete choice probabilities. Instead of parametric choice models such as Multinomial Logit (MNL), they only imposed certain consistency and separability conditions on the choice probabilities, namely, a more preferred item is chosen with strictly higher probabilities. Under this modeling framework, they studied a best-item identification problem under the fixed-confidence setting, i.e., aiming to minimize the feedback required to guarantee a desired level of confidence. Leveraging an information-theoretic measure dating back to the work of \cite{chernoff1959sequential}, they proposed a randomized policy called the \textsc{Myopic Tracking Policy} (\textsc{MTP}) and showed that it is worst-case asymptotically optimal. Their work also highlighted a useful trade-off in this problem: larger display sets increase coverage by comparing more items simultaneously but may reduce the precision of individual comparisons. Conversely, smaller sets (e.g., pairwise displays) enhance precision but limit coverage.

\subsection{Summary of Contributions}

While \cite{feng2022robust} provided a principled approach to the optimal learning problem, \textsc{MTP} has notable limitations. One pressing issue is that it requires repeatedly solving combinatorial optimization problems throughout the time horizon, which restricts the scalability of the algorithm. Furthermore, the theoretical guarantee of \textsc{MTP} has two important limitations: (i) it focuses on the hardest-to-learn instances with limited insights for general cases, and (ii) it allows a residual term on the order of $o(\log(1/\delta))$, where $\delta$ is the target error probability. These two limitations imply that the guarantees of \textsc{MTP} may be weak for general instances and when the target error probability is only moderately small.

The first part of our paper revisits the best-item identification problem, also referred to as ``learning-to-select.'' We propose a surprisingly simple algorithm, \textsc{Nested Elimination} (\textsc{NE}), which significantly improves upon earlier approaches by (i) being computationally simpler and (ii) offering stronger theoretical guarantees. Our main contributions are as follows:

\begin{enumerate}[label = (\roman*)]
    \item \textit{Simpler Implementation.} \textsc{NE} employs a ``nested'' structure, shrinking display sets on a pathwise basis. This is combined with a carefully designed (but easy-to-implement) sequence of hitting times that determine when and how suboptimal items are eliminated. By avoiding the need to solve combinatorial optimization problems, \textsc{NE} achieves a running time reduction of up to \textit{three orders of magnitude} compared to \textsc{MTP}; see Section~\ref{section_experiment1}.

	\item \textit{Stronger Theoretical Guarantee.} We provide a thorough theoretical analysis of \textsc{NE}'s performance from multiple perspectives. For \textit{every} preference instance $ f $ (not just the worst-case one) and every error tolerance \( \delta \), we derive a non-asymptotic and instance-specific bound on the sample complexity of \textsc{NE}; see Theorem~\ref{theorem_fixedconfidence}. Notably, this bound can be written in the form of $$ \frac{\log(1/\delta)}{I^{\mathrm N}(f)} + C_f, $$ 
	where $ I^{\mathrm N}(f) $ is an explicit function of the instance $ f $ and $ C_f $ is a constant \textit{independent of} $ \delta $. 
	
	This bound universally outperforms that of \textsc{MTP}, where the improvement can be of order \( \Omega(\log(1/\delta)) \); see Section \ref{subsection_best_comparisons} for more detailed discussion. Furthermore, by comparing with the information-theoretic lower bound, we show that \textsc{NE} achieves higher-order worst-case optimality than \textsc{MTP}, where the ``sensitivity'' of the optimality criterion sharpens from $  O(\log(1/\delta)) $ to $ O(1) $ (see Proposition~\ref{prop_OA} and the discussion thereafter).\tbl{We further characterize $ \IN(f) $ across representative subclasses of $ \mathcal M_p $ in Section~\ref{sec:understanding-NE-beyond-OA} and shed light on \textsc{NE}'s performance beyond the worst case.}
\end{enumerate}

\vskip .2 cm

Our analysis develops a collection of technical insights and methods
for constructing nested procedures. In the second part of the paper, we apply them to the more challenging full-ranking identification problem, which we refer to as ``learning-to-rank.'' We introduce a divide-and-conquer algorithm named \textsc{Nested Partition} (\textsc{NP}), as detailed in Section~\ref{subsection_ranking_algorithm}. The elimination procedure \textsc{NP} mirrors that of the well-known Quicksort algorithm \citep{hoare1962quicksort} and similarly recursively partitions the active set into two parts, where items in one part are deemed superior to those in the other. 
Similar to the analysis of \textsc{NE}, we theoretically establish \textsc{NP}'s sample complexity in the form of
$$ \frac{\log(1/\delta)}{\JN(f)} + C_f', $$ 
where $ \JN(f) $ is an explicit function of the instance $ f $ and $ C_f' $ is a constant \textit{independent of} $ \delta $; see Theorem \ref{theorem_ranking_fixedconfidence}. By comparing with the information-theoretic lower bound for the full-ranking identification problem, we show that \textsc{NP} attains\tbl{the same worst-case asymptotic optimality} for the ranking problem; see Propositions~\ref{prop_OA_ranking} and \ref{prop_worstcase_ranking} and Theorem~\ref{theorem_lower_bound_ranking}.

\tbl{The nested approach serves as a versatile foundation for other variations of the problem. The remainder of the paper thus further develops it by incorporating a capacity constraint $ \kappa $ for the display sets, especially when $ \kappa \ll K $. For learning-to-select, we introduce \textsc{Hierarchical Nested Elimination} (\textsc{HNE}), which arranges the items into a tournament and calls \textsc{NE} as a local comparison oracle. For learning-to-rank, we introduce \textsc{Nested Sort-and-Verify} (\textsc{NSV}). It iterates between sorting with \textsc{MergeSort} using \textsc{NE} on pairs as a noisy comparison oracle, and verifying the candidate ranking on its adjacent pairs at higher accuracy. In each case, we (i) derive a corresponding hardness quantity, (ii) establish a $ \delta $-PAC sample-complexity guarantee, and (iii) prove an information-theoretic lower bound; see Theorems~\ref{theorem_fixedconfidence_capacity} and~\ref{thm_worstcase_capacity} for selection, and Theorems~\ref{theorem_fixedconfidence_capacity_ranking} and~\ref{thm_worstcase_capacity_ranking} for ranking. Both algorithms are worst-case asymptotically optimal in the pairwise regime $ \kappa=2 $. For $ \kappa \geq 3 $, \textsc{HNE} remains minimax optimal up to a factor that depends on $ \kappa $ alone (Proposition~\ref{prop:constant optimality}), and \textsc{NSV} is worst-case optimal within a universal factor of $ 1+p<2 $.}

\vspace{0.2 cm}
\noindent\textbf{Methodological Innovations.} We also find it helpful to briefly explain our main technical challenges and how we overcome them methodologically. Let us start with the challenges.
\begin{enumerate}[label = (\roman*)]
	\item While the nested approach is intuitive and provides simple structures, it is unclear \textit{a priori} whether nested structures should be optimal, and if so, in what sense.
	\item Even within the realm of nested procedures, many moving parts need to be fixed. For example, when should an item be eliminated, and if so, which one? This ultimately boils down to a sequence of stopping problems, and it is unclear \textit{a priori} what those stopping times are. 
	\item Although \textsc{NE} and \textsc{NP} are easy to implement, their analysis is nontrivial. Specifically, the history-dependent elimination criteria and the need to ``transfer'' information across assortments make it difficult to decouple the analysis across stages or items. This makes the system dynamics challenging to characterize.
\end{enumerate}

We address the first challenge by establishing a connection between our nested approach and the nested structure of the optimal solution to the max-min problem associated with the Chernoff-type information-theoretic measure. In other words, our approach of making the display sets shrink pathwise is not ad hoc. Rather, it relates to the fact that the optimal allocation among display sets is ``naturally'' nested, at least under the worst-case instances. We address the second challenge by establishing a connection between our elimination criteria and a certain type of sequential probability ratio test (SPRT). It turns out that with the right perspective (e.g., the right hypotheses to test as well as the right choice model classes), we can design simple elimination criteria in a principled way. We elaborate on those ideas in greater detail in Sections \ref{sec:key technical insights} and \ref{section_Methodology}.

Finally, to address the third challenge, we represent the system dynamics with a sequence of multi-dimensional random walks, where the initial state of the later-stage random walk depends on the ending state of the earlier stage; see Figures~\ref{fig:process_NE} and \ref{process_NP} for illustrations. This allows us to reduce the analysis to characterizing the hitting time and hitting distribution of the random walk at every stage. Using tools such as martingale theory, we are thus able to conduct a tight analysis, which is ultimately reflected in the residual terms on the order of $ O(1) $ in Theorems~\ref{theorem_fixedconfidence} and \ref{theorem_ranking_fixedconfidence}.

As can be seen, our algorithm design ideas and proof techniques differ from the classical successive elimination-based algorithms for multi-armed bandit problems, typically based on estimating the expected reward of each arm (\citealp{even2006action,kalyanakrishnan2010efficient,karnin2013almost}). Our approach is also distinctively different from the popular approach based on the track-and-plug-in strategies inspired by the information-theoretic lower bound in pure-exploration problems \citep{chernoff1959sequential,garivier2016optimal, feng2022robust}. We believe our approach holds independent significance and serves as a useful foundation for the development of online learning algorithms for various purposes.

\subsection{Literature Review}

Our problem could be cast as a \textit{pure exploration} problem with structured (e.g., choice-based) feedback. In this sense, it could be viewed as a variant of the best arm identification problem \citep{even2006action,audibert2010best,karnin2013almost,garivier2016optimal}, which further finds its root in active sequential hypothesis testing (\citealp{chernoff1959sequential,naghshvar2013active}). Specifically, the work of \citet{garivier2016optimal} provided an information-theoretic characterization of the expected sample complexity of best arm identification, and achieved the minimal complexity asymptotically through a track-and-plug-in strategy.
While our tasks of best-item and full-ranking identification can be conceptualized as pure exploration problems, the underlying model significantly deviates from standard multi-armed bandits. Specifically, the decision variable in our context is a subset of items referred to as a \textit{display set}, instead of a single item, and the observation is an item rather than a stochastic reward. Therefore, we refrain from using the terminology ``arm'' to prevent ambiguity.

Related to the literature on best-item/full-ranking identification, there also exists a strand of research that incorporates the choice-based feedback model into the paradigm of machine learning. While much of this literature focuses on investigating noisy pairwise comparison models \citep{braverman2008noisy,  ailon2012active, wauthier2013efficient, shah2018simple, heckel2019active}, there are a few exceptions that have delved into learning from multiway comparisons, akin to our model.
In particular, \citet{chen2018nearly} studied the problem of top-item identification under a Luce-type choice model, which is different from the class of choice models we consider in this work, as described in Section~\ref{section_setup}. Additionally, they considered a different asymptotic regime. That is, they fixed the confidence level while letting other instance-specific parameters (such as the number of items) tend to infinity. In contrast, our approach maintains a fixed instance and lets the confidence level tend to zero, which is more commonly adopted in the literature on pure exploration. 
Furthermore,  \citet{saha2020best} considered the problem of identifying a near-optimal item under a random utility-based discrete choice model, where each item is associated with an unknown random utility score. Nevertheless, they fixed the size of the display sets, whereas we allow for display sets of varying sizes. 
Finally, we remark that, to the best of our knowledge, only the results in \citet{feng2022robust} are comparable to ours.
We finally note that a preliminary version of this work appeared in the \emph{Proceedings of the 40th International Conference on Machine Learning (ICML 2023)}; see \citet{yang23b}. In this journal version, we have significantly expanded upon the earlier work; see Appendix~\ref{appendix:conference} for a brief summary of incremental contributions.

\section{Problem Setup and Preliminaries}
\label{section_setup}

\textbf{Preferences and Choice-Based Feedback.} We consider a choice-based feedback model in which a customer randomly selects one item from the display set presented by the company (or agent). We denote the universe of available items as $[K]:= \{1,2,\ldots, K\}$. The \textit{ranking} over the items is represented by a bijection $ \sigma: [K] \to [K] $, so that $ \sigma(i) = k $ means item $ i $ is in the $ k $th highest position. The company uses \textit{display sets} as an informational lever to collect the customer's feedback efficiently. The collection of all the possible display sets is $\mathcal{S}:=\{S \subseteq [K],|S| \geq 2\}$.\footnote{Note that the case where the display set is a singleton is completely uninformative.} The choice behavior is modeled by the probability $f(i|S)$ that item $i$ is chosen from the display set $S$ for every $S\in\mathcal S$ and $i\in S$. We refer to the  collection of choice probabilities $ \{f(i|S): i\in S, S\in\mathcal S\} $ as a \emph{preference} instance $f$, which is unknown to the company. Clearly, the choice probabilities will greatly affect the efficiency of the learning problem. In this regard, we follow the notation of \citet{feng2022robust} and consider a broad class of preference instances, termed the \textit{$p$-Separable} family $\mathcal M_p$; see Definition~\ref{definition_p_family} below.

\begin{definition}[$p$-Separable family]
Let $p\in(0,1)$ be a fixed dispersion parameter. A preference $f$ belongs to the $p$-Separable family $\mathcal M_p$ if:
\label{definition_p_family}
\begin{enumerate}[label = (\roman*), noitemsep, nolistsep]
\item For any $S\in\mathcal S$, $f(i | S)>0$  if and only if $i\in S$;
\item For any $S\in\mathcal S$, $\sum_{i\in S}f(i | S)=1$;
\item There exists a global ranking $\sigma_f: [K]\to [K]$ such that for any $S\in\mathcal S$ and $i, i'\in S$, $f(i|S)\le p f(i'|S) $ if  $\sigma_f(i')<\sigma_f(i)$.
\end{enumerate}    
\end{definition}

\vspace{0.1 cm}
\begin{remark}
	{\sf  It is worth noting that the \textit{$p$-Separable} family $\mathcal M_p$ of preference instances is relatively general. Essentially, we assume that the choice probabilities corresponding to $ f $ are (statistically) consistent with some (unknown) ranking of items. In addition, the choice probabilities are separable by at least a factor of $ p $. In this way, the underlying ranking, and hence the top-ranked item, and so on, is uniquely defined. Many common choice models, such as the multinomial logit (MNL) model and the Mallows choice model, could be incorporated into this framework. See Remark~2 of \citet{feng2022robust} for more discussion. \hfill $ \diamond $}
\end{remark}

\vskip 0.1 cm

\tbl{
	\begin{remark}
		\label{remark:mixture}
		{\sf We note that the existence of a single global ranking in Definition~\ref{definition_p_family} is central to our formulation, as it is what renders the learning target a well-defined functional of $ f $. One might hope to dispense with it by positing a mixture model over customer type clusters, $ f=\sum_{\ell=1}^{L}w_\ell f_\ell $ with each $ f_\ell\in\mathcal M_{p_\ell} $ carrying its own consensus ranking $ \sigma_\ell $, and by targeting the $ \sigma_\ell $'s instead. The consensus rankings at the cluster levels, however, are not identifiable from $f$ alone in general. In Proposition~\ref{prop:nonidentifiability} in Appendix~\ref{append:mixture-nonidentifiability}, we construct a well-behaved preference $ f\in\mathcal M_{27/29} $, which admits two cluster decompositions whose consensus rankings and top-ranked items are irreconcilable. Without side information such as observable or inferable cluster labels, component-level rankings cannot in general be recovered, even if $f$ is known exactly.\hfill $ \diamond $}
	\end{remark}
}

\vspace{0.1 cm}
\begin{remark}
\label{remark:p_conservative}
{\sf The separation parameter $p$ measures the ``noise level'' of the choice-based feedback model. Throughout the paper, we perform our analysis treating the value of $ p $ as known and given. However, note that $\mathcal M_p \subseteq \mathcal M_{p'}$ for all $p < p'$. Therefore, if only a conservative estimate (i.e., an upper bound) of $p$, say, $ p' $ is available, our theoretical performance guarantees for the algorithm still hold after replacing $ p $ with $ p' $.\hfill $ \diamond $}
\end{remark}

\vspace{0.1 cm}

For ease of exposition, we assume throughout this paper that the unknown global ranking $\sigma_f$ of the underlying preference $f$  is the identity ranking $\sigma_* := (1,2,\ldots, K)$ without loss of generality. Accordingly, item $1$ is always the top-ranked item.

\vspace{0.2 cm}
\noindent\textbf{Problem Formulation. } 
The company aims to achieve a specific goal by displaying subsets of the item set $[K]$ to customers with an unknown consensus preference $f$ sequentially and adaptively. Specifically, at each time step $t\in \mathbb N^+:=\{1,2,3,\ldots\}$, the company chooses one display set $S_t\in \mathcal S$ and presents it to one customer. Then the customer selects an item $X_t \in S_t$ according to the underlying probability distribution, $f(\cdot|S_t)$.

In this study, we consider two distinct yet related objectives for the company: learning-to-select (best-item identification) and learning-to-rank (full-ranking identification). Specifically, the company employs an \emph{online} policy $ {\uppi}$ to (i) decide the display set $S_t$ to present at each time step $t$; (ii) select a time $\tau$ to stop the interactions, and (iii) ultimately make a recommendation. Here, a recommendation is represented by $i_{\mathrm{out}}$ if the goal is learning-to-select; and $\sigma_{\mathrm{out}}$ if the goal is learning-to-rank. More formally, let $\mathcal F_t$ denote the sigma-field generated by the history of display sets and customers' choices up to and including time $t$, i.e., $(S_1,X_1,\ldots,S_t,X_t)$. Therefore, the online algorithm ${\uppi}$ comprises three components:
\begin{itemize}[noitemsep, nolistsep]
	\item The \emph{display rule} selects $S_t$ (with possible randomization), which is adapted to the filtration $\mathcal F_{t-1}$;
	\vspace{3pt}
	\item The \emph{stopping rule} determines a stopping time\footnote{In this work, we slightly abuse the terminology \emph{stopping time}, although the context should make our usage  clear. In fact, $\tau$ is both a stopping time with respect to the corresponding filtration and the time step to terminate the algorithm.} $\tau$, which is adapted to the filtration $(\mathcal F_t)_{t=1}^{\infty}$;
	\vspace{3pt}
	\item The \emph{recommendation rule} produces a candidate best item  $i_{\mathrm{out}}$ or full ranking $\sigma_{\mathrm{out}}$, which is $\mathcal F_{\tau}$-measurable.
\end{itemize}

To facilitate comparisons with previous work, we will also adopt the \emph{fixed-confidence} setting in the theoretical analysis. 
In the fixed-confidence setting, a confidence level $\delta\in(0,1)$ is given. Then the company is required to identify the best item or the full ranking with probability at least $1- \delta$  using the fewest time steps (i.e., samples). 

\vskip 0.1 cm
\begin{definition}[$\delta$-PAC policy]
	\label{definition_delta_pac}
Consider either the problem of best-item or full-ranking identification. For a prescribed confidence level $\delta\in (0,1)$, an online policy ${\uppi}$ is said to be $\delta$-PAC ({\em probably approximately correct)} if for all preferences $f\in\mathcal M_p$, it terminates within a finite time almost surely (i.e., $\mathbb P (\tau<\infty) = 1$) and the probability of error is no more than $\delta$ (i.e., $\mathbb P (i_{\mathrm{out}} \neq 1 ) \le \delta$ for best-item identification, or $\mathbb P (\sigma_{\mathrm{out}} \neq \sigma_* ) \le \delta$ for full ranking). Furthermore, for a class of policies $ \Pi = \{{\uppi}_\delta\}_{\delta \in (0,1)} $ parameterized by $ \delta $, we say it is \textit{PAC} if $ {\uppi}_\delta $ is $\delta$-PAC for all $ \delta $.
\end{definition}
\vskip 0.1 cm 

In this regard, our overarching objective is to design and analyze $\delta$-PAC policies while minimizing their expected sample complexities $\E [\tau]$ for the best-item and full-ranking identification problems respectively.

\vspace{0.2 cm}
\noindent\textbf{Other Notations.}
For any display set $S\in \mathcal S$ and its subset $S'\subseteq S$, we define $f(S' |S):=\sum_{i\in S'}f(i|S)$, which is the probability that a customer with preference $f$ chooses one item in the subset $S'$ when presented with display set $S$.  
Consider any multivariate function $g: \mathbb R \times \mathbb R ^{n-1} \to \mathbb R$, and any univariate function $h: \mathbb R \to \mathbb R$. For any fixed $y\in\mathbb R ^{n-1}$, we say $g(x,y)=O_x(h(x))$ (resp. $\Omega_x(h(x))$) if there exists a positive constant $c$ and a constant $x_0$ (possibly dependent on parameter $y$) such that $|g(x,y)|\le c\cdot h(x)$ (resp. $|g(x,y)|\ge c\cdot h(x)$) for all $x\ge x_0$. Alternatively, we say $g(x,y)=o_x(h(x))$ (resp. $\omega_x(h(x))$) if for any positive constant $c$, there exists a constant $x_0$ (possibly dependent on parameter $y$) such that $ |g(x,y)|<c\cdot h(x)$ (resp. $|g(x,y)|> c\cdot h(x)$) for all $x\ge x_0$.

\section{The Learning-to-Select (Best-Item Identification) Problem} 
\label{section_best}

\tbl{This section studies the best-item identification problem from choice-based feedback. In Section~\ref{subsection_best_algorithm}, we introduce \textsc{Nested Elimination} (\textsc{NE}), a structurally simple and computationally efficient elimination algorithm. Section~\ref{subsection_best_results} analyzes its performance, where we show that it is $\delta$-PAC, establish a non-asymptotic and instance-specific bound on its expected sample complexity, and, by comparing this bound against the information-theoretic lower bound, prove that \textsc{NE} is worst-case asymptotically optimal. Section~\ref{sec:understanding-NE-beyond-OA} moves beyond the worst case, characterizing how the sample complexity of \textsc{NE} varies across several representative classes of instances. Section~\ref{subsection_best_comparisons} then compares \textsc{NE} with policies from previous work---most notably \textsc{MTP}---from different angles. Finally, Section~\ref{sec:key technical insights} distills the two ideas behind \textsc{NE}, its nested display structure and its SPRT-based elimination rule, both of which we build on for the full-ranking problem.}

\subsection{The \textsc{Nested Elimination} Algorithm}
\label{subsection_best_algorithm}

\begin{figure}[htbp]
	\centering
	\includegraphics[width=.6\textwidth]{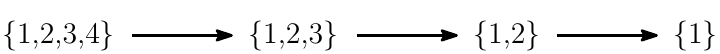}
	\vspace{0.2 cm}
	\caption{\textbf{A possible trajectory of $ S_{\mathrm{active}} $ under \textsc{NE}.} The item with the lowest vote is eliminated according to the criterion \eqref{eq:stopping rule}. }     
	\label{fig_NE_illustration}
\end{figure}

As the name suggests, our algorithm is elimination-based. It maintains an \textit{active} set, denoted by $S_{\mathrm{active}}$, that starts as the full set and shrinks over time. At each time step $t$, \textsc{NE} displays $S_{\mathrm{active}}$ to the next customer, and observes the choice $X_t \in S_{\mathrm{active}}$. The central part of the algorithm concerns a \textit{simple} rule to determine which items are eliminated and when, where the meaning of ``simplicity'' is two-fold.
\begin{enumerate}
	\item First, this rule is based on simple sufficient statistics. It maintains a system of \emph{voting scores} for every item $ i \in [K] $, denoted by $W_t(i)$, which counts the number of times that item $i$ is chosen up to time $t$ regardless of display set history.
	\item Second, the elimination criterion is easy to implement. Specifically, let $ \pi_t $ be a ranking over the active items based on their scores. Formally, $ \pi_t $ is a bijection from $ [|S_{\mathrm{active}}|] $ to  $ S_{\mathrm{active}} $ so that the $i^{\mathrm{th}}$ most voted item is  $\pi_t(i)$ for every $i\in [|S_{\mathrm{active}}|]$.  The active set shrinks to a smaller one with size $ k $ ($ k <|S_{\mathrm{active}}|$ ) if the voting scores of the top-$ k $ items satisfy the following condition:
	\begin{align}\label{eq:stopping rule}
	\sum_{i=1}^k W_t(\pi_t(i))-kW_t(\pi_t(k+1))\ge M.
	\end{align}
	Roughly speaking, the bottom-ranked items are eliminated if their scores are ``far exceeded'' by the top-$ k $ most voted items.\footnote{In fact, Condition \eqref{eq:stopping rule} can be further simplified by nominally only eliminating one item at a time. That is, we can take $ k = |S_{\mathrm{active}}| -1 $  without loss of generality (but allowing multiple eliminations between observations). This equivalent description of \textsc{NE} is formally summarized in Algorithm~\ref{algo2} in the appendix.}

	 In the equation above, there is a tuning parameter $M>0$, which plays an essential role in controlling the accuracy of the eliminations. We will discuss the choice of parameter $M$ in Theorem~\ref{theorem_fixedconfidence} further. As a general rule, the larger the parameter $M$, the more effective the eliminations are in preserving the best item. This, in turn, leads to a lower probability of outputting suboptimal items.
\end{enumerate}
As the algorithm progresses, only one item $i_{\mathrm{out}}$ eventually remains in the active item set $S_{\mathrm{active}}$. That will be the output of \textsc{NE}. We provide a pseudocode description of \textsc{NE} in  Algorithm~\ref{algo1}, as well as an illustration in Figure~\ref{fig_NE_illustration}.

\begin{algorithm}[t]
	\caption{Nested Elimination (\textsc{NE})}
	\label{algo1}
	\hspace*{0.02in} {\bf Input:} Tuning parameter $M>0.$
 
        \hspace*{0.02in} {\bf Output:} The only element $i_{\mathrm{out}}$ of $S_{\mathrm{active}}$.
	\begin{algorithmic}[1]
		\STATE Initialize voting score $W_0(i)\leftarrow 0$ for all $i\in[K]$, active item set $S_{\mathrm{active}}\leftarrow [K]$, $t \leftarrow 0$.
		\WHILE{$|S_{\mathrm{active}}|>1$}
		\STATE Update the timer: $t \leftarrow t+1$.
		\STATE Display  $S_{\mathrm{active}}$ and observe the choice $X_t \in S_{\mathrm{active}}$. Update voting scores based on $ X_t $:
		\begin{equation*}
		W_t(i) \leftarrow \begin{cases}
		W_{t-1}(i) + 1 & \text{if } i = X_t \\
		W_{t-1}(i) & \text{if }  i \neq X_t.
		\end{cases}
		\end{equation*}
		\vspace{-.8 cm}
		\STATE{Update the active set:}
		\\ (i) Sort the active items based on their voting scores: find  $ \pi_t  $  so that $ W_t(\pi_t(1))\ge W_t(\pi_t(2))\ge \cdots\ge W_t(\pi_t(|S_{\mathrm{active}}|)). $
		\\ \hypertarget{algo1_elimination}{(ii)} Find the smallest $k$ such that \eqref{eq:stopping rule} is satisfied. If such $k$ exists, update the active set by only keeping the top $ k $ items: $ S_{\mathrm{active}}  \leftarrow \{\pi_t(1), \ldots, \pi_t(k)\} $.
		\ENDWHILE
	\end{algorithmic}
\end{algorithm}

\begin{figure}[htbp]
	\centering
	\hspace{-35pt}
	\includegraphics[width=.8\textwidth]{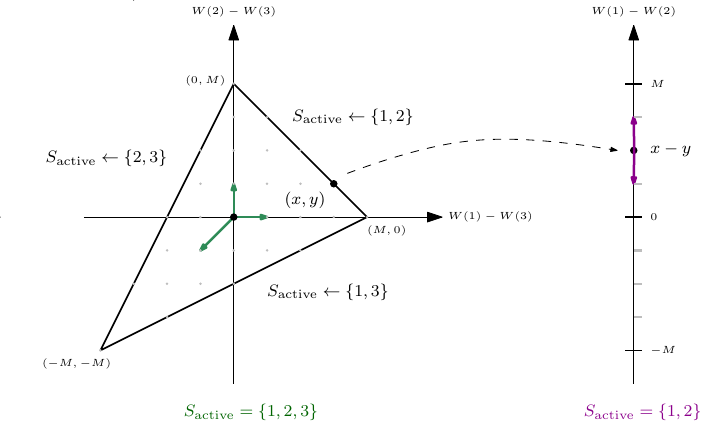}
	\caption{\textbf{A visualization of the system dynamics under \textsc{NE}.} 
		Let \( K = 3 \). In the first stage, the active set is \( [3] = \{1, 2, 3\} \). The system dynamics are visualized by projecting the state variables \( \{W(i)\} \) onto the two-dimensional space spanned by \( \big(W(1) - W(3),\, W(2) - W(3)\big) \). This projection results in a random walk that begins at the origin and evolves according to an i.i.d.\ sequence with possible increments of \( (1, 0) \), \( (0, 1) \), and \( (-1, -1) \), occurring with probabilities \( f(1 | [3]) \), \( f(2 | [3]) \), and \( f(3 | [3]) \), respectively. The first stage finishes when the random walk reaches the boundary of a triangle defined by vertices \( (0, M) \), \( (M, 0) \), and \( (-M, -M) \). Each face of the triangle corresponds to the elimination of one item. In the illustrated path, item~3 is eliminated, and the active set updates to \( [2] = \{1, 2\} \), which is an event with high probability under any OA preference instance.
		\\ In the second stage, the state variables are further projected into the one-dimensional space spanned by \( W(1) - W(2) \). That results in a one-dimensional random walk, starting from the endpoint inherited from the first stage. It evolves by increments of \( +1 \) or \( -1 \) with probabilities \( f(1 | [2]) \) and \( f(2 | [2]) \), respectively. The second stage ends when the random walk reaches the endpoints \( M \) or \( -M \), which corresponds to the selection of item~1 or item~2, respectively.
	} 
	\label{fig:process_NE}
	\vspace{4pt}
\end{figure}

\begin{remark}
	\label{remark_randomwalk}
	{ \sf One key observation from \textsc{NE} is that at every stage (i.e., during the time steps between item eliminations), the ``active'' voting scores \( \{W_t(i) : i \in S_{\text{active}}\} \) evolve according to a (biased) multi-dimensional random walk. The elimination criterion corresponds to the hitting time of the boundary of a polytope; see Figure~\ref{fig:process_NE} for an illustration. As a result, analyzing \textsc{NE} reduces to studying the expected hitting time and the hitting distribution at each stage. This structure provides significant insights into the system dynamics, allowing us to leverage tools such as martingale theory to derive non-asymptotic bounds.\footnote{\label{fn:gambler}For instance, when \( K = 2 \), the random walk simplifies to the well-known (one-dimensional) \emph{gambler's ruin} problem. In this problem, the player wins one dollar with probability \( f(1 \mid [2]) \geq \tfrac{1}{1 + p} \) and loses one dollar with probability \( f(2 \mid [2]) \leq \tfrac{p}{1 + p} \) at each step, quitting when either \( M \) dollars are won or lost. The error probability in our problem (i.e., \textsc{NE} outputting the incorrect item) corresponds to the probability that the player ends up losing, while the sample complexity corresponds to the expected number of steps before the player quits. In this simplest case, both quantities can be characterized in closed form.}
	\hfill $ \diamond $}
\end{remark}

\subsection{Theoretical Analysis of \textsc{NE}}
\label{subsection_best_results}

Let us start by introducing some notation. The sample complexity of \textsc{NE} is characterized by a novel instance-specific \emph{hardness quantity}. \tbl{We denote it by \(\IN(f)\) and give a closed-form expression below.} In addition, we define  $\beta_K := 2^{K-1}-1$, a constant independent of $ \delta $. Finally, we assume that the parameter $M$ is an integer without loss of generality. In general, $M$ appearing in the analysis should be replaced by $\lceil M \rceil$ without affecting other expressions. We now present our first main result below.

\vskip 0.2 cm
\begin{theorem}[Sample complexity of \textsc{NE}] 
	\label{theorem_fixedconfidence}
	For every confidence level $\delta\in(0,1)$, \textsc{NE} is $\delta$-PAC with the parameter value
	\begin{equation}
	\label{equation_M}
	M = \frac{\log(1/\delta)+\log(\beta_K)} {\log(1/p)}.
	\end{equation}
	Furthermore, for every preference instance $f\in\mathcal M_p$, there is a constant $ C_f $ independent of $ \delta $ such that 
	\begin{equation}
	\label{equation_fixedconfidence}
	\E [\tau] \le \frac{\log(1/\delta)}{I^{\mathrm N}(f)}+ C_f.
	\end{equation}
\end{theorem}
\vskip 0.2 cm

Theorem~\ref{theorem_fixedconfidence} shows a few things simultaneously. First, it establishes that \textsc{NE} is $\delta$-PAC for suitable choices of the input parameter $M$. Second, it provides a sharp, instance-specific characterization of \textsc{NE}'s sample complexity via the hardness measure $I^{\mathrm N}(f)$. Roughly speaking, it measures the average informativeness of a vote under \textsc{NE} in the long run. 

We highlight that a notable feature of our result is that the residual term is bounded in $\delta$. In contrast, standard Chernoff-type analyses typically yield residual terms of order $o(\log(1/\delta))$; see Section~\ref{subsection_best_comparisons} for further discussion. \tbl{Another notable feature of the result is that it provides a closed-form characterization for $ I^{\mathrm N}(f) $, which we will elaborate on more later. } The proof of Theorem~\ref{theorem_fixedconfidence} is deferred to Appendix~\ref{appendix_theorem_fixedconfidence}. As a brief sketch, it builds on a series of intermediate results that may be of independent interest. In Proposition~\ref{proposition_tau_general}, we show that the expected stopping time of \textsc{NE} with input parameter $M$ can be characterized by
$
\frac{\log(1/p)}{I^{\mathrm N}(f)} M
$
up to a negligible term as $M$ becomes large. Then we establish in Proposition~\ref{proposition_error} that the error probability of \textsc{NE} is uniformly upper bounded by $p^M\beta_K $.  As mentioned in Remark~\ref{remark_randomwalk}, our analysis leverages the analytical tractability afforded by the simple structure of \textsc{NE}, which boils down to a sequence of random walks.

\vspace{0.2 cm}
\tbl{
	\noindent\textbf{The Hardness Quantity $ \IN (f) $.}
	Part of Theorem~\ref{theorem_fixedconfidence} is the explicit formula for the hardness measure $ \IN(f) $ itself, which we present now. Let us define
	\begin{align*}
	\Delta^{i}_{r}   := f(K-r+1|[K-i+1])-f(K-r+2|[K-i+1])
	\end{align*}
	for every $ r \in [K-1] $ and $ i \in [r-1] $, which is the rate at which items $ K-r+1 $ and $ K-r+2 $ separate in votes during stage $ i $. Following Algorithm~\ref{algo1}, we then recursively define
	\begin{align}\label{equation_definition_Dfr_NE}
	D(f,1) := \frac 1 {1- Kf(K|[K])} \quad \text{ and } \quad D(f, r) := \frac {(K-r+1)\sum_{i=1}^{r-1}  \Delta^{i}_{r}  D(f,i)} {1-(K-r+1)f(K-r+1|[K-r+1])}
	\end{align}
	for all $r\in [K-1] \setminus \{ 1\}$. Intuitively, $ D(f,r) $ represents the time that Algorithm~\ref{algo1} takes to complete stage $ r $ after normalizing $ M $ to be one, in an idealized setting with deterministic fractional votes, where item $ i $ receives $ f(i|[K-r+1]) $ votes in every period.\footnote{\tbl {To see this, recall from \eqref{eq:stopping rule} that stage $ r $ ends when the margin $ \sum_{i=1}^{K-r}(W(i)-W(K-r+1)) $ reaches $ M $, accruing at the constant rate $ 1-(K-r+1)f(K-r+1|[K-r+1]) $. For $ r=1 $ the margin starts from zero, so accumulating all of $ M $ gives $ D(f,1) $. For $ r\ge2 $, stage $ r $'s margin inherits the scores from stage $ r-1 $. Rearranging that stage's stopping condition shows the inherited margin falls short of $ M $ by $ (K-r+1)(W(K-r+1)-W(K-r+2)) $, which stage $ r $ must still accumulate. To evaluate $ W(K-r+1)-W(K-r+2) $, note this pairwise difference accrues at rate $ \Delta^i_r $ during each earlier stage $ i\in[r-1] $, of length $ MD(f,i) $, and hence equals $ M\sum_{i=1}^{r-1}\Delta^i_r D(f,i) $. Dividing the residual by the accrual rate gives the length of stage $ r $; normalizing by $ M $ recovers \eqref{equation_definition_Dfr_NE}.}} We now define the hardness quantity under \textsc{NE} by
	$$\IN(f):= \log\left(\tfrac 1 p \right) \left[\sum\nolimits_{r=1}^{K-1} D(f,r)\right]^{-1}, $$
	where the factor $ \log\left(1/p\right) $ converts the raw voting scores into statistical confidence by taking into account the $ p $-separability of the customer choices. The two ingredients of \eqref{equation_fixedconfidence} can thus be combined transparently. In the proof we show that, after taking into account the stochastic votes, \textsc{NE} runs for roughly $ M \sum_{r=1}^{K-1} D(f,r) $ periods. Meanwhile, the $ \delta $-PAC requirement \eqref{equation_M} calls for $ M \approx \log(1/\delta)/\log(1/p) $, and hence $ \E[\tau] \approx \log(1/\delta)/\IN(f) $.
}

\vspace{0.2 cm}
\noindent\textbf{Information-Theoretic Lower Bound.} We next compare \textsc{NE}'s sample-complexity guarantee with the
information-theoretic lower bound for best-item identification.
Let $\mathcal P (\mathcal S)$ denote the collection of all probability distributions over $\mathcal S$. For any display set $S\in \mathcal S$ and probability distribution ${\lambda\in \mathcal P (\mathcal S)}$, we write
the KL divergence between preferences $f$ and $f'$ with respect to $S$ and $\lambda$ as
$$
D_S(f \| f') := \sum_{i\in S}  f(i|S) \log \tfrac {f(i|S)} {f'(i|S)} \, \quad \text{ and } \quad \,  D_\lambda(f \| f') :=
 \sum_{S\in \mathcal S} \lambda(S)  D_S(f \| f')
$$
respectively. For any preference $f\in\mathcal M_p$, we define  $\overline{\mathcal{M}}_p(f) := \{f'\in \mathcal M_p: \sigma_{f'}^{-1}(1)\neq \sigma_f^{-1}(1)\}$, which represents the set of alternative preferences with different best items. 
\cite{feng2022robust} developed the following information-theoretic lower bound for this learning-to-select problem.

\begin{fact}\label{fact_lower_bound}
	For any preference $f\in\mathcal M_p$, let 
		\begin{align}\label{eq:max-min}
		I_*(f) := \sup_{\lambda\in \mathcal P (\mathcal S)\!} \ \inf_{f'\in \overline{\mathcal{M}}_p(f)}  D_\lambda(f \| f').
		\end{align}
	Then any $\delta$-PAC best-item identification policy satisfies
	$
	\E [\tau] \ge  \frac {\log(1/\delta) - \log 2.4} {I_*(f)}.
	$
\end{fact}

\tbl{
\vspace{0.2 cm}
\noindent\textbf{The Hardest-to-Learn Instances and Worst-Case Optimality of \textsc{NE}.}
The quantity $I_*(f)$ characterizes the intrinsic difficulty of identifying
the best item under preference instance $f$, but is generally
difficult to evaluate. To that end, \citet{feng2022robust} identified a class of hardest-to-learn preferences that minimize $I_*(f)$, termed the ``Ordinal Attraction'' (OA) model. Specifically, for any ranking $\sigma$, the OA preference instance 
$f_\sigma^{\mathrm{OA}}$ is given by
\[
f^{\mathrm{OA}}_\sigma(i \mid S) = \tfrac{p^{\sigma(i \mid S)-1}}{1 + p + p^2 + \cdots + p^{|S|-1}} \quad \text{for all } S \in \mathcal{S} \text{ and } i \in S.
\]
In other words, the choice probability of an item decays geometrically with its \textit{relative ranking} within the display set.  Let $f^{\mathrm{OA}}$ be the OA instance associated with the identity
ranking and define $
I_*^{\mathrm{OA}}
:=
I_*(f^{\mathrm{OA}}).
$
It is then known that
$
I_*^{\mathrm{OA}} =
\min_{f\in\mathcal M_p}I_*(f).
$

We next show that the same OA instance is also the hardest instance for
\textsc{NE}, and that the corresponding hardness quantity coincides with the
information-theoretic benchmark.}

\begin{proposition}[Minimal value of $I^{\mathrm N}$] \label{prop_OA}
	The value of $ I^{\mathrm N}(f) $ attains its minimum at the OA preference instance $ f^{\mathrm{OA}} $, where it coincides with the minimal value of $ I_\ast(f) $. Formally,
	$$
	f^{\mathrm{OA}} \in \argmin\nolimits_{f\in \mathcal M_p} I^{\mathrm N}(f) \quad \text{and} \quad \min_{f\in \mathcal M_p} I^{\mathrm N}(f) = \min_{f\in \mathcal M_p} I_*(f) = I_*^{\mathrm{OA}}.
	$$
\end{proposition}

We defer its proof to Appendix~\ref{appendix_prop_OA}. Combining Theorem~\ref{theorem_fixedconfidence},
Fact~\ref{fact_lower_bound}, and Proposition~\ref{prop_OA} shows that
\textsc{NE} attains the optimal worst-case leading coefficient. Moreover,
because the residual term in Theorem~\ref{theorem_fixedconfidence} is
bounded in $\delta$, \textsc{NE} satisfies the following
optimality criterion
\begin{align}
\textsc{NE}
\in
\argmin_{\Pi\text{ is PAC}}\ 
\sup_{f\in\mathcal M_p}\ 
\limsup_{\delta\downarrow0}\ 
\frac{
	\mathbb E_f[\tau]
	-
	\tfrac{\log(1/\delta)}{I_*^{\mathrm{OA}}}
}{
	\omega_{1/\delta}(1)
},
\label{eq:optimality of NE}
\end{align}
for any \textit{arbitrarily slowly} growing function $\omega_{1/\delta}(1)$.

\tbl{
\subsection{Discussion: Understanding \textsc{NE}'s Performance Beyond OA Instances} 
\label{sec:understanding-NE-beyond-OA}

While the hardness quantity $\IN(f)$ under OA preferences provides useful insight into the worst-case sample complexity of \textsc{NE}, it is also helpful to understand the behavior of $ \IN $ beyond the worst-case. In this subsection, we discuss three representative classes of preference instances.   The examples progress from instances that exactly attain the worst-case benchmark,
to MNL instances that remain within a constant factor of it (when the gap between the top-two items is small), and finally
to instances on which \textsc{NE} can perform arbitrarily better (when there is a superstar item). Without loss, we will assume that the underlying ranking of $ f $ is the identity ranking  for ease of notation.

\vskip 0.2 cm
\textbf{Example 1: An Equivalence Class Based on Top-$k$ Display Sets.} We make a simple observation from the inductive definition of $\IN(f)$: it depends on the preference instance $f$ only through the choice probabilities associated with the nested collection of top-$k$ display sets, $\mathcal{N}:= \{[k] : k \in \{2,\ldots,K\}\}$. This observation implies that any preference instance $f$ satisfies $ \IN(f) = I_*^{\mathrm{OA}} $ as long as it matches the OA choice probabilities on these nested top-$k$ display sets, namely, if for every $k\in\{2,\ldots,K\}$, 
\begin{align*}
f(i\mid [k])
=
\tfrac{p^{i-1}}{1+p+p^2+\cdots+p^{k-1}}
\quad \text{for all } i\in\{1,\ldots,k\}.
\end{align*}
Besides the OA instance itself, the condition above includes two important choice-model families as special cases: (i) the Mallows choice model (see \citealp{desir2021mallows}); and (ii) a subclass of the \emph{Multinomial Logit} (MNL) model in which the attraction scores form a geometric sequence, namely $v_i=p^{i-1}$ for every item $i$. 

The performance of \textsc{NE} under all preferences in this family can thus be characterized by  $ I_*^{\mathrm{OA}} $, which admits the following closed-form expression (\citealp{feng2022robust})
\begin{align}
I_*^{\mathrm{OA}}  =
(1-p)\log\left(\tfrac1p\right)
\left(
1+
\sum_{n=2}^K
\tfrac{p^{n-1}}
{1+2p+\cdots+(n-1)p^{n-2}}
\right)^{-1}.
\label{eq:hardest to learn}
\end{align}
For future reference, we also adopt the constant-factor bounds of $ I_*^{\mathrm{OA}} $:
\begin{align}
\label{eq:approximating IOA}
\log\left(\tfrac1p\right)(1-p)\tfrac{1}{1+2p}
\leq
I_*^{\mathrm{OA}}
\leq
\log\left(\tfrac1p\right)(1-p)\tfrac{1}{1+p}.
\end{align}
In other words, $I_*^{\mathrm{OA}}$ can be characterized by $\log(1/p)(1-p)$ up to a constant factor, which we adopt for simple comparisons with other hardness quantities.

\vspace{0.1cm}
\textbf{Example 2: MNL Preferences.} The Multinomial Logit model is among the most popular models of choice. In light of its popularity, we investigate the hardness measure $\IN(f)$ for the family of MNL instances. Formally, let $\mathcal{M}_p^{\mathrm{MNL}}$ denote the class of $p$-separable MNL instances. Each preference instance $f^{\mathrm{MNL}}\in\mathcal{M}_p^{\mathrm{MNL}}$ specifies choice probabilities of the form
\begin{align*}
f^{\mathrm{MNL}}(i\mid S)
=
\tfrac{v_i}{\sum_{j\in S}v_j}
\quad \text{for all } S\in\mathcal S \text{ and } i\in S,
\end{align*}
where $v_1,\ldots,v_K$ are the attraction scores and satisfy the $p$-separability condition
$
v_j/ v_i \leq p
$ whenever $ i < j  $. The tight separability parameter of an MNL instance is thus
$
p \;=\; \max_{\ell\in[K-1]} \tfrac{v_{\ell+1}}{v_\ell},
$
which can be read directly off the sorted attraction-score vector.

\noindent \underline{Small gap between $v_1$ and $v_2$.} To build intuition, consider the special case $K=3$. Without loss of generality, suppose $v_1=1$, $v_2=p_2$, and $v_3=p_2p_3$, where $p_2,p_3\in(0,p]$. In this case, the hardness measure can be expressed as
\begin{align*}
\IN(f^{\mathrm{MNL}})
=
\log\left(\tfrac{1}{p}\right)
\cdot
\tfrac{(1-p_2)(1+p_2-2p_2p_3)}
{(1+p_2)^2-p_2p_3-3p_2^2p_3}.
\end{align*}
It is straightforward to verify that $\IN(f^{\mathrm{MNL}})$ is decreasing in $p_3$. Comparing the value at $p_3=0$ and that at $ p_3 = 1 $, respectively,  we obtain
\begin{align*}
\log\left(\tfrac{1}{p}\right)(1-p_2)\tfrac{1}{1+2p_2}
\;\leq\;
\IN(f^{\mathrm{MNL}})
\;\leq\;
\log\left(\tfrac{1}{p}\right)(1-p_2)\tfrac{1}{1+p_2}.
\end{align*}
As a result, up to a constant factor, $\IN(f^{\mathrm{MNL}})$ is governed by $\log(1/p)(1-p_2)$, which depends only on the relative gap between the top two attraction scores and is independent of $v_3$. The next result extends this intuition to the more general case.

\begin{lemma}
	\label{lma:MNL-small-top-gap}
	Given a preference instance $f^{\mathrm{MNL}}\in\mathcal{M}_p^{\mathrm{MNL}}$ with $v_2/v_1=p$, we have
	$
	\IN(f^{\mathrm{MNL}})
	\leq
	\log\left(\tfrac{1}{p}\right)\frac{1-p}{1+p}.
	$
	This combined with \eqref{eq:approximating IOA} implies that
	$
	I_*^{\mathrm{OA}}
	\leq
	\IN(f^{\mathrm{MNL}})
	\leq
	(1+2p)I_*^{\mathrm{OA}}.
	$
\end{lemma}
In other words, as long as the $ p $-separability condition is binding at the top-two ranked items, $\IN(f^{\mathrm{MNL}})$ differs from $I_*^{\mathrm{OA}}$ up to a constant factor, regardless of the gaps among the remaining items.

\noindent \underline{Large gap between $v_1$ and $v_2$.} In contrast, if $v_1/v_2$ is large, then $\IN(f^{\mathrm{MNL}})$ can be significantly larger than $I_*^{\mathrm{OA}}$, even when the gaps among the lower-ranked items are narrow. Intuitively, when the top item is a ``superstar'' item, \textsc{NE} can identify it quickly and terminate early, which leads to substantially lower sample complexity. This phenomenon can be described more generally, which we defer to Example~3.

\vspace{0.1cm}
\textbf{Example 3: Superstar-Item Preferences.}  The aforementioned phenomenon is formalized below.

\begin{lemma} \label{lma:dominating-item} 
	Consider a preference instance $f\in\mathcal{M}_p$, let $ R = R(f) := \min_{S\in \mathcal{N}} \frac{f(1\mid S)}{f(2\mid S)} $ be the minimum choice-probability ratio between the top-two ranked items across all display sets in $ \mathcal{N} $. If $R\ge K$, then $ \IN(f) \ge \log\left(\tfrac{1}{p}\right) \tfrac{R-1}{R+K-1} \prod_{m=2}^{K-1} \left(1-\tfrac{m}{R}\right). $ Consequently, for every $\varepsilon>0$, we have \[ \IN(f) \ge \tfrac{(1+p)(1-\varepsilon)}{1-p} \,I_*^{\mathrm{OA}} \] when $R$ is sufficiently large. 
\end{lemma}

Lemma~\ref{lma:dominating-item} shows that $\IN(f)$ can exceed $I_*^{\mathrm{OA}}$ by an arbitrarily large factor as $p\uparrow 1$. Thus, when the top-ranked item has a sufficiently strong advantage over the remaining items, \textsc{NE} performs much better than its worst-case guarantee. 
}

\subsection{Discussion: Comparisons with Previous Policies}
\label{subsection_best_comparisons}
\vspace{0.2 cm}
\noindent\textbf{Comparison with \textsc{MTP}.}
\tbl{ 
\citet{feng2022robust} designed a Chernoff-type randomized strategy,  \textsc{MTP}, which caters to the OA instances and is shown to be worst-case asymptotically optimal, i.e., 
\begin{align}\label{eq:optimality of MTP}
\textsc{MTP} \in \argmin_{\Pi \text{ is } \text{PAC}} \ \  \sup_{f \in \mathcal M_p} 
\ \  \limsup_{\delta \downarrow 0}  \ \  \frac{\mathbb{E}_f[\tau]}{\log(1/\delta)}.
\end{align}
It is thus our main benchmark algorithm. 
}
Let us compare \textsc{NE} with \textsc{MTP} in terms of both the algorithm design and their theoretical guarantees. From the algorithmic side, \textsc{NE} is quite easy to implement, as we directly exploit a nested display structure.  At each time step, its display rule is to simply and consistently show the active item set $S_{\mathrm{active}}$. Its stopping rule only requires sorting the voting scores of the active items plus a verification step \eqref{eq:stopping rule}. In comparison, \textsc{MTP} involves solving two combinatorial optimization problems
at every time step: one for maximum likelihood estimation and the other to track the generalized likelihood ratio process. In fact, it is clear from the numerical studies in Section \ref{section_experiment} that the runtime of \textsc{NE} typically improves over \textsc{MTP} by \textit{three orders of magnitude}, especially for large $ K $.

\begin{figure}[t]
	\centering
	\definecolor{mahogany}{rgb}{0.75, 0.25, 0.0}
	\definecolor{blue(pigment)}{rgb}{0.2, 0.2, 0.6}
	\definecolor{dartmouthgreen}{rgb}{0.05, 0.5, 0.06}
	\tikzset{every picture/.style={line width=0.85pt}}
	\begin{tikzpicture}[x=0.75pt,y=0.75pt,yscale=-1,xscale=1,scale = 0.8]
	\draw  (53,327.12) -- (501,327.12)(53,20.64) -- (53,327.12) -- cycle (494,322.12) -- (501,327.12) -- (494,332.12) (48,27.64) -- (53,20.64) -- (58,27.64)  ;
	
	\draw  [mahogany] (269,67.28) .. controls (345,67.28) and (281,291.59) .. (455,294.92) ;
	\draw [mahogany]   (269,67.28) .. controls (193,67.28) and (257,291.59) .. (83,294.92) ;
	
	\draw [blue(pigment)]  (269,67.28) .. controls (345,67.28) and (281,220.47) .. (455,222.74) ;
	\draw  [blue(pigment)] (269,67.28) .. controls (193,67.28) and (257,220.47) .. (83,222.74) ;
	
	\draw  [dartmouthgreen]  (76.75,67.28) -- (460,67.28) ;
	
	\draw  [dotted]  (269,67.28) -- (269,327.12) ;
	
	\fill (269,67.28) circle (3pt) ;
	
	\draw (64,00) node [anchor=north west][inner sep=0.75pt]  [xscale=0.930,yscale=0.930]  {$ \scalemath{1.05}{\lim_{\delta \downarrow 0}  \frac{\E [\tau]} {\log( 1/ \delta)}}$};
	
	\draw (460,340) node [anchor=north west][inner sep=0.75pt]  [xscale=0.930,yscale=0.930]  {$f\in \mathcal M_p$};
	
	\draw (175,340) node [anchor=north west][inner sep=0.75pt]  [xscale=0.930,yscale=0.930]  {\scriptsize{\textit {Ordinal Attraction} (OA) Instances} };

    \draw (255,15) node [anchor=north west][inner sep=0.75pt]  [xscale=0.930,yscale=0.930]  {$\scalemath{1.05}{\frac 1 {I_*^{\mathrm{OA}}}}$ (worst case \& Feng et al., 2022)};
	
	\draw (365,147) node [anchor=north west][inner sep=0.75pt]  [xscale=0.930,yscale=0.930]  {$\scalemath{1.05}{\frac 1 { \IN(f)}}$ (ours)};
	
	\draw (380,233) node [anchor=north west][inner sep=0.75pt]  [xscale=0.930,yscale=0.930]  {$\scalemath{1.05}{\frac 1 {I_*(f)}}$ (lower bound)};
	\end{tikzpicture}
	\caption{\textbf{A conceptual illustration of the theoretical contributions of \textsc{NE}.} The horizontal axis represents different preference instances $f$, while the vertical axis represents the asymptotic expected sample complexity.
	}
	\label{figure_theoretical_comparison}
\end{figure}

\tbl{
On the theoretical side, \textsc{NE}'s guarantee is uniformly sharper than that for \textsc{MTP} over all instances $f\in\mathcal M_p$. By \eqref{equation_fixedconfidence}, the expected sample complexity of \textsc{NE} is
\begin{equation}
\label{equation_fixedconfidence_NE}
\E [\tau] \le \frac{\log(1/\delta)}{ \IN(f)}+ O_{1/\delta}(1),
\end{equation}
whereas that of \textsc{MTP} is
$
\E [\tau] \le \frac{\log(1/\delta)}{I_*^{\mathrm{OA}}} + o_{1/\delta}\left(\log\left(\tfrac 1 \delta\right)\right).
$
Since $\IN(f)\ge I_*^{\mathrm{OA}}$ by Proposition~\ref{prop_OA}, the improvement takes one of two forms. When $\IN(f) > I_*^{\mathrm{OA}}$, the improvement is in the leading term and is on the order of $ \Omega_{1/\delta}\left(\log\left(1/\delta\right)\right) $. When $\IN(f) = I_*^{\mathrm{OA}}$, the leading terms coincide and the improvement moves to the residual, tightening $o_{1/\delta}(\log(1/\delta))$ to $O_{1/\delta}(1)$.
\footnote{The $o_{1/\delta}(\log(1/\delta))$ term for MTP admits no explicit expression, as \textsc{MTP}'s track-and-plug-in approach \citep{garivier2016optimal} only targets the asymptotic regime. Benefiting from the simplicity of \textsc{NE}, our analysis is instead rooted in the non-asymptotic regime and yields a tighter residual.} We refer the reader to Figure~\ref{figure_theoretical_comparison} for a graphical illustration.

The same improvement can be read at the level of optimality criteria. Rewriting \textsc{MTP}'s guarantee \eqref{eq:optimality of MTP} equivalently as
\begin{align*}
\textsc{MTP} \ \in\  \argmin_{\Pi \text{ is } \text{PAC}} \ \ \sup_{f \in \mathcal M_p} \ \ \limsup_{\delta \downarrow 0} \ \ \tfrac{\mathbb{E}_f[\tau] - \tfrac{\log (1/\delta)}{I_*^{\mathrm{OA}}}}{\log(1/\delta)},
\end{align*}
it differs from \textsc{NE}'s criterion \eqref{eq:optimality of NE} in the normalizer, as \textsc{NE} divides the same residual by an arbitrarily slowly growing $\omega_{1/\delta}(1)$ rather than by $\log(1/\delta)$. The far smaller normalizer makes \textsc{NE}'s criterion strictly more sensitive, so \textsc{NE} attains higher-order worst-case asymptotic optimality. 
}

\vskip 0.2 cm
\tbl{
\noindent \textbf{Comparison with the Full-Display Policy.} It is also natural to compare \textsc{NE} against the simplest conceivable alternative: the \emph{full-display} policy, which offers the entire assortment $ [K] $ in every round and selects the item with the largest empirical choice frequency. Through a standard concentration argument, this baseline also achieves an $ O(\log(1/\delta)) $ sample complexity with an appropriate stopping rule. However, its leading coefficient can be significantly worse than that of \textsc{NE}. In Appendix~\ref{appendix_full_display}, we show that the advantage of \textsc{NE} over the full-display policy can be arbitrarily large when $ p $ is close to one and $ K $ is large. The intuition is that displaying all $ K $ items dilutes each choice across many non-top alternatives, rendering every observation only weakly informative about the best item. In contrast, \textsc{NE} progressively shrinks the displayed set and concentrates samples where they are most discriminating.
}

\subsection{Discussion: Key Technical Insights}\label{sec:key technical insights}

\tbl{Beyond the performance guarantees, we highlight two technical insights underlying \textsc{NE} that resurface when we turn to the learning-to-rank problem.

\vspace{0.2 cm}
\noindent\textbf{Insights Behind the Nested Structure.}
The sequential-elimination design is motivated by a deeper connection to the nested structure of the optimal allocation in the information-theoretic lower bound. More formally, let $ \lambda_*^{\mathrm{OA}} := \arg\max_{\lambda \in  \mathcal P(\mathcal S)} \min_{f' \in \overline{\mathcal{M}}_p(f^{\mathrm{OA}})} D_\lambda \left( f^{\mathrm{OA}} \| f' \right). $ Then it is known (\citealp{feng2022robust}) that there exists $\{\lambda_2^*, \dots, \lambda_K^*\}$ such that
\[
\lambda_*^{\mathrm{OA}}(S) = 
\begin{cases} 
\lambda_k^* & \text{if } S = [k] \text{ for some } k \in \{2, \dots, K\} \\
0 & \text{otherwise}.
\end{cases}
\]
In other words, the optimal allocation for the max-min problem is naturally nested. More broadly, optimal allocations are widely used in pure exploration problems, but most methods treat them as a black box, including plug-in-and-randomize strategies such as \textsc{MTP}. Instead, \textsc{NE} \textit{incorporates} this nested structure into the policy, and shrinks the display set on a pathwise basis. To our knowledge, this is the first connection drawn between a sequential-elimination structure and the Chernoff-type information measures.

\vspace{0.2 cm}
\noindent\textbf{Insights Behind the Stopping Rule: An SPRT Perspective.} Given the nested structure of the display policy, the question then becomes which item to eliminate, and when. In short, \textsc{NE}'s elimination criterion at each stage can be read as a Sequential Probability Ratio Test (SPRT) tailored to the OA instances. 

Specifically, the learner treats $ f $ as belonging to the hardest-to-learn OA preference class and only tries to differentiate it from other OA instances with different underlying rankings. At each stage, the maximum likelihood estimate (MLE) of the ranking $ \sigma_f $ is simply the order of the voting scores, which identifies the most plausible item to eliminate. Hence the elimination candidate is always the item with the lowest voting score in $ S_{\mathrm{active}} $, say, $ \hat{k} $. To decide whether to eliminate it or to collect more samples, \textsc{NE} effectively performs a generalized log-likelihood-ratio test of the hypothesis $ H_0:\text{Item } \hat{k} \text{ is not top-ranked } $ against the hypothesis $ H_1:\text{Item } \hat{k} \text{ is top-ranked}. $  The elimination criterion \eqref{eq:stopping rule} is equivalent to testing whether this ratio in favor of $ H_0 $ exceeds a fixed threshold; further details are given in Appendix~\ref{section_Interpretation}. In short, under the SPRT perspective, \textsc{NE}'s elimination logic is to eliminate an item as soon as it is confident the item is \textit{not top-ranked}.\footnote{While intuitive, it is worth noting that this principle is not the only plausible approach. For instance, one might alternatively choose to eliminate an item when it is almost certain that the item is \textit{bottom-ranked}. However, the optimality of \textsc{NE} suggests that this alternative approach is overly conservative and fails to achieve the optimal display set allocation.}
}

\section{The Learning-to-Rank (Full-Ranking Identification) Problem} 
\label{section_fullranking}

\tbl{In this section, we move on to the problem of full-ranking identification from choice-based feedback, which is notably more challenging, as it necessitates the determination of the ranking for all items. To this end, Section~\ref{section_Methodology} starts with a discussion on how we could leverage the intuition, insight, and methodology pioneered by \textsc{NE}. Section~\ref{subsection_ranking_algorithm} then introduces
	\textsc{Nested Partition} (\textsc{NP}), a Quicksort-like algorithm
	built on this structure.
	Section~\ref{subsection_ranking_results} establishes the
	$\delta$-PAC correctness of \textsc{NP} and a non-asymptotic,
	instance-specific bound on its expected sample complexity through the
	hardness quantity $\JN(f)$.
	Section~\ref{sec:optimality of NP} compares this guarantee
	with the information-theoretic lower bound and shows that \textsc{NP}
	is worst-case asymptotically optimal.
	Finally, Section~\ref{sec:understanding-NP-beyond-OA} moves beyond the
	worst case and characterizes the behavior of $\JN(f)$ across several
	representative classes of instances.}

\subsection{From Best-Item to Full-Ranking Identification}\label{section_Methodology}

Before we dive into the technical details, we would like to first discuss the high-level ideas, namely, how the nested structure is derived from the information-theoretic lower bound, as well as how the elimination criterion is connected to the SPRT principle.

\vspace{0.2 cm}
\noindent\textbf{Nested Structure for Ranking-Identification. } Recall that a fundamental idea of \textsc{NE} is to exploit the ``natural'' nested structure implied by the information-theoretic lower bound.  Does a similar structure exist for the learning-to-rank problem, at least when tailored to the OA instances? The next result provides a positive answer. Formally speaking, let us introduce ${\widetilde{\mathcal{M}}}_p^{\mathrm{OA}}(f) := \{f'\in \mathcal M_p^{\mathrm{OA}}: \sigma_{f'}\neq \sigma_f\}$ as the collection of alternative preferences with distinct rankings within the class of OA preferences. The ``natural'' nested structure for ranking identification is stated in the result below.

\vspace{0.2 cm}
\begin{proposition} \label{prop_OA_ranking_bound}
It holds that 
	\begin{equation}
	\label{eq_OA_J_maxmin}
	J_*^{\mathrm{OA}} \ :=\  \sup_{\lambda\in \mathcal P (\mathcal S)} \ \inf_{f'\in {\widetilde{\mathcal{M}}}_p^{\mathrm{OA}}(f^{\mathrm{OA}})}  \  D_\lambda(f^{\mathrm{OA}} \| f') \ =\  \log \left(\tfrac{1}{p}\right) \cdot \tfrac {1-p} {K-1+p}.
	\end{equation}
	In addition, the optimal solution to the outer maximization problem can be written as:
		\begin{equation}\label{eq:optimal OAM allocation for ranking}
		\begin{aligned}
		\lambda^*(S) := \begin{cases}
		(1-p^{K-n+1} )/(K-1+p) & \quad \text { if } S=\{n,\ldots,K \} \text { for some } n \in [K-1]\setminus \{1\} \\
		(1-p^K)/[(1-p)(K-1+p)]& \quad \text { if } S=[K] \\ 
		0 &\quad \text { otherwise. }\end{cases}
		\end{aligned}
		\end{equation}
\end{proposition}
\vspace{0.2 cm}

The proof of Proposition~\ref{prop_OA_ranking_bound} is detailed in Appendix~\ref{appendix_prop_OA_ranking_bound}. This proposition says that when tailored to the worst-case OA preference $  f^{\mathrm{OA}} $, the optimal allocation $ \lambda^\ast $ of the display sets is supported on a \textit{nested} collection of $K-1$ display sets. That is, $\lambda^*(S) >0 $ if and only if $S=\{n,\ldots,K \}$  for some $n \in [K-1]$. This is similar to the learning-to-select setting, although interestingly, the nested display sets shrink in a reverse direction (i.e., from top-ranked to bottom-ranked).

Following the same high-level logic as \textsc{NE}, the nested structure would suggest the following display set dynamics when tailored to the OA preferences: the company should start with the full display set, and then sequentially single out the highest-voted item when it dominates the rest of the active set by a large margin. The sequence of stopping times should be calibrated according to a generalized log-likelihood ratio test. The output ranking would then be based on the order of elimination, with
items eliminated earlier ranked higher. In this paper, we convert this high-level idea into a strategy we call \textsc{Nested Partition} (\textsc{NP}). Briefly speaking, it recursively partitions the active set into two subsets, referred to as $S_{\mathrm{high}}$ and $S_{\mathrm{low}}$, respectively, whenever the votes of items in $S_{\mathrm{high}}$ dominate those in  $S_{\mathrm{low}}$ by a high margin. The output ranking is then based on the partitions when every partitioned set becomes a singleton. As such, \textsc{NP} generalizes \textsc{NE-Ranking}  since the latter always makes $S_{\mathrm{high}}$ a singleton.\footnote{Notably, there could be potentially other ways to convert the high-level intuition, too. For example, one possible alternative implementation would be to ensure the nested structure on a pathwise basis. We refer to this strategy as \textsc{NE-Ranking} and detail it in Appendix~\ref{appendix_NE_ranking}. Although the system dynamics of those strategies admit the same deterministic approximation under the OA preferences, we choose \textsc{Nested Partition} (\textsc{NP}) for two reasons. The first one is more theoretical. The partition construction of \textsc{NP} leads to closed polytopes while \textsc{NE-Ranking} leads to a disjoint union of cones as hitting boundaries for the corresponding random walks. That makes \textsc{NP} more amenable to theoretical performance guarantees. Second, we observe that \textsc{NP} typically has better non-asymptotic empirical performance than \textsc{NE-Ranking}, which is demonstrated in Section \ref{section_experiment2}.}

\vspace{0.2 cm}
\noindent\textbf{The SPRT Perspective. } The elimination criterion for \textsc{NP} also finds its root in an SPRT, albeit with different hypotheses. Restricting the preference instances to the OA family, \textsc{NP}'s criterion to partition the active set into  $S_{\mathrm{high}}$ and $S_{\mathrm{low}}$ is equivalent to a generalized likelihood ratio test regarding whether all items in $S_{\mathrm{high}}$ dominate those in $S_{\mathrm{low}}$ in ranking.
We refer to Appendix \ref{sub:likelihood test for NP} for full details of the verification.

\subsection{The \textsc{Nested Partition} Algorithm}
\label{subsection_ranking_algorithm}

\begin{figure}[t]
	\begin{minipage}{\textwidth}
		\begin{algorithm}[H]
			\caption{\textsc{Nested Partition} (Master Routine)}
			\label{algo_ranking}
			{\bf Input:} Tuning parameter $M>0$.\\
			{\bf Output:} The full ranking $\sigma_{\mathrm{out}}$.
			\begin{algorithmic}[1]
				\STATE Initialize global variables: voting scores $W_0(i)\leftarrow 0$ for all $i\in[K]$ and timer $t\leftarrow 0$.
				\STATE $\pi_{\mathrm{out}} \leftarrow \textsc{Partition}([K])$.
				\STATE \textbf{return} $\sigma_{\mathrm{out}} \leftarrow \pi^{-1}_{\mathrm{out}}$.
			\end{algorithmic}
		\end{algorithm}
		\vspace{-1cm}
		\begin{algorithm}[H]
			\caption{\textsc{Partition}($S_{\mathrm{active}}$) (Subroutine)}
			\label{algo_partition}
			{\bf Input:} Active set $S_{\mathrm{active}}$.\\
			{\bf Output:} A permutation over the active set $\pi:[\,|S_{\mathrm{active}}|\,]\to S_{\mathrm{active}}$.
			\begin{algorithmic}[1]
				\STATE \textbf{if} $|S_{\mathrm{active}}|=1$ \textbf{then return} $S_{\mathrm{active}}$.
				\LOOP
				\STATE $t\leftarrow t+1$; display $S_{\mathrm{active}}$ and observe the choice $X_t\in S_{\mathrm{active}}$.
				\vskip 0.2 cm
				\STATE  Update voting scores based on $ X_t $:
				$
				W_t(i) \leftarrow \begin{cases}
				W_{t-1}(i) + 1 & \text{if } i = X_t \\
				W_{t-1}(i) & \text{if }  i \neq X_t.
				\end{cases}
				$
				\vskip 0.2 cm
				\STATE Sort the active items as $\pi_t:[\,|S_{\mathrm{active}}|\,]\to S_{\mathrm{active}}$ with $W_t(\pi_t(1))\ge\cdots\ge W_t(\pi_t(|S_{\mathrm{active}}|))$.
				\IF{there exists $k$ with $W_t(\pi_t(k))-W_t(\pi_t(k+1))\ge M$}
				\STATE  $S_{\mathrm{high}}\leftarrow\{\pi_t(1),\dots,\pi_t(k)\}$, $\;S_{\mathrm{low}}\leftarrow S_{\mathrm{active}}\setminus S_{\mathrm{high}}$ (\textit{so that \eqref{eq:partition rule} is satisfied})

				\STATE \textbf{return} $\big(\textsc{Partition}(S_{\mathrm{high}}),\,\textsc{Partition}(S_{\mathrm{low}})\big)$ (\textit{concatenate left and right recursions})
				\ENDIF
				\ENDLOOP
			\end{algorithmic}
		\end{algorithm}
	\end{minipage}
\end{figure}

We summarize the procedure of \textsc{NP} in two pseudocode environments: the master routine in Algorithm~\ref{algo_ranking} and the core recursive subroutine \textsc{Partition} in Algorithm~\ref{algo_partition}. As its name suggests, the subroutine \textsc{Partition}'s goal is to partition the active set $  S_{\mathrm{active}} $ into an ordered pair of two subsets, thus separating the higher-voted items from others. At each time step $ t $ during \textsc{Partition}, the entire active item set $S_{\mathrm{active}}$ is displayed to customers. Throughout the procedure, the algorithm maintains the voting scores
$\{W_t(i)\}_{i\in[K]}$, defined as in \textsc{NE}. The active set is partitioned into two subsets, $(S_{\mathrm{high}}, S_{\mathrm{low}})$, once the top-voted items lead the bottom-voted ones by a margin of at least $M$. More precisely, 
\begin{align}
\label{eq:partition rule}
\min_{i \in S_{\mathrm{high}}} W_t(i) - \max_{i \in S_{\mathrm{low}}} W_t(i) \ge M.
\end{align}
 Similar to the parameter $M$ in \textsc{Nested Elimination} for best-item identification, $M$ controls the partition's accuracy, a topic we will explore further in subsequent analysis.\footnote{It is worth mentioning that due to the sequential nature of the process, the definition of $ (S_{\mathrm{high}}, S_{\mathrm{low}}) $ is unique when \eqref{eq:partition rule} becomes first satisfied.} Incorporating the subroutine \textsc{Partition}, our \textsc{NP} algorithm works in a way very similar to the Quicksort algorithm \citep{hoare1962quicksort}, a widely recognized sorting method. The algorithm begins by executing \textsc{Partition} on the entire item set $[K]$. When the partition criterion is met, $[K]$ is divided into two subsets, and the subroutine is recursively applied to each subset. This process continues until each subset becomes a singleton.

The execution of \textsc{NP} generates a binary tree structure, which we illustrate in Figure~\ref{figure_rankingtree_example}. In this representation, items on the left are considered superior to those on the right, as determined by the algorithm. Each item is assigned to a unique leaf node, and its horizontal placement in the tree reflects its output ranking. This tree-type structure will be revisited in our analysis of \textsc{NP}.

\begin{figure}[htbp]
\begin{center}
\begin{tikzpicture}[
  level/.style={level distance=15mm},
  level 1/.style={sibling distance=80mm},
  level 2/.style={sibling distance=40mm},
  level 3/.style={sibling distance=20mm},
  arrow/.style={->, line width=1.5pt, >=stealth}
]

\node {$\{1,2,3,4,5,6,7\}$}
  child[arrow] {node {$\{1,2,3\}$}
    child[arrow] {node {$\{1\}$}}
    child[arrow] {node {$\{2,3\}$}
      child[arrow] {node {$\{2\}$}}
      child[arrow] {node {$\{3\}$}}
    }
  }
  child[arrow] {node {$\{4,5,6,7\}$}
    child[arrow] {node {$\{4,5\}$}
      child[arrow] {node {$\{4\}$}}
      child[arrow] {node {$\{5\}$}}
    }
    child[arrow] {node {$\{6,7\}$}
      child[arrow] {node {$\{6\}$}}
      child[arrow] {node {$\{7\}$}}
    }
  };
\end{tikzpicture}
\caption{\textbf{A possible trajectory of $ S_{\mathrm{active}} $ under \textsc{NP} represented by a binary tree.} A partition separates the highest-voted items from the lowest-voted ones according to \eqref{eq:partition rule}.} 
\label{figure_rankingtree_example}
\end{center}
\end{figure}

\begin{figure}[htbp]
	\centering
	\includegraphics[width=.7\textwidth]{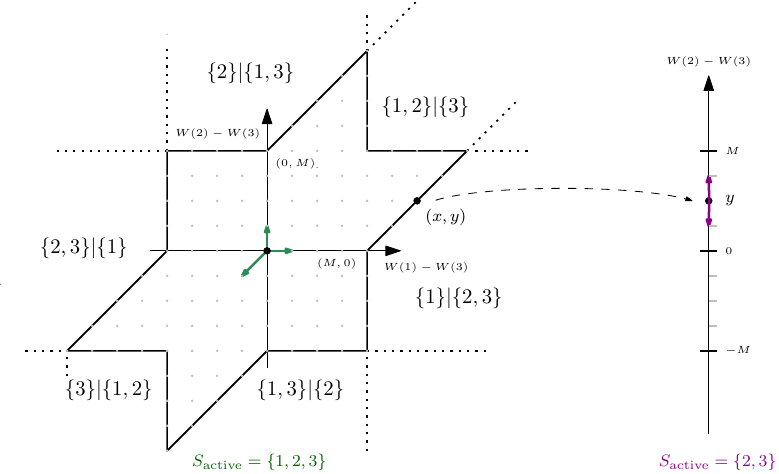}
	\caption{\textbf{A visualization of the system dynamics under \textsc{NP}.} 
		Let \( K = 3 \). In the initial stage, the active set is \( [3] = \{1, 2, 3\} \). The projected state variables \( \big(W(1) - W(3),\, W(2) - W(3)\big) \) and the random walk dynamics are the same as \textsc{NE} illustrated in Figure \ref{fig:process_NE}. What differentiates \textsc{NE} and \textsc{NP} is the hitting boundaries. Under \textsc{NP}, the first stage finishes when the random walk hits the boundary of the concave polygon defined by vertices $ (M, 0) $, $ (2M, M) $, $ (M,M) $, $ (M, 2M) $, $ (0,M) $, $ (-M, M) $, $ (-M, 0) $, $ (-2M, -M) $, $ (-M, -M) $, $ (-M, -2M) $, $ (0, -M) $, and $ (M, -M) $. The 12 faces are further divided into 6 different partition possibilities. For example, $ \{1,2\}|\{3\} $ means $ S_{\mathrm{high}} = \{1,2\}$ and $ S_{\mathrm{low}} = \{3\}$. In the illustrated path, the first stage finishes with $ S_{\mathrm{high}} = \{1\}$ and $ S_{\mathrm{low}} = \{2,3\}$, which is an event with high probability under any OA preference instance.\\
		Since $ S_{\mathrm{high}} $ is a singleton, it suffices to look at $ \{2,3\} $ as the active set in the next stage. Here, the state variables are further projected into the one-dimensional space spanned by \( W(2) - W(3) \). That results in a one-dimensional random walk, which is the same as that under \textsc{NE}. Depending on which endpoint the random walk hits, the resulting ranking is either $ (1,2,3) $ or $ (1,3,2) $.
 } 
	\label{process_NP}
\end{figure}

\vskip 0.1 cm
\begin{remark}
\label{remark_dynamics}
    {\sf The behavior of voting scores within the subroutine \textsc{Partition} can be viewed as a (biased) multi-dimensional random walk similar to the case in \textsc{Nested Elimination} explained in Remark~\ref{remark_randomwalk}. The two algorithms differ in their hitting boundaries. We illustrate its system dynamics in Figure~\ref{process_NP}. 
 }
\end{remark}

\subsection{Theoretical Analysis of \textsc{NP}}
\label{subsection_ranking_results}

In this subsection, we theoretically analyze the performance of 
the \textsc{Nested Partition} algorithm. Specifically, we demonstrate the $\delta$-PAC correctness of \textsc{NP}, and provide an upper bound on its expected sample complexity. Furthermore, we establish the information-theoretic lower bound on the sample complexity for the full-ranking identification problem, and thereby illustrate that our algorithm attains worst-case asymptotic optimality.

Let the tuning parameter $ M $ be fixed. In order to assess the complexity of ranking from choice-based feedback with respect to our algorithm, \tbl{for each \(f\in\mathcal M_p\), we introduce a hardness quantity
denoted by \(\JN(f)\), which will be explained in detail below}.
Our first main result below characterizes the sample complexity of \textsc{NP} through the quantity $ \JN(f) $.

\vspace{0.2cm}
\begin{theorem}[Sample complexity of \textsc{NP}] 
	\label{theorem_ranking_fixedconfidence}
 For every confidence level $\delta\in(0,1)$, \textsc{NP} is $\delta$-PAC with  parameter
	\begin{equation}
	\label{equation_ranking_M}
	M = \frac{\log(1/\delta)+\log(K-1)} {\log(1/p)}.
	\end{equation}
	Furthermore, for every preference instance $f\in\mathcal M_p$, there is a constant $ C_f' $ independent of $ \delta $ such that 
	\begin{equation}
	\label{equation_ranking_fixedconfidence}
	\E [\tau] \le \frac{\log(1/\delta)}{\JN(f)}+ C_f'.
	\end{equation}
\end{theorem}
\vspace{0.2cm}

The proof of Theorem~\ref{theorem_ranking_fixedconfidence} is deferred to Appendix~\ref{appendix_theorem_ranking_fixedconfidence}. Again, we would like to emphasize that the instance-specific sample complexity guarantee allows for a residual term  \textit{independent of} $ \delta $, which is different from the typical literature. We also remark that although the proof structure for Theorem~\ref{theorem_ranking_fixedconfidence} parallels that of Theorem~\ref{theorem_fixedconfidence}, the specific details exhibit noteworthy variations owing to their inherent complexity. We leverage two kinds of binary tree representations in the analysis: one deterministic and the other stochastic. To characterize the overall expected sample complexity, we provide both upper and lower bounds on the duration that the algorithm allocates to each node of the tree, which necessitates novel techniques for martingale constructions.

\vskip 0.2 cm
\tbl{
	\noindent\textbf{The Hardness Quantity: Explicit Expressions.} Similar to the learning-to-select problem, Theorem~\ref{theorem_ranking_fixedconfidence} characterizes the sample complexity through the hardness measure $ \JN(f) $ itself in an explicit formula, which we explain now. 
	
	Without loss, suppose that the ground-truth ranking is $ \sigma = (1,2, \ldots, K) $. For every \textit{contiguous} display set, i.e., those of the form $ S = [a,b]:= \{a,a+1, \ldots, b\} $, and score vector $w$,  let 
	\begin{equation*}
	\bar \tau(f,S,w) := \min_{i\in S, i\neq \max(S)} \tfrac {1-w(i)+w(i+1)} {f(i|S)-f(i+1|S)} \quad \text{ and } \quad
	i^*(f,S,w) := \min\argmin_{i\in S, i\neq \max(S)} \tfrac {1-w(i)+w(i+1)} {f(i|S)-f(i+1|S)} .
	\end{equation*}
	Intuitively, $\bar \tau(f,S,w)$ represents the normalized time that Algorithm~\ref{algo_partition} takes, in an idealized setting with deterministic fractional votes, to reach the next recursive stage when $M=1$, the active display set is $S$, and the voting score is $w$.  Correspondingly, $ i^*(f,S,w) $ is the \textit{split index} of this stage. After that, the recursion processes the two display sets $ \bar S_{\mathrm{high}}=[a,i^*] $, $ \bar S_{\mathrm{low}}=[i^*+1,b], $ and the score vector is updated by $w'(i) := w(i) + \bar\tau(f,S,w)  f(i |S)  \mathbb{I}\{i \in S\}$ for every item $ i $. Following Algorithm~\ref{algo_partition}, we thus recursively define
	\begin{equation}
	\label{equation_DfSW}
	D(f, S, w) :=  \begin{cases}
	0 & \text{if } |S| =1 \\
	\bar \tau(f,S,w) +  D(f, \bar  S_{\mathrm{high}}, w') +  D(f, \bar S_{\mathrm{low}}, w') & \text{otherwise}\\
	\end{cases}
	\end{equation}
	Finally, the hardness quantity $\JN(f)$ can be expressed as
	\begin{equation}
	\label{equation_JNf_definition}
	{\JN(f)}= \log\left(\tfrac 1 p \right) \left[ D(f, [K], \mathbf 0) \right]^{-1},
	\end{equation}
	where the factor $ \log\left(1/p\right) $ converts the raw voting scores into statistical confidence by taking into account the $ p $-separability of the customer choices. Similarly to the best-item identification problem, the two ingredients of \eqref{equation_ranking_fixedconfidence} can thus be combined in the following way: on the one hand, \textsc{NP} runs for roughly $ M \left[ D(f, [K], \mathbf 0) \right] = M \log(1/p)/ \JN(f)$ periods. Meanwhile, the $ \delta $-PAC requirement  calls for $ M \approx \log(1/\delta)/\log(1/p) $, and hence $ \E[\tau] \approx \log(1/\delta)/\JN(f) $.
}

\subsection{Worst-Case Asymptotic Optimality of \textsc{NP}}\label{sec:optimality of NP}
Having established the sample complexity bound of \textsc{NP}, a natural inquiry arises regarding its optimality properties, e.g., how its sample complexity compares with the lower bound. 
Drawing inspiration from the analysis of \textsc{NE}, we analyze $ \JN(f) $ through the hardness quantity tailored to the OA preference $ f^{\mathrm{OA}} $. To this end, recall from \eqref{eq_OA_J_maxmin} that 
$
J_*^{\mathrm{OA}}   =   \log \left(\tfrac{1}{p}\right) \cdot \tfrac {1-p} {K-1+p}.
$ is the optimal value of the max-min problem $ \sup_{\lambda\in \mathcal P (\mathcal S)} \ \inf_{f'\in {\widetilde{\mathcal{M}}}_p^{\mathrm{OA}}(f^{\mathrm{OA}})}  D_\lambda(f^{\mathrm{OA}} \| f') $. Proposition~\ref{prop_OA_ranking} below shows that the value of $ \JN(f) $ attains its minimum at the OA preference instance $ f^{\mathrm{OA}} $, where the minimum value exactly equals $ J_*^{\mathrm{OA}} $.

\tbl{
\begin{proposition}[Worst-case guarantee of \textsc{NP}]
	\label{prop_OA_ranking}
	 It holds that
	\[
	\min_{f\in\mathcal M_p}\JN(f)
	=
	\JN(f^{\mathrm{OA}})
	=
	J_*^{\mathrm{OA}}.
	\]
	Combined with Theorem~\ref{theorem_ranking_fixedconfidence}, this result implies that for every $f\in\mathcal M_p$, 
	$
	\mathbb E_f[\tau]
	\le
	\frac{\log(1/\delta)}{J_*^{\mathrm{OA}}}+C_f',
	$
	where $C_f'$ is independent of $\delta$.
\end{proposition}
}

\vspace{0.2cm}

The proof of Proposition~\ref{prop_OA_ranking} is postponed to Appendix~\ref{appendix_prop_OA_ranking}. 
On the other hand, we now establish the information-theoretic lower bound for the full-ranking identification problem.
For any preference $f\in\mathcal M_p$, we define ${\widetilde{\mathcal{M}}}_p(f) := \{f'\in \mathcal M_p: \sigma_{f'}\neq \sigma_f\}$, which represents the set of alternative preferences with distinct rankings. The ensuing non-asymptotic and instance-specific lower bound on $\E [\tau]$ is presented in Theorem~\ref{theorem_lower_bound_ranking} below. See Appendix~\ref{appendix_theorem_lower_bound_ranking} for the proof of Theorem~\ref{theorem_lower_bound_ranking}.

\vspace{0.2cm}
\begin{theorem}[Lower bound for full-ranking identification]
	\label{theorem_lower_bound_ranking}
	For any preference $f\in\mathcal M_p$, let 
	\begin{align}\label{eq:max-min_ranking}
	J_*(f) := \sup_{\lambda\in \mathcal P (\mathcal S)} \inf_{f'\in {\widetilde{\mathcal{M}}}_p(f)}  D_\lambda(f \| f').
	\end{align}
	Then any $\delta$-PAC full-ranking policy satisfies
	$
	\E [\tau] \ge  \frac {\log(1/\delta) - \log 2.4} {J_*(f)}.
	$
\end{theorem}
\vspace{0.2cm}

In analogy to the case of best-item identification, $J_*(\cdot)$ quantifies the intrinsic complexity of the full-ranking identification problem through the information-theoretic lower bound. It is defined through a max-min problem \eqref{eq:max-min_ranking} that is similar to \eqref{eq:max-min}, yet with a change of feasible region from $ \overline{\mathcal{M}}_p(f)  $ to $ \widetilde{\mathcal{M}}_p(f) $. That reflects the difference in hypothesis class between \textit{best-arm} and \textit{full-ranking} identification. Similarly to the learning-to-select problem, the formidable complexity of the \eqref{eq:max-min_ranking} renders it impractical to utilize directly to design efficient online policies. In contrast, the structural insights under the OA preference instances lead to the simple design of \textsc{NP}.  

As a direct consequence of Theorem~\ref{theorem_lower_bound_ranking}, the result below characterizes the worst-case performance of \textit{any} $\delta$-PAC full-ranking identification policy in a way that is relatable to Proposition \ref{prop_OA_ranking}.

\vspace{0.2cm}
\begin{proposition}[Worst-case analysis of general policies]\label{prop_worstcase_ranking}
	It holds that 
	$$
	\min_{f\in \mathcal M_p} J_*(f) \le  J_*^{\mathrm{OA}} .
	$$
	As a result, for any $\delta$-PAC full-ranking identification policy, there exists a preference instance $ f\in \mathcal M_p  $ such that
	$\E [\tau] \ge \frac{\log(1/\delta)}{J_*^{\mathrm{OA}}}+ O_{1/\delta}(1).$
\end{proposition}
\vspace{0.2cm}
The proof of Proposition~\ref{prop_worstcase_ranking} is postponed to Appendix~\ref{appendix_prop_worstcase_ranking}.  \tbl{Now, combining all the pieces in Proposition~\ref{prop_OA_ranking}, Theorem~\ref{theorem_lower_bound_ranking}, and Proposition~\ref{prop_worstcase_ranking}, we note that \textsc{NP} attains the optimal worst-case leading coefficient in the learning-to-rank problem. In addition, since the residual term is bounded in $ \delta $, \textsc{NP} satisfies the same optimality criterion as \textsc{NE} does in the learning-to-select problem. That is, 
$$
\textsc{NP}
\in
\argmin_{\Pi\text{ is PAC}}\ 
\sup_{f\in\mathcal M_p}\ 
\limsup_{\delta\downarrow0}\ 
\frac{
	\mathbb E_f[\tau]
	-
	\log(1/\delta)/J_*^{\mathrm{OA}}
}{
	\omega_{1/\delta}(1)
},
$$
for any \textit{arbitrarily slowly} growing function $\omega_{1/\delta}(1)$.}

\tbl{
\vspace{0.1 cm}
\begin{remark}
	{\sf Similar to the best-item identification problem, one could compare \textsc{NP} against the full-display policy. For the learning-to-rank problem the advantage of \textsc{NP} is in fact more pronounced. In Appendix~\ref{appendix_full_display}, we show that \textit{any} $\delta$-PAC full-display policy satisfies $ \sup_{f\in\mathcal M_p} \liminf_{\delta\downarrow0}\
		\frac{\mathbb E_f[\tau]}{\log(1/\delta)}\;=\;+\infty
		$, regardless of the stopping and decision rules, which stands in contrast with the performance of \textsc{NP}.  The intuition is again that full-display loses the benefit of \textit{focus}. Specifically, recovering the full ranking under full-display hinges on ordering the bottom pair of items, which are chosen so rarely that each sample is nearly uninformative about the tail of the ranking. In contrast, \textsc{NP} funnels its comparisons into small contiguous blocks in which adjacent items compete head-to-head, thereby speeding up the learning. \hfill$\diamond$}
\end{remark}
}

\tbl{
\subsection{Discussion: Understanding \textsc{NP}'s Performance Beyond OA Instances}\label{sec:understanding-NP-beyond-OA}

Similar to the best-item identification case, let us now provide a few examples where $ \JN(f) $ can be characterized beyond the OA instances. Without loss, we will assume that the underlying ranking is the identity ranking for ease of notation. 
	
	\vskip .2 cm
	\noindent \textbf{Example 1: An Equivalence Class Based on Ranking-Contiguous Display Sets.}
	Since $ \JN(f) $ depends on the preference instance $f$ only through the choice
	probabilities on \textit{contiguous} display sets, any two instances $ f $ and $ f' $ satisfy
	$ \JN(f) = \JN(f') $ whenever $ f(\cdot \mid [a,b]) = f'(\cdot \mid [a,b]) $
	for every display set of the form $ [a,b] $. In other words, $ \JN $ is defined up to the equivalence class based on contiguous display sets. In particular, a preference instance $f$ satisfies $ \JN(f) = J_*^{\mathrm{OA}} $ as long as
	\begin{align*}
	f(\ell \mid S) = \tfrac{p^{\ell-a}}{1 + p + p^2 + \cdots + p^{b-a}},
	\quad \text{ for every contiguous display set } S = [a,b] \text{ and } \ell \in S.
	\end{align*}
	
	This characterization subsumes a few concrete subfamilies of preference instances, including (i) the OA instance itself; (ii) the \emph{Mallows model} with dispersion parameter $ p $; and (iii) the subclass of the \emph{Multinomial Logit (MNL)} model whose attraction scores form the
	geometric sequence $ (1, p, \ldots, p^{K-1}) $.

	\vskip .2 cm
	\noindent \textbf{Example 2: MNL preferences.}
	As in the best-item identification problem, we examine the hardness quantity for an MNL instance
	$\fMNL \in \mathcal{M}_p^\mathrm{MNL}$. To build intuition, consider the $ K = 3 $ case, and suppose
	without loss of generality that $v_1 = 1$, $v_2 = p_2$, and $v_3 = p_2 p_3$, where
	$p_2, p_3 \in (0, p]$. After some algebra, the hardness measure can be written as
	\begin{align*}
	\JN(\fMNL) =
	\begin{cases}
	\log\left(\tfrac{1}{p}\right) \tfrac{(1-p_2)(1-p_3)}{2-p_2-p_2p_3} &\quad 1 - 2p_2 + p_2p_3 \geq 0 \\
	\log\left(\tfrac{1}{p}\right) \tfrac{(1-p_2)(1-p_3)}{1+p_2-2p_2p_3} &\quad  1 - 2p_2 + p_2p_3 \leq 0 \\
	\end{cases}
	\end{align*}
	
	Notably, $ \JN(\fMNL) $ depends critically on \textit{both} $ p_2 $ and $ p_3 $: it diminishes when
	either $ p_2 $ or $ p_3 $ approaches one. This stands in contrast to
	the best-item identification setting, where $\IN(\fMNL)$ is governed up to a constant factor by
	$\log(\tfrac{1}{p})(1 - p_2)$. The following result generalizes this observation to a generic MNL instance.

	\begin{proposition}
		\label{prop:MNL_NP}
		Consider an MNL preference instance \(\fMNL \in \mathcal{M}_p\) with
		attraction parameters
		$
		v_1>v_2>\cdots>v_K>0,
		$ and let $
		p_i:=\frac{v_i}{v_{i-1}}
		$ for $ i\in \{2, \ldots, K\}.
		$ Then 
		\[
		\log(\tfrac{1}{p})\left(\sum_{i=2}^{K}\tfrac{1+p_i}{1-p_i}\right)^{-1}
		\leq
		\JN(\fMNL)
		\leq
		(1+p)\log(\tfrac{1}{p})\left(\sum_{i=2}^{K}\tfrac{1+p_i}{1-p_i}\right)^{-1}.
		\]
	\end{proposition}

	Proposition~\ref{prop:MNL_NP} above shows that $  \JN(\fMNL) $ is governed by the term $ \log(\tfrac{1}{p})\left(\sum_{i=2}^{K}\tfrac{1+p_i}{1-p_i}\right)^{-1} $ up to a factor of at most $1+p$. Roughly speaking, every item $ i $ contributes an additive $  \tfrac{1+p_i}{1-p_i} /\log(\tfrac1p)$ to the inverse rate $1/\JN(\fMNL) $, the leading coefficient of \textsc{NP}'s sample complexity. If all gaps are binding under the $p$-separability condition,  i.e., $ p_i = p $ for all $ i $,  then we recover the geometric MNL instance, and its worst-case hardness quantity $ J_*^{\mathrm{OA}} = \log \left(\tfrac{1}{p}\right) \cdot \tfrac {1-p} {K-1+p}
	\approx \log \left(\tfrac{1}{p}\right) \cdot \tfrac{1-p}{(K-1)(1+p)} $ up to a constant. 
	
	The results above reveal two main insights. First,  the gap between \textit{every} pair of adjacent items matters, rather than only the top gap, and the contribution of each gap is nearly equal regardless of its position. That is qualitatively different from the best-item identification problem.  Second,  if only a small number of gaps are narrow and others are wide, then $ \JN(\fMNL) $ can be much larger than the worst-case guarantee $ J_*^{\mathrm{OA}} $. For example, suppose only one gap is binding under $p$-separability (i.e., $p_\ell = p$ for some $\ell$) and other gaps are well separated (i.e., there exists a fixed $\rho < p$ such that  $p_i \leq \rho$ for all $i \neq \ell$ ). Then Proposition~\ref{prop:MNL_NP} gives 
	\begin{align*}
		\JN(f) = \Omega \left( \log(\tfrac{1}{p}) (K + \tfrac{1}{1-p})^{-1}\right) \quad \gg \quad J_*^{\mathrm{OA}} = \Theta \left( \log(\tfrac{1}{p}) (\tfrac{K}{1-p})^{-1}\right)
	\end{align*}
	as $K$ grows and $p$ approaches one; see Appendix~\ref{append:understanding JN} for more details, where we also generalize this observation to a wider class of instances beyond the MNL family. This difference between $ K+\tfrac{1}{1-p} $ and $ \tfrac{K}{1-p} $ reflects that under structurally easier instances,  \textsc{NP} can substantially outperform its worst-case guarantee.
}

\vskip 0.3 cm

\tbl{
	\section{Extension: Incorporating Capacity Constraints}\label{sec_extension_capacity}
	
	In many practical applications, the decision-maker may face a physical or cognitive limit on the number of items that can be displayed simultaneously.  In this section, we extend the nested approach to the capacity-constrained setting. That is, given a capacity constraint $\kappa$, where $2 \le \kappa \leq K$, we restrict the feasible collection of display sets to
	$
	\mathcal{S}^{\kappa}:=\{S\subseteq [K]: 2\le |S|\le \kappa\}.
	$
	For both the best-item and full-ranking identification problems, we use a \emph{divide-and-conquer} decomposition that resolves capacity-constrained subproblems with a nested oracle.
}

\tbl{	
	\subsection{The Learning-to-Select Problem with Capacity Constraints}\label{extension_selection}

	We first consider the best-item identification problem. To adapt \textsc{NE} into the capacitated setting, we introduce the \textsc{Hierarchical Nested Elimination} (\textsc{HNE}) algorithm. 
	It uses a divide-and-conquer structure, where in each round, it partitions the current active set into feasible batches of size at most $\kappa$, runs \textsc{NE} within each batch to identify a local winner, and passes these local winners to the next round. 
	The procedure repeats until only one item remains. The formal procedure is summarized in Algorithm~\ref{algo_cap}.
	
	\begin{algorithm}[h]
		\caption{Hierarchical Nested Elimination (\textsc{HNE})} 
		\label{algo_cap}
		\hspace*{0.02in} {\bf Input:} Full item set $[K]$, capacity constraint $\kappa$ (where $\kappa \leq K$), and tuning parameter $M > 0$. 
		
		\hspace*{0.02in} {\bf Output:} The only element $i_{\mathrm{out}}$ of $S_{\mathrm{active}}$.
		\begin{algorithmic}[1]
			\STATE Initialize active item set $S_{\mathrm{active}}\leftarrow [K]$.
			\WHILE{$|S_{\mathrm{active}}|>1$}
			\STATE If $|S_{\mathrm{active}}| > \kappa$, randomly form $m = \left \lfloor |S_{\mathrm{active}}| / \kappa \right \rfloor$ disjoint batches of size $\kappa$ from $S_{\mathrm{active}}$; otherwise, treat $S_{\mathrm{active}}$ as a single batch.
			\STATE Run \textsc{Nested Elimination} with parameter $M$ on each batch.
			\STATE Update $S_{\mathrm{active}}$ to consist of the winners from all batches along with the items not assigned to any batch.
			\ENDWHILE
		\end{algorithmic}
	\end{algorithm}

	To facilitate the theoretical analysis of \textsc{HNE}, we first characterize the structural parameters of the algorithm. To identify the single best item from the initial set of size $K$, the algorithm must eliminate a total of $K-1$ items. 
	Observing that each execution of the \textsc{NE} subroutine eliminates exactly $\kappa-1$ items (with the exception of the final call for rounding reasons), the total number of subroutine calls is $N_{\mathrm{call}} :=\lceil \frac {K-1} {\kappa - 1} \rceil$. We are thus ready to write down the hardness of the problem under capacity constraints by adapting the worst-case hardness quantity from the unconstrained setting. Let
	\begin{align*}
	I_{*, \kappa}^{\mathrm{OA}} :=  (1-p)\,\log\!\left(\tfrac{1}{p}\right)
	\left(
	1 + \sum\nolimits_{n=2}^{\kappa} \tfrac{p^{n-1}}{1 + 2p + \cdots + (n-1)p^{n-2}}
	\right)^{-1}
	\end{align*}
	be the value of $I_*^{\mathrm{OA}}$ (defined in \eqref{eq:hardest to learn}) with $\kappa$ items. Recall from \eqref{eq:approximating IOA} that it is not sensitive to $ \kappa $, i.e., $\log\left(\tfrac{1}{p}\right)(1-p)\tfrac{1}{1+2p}
	\;\leq\;
	I_{*, \kappa}^{\mathrm{OA}}
	\;\leq\;
	\log\left(\tfrac{1}{p}\right)(1-p)\tfrac{1}{1+p}$. We then define the hardness quantity for \textsc{HNE} as the worst-case hardness per subroutine call, scaled by the total number of calls:
	$$
	\IHNE:= \tfrac {{I_{*, \kappa}^{\mathrm{OA}}}} {N_{\mathrm{call}}}.
	$$
	With these definitions in place, we formally state the sample complexity and correctness guarantees of \textsc{HNE} in Theorem~\ref{theorem_fixedconfidence_capacity}.
	
	\vspace{0.2 cm}
	\begin{theorem}[Sample complexity of \textsc{HNE}] 
		\label{theorem_fixedconfidence_capacity}
		For every confidence level $\delta\in(0,1)$, \textsc{HNE} is $\delta$-PAC with the parameter value
		\begin{equation}
		\label{equation_M_capacity}
		M = \tfrac{\log(1/\delta)+\log(N_{\mathrm{call}}\beta_{\kappa})}{\log(1/p)}.
		\end{equation}
		Furthermore, for every preference instance $f\in\mathcal M_p$, there is a constant $ C_f'' $ independent of $ \delta $ such that 
		\begin{equation}
		\label{equation_fixedconfidence_capacity}
		\E [\tau] \le \frac{\log(1/\delta)}{\IHNE}+ C_f''.
		\end{equation}
	\end{theorem}
	\vspace{0.2 cm}

	In Theorem~\ref{thm_worstcase_capacity} below, we provide a lower bound analysis to demonstrate the near-optimality of \textsc{HNE} in the worst case.
	
	\vspace{0.2 cm}
	\begin{theorem}[Worst-case analysis of general policies] 
		\label{thm_worstcase_capacity}
		For any $\delta$-PAC best-item identification policy with capacity constraint $\kappa$, there exists a preference instance $ f\in \mathcal M_p  $ such that
		$$\E [\tau] \ge \frac{\log(1/\delta)}{I_{\kappa}^{\mathrm{worst}}}+ O_{1/\delta}(1), \quad \text{ where } \quad I_{\kappa}^{\mathrm{worst}} := \tfrac{ \log(1/p) \left[(\kappa-1)(1+p^\kappa) - \frac{2(p-p^{\kappa})}{1-p}\right]   } { (K-1)(1-p^\kappa) }.
		$$
	\end{theorem}
	\vspace{0.2 cm}

	\noindent \textbf{Worst-case optimality of \textsc{HNE}.}  Now we examine the optimality of \textsc{HNE} by comparing the upper and lower bounds. When restricted to pairwise comparisons ($\kappa = 2$), the total number of subroutine calls is $N_{\mathrm{call}} = K-1$. By substituting $\kappa=2$ into the hardness quantities, we have 
	$
	I_{\kappa}^{\mathrm{worst}} = \IHNE =\frac{(1-p)\log(1/p)}{(K-1)(1+p)}.
	$
	As a result, the hardness quantity of \textsc{HNE} exactly matches that of the lower bound, which implies that \textsc{HNE} achieves asymptotic worst-case optimality for pairwise comparisons. Next, we fix a general $ \kappa \geq 2 $ and establish the following result.
	\begin{proposition}\label{prop:constant optimality}
		For every $ p \in (0,1) $ and $ 2 \le \kappa \le K $, it holds that $ \frac{\IHNE}{I_{\kappa}^{\mathrm{worst}}} \ge  \frac{2(\kappa-1)} {3\kappa^2},$
		which is a constant independent of $ p $ and $ K $.
	\end{proposition}

	As a result, for every fixed $ \kappa $, \textsc{HNE} is worst-case asymptotically  optimal up to a constant. We note that in many real-world scenarios, the capacity $\kappa$ is a small constant relative to the total number of items $K$. As a result, we find it particularly interesting to treat $ \kappa $ as a fixed quantity as we scale up $ K $. 
}	

\tbl{
\subsection{The Learning-to-Rank Problem with Capacity Constraints}
	\label{extension_rank}
	
	We now consider full-ranking identification under the capacity constraint~$\kappa$. We design an algorithm, termed \textsc{Nested Sort-and-Verify} (\textsc{NSV}), that is easy to implement and only relies on pairwise comparisons, and hence satisfies the capacity constraints for all $ \kappa \geq 2$. This algorithm consists of two layers. 
	The first layer includes a divide-and-conquer structure that combines \textsc{NE} and \textsc{MergeSort} \citep{parberry1998mergesort}. The former is what we studied earlier, and used here as a noisy comparison oracle that outputs the preferred item with controlled error probability. Since only pairs are involved, we write $\textsc{NE}_2(i,j;m)$ for \textsc{NE} applied to items $ \{i,j\} $ with tuning parameter~$m$. The latter is a classic sorting algorithm based on pairwise comparisons that determines a ranking, with worst-case guarantees on the number of calls. Overall, this layer outputs a (rough) candidate ranking where both the sample size and error probability can be controlled.

	The first layer alone, however, cannot achieve the optimal leading coefficient. As we will show later, the leading coefficient should grow \textit{linearly} in $ K $ whereas a mere combination of \textsc{NE} and \textsc{MergeSort} only leads to a leading coefficient on the order of $ K \log K $. Therefore, we introduce a second \textit{verification} layer that uses \textsc{NE} to comb through the adjacent pairs of the candidate ranking, which visits only a linear number of items. The verification layer is set with much higher accuracy to tell whether the candidate ranking is correct. If the verification fails, the algorithm restarts. The formal procedure is summarized in Algorithm~\ref{algo_verified_ne_mergesort}.

	\begin{algorithm}[t]
		\caption{Nested Sort-and-Verify (\textsc{NSV})}
		\label{algo_verified_ne_mergesort}
		\hspace*{0.02in}
		{\bf Input:} Items $[K]$, confidence level $\delta$, and separation
		parameter $p$.
		
		\hspace*{0.02in}
		{\bf Output:} A ranking $\sigma_{\mathrm{out}}$.
		\begin{algorithmic}[1]
			\STATE Compute $\eta_\delta$, $m_\delta$, and $M_\delta$ according
			to \eqref{eq:verified-ne-parameters}.
			\REPEAT
			\STATE (\emph{Sort.}) Run \textsc{MergeSort} on $[K]$, answering each
			comparison query between $i$ and $j$ with a fresh execution of
			$\textsc{NE}_2(i,j;m_\delta)$, the returned item being ranked
			higher. Let $\widehat\pi=(\widehat\pi(1),\ldots,\widehat\pi(K))$ be the resulting sorted permutation.
			\STATE (\emph{Verify.}) For $r=1,\ldots,K-1$, run a fresh
			execution of
			$\textsc{NE}_2\big(\widehat\pi(r),\widehat\pi(r+1);M_\delta\big)$.
			The verification \emph{passes} if every execution returns
			$\widehat\pi(r)$; otherwise it \emph{fails}.
			\UNTIL{the verification passes}
			\STATE \textbf{return}
			$\sigma_{\mathrm{out}}\leftarrow\widehat\pi^{-1}$.
		\end{algorithmic}
	\end{algorithm}

	To facilitate the theoretical analysis of \textsc{NSV}, let us describe its structural components and a few central observations. In the \textit{sort} phase, the \textsc{MergeSort} algorithm is known to incur at most $K\log_2 K$ comparisons in the worst case, regardless of the comparison outcomes \citep{parberry1998mergesort,sedgewick2011algorithms}. If every pairwise nested comparison is correct, \textsc{MergeSort} produces the correct ranking. In addition, our earlier random-walk analysis of \textsc{NE} provides tight error-probability and sample-complexity guarantees that we leverage. In the \textit{verification} phase, the algorithm re-compares each of the $K-1$ adjacent pairs in the candidate ranking. The two phases use different tuning parameters for accuracy: $m_\delta = O(\log\log\frac{1}{\delta})$ for the sort phase, and $M_\delta = O(\log\frac{1}{\delta})$ for the verification phase. That makes the sort-phase cost asymptotically negligible relative to the verification-phase cost, so that the leading sample complexity is governed by the $K-1$ verification comparisons. With these observations in place, we formally state the performance guarantee of \textsc{NSV} in Theorem~\ref{theorem_fixedconfidence_capacity_ranking}.
	
	\vspace{0.2 cm}
	\begin{theorem}[Sample complexity of \textsc{NSV}]
		\label{theorem_fixedconfidence_capacity_ranking}
		For every $K\ge2$, $2\le\kappa\le K$, $p\in(0,1)$,
		and $\delta\in(0,1)$,
		Algorithm~\ref{algo_verified_ne_mergesort} is $\delta$-PAC with the parameter values
		\begin{equation}
		\label{eq:verified-ne-parameters}
		\eta_\delta
		:=
		\min\left\{
		\tfrac14,\,
		\tfrac{K\log_2K}{(K-1)\log(e/\delta)}
		\right\}, \quad 
		m_\delta
		:=
		\left\lceil
		\tfrac{\log(K\log_2K/\eta_\delta)}{\log(1/p)}
		\right\rceil, \quad  \text{and}
		\quad
		M_\delta
		:=
		\left\lceil
		\tfrac{\log(4(K-1)/\delta)}{\log(1/p)}
		\right\rceil.
		\end{equation}
		Furthermore, its expected sample complexity is uniformly bounded by
		\begin{equation}
		\label{eq:verified-ne-sample}
		\sup_{f\in\mathcal M_p}\mathbb E_f[\tau]
		\le
		\tfrac{
			K\log_2K(1+p) 
		}{
			(1-p)(1-\eta_\delta)(1-\delta/4)
		} m_\delta + \tfrac{(K-1) (1+p) 
		}{
			(1-p)(1-\eta_\delta)(1-\delta/4)
		}M_\delta.
		\end{equation}
		Consequently,
		\begin{equation}
		\label{eq:verified-ne-leading}
		\sup_{f\in\mathcal M_p} \E_f[\tau] \le \tfrac{\log(1/\delta)}{\JNSV} + O\!\left(\log\log\tfrac{1}{\delta}\right), \quad \text{ where } \quad \JNSV := \tfrac{\log(1/p)(1-p)}
		{(K-1)(1+p)}.
		\end{equation}
		The quantity $\JNSV$ does not depend on the capacity $\kappa$, since \textsc{NSV} displays only pairs and is therefore feasible for every $\kappa\ge2$.
	\end{theorem}
	\vspace{0.2 cm}
	
	In Theorem~\ref{thm_worstcase_capacity_ranking} below, we provide a lower bound analysis to demonstrate the near-optimality of \textsc{NSV} in the worst case.
	
	\vspace{0.2 cm}
	\begin{theorem}[Worst-case lower bound for capacity-constrained ranking] 
		\label{thm_worstcase_capacity_ranking}
		For any $\delta$-PAC full-ranking policy with capacity constraint $\kappa$, there exists a preference instance $ f\in \mathcal M_p  $ such that
		$$\E [\tau] \ge \frac{\log(1/\delta)}{J_{\kappa}^{\mathrm{worst}}}+ O_{1/\delta}(1), \quad \text{ where } \quad 
		J_{\kappa}^{\mathrm{worst}} = \tfrac{ \log(1/p) (1-p)(1-p^{\kappa -1 }) } { (K-1)(1-p^\kappa) }.
		$$
	\end{theorem}
	\vspace{0.2 cm}
	
	\noindent \textbf{Near-optimality of \textsc{NSV}.} Now we examine the optimality of \textsc{NSV} by comparing the upper and lower bounds. Note that the comparison of leading coefficients satisfies
	\begin{equation}
	\label{eq:verified-ne-optimality-ratio}
	1 \;\le\; \frac{J_{\kappa}^{\mathrm{worst}}}{\JNSV}
	\;=\;
	\frac{(1+p)(1-p^{\kappa-1})}{1-p^\kappa}
	\;<\; 1+p \;<\; 2.
	\end{equation}
	When restricted to pairwise comparisons ($\kappa = 2$), substituting into the hardness quantities yields $J_{2}^{\mathrm{worst}} = \JNSV$, so \textsc{NSV} achieves exact asymptotic worst-case optimality. For general $\kappa \ge 2$, the ratio \eqref{eq:verified-ne-optimality-ratio} is bounded above by $1+p$. As a result, \textsc{NSV} is worst-case asymptotically optimal up to a universal factor of $2$.
}

\tbl{
	\vspace{0.1 cm}
	\begin{remark}
		\label{remark:cumulative regret}
		{\sf The pure-exploration problem we study also has implications for the cumulative regret of both tasks. Suppose that, alongside choosing display sets, the firm takes a downstream action $ a_t $ (say, which product to launch, or in which order to list the items), incurring a per-period loss $ L(a_t,\sigma_f)\in[0,\bar\Delta] $ that equals zero at the hindsight-optimal action. Consider an explore-then-exploit policy that runs \textsc{NE} at confidence level $ \delta $ and then commits over an operating horizon of length $ T $. Bounding the loss before commitment by Theorem~\ref{theorem_fixedconfidence} and the loss after commitment by the $ \delta $-PAC guarantee gives
			$\mathrm{Regret}_T \;\le\; \bar\Delta\,\E[\tau\wedge T] + \bar\Delta T\delta \;\le\; \bar\Delta\left(\tfrac{\log(1/\delta)}{\IN(f)} + C_f + T\delta\right). $
			Hence the choice $ \delta = 1/T $ yields 
			$$ \mathrm{Regret}_T \le \frac{\bar\Delta}{\IN(f)} \log(T) + O(1). $$ 
			The same argument applies to full-ranking identification via Theorem~\ref{theorem_ranking_fixedconfidence}, with $ \IN(f) $ replaced by $ \JN(f) $, and to the capacitated algorithms of Section~\ref{sec_extension_capacity} via $ \IHNE $ and $ \JNSV $. Sharpening the leading coefficient of $ \log(1/\delta) $ therefore sharpens the leading coefficient of the $ \log T $ regret bound above. \hfill$\diamond$}
	\end{remark}
}

\section{Numerical Experiments}
\label{section_experiment}
In this section, we empirically evaluate the performance of our algorithms. In Section~\ref{section_experiment1}, we investigate the best-item identification problem and compare \textsc{NE} with \textsc{MTP} \citep{feng2022robust}, focusing on their stopping times and runtimes. Subsequently, in Section~\ref{section_experiment2}, we conduct a numerical examination of the full-ranking identification problem, confirming the efficacy of our algorithm \textsc{NP}. For each experiment, the statistics of different methods are averaged over $512$ independent trials. The corresponding standard errors are also displayed as the (tiny) error bars in the figures. Additional implementation details and numerical results can be found in Appendix~\ref{appendix_experiments}.\footnote{Code is available through the\tbl{up-to-date} link \url{https://anonymous.4open.science/r/ranking_selection_choice_nested-083E/README.md}.}

\begin{figure}[tb]
	\begin{minipage}{1.0\columnwidth}
		\centering
		\hspace{-6pt}
		\subfigure[$K=5, p=0.9$.]{
			\begin{minipage}[b]{0.25\textwidth}
				\includegraphics[width=1\textwidth]{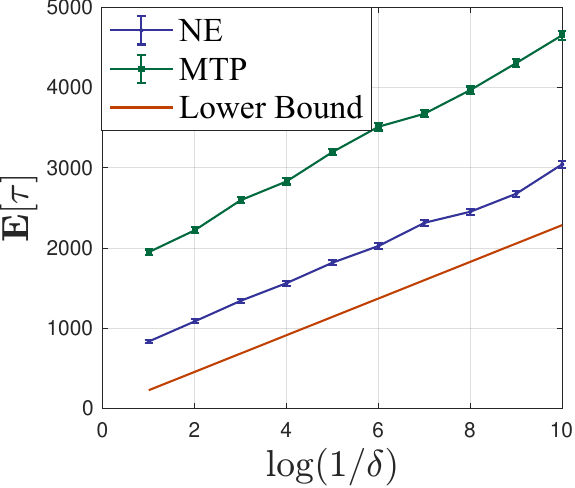} 
			\end{minipage}
		}
		\hspace{-6pt}
		\subfigure[$K=10, p=0.9$.]{
			\begin{minipage}[b]{0.25\textwidth}
				\includegraphics[width=1\textwidth]{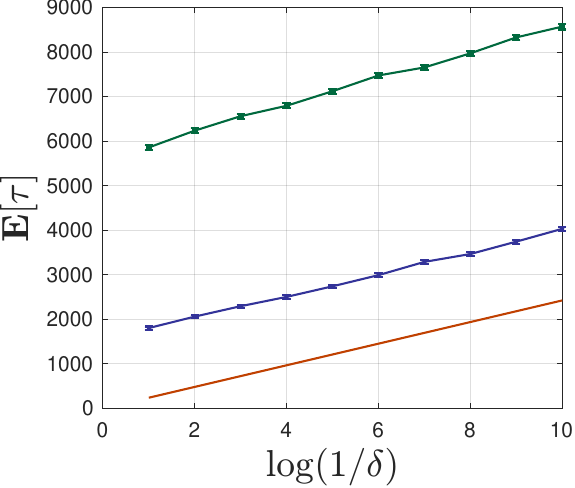} 
			\end{minipage}
		}
            \hspace{-6pt}
		\subfigure[$K=15, p=0.9$.]{
			\begin{minipage}[b]{0.25\textwidth}
				\includegraphics[width=1\textwidth]{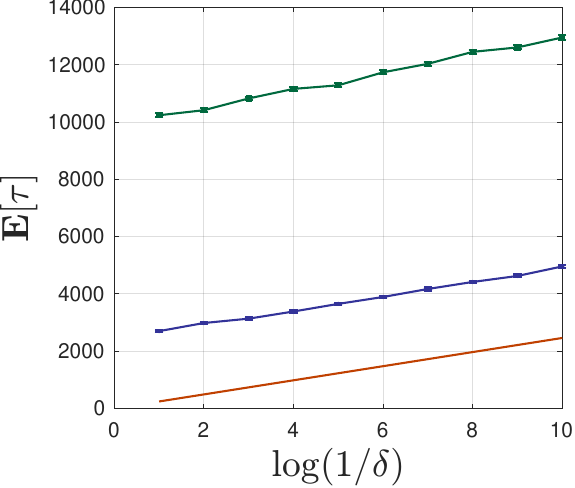} 
			\end{minipage}
		}
		\\ \hspace{-6pt}
		\subfigure[$K=5, p=0.6$.]{
			\begin{minipage}[b]{0.25\textwidth}
				\includegraphics[width=1\textwidth]{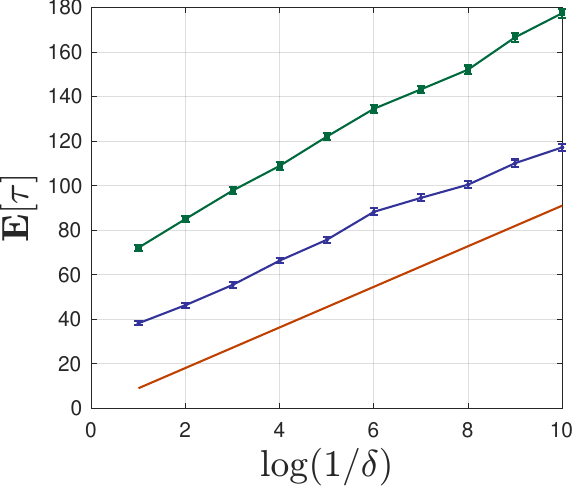} 
			\end{minipage}
		}
            \hspace{-6pt}
		\subfigure[$K=10, p=0.6$.]{
			\begin{minipage}[b]{0.25\textwidth}
				\includegraphics[width=1\textwidth]{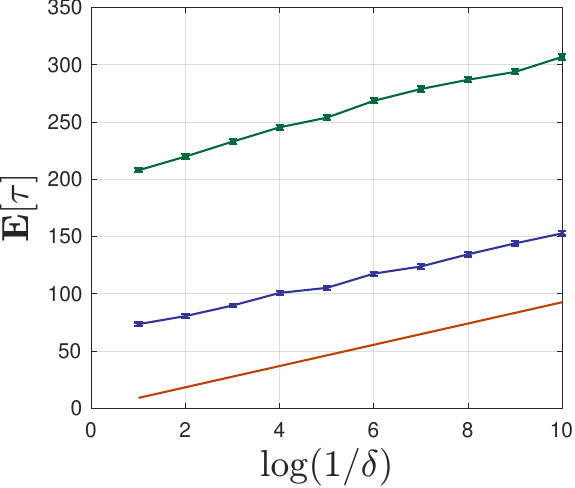} 
			\end{minipage}
		}
		\hspace{-6pt}
		\subfigure[$K=15, p=0.6$.]{
			\begin{minipage}[b]{0.25\textwidth}
				\includegraphics[width=1\textwidth]{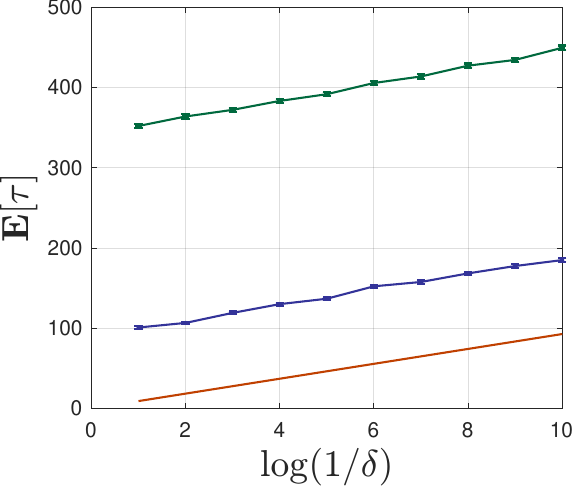} 
			\end{minipage}    
		}
	\end{minipage}
        \vspace{3pt}
	\caption{\textbf{Empirical stopping times of \textsc{NE} and \textsc{MTP} under the hardest-to-learn (i.e., OA) preference instances.} Different values of $\delta$, $ K $, $ p $ are considered.} \label{figure_worst_case}
\end{figure}

\subsection{The Best-Item Identification Problem}
\label{section_experiment1}
First, we consider the worst-case preferences in $  \mathcal M_p^{\mathrm{OA}}$. Recall that $\mathcal M_p^{\mathrm{OA}}$ represents the ``hardest-to-learn'' preferences that minimize both hardness quantities $ \IN (\cdot)$ and $ I_\ast (\cdot)$; see \eqref{eq:hardest to learn} and Proposition \ref{prop_OA}. We conduct our experiments with different target confidence levels $\delta$, as well as values of $K$ and $p$. We plot the empirical average stopping times of \textsc{NE} and
\textsc{MTP}, computed over the simulation trials, against
\(\log(1/\delta)\). The results are summarized in Figure~\ref{figure_worst_case}.\footnote{The empirical error probability is consistently lower than the corresponding target confidence level $\delta$ because we use the value of $ M $ in \eqref{equation_M} with theoretical guarantees. This choice of $ M $ is asymptotically tight for small $ \delta $; see Remark~\ref{remark:tightness of error probability}.}
In addition, we report the empirical means of the CPU runtimes for the whole procedure\footnote{All our experiments are implemented in MATLAB and parallelized on an Intel Xeon Gold 6244 CPU (3.60 GHz).} for $\delta = 0.01$ in Table~\ref{table_time}. 

\begin{table}
	
	\begin{center}  
		\begin{tabular}{crrrr}
			\toprule
			&  \multicolumn{2}{c}{$p=0.9$} &    \multicolumn{2}{c}{$p=0.6$}            \\
			\cmidrule(r){2-3}  \cmidrule(r){4-5}
			$K$& \multicolumn{1}{c}{$\textsc{NE}$}  & \multicolumn{1}{c}{$\textsc{MTP}$} & \multicolumn{1}{c}{$\textsc{NE}$}  & \multicolumn{1}{c}{$\textsc{MTP}$}  \\
			\midrule
			5  &0.0773 & 23.4022& 0.0035 & 0.8957 \\ 
			10 &0.1297 &108.4158& 0.0050 & 3.5353  \\
			15 & 0.1376 &400.5358& 0.0064 & 13.7457 \\
			\bottomrule
		\end{tabular}
	\end{center}
	\vskip 0.2 cm
	
	\caption{\textbf{Average CPU runtime (secs).} Here  $\delta = 0.01$, and the underlying preference is $ f^{\mathrm{OA}} $.}
    \label{table_time}
\end{table}

\begin{figure}[htbp]
	\centering
	\vspace{2pt}
	\begin{minipage}{1.0\columnwidth}
		\centering
		\hspace{-1pt}
		\subfigure[Preference $f_1$]{
			\begin{minipage}[b]{0.25\textwidth}
				\includegraphics[width=1\textwidth]{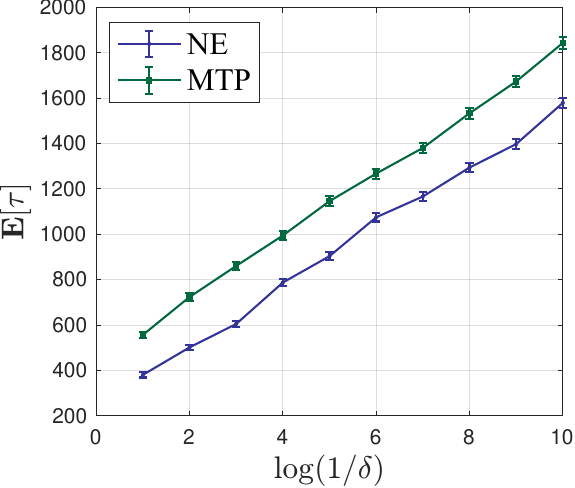} 
			\end{minipage}
		}
		\subfigure[Preference $f_2$]{
			\begin{minipage}[b]{0.25\textwidth}
				\includegraphics[width=1\textwidth]{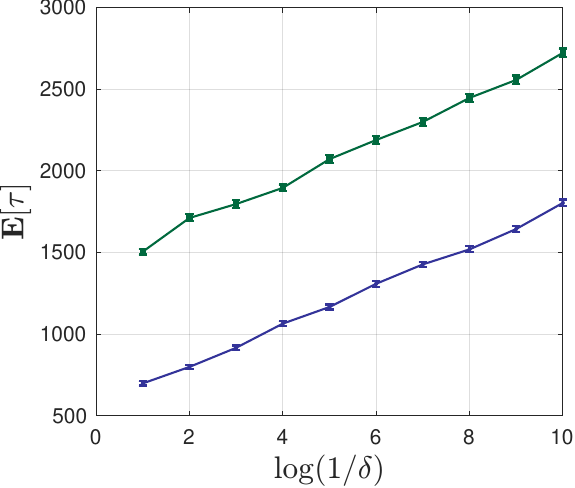}  
			\end{minipage}
		}
	\end{minipage}
	\caption{\textbf{Empirical stopping times of \textsc{NE} and \textsc{MTP} under the real data.} The two (non-worst-case) preferences $f_1$ and $f_2$ are calibrated from the Netflix Prize and Debian Logo datasets, respectively.} 
	\label{figure_general}
	\vspace{4pt}
\end{figure}

Next, we examine two general (non-worst-case) preferences $f_1$ and $f_2$, which are calibrated using MNL models fitted to the Netflix Prize and
Debian Logo data sets, respectively. The number of items for preference $f_1$ is $4$, while  $f_2$ has $8$ items. We set $p=0.9$ for both preferences; see Appendix~\ref{appendix_experiments} for detailed information. 
Figure~\ref{figure_general} shows the experimental results under the two general preferences. 
Let us summarize our observations from Figures \ref{figure_worst_case} and \ref{figure_general} as well as Table~\ref{table_time}:

\begin{enumerate}[label = (\roman*)]
\item \textbf{Sample Efficiency. }  \textsc{NE} consistently outperforms \textsc{MTP} in terms of empirical stopping times across all levels of $\delta$.  Notably, in the non-asymptotic regime where $\delta$ is moderately small, \textsc{NE} is significantly superior, indicating its greater practicality in real-world applications.
	
\item \textbf{Computational Efficiency. } \textsc{NE} is very computationally efficient and demonstrates a substantial advantage with regard to CPU runtimes as the problem scale increases. In contrast, \textsc{MTP} becomes computationally intractable with a reasonable amount of computing resources for large values of $K$ due to the time-consuming integer optimization problems at each time step. 
The results show that the runtime of \textsc{NE} typically improves upon \textsc{MTP} by \textit{three orders of magnitude}, especially for large values of $K$.
\end{enumerate}

\subsection{The Full-Ranking Identification Problem}
\label{section_experiment2}
To demonstrate the superiority of our \textsc{Nested Partition} algorithm for the full-ranking identification problem, we compare it against two baseline algorithms.  The first is \textsc{NE-Ranking}, an elimination-based algorithm outlined in Algorithm~\ref{algo_NE_ranking}.
 The second is termed \textsc{Repeated-NE}. As its name suggests, this algorithm repeatedly employs \textsc{NE}  with a confidence level of $\delta/(K-1)$. In each iteration of the \textsc{NE} subroutine, the best item within the active item set is eliminated. There are a total of $K-1$ iterations, ensuring that the overall error probability of the ranking is no greater than $\delta$.
\begin{figure}
	\centering
	\subfigure[$K=5, p=0.9$ ]{\includegraphics[width=0.25 \textwidth]{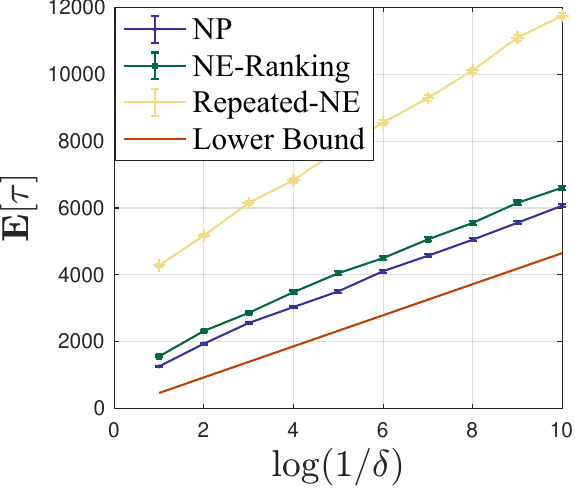}}
	\subfigure[$K=10, p=0.9$ ]{\includegraphics[width=0.25 \textwidth]{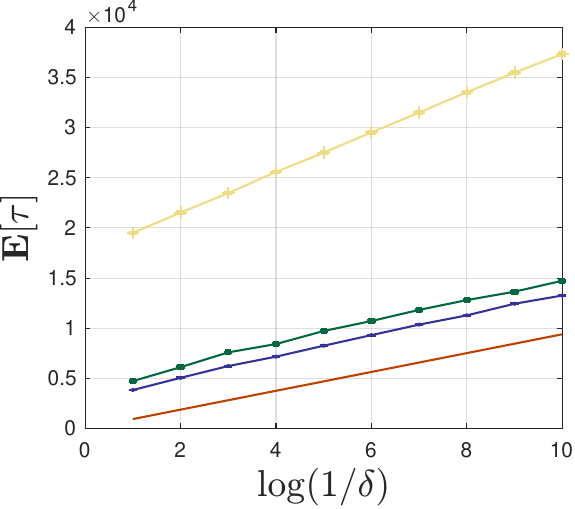}}
	\subfigure[$K=15, p=0.9$ ]{\includegraphics[width=0.25 \textwidth]{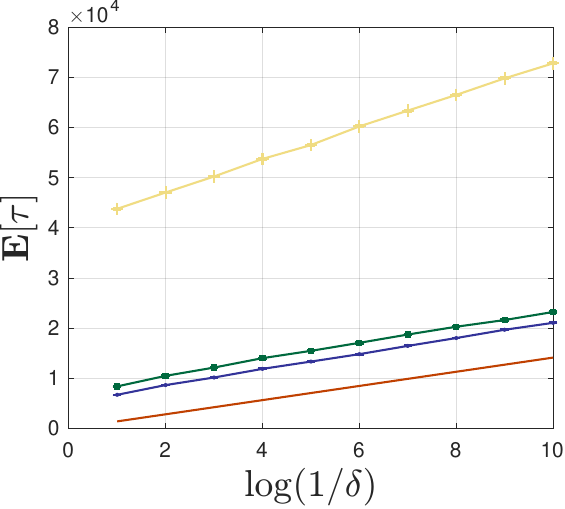}}
	
	\subfigure[$K=5, p=0.6$ ]{\includegraphics[width=0.25 \textwidth]{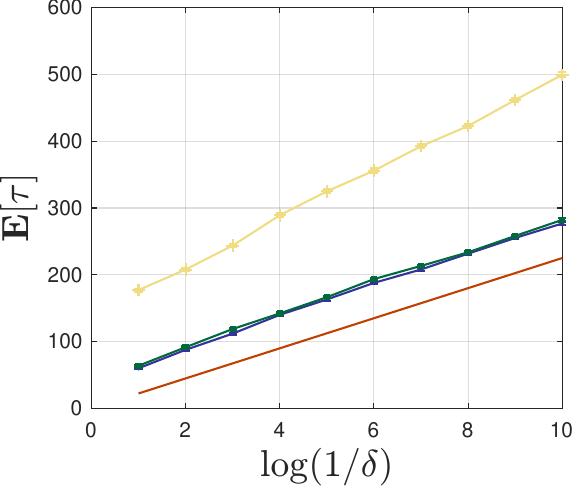}}
	\subfigure[$K=10, p=0.6$ ]{\includegraphics[width=0.25 \textwidth]{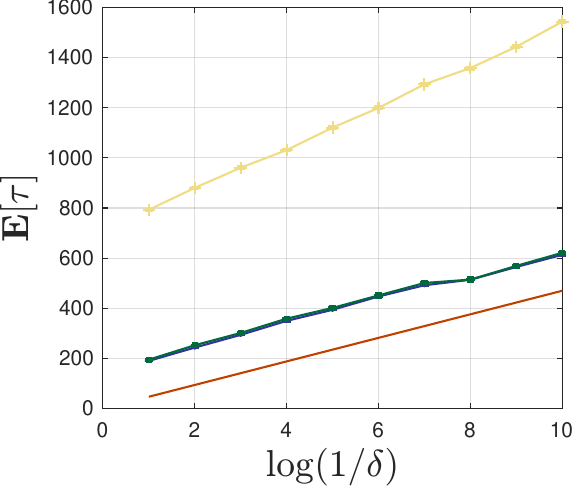}}
	\subfigure[$K=15, p=0.6$ ]{\includegraphics[width=0.25 \textwidth]{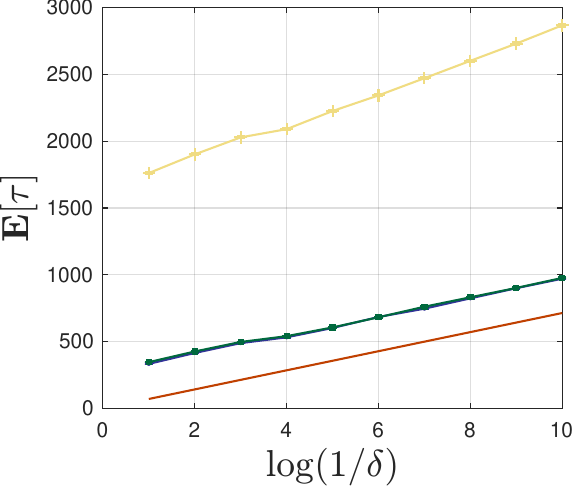}}
	\vspace{3pt}
	\caption{\textbf{Empirical stopping times of \textsc{NP}, \textsc{NE-Ranking}, and \textsc{Repeated-NE} under the hardest-to-learn (i.e., OA) preference instances.} Different values of $\delta$, $ K $, $ p $ are considered. } \label{figure_worst_case_ranking}
\end{figure}

\begin{figure}
	\centering
	\hspace{-1pt}
	\subfigure[Preference $f_1$]{
		\begin{minipage}[b]{0.25\textwidth}
			\includegraphics[width=1\textwidth]{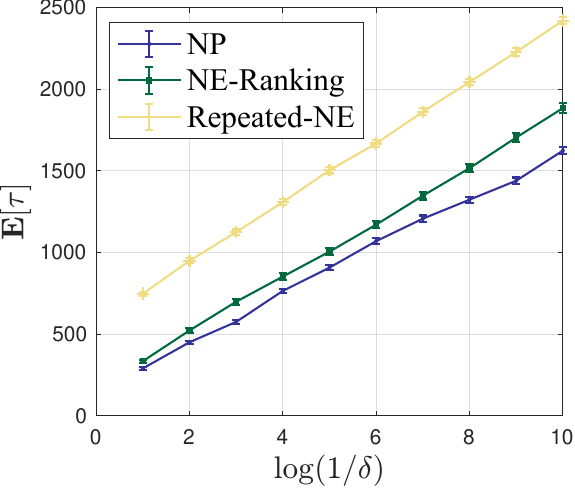} 
		\end{minipage}
	}
	\hspace{-1pt}
	\subfigure[Preference $f_2$]{
		\begin{minipage}[b]{0.25\textwidth}
			\includegraphics[width=1\textwidth]{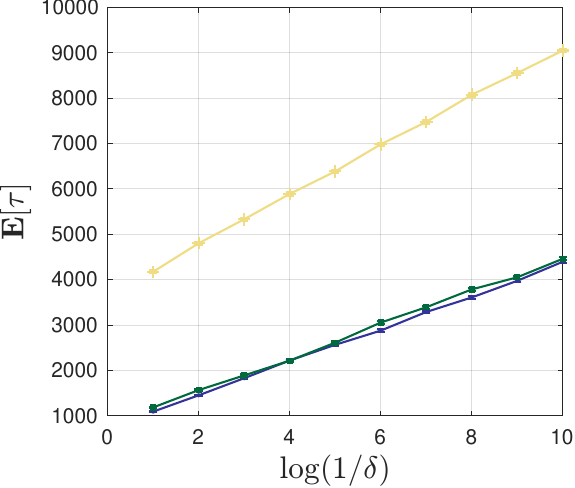}  
		\end{minipage}
	}
	\caption{\textbf{Empirical stopping times of \textsc{NP}, \textsc{NE-Ranking}, and \textsc{Repeated-NE} under real data.} The two (non-worst-case) preferences $f_1$ and $f_2$ are calibrated from the Netflix Prize and Debian Logo datasets, respectively. } 
	\label{figure_general_ranking}
\end{figure}

The experimental results under the OA preferences and general preferences are shown in Figures~\ref{figure_worst_case_ranking} and \ref{figure_general_ranking}, respectively. Our primary findings are outlined as follows:
\begin{enumerate}[label = (\roman*)]
	\item \textsc{NP} significantly outperforms the \textsc{Repeated-NE} baseline, particularly for large values of $K$. As \textsc{Repeated-NE} merely represents a direct application of the algorithm for best-item identification, our finding underscores the importance of investigating the full-ranking identification problem in its own right. 
	
	\item The performance of \textsc{NP} is never worse than that of \textsc{NE-Ranking}. In many cases their performance is nearly identical, which is not surprising since they derive from the same idea explained after Proposition~\ref{prop_OA_ranking_bound}. In fact, they admit the same deterministic approximation under OA preference instances. In other cases (e.g., $p=0.9$), \textsc{NP} is noticeably better. This demonstrates that the sophistication of recursive partitioning not only brings benefits in asymptotic theoretical guarantees, but also in non-asymptotic empirical performance.
	
	\item Of particular interest are the results under the OA preferences shown in Figure~\ref{figure_worst_case_ranking}. With regard to the growth trend, as $\log(1/\delta)$ increases, the slopes of the curves corresponding to \textsc{NP} and \textsc{NE-Ranking} in each subfigure approach that of the asymptotic lower bound, which is exactly equal to $1 / J_*^{\mathrm{OA}} $. Conversely, the slope of the curve corresponding to  \textsc{Repeated-NE} is consistently larger than that of the lower bound, indicating its suboptimal asymptotic performance. This highlights that under OA preferences, the expected stopping time of \textsc{NP} matches the lower bound asymptotically, thereby corroborating our theoretical findings. As for \textsc{NE-Ranking}, although lacking theoretical guarantees, we conjecture it is also worst-case asymptotically optimal.
\end{enumerate}

\section{Conclusions and Future Work}
\label{section_conclusion}
In this paper, we study online preference learning under choice-based feedback. Specifically, we address the best-item identification problem and introduce the \textsc{Nested Elimination} (\textsc{NE}) algorithm, which represents a significant advancement over prior research. Additionally, we initiate exploration into the full-ranking identification problem from choice-based feedback, for which we propose the \textsc{Nested Partition} (\textsc{NP}) algorithm. Leveraging the foundational principles of \textsc{NE}, \textsc{NP} is a step toward a more advanced approach. Both algorithms feature straightforward design and implementation, making them practical solutions for various applications. Our theoretical analysis and numerical experiments clearly demonstrate the computational and sample efficiency of our algorithms.\tbl{We further extend the nested approach to the capacity-constrained setting, where we develop
	\textsc{HNE} for best-item identification and \textsc{NSV} for full-ranking identification, each
	with a corresponding information-theoretic lower bound.}

There are a few opportunities for future work. First, we consider the \textit{fixed-confidence} formulation of the learning problem. A promising future direction would be to investigate the \emph{fixed-budget} setting, where the total number of time steps is strictly bounded, by combining the ideas from the multi-armed bandit literature. 
Second, this paper considers a setting for a fixed separation parameter $ p < 1 $ (or at least when a conservative estimate of $p$ is available). It will be interesting to design an algorithm that is fully agnostic to the value of $ p $ as well. \tbl{Third, the optimality guarantees of the nested approach are established in the worst-case (minimax) sense. It would be valuable to explore how \textsc{NE} and \textsc{NP}, or variants thereof, compare to the information-theoretic lower bounds on an instance-specific basis, although a major difficulty is a better understanding of $ I_*(f) $ and $ J_*(f) $ in general.}


\SingleSpacedXI
\bibliographystyle{informs2014} 
\bibliography{references} 

%
%
%
\newpage 
\ECSwitch

\renewcommand{\theHsection}{A\arabic{section}} 

\begin{APPENDICES}
\section{Analysis of Nested Elimination}
\label{appendix_analysis_NE}
\subsection{Preliminaries}
\label{subsection_best_Preliminaries}

\noindent\textbf{An Equivalent Formulation of Nested Elimination. } Note that multiple items can be eliminated within a single time step under Algorithm~\ref{algo1}. In this regard, it is straightforward to verify that if there exists some value of $k$ such that the elimination criterion is satisfied, i.e., $\sum_{i=1}^k W_t(\pi_t(i))-kW_t(\pi_t(k+1))\ge M$, then for all integer $ k'\in [k, |S_{\mathrm{active}}|-1]$,
$\sum\nolimits_{i=1}^{k'} W_t(\pi_t(i))-k'W_t(\pi_t(k'+1))\ge M.$
Thus, the outcomes of the eliminations will not be altered if we only allow eliminating the items one by one, starting with the least-voted item, still within one time step.
For convenience of analysis, we present an equivalent description of \textsc{NE} in Algorithm~\ref{algo2}, which only allows eliminating the items one by one, starting with the least voted item, within one time step.

\begin{algorithm}[htbp]
\caption{Nested Elimination (allowing eliminating one item at one time)}
\label{algo2}
\hspace*{0.02in} {\bf Input:} Tuning parameter $M>0.$

\hspace*{0.02in} {\bf Output:} The only element $i_{\mathrm{out}}$ of $S_{\mathrm{active}}$.
\begin{algorithmic}[1]
\STATE Initialize voting score $W_0(i)\leftarrow 0$ for all $i\in[K]$, active item set $S_{\mathrm{active}}\leftarrow [K]$, $t \leftarrow 0$
\WHILE{$|S_{\mathrm{active}}|>1$}
\STATE Sort the remaining items based on their voting scores. That is, find a permutation $\pi_t: {[|S_{\mathrm{active}}|]} \to S_{\mathrm{active}} $ such that $W_t(\pi_t(1))\ge W_t(\pi_t(2))\ge \cdots\ge W_t(\pi_t(|S_{\mathrm{active}}|))$.
\IF{$\sum_{i=1}^{|S_{\mathrm{active}}|-1} W_t \big(\pi_t(i) \big)-(|S_{\mathrm{active}}|-1)
	W_t \big(\pi_t(|S_{\mathrm{active}}|)\big)\ge M$}
\STATE $S_{\mathrm{active}}\leftarrow \{ \pi_t(1),\ldots,\pi_t(|S_{\mathrm{active}}|-1)\}$
\ELSE
\STATE Update the timer: $t \leftarrow t+1$.
\STATE Display the active set $S_{\mathrm{active}}$, and observe the choice $X_t \in S_{\mathrm{active}}$.
\STATE Update voting scores based on $ X_t $:
\begin{equation*}
   W_t(i) \leftarrow \begin{cases}
    W_{t-1}(i) + 1 & \text{if } i = X_t \\
    W_{t-1}(i) & \text{if }  i \neq X_t.\\
    \end{cases}
\end{equation*}
\ENDIF
\ENDWHILE
\end{algorithmic}
\end{algorithm}

In Algorithm~\ref{algo2}, the whole procedure can be divided into $K-1$ stages according to the number of active items. For any stage $r \in [K-1]$, we denote the active item set of size $K-r+1$ as $S_r = \{ S_r^1, S_r^2, \ldots, S_r^{K-r+1}\}$, where the corresponding true ranking satisfies $S_r^1< S_r^2< \ldots< S_r^{K-r+1}$. In particular, $S_1 = [K]$. For convenience, we also set $S_K$ as the singleton $S_{\mathrm{active}}$ when the algorithm terminates, and refer to the item that is eliminated in stage $r$ as $a_r$, i.e., $a_r:= S_r \setminus S_{r+1}$.

\subsection{Proof of Theorem~\ref{theorem_fixedconfidence}}
\label{appendix_theorem_fixedconfidence}

Before providing the formal proof of Theorem~\ref{theorem_fixedconfidence}, we introduce several key intermediate results, which we believe also have some independent significance.

Proposition~\ref{proposition_tau_general} below states that the expected stopping time of \textsc{NE} with input parameter $M$ is asymptotically upper bounded by ${\log(1/p)M}/{I^{\mathrm N}(f)}$ as the tuning parameter $M$ tends to infinity. Refer to Appendix~\ref{appendix_proposition_tau_general} for the  proof of Proposition~\ref{proposition_tau_general}. 

\vspace{0.2 cm}
\begin{proposition}[Expected stopping time of \textsc{NE} ]
	\label{proposition_tau_general}
	For any customer preference $f\in\mathcal M_p$,  \textsc{NE} ensures that
	\begin{equation*}
	\E [\tau] \le \frac{\log(1/p)M} {I^{\mathrm N}(f)} + o_M(1)
	\end{equation*}
	where the  $o_M(1)$ term is specified in Equation~\eqref{equation_oM1} in the corresponding proof.
\end{proposition}
\vspace{0.2 cm}

In addition to the expected stopping time, the other important performance metric is the error probability. Proposition~\ref{proposition_error}, proved in Appendix~\ref{appendix_proposition_error}, provides an upper bound on the error probability of \textsc{NE}.

\vspace{0.2 cm}
\begin{proposition}[Error probability of \textsc{NE}]
	\label{proposition_error}
	For any customer preference $f\in\mathcal M_p$, \textsc{NE} outputs an item $i_{\mathrm{out}} $ satisfying
	\begin{equation*}
	\mathbb P (i_{\mathrm{out}} \neq 1 ) \le  \beta_K\cdot p^M. 
	\end{equation*}
\end{proposition}
\vspace{0.2 cm}

Note that the upper bound demonstrated in Proposition~\ref{proposition_error} does not depend on the specific preference instance $f$. In particular, it decays exponentially in the exogenous parameter $M$.

\vspace{0.1 cm}
\begin{remark}\label{remark:tightness of error probability}
	{\sf The exponential decay rate of the error probability in Proposition~\ref{proposition_error} is tight. More precisely, we have 
	$$
	\log(p) \overset{(a)}{\le} \lim_{M\to \infty }\frac {\log( \mathbb P (i_{\mathrm{out}} \neq 1 ) )} {M} \overset{(b)}{\le}\log (p),
	$$
	where (a) is a consequence of the lower bound in Fact~\ref{fact_lower_bound}, and (b) follows from Proposition~\ref{proposition_error} directly. \hfill$ \diamond $}
\end{remark}
\vspace{0.1 cm}

With all the necessary results in hand, we proceed to present the proof of Theorem~\ref{theorem_fixedconfidence}. 

\proof{Proof of Theorem~\ref{theorem_fixedconfidence}. }Consider any fixed confidence level $\delta\in(0,1)$. On account of Proposition~\ref{proposition_error}, with  parameter $M = \frac{\log(1/\delta)+\log(\beta_K)} {\log(1/p)}$,  \textsc{NE} outputs an item $i_{\mathrm{out}} $ satisfying
\begin{equation*}
    \mathbb P (i_{\mathrm{out}} \neq 1 ) \le  \beta_K\cdot p^M= \delta
\end{equation*}
for any customer preference $f\in\mathcal M_p$. Therefore, according to Definition~\ref{definition_delta_pac}, to confirm \textsc{NE} is $\delta$-PAC, it remains to show  $\mathbb P (\tau<\infty) = 1$. But that is directly implied by Proposition~\ref{proposition_tau_general}, which indicates that $\mathbb E [\tau] < \infty$. Therefore, \textsc{NE} is $\delta$-PAC. With $M = \frac{\log(1/\delta)+\log(\beta_K)} {\log(1/p)}$, we may invoke Proposition~\ref{proposition_tau_general} again
and bound the expected sample complexity as follows:
\begin{align*}
\E [\tau] \ \le\  \frac{\log(1/p)M}{I^{\mathrm N}(f)}  + o_M(1)  \ =\  \frac{\log(1/\delta)}{I^{\mathrm N}(f)}+\frac{\log(\beta_K)}{I^{\mathrm N}(f)} + o_M(1) \ =\  \frac{\log(1/\delta)}{I^{\mathrm N}(f)} + O_{1/\delta}(1),
\end{align*}
where the last step comes from the fact that $ \beta_K $ and $ p $ are fixed as constants as $ 1/\delta $ grows. 
\Halmos
\endproof

\subsection{Proof of Proposition~\ref{proposition_tau_general}}
\label{appendix_proposition_tau_general}

Let us start with a technical lemma.
\begin{lemma}
	For any stage $r\in[K-1]$, we introduce the event 
	$$
	\mathcal E _r = \{ a_j = K-j+1 \textup{ for all } j\in[r-1] \} \in \mathcal{F}_{T_{r-1}},
	$$
	which means that \textsc{NE} eliminates the worst item correctly in each of the previous $r-1$ stages. Then for any customer preference $f\in\mathcal M_p$ and any stage $r \in [K-1]\setminus \{1\}$, there exists a constant $ q_0 \in (0,1) $, independent of $ M $, such that
	$$
	\mathbb P(\mathcal E _{r-1}\setminus \mathcal E _{r}) \le Q_*^r := (K-r+1)q_0^M.
	$$
	In other words, the probability $ \mathbb P(\mathcal E _{r-1}\setminus \mathcal E _{r}) $ decays exponentially fast in $ M $.
	\label{lemma_Error_r}
\end{lemma}

\proof{Proof of Proposition~\ref{proposition_tau_general}. }
Recall that Algorithm~\ref{algo1} and Algorithm~\ref{algo2} are equivalent with respect to the final outputs. Since the latter gives us convenience in the analysis, we will adopt Algorithm~\ref{algo2} in this proof.

For any stage $r\in [K-1]$, its cumulative time is denoted as $T_r$, i.e., 
\begin{align*}
T_r := \inf\left\{t\ge 1 \,\middle\vert\, \sum_{i=1}^{K-r} W_t(\pi_t(i))-(K-r)W_t(\pi_t(K-r+1))\ge M \right\},
\end{align*}
which is a stopping time by definition. For ease of notation, we also set $T_0 = 0$. For any stage $r\in [K-1]$, we denote its number of time steps as $\tau_r := T_r-T_{r-1}$. As such, we are interested in bounding the expected stopping time $\E [\tau]  = \E[T_{K-1}]$.

\vspace{0.2 cm}
\noindent\textbf{Step 1 (Decomposition of the expected stopping time).}  For any stage $r\in[K-1]$, the event
$$
\mathcal E _r = \{ a_j = K-j+1 \textup{ for all } j\in[r-1] \},
$$
means that \textsc{NE} eliminates the worst item correctly in each of the first $r-1$ stages. In particular, $\mathcal E_1$ is always true and $\Omega = \mathcal E _{1}\supseteq \mathcal E _{2}\supseteq \cdots \supseteq  \mathcal E _{K-1}$.  Note that if $\mathcal E _r$ holds, then the active set $S_r$ in stage $r$ must be exactly $[K-r+1]$. Due to linearity of expectation, we can decompose the expected stopping time as follows:
\begin{align*}
    \E [\tau] \, =\,  \sum_{r=1}^{K-1} \E[\tau_r ] \, =\,
    \sum_{r=1}^{K-1} \E[\tau_r \cdot \mathbbm{1}\{\mathcal E _r\} ] +  \sum_{r=1}^{K-1} \E[\tau_r \cdot \mathbbm{1}\{\mathcal E _r^c\} ]\ \leq\  \sum_{r=1}^{K-1}  \E[\tau_r \mid \mathcal{E}_r ] + \sum_{r=1}^{K-1}  \E[\tau_r \cdot \mathbbm{1}\{\mathcal E _r ^c\} ].
\end{align*}
We will also introduce two pieces of shorthand notation
\begin{align*}
T^{\dagger} = \sum_{r=1}^{K-1} \E[\tau_r \mid \mathcal E _r ] \quad \text{ and } \quad 
     T^{\ddagger} = \sum_{r=1}^{K-1} \E[\tau_r \cdot \mathbbm{1}\{\mathcal E _r^c\} ].
\end{align*}

In the following steps, we will bound $T^{\dagger}$ and $T^{\ddagger}$ separately. Specifically, we will show $T^{\dagger} = \frac{\log(1/p)M}{I^{\mathrm N}(f)} + o_M(1)$ and $T^{\ddagger} =  o_M(1)$, which will finish the proof.

\vspace{0.2 cm}
\noindent\textbf{Step 2 (Bounding $T^{\dagger}$).} We will analyze the dynamics of the score vector $ \bm W_{t} := (W_{t}(1), \ldots, W_{t}(K)) $ stage by stage. Consider any  stage $r \in [K-1] $ and suppose that the event $ \mathcal{E}_r \in \mathcal{F}_{T_{r-1}} $ holds, which means that the items $ \{K-r+2, \ldots, K\} $ have been (correctly) eliminated in all previous stages and the display set $ [K-r+1] $ is offered throughout stage $ r $. As a result, the score vector starts with value $ \bm W_{T_{r-1}} $, and at every step, it increases by $ \bm e_i $ with probability $ f(i|[K-r+1]) $ for every $ i \in [K-r+1] $ until $ T_r $. We will analyze both the stopping time and stopping distribution of $ \bm W_{t} $ stage by stage.

\underline{\underline{ The first stage ($ r = 1 $).}} To help build intuition and establish the base case, let us start from the first stage, where the full display set $ [K] $ is offered every time. Therefore, the score vector $ \bm W_{t} := (W_{t}(1), \ldots, W_{t}(K)) $ behaves like a random walk. It starts with the zero vector $ \bm 0 $. At every step, $ \bm W_t $ is increased by $ \bm e_i $ with probability $ f(i|[K]) $ for every $ i \in [K] $ until time $ T_1 $.

Motivated by the stopping criterion of the first stage, consider the stochastic process $ \{\sum_{i=1}^{K-1} W_{t}(i)-(K-1)W_{t}(K)\} $. It starts with zero, increases by one with probability $ \sum_{i=1}^{K-1} f(i|[K]) $ and decreases by $ K-1 $ with probability $ f(K|[K]) $. Therefore, its drift (i.e., expected increment every step) equals 
\begin{align*}
\sum_{i=1}^{K-1} f(i|[K])-(K-1)f(K|[K]) = {1- Kf(K|[K])} = \frac{1}{D(f,1)},
\end{align*}
where the expression of $ D(f,1) $ comes from \eqref{equation_definition_Dfr_NE}. We can use this random walk to analyze the expected duration of the first stage. Notice that the worst item in $S_1$ (i.e., item $K$) is not necessarily the one that is eliminated in the first stage, and $\pi_{T_1}$ might not be consistent with the ground truth $
\sigma_*$.  
Thus, we have
\begin{align*}
   \sum_{i=1}^{K-1} W_{T_1}(i)-(K-1)W_{T_1}(K) \le  \sum_{i=1}^{K-1} W_{T_1}(\pi_{T_1}(i))-(K-1)W_{T_1}(\pi_{T_1}(K))\stackrel{(a)}{=}M,
\end{align*}
where part (a) is due to the first stage's stopping rule plus the fact that there is exactly one vote every step, and hence there is no overshoot of the threshold $ M $. By taking expectation on both sides, we can get 
\begin{align*}
    M \ \ge\ \E\left [ \sum_{i=1}^{K-1} W_{T_1}(i)-(K-1)W_{T_1}(K) \mid \mathcal{E}_1 \right] 
    \ =\  \frac { 1 } {D(f,1)} \E[\tau_1 \mid \mathcal E _1 ],
\end{align*}
where the first equality follows from the optional stopping theorem, and the fact that $\{\sum_{i=1}^{K-1} W_{t}(i)-(K-1)W_{t}(K)- \frac{t}{D(f,1)}\}_t$ is a martingale up to the stopping time $ T_1 $. Rearranging the terms, we conclude that
\begin{align}
{ \E[\tau_1 \mid \mathcal{E}_1 ] }  \le {D(f,1)} M.
\label{equation_theorem_tau_result1}
\end{align}

\vspace{0.2 cm}
\underline{\underline{Any subsequent stage ($ r > 1 $).}} Consider any subsequent stage $r \in [K-1]\setminus \{1\}$ and suppose that the event $ \mathcal{E}_r \in \mathcal{F}_{T_{r-1}} $ holds, which means that the items $ \{K-r+2, \ldots, K\} $ have been (correctly) eliminated in all previous stages and the display set $ [K-r+1] $ is offered throughout the stage. As a result, the score vector starts with value $ \bm W_{T_{r-1}} $, and at every step, it increases by $ \bm e_i $ with probability $ f(i|[K-r+1]) $ for every $ i \in [K-r+1] $ until $ T_r $. Similarly to the previous arguments, we will analyze both the stopping time (i.e., $ \mathbb{E}[\tau_r | \mathcal{E}_r] $) and stopping distribution of the scores $ \bm W_{T_r} $ conditional on $ \mathcal{E}_r $.

Invoking the dynamics of the random walk $ \bm W_t $ in stage $ r $, the process $ \left\{\sum_{i=1}^{K-r} W_{t}(i)  -  (K-r)W_{t}(K-r+1)  \right\} $
is a random walk that, independent of the history $ \mathcal{F}_{T_{r-1}} $,  increases by one with probability $ \sum_{i=1}^{K-r} f(i|[K-r+1]) $ and decreases by $ (K-r) $ with probability $ f(K-r+1|[K-r+1]) $. Therefore, it holds that for every $ \bm W_{T_{r-1}} \in \mathcal{E}_r $,
\begin{align}
M &\stackrel{(a)}{\ge}\ \E\left [ \sum_{i=1}^{K-r} W_{T_r}\left(i\right)-(K-r)W_{T_r}\left({K-r+1}\right) \mid \bm W_{T_{r-1}} \right] \notag \\
&=\E\left [ \sum_{i=1}^{K-r} \left(W_{T_r}\left(i\right) - W_{T_{r-1}}\left(i\right) \right)-(K-r)\left(W_{T_r}\left({K-r+1}\right) -W_{T_{r-1}}\left({K-r+1}\right) \right) \mid \bm W_{T_{r-1}} \right] \notag \\ &\phantom{=} + \sum_{i=1}^{K-r} W_{T_{r-1}}\left(i\right)-(K-r)W_{T_{r-1}}\left({K-r+1}\right) \notag \\
&\stackrel{(b)}{=}  \left( \sum_{i=1}^{K-r} f(i|[K-r+1])-(K-r)f(K-r+1|[K-r+1]) \right) \E[\tau_r | \bm W_{T_{r-1}}] \notag \\
&\quad + M - (K-r+1) (W_{T_{r-1}}({K-r+1})-W_{T_{r-1}}({K-r+2})). \notag \\
&\stackrel{(c)}{=}  \left( \sum_{i=1}^{K-r} f(i|[K-r+1])-(K-r)f(K-r+1|[K-r+1]) \right) \E[\tau_r | \bm W_{T_{r-1}}] \notag \\
&\quad + M - (K-r+1) \Z_{T_{r-1}}(K-r+1). \notag 
\end{align}
In the derivations above, step (a) comes from the stopping criterion of stage $ r $. Step (b) comes from the optional stopping theorem for the incremental process, plus the fact that item $K-r+2$ is eliminated at stage $ (r-1) $, and as a result, 
$
\sum_{i=1}^{K-r+1} W_{T_{r-1}}\left(i\right)-(K-r+1)W_{T_{r-1}}\left({K-r+2}\right) = M
$ according to the stopping criterion of stage $ r-1 $. Finally, in step (c), we introduce
\begin{align*}
\Z_{t}(\ell) := (\bm e_\ell - \bm e_{\ell+1})^T \bm W_t = W_t(\ell) - W_t(\ell+1)
\end{align*}
for every $ \ell \in [K-1] $ as the score difference between item $ \ell $ and item $ \ell + 1 $ for shorthand notation. In particular, $ \Z_{T_{r-1}}(K-r+1) $ means the score difference between the two least preferred items according to the ground truth preference (i.e., items $ K-r+1 $ and $ K-r+2 $) at time $ T_{r-1} $ (i.e., at the end of the previous stage $ r-1 $). Now, we rearrange terms and have
\begin{align*}
    \E[\tau_r \mid \bm W_{T_{r-1}}] &\le \frac{K-r+1} {\sum_{i=1}^{K-r} f(i|[K-r+1])-(K-r)f(K-r+1|[K-r+1]) } \Z_{T_{r-1}}(K-r+1)
\end{align*}
By taking expectation over $ \bm W_{T_{r-1}} $ conditional on $\mathcal E_r$, we get
 \begin{align}
 \label{equation_theorem_tau4}
    \E[\tau_r \mid \mathcal E_r] 
    &\le \frac{K-r+1} {1-(K-r+1)f(K-r+1|[K-r+1]) } \E[\Z_{T_{r-1}}(K-r+1) \mid \mathcal E_r].
\end{align}

As can be seen, the time duration in stage $ r $ depends on the distribution of $ \Z_{T_{r-1}}(K-r+1) $, which further depends on the stopping distribution of the previous stage $ r-1 $. To analyze $ \E[\Z_{T_{r-1}}(K-r+1)\mid \mathcal E_r] $, we first notice that $ \E[\Z_{T_{r-1}}(K-r+1)\mid \mathcal E_r]  \approx \E[\Z_{T_{r-1}}(K-r+1)\mid \mathcal E_{r-1}]$ because $ \mathcal E_{r-1} \approx \mathcal E_r $ up to a small probability. More precisely, by the law of total probability, 
\begin{align*}
&\E[\Z_{T_{r-1}}(K-r+1)\mid \mathcal E_{r-1}] \\
\ =\  &\E[\Z_{T_{r-1}}(K-r+1)\mid \mathcal E_{r}] \Pr(\mathcal{E}_r|\mathcal{E}_{r-1}) + \E[\Z_{T_{r-1}}(K-r+1)\mid \mathcal E_{r-1} \setminus \mathcal{E}_r] \Pr(\mathcal{E}_r^c|\mathcal{E}_{r-1}) \\
\ \geq\  &\E[\Z_{T_{r-1}}(K-r+1)\mid \mathcal E_{r}] \Pr(\mathcal{E}_r|\mathcal{E}_{r-1})\\
\ \geq\  &\E[\Z_{T_{r-1}}(K-r+1)\mid \mathcal E_{r}] \Big(1 - \Pr(\mathcal{E}_{r-1}\setminus \mathcal{E}_{r})\Big).
\end{align*}
Rearranging terms, we have
\begin{align}\label{equation_theorem_tau1}
\E[\Z_{T_{r-1}}(K-r+1)\mid \mathcal E_{r}] \leq \E[\Z_{T_{r-1}}(K-r+1)\mid \mathcal E_{r-1}] + M \Pr(\mathcal{E}_{r-1} \setminus \mathcal{E}_{r}),
\end{align}
where the inequality is because  $ |\Z_{T_{r-1}}(K-r+1)| \leq M$ almost surely for otherwise the algorithm would have already been terminated.

To analyze $ \E[\Z_{T_{r-1}}(K-r+1)\mid \mathcal E_{r-1}] $, note that in stage $ r-1 $ and conditional on event  $\mathcal E _{r-1} \in \mathcal{F}_{T_{r-2}} $, the active set $ [K-r+2] $ is consistently offered. Therefore, in stage $ r-1 $, the process $ \{\Z_{t}(K-r+1)\} $ starts with $ \Z_{T_{r-2}}(K-r+1) $ and behaves according to a random walk that increases by one with probability $ f(K-r+1|[K-r+2]) $ and decreases by one with probability $ f(K-r+2|[K-r+2]) $ every period. Its drift (i.e., expected increment every step) at stage $ r-1 $ equals
\begin{align*}
\Delta^{r-1}_{r}  = f(K-r+1|[K-r+2])-f(K-r+2|[K-r+2]).
\end{align*}
By the optional stopping theorem, for every $ \Z_{T_{r-2}}(K-r+1) \in \mathcal{E}_{r-1} $, we have
\begin{align*}
&\E[\Z_{T_{r-1}}(K-r+1)\mid \Z_{T_{r-2}}(K-r+1)] 
= \Delta^{r-1}_{r}  \cdot \E[\tau_{r-1}\mid \Z_{T_{r-2}}(K-r+1)]  + \Z_{T_{r-2}}(K-r+1).
\end{align*}
Take expectation over $ \Z_{T_{r-2}}(K-r+1) $ conditional on $ \mathcal{E}_{r-1} $, and we have
\begin{align}
    &\phantom{=\ } \E[\Z_{T_{r-1}}(K-r+1)\mid \mathcal{E}_{r-1} ] = \Delta^{r-1}_{r}\cdot \E[\tau_{r-1}\mid \mathcal{E}_{r-1} ]  +
    \E[ \Z_{T_{r-2}}(K-r+1)\mid \mathcal{E}_{r-1} ].\label{equation_theorem_tau2}
\end{align}
Combine \eqref{equation_theorem_tau1} and \eqref{equation_theorem_tau2}, and we have
\begin{align*}
\E[\Z_{T_{r-1}}(K-r+1) \mid \mathcal E _r] 
\le \   \E[ \Z_{T_{r-2}}(K-r+1)\mid \mathcal{E}_{r-1} ] + \Delta^{r-1}_{r} \cdot \E[\tau_{r-1}\mid \mathcal E _{r-1}]   +  M \mathbb P(\mathcal E _{r-1}\setminus \mathcal E _{r}).
\end{align*}
Apply the same process to all further previous stages $ r-2, r-3, \ldots, 1 $ using backward induction, and we have that for all $ j \in [r-1] $,
\begin{align*}
\E[\Z_{T_{r-j}}(K-r+1) \mid \mathcal E_r] 
\le \   \E[ \Z_{T_{r-j-1}}(K-r+1)\mid \mathcal{E}_{r-j} ] + \Delta^{r-j}_{r} \cdot \E[\tau_{r-j}\mid \mathcal E _{r-j}]   +  M \mathbb P(\mathcal E _{r-j}\setminus \mathcal E _{r-j+1}).
\end{align*}
Equivalently, by taking $ i = r-j $, we have for $ i \in [r-1] $,
\begin{align*}
\E[\Z_{T_{i}}(K-r+1) \mid \mathcal E_r] 
\le \   \E[ \Z_{T_{i-1}}(K-r+1)\mid \mathcal{E}_{i} ] + \Delta^{i}_{r} \cdot \E[\tau_{i}\mid \mathcal E _{i}]   +  M \mathbb P(\mathcal E _{i}\setminus \mathcal E _{i+1}).
\end{align*}
Therefore,
\begin{align*}
&\ \E[\Z_{T_{r-1}}(K-r+1) \mid \mathcal E _r] \\
\le &\   \E[ \Z_{T_{r-2}}(K-r+1)\mid \mathcal{E}_{r-1} ] + \Delta^{r-1}_{r} \cdot \E[\tau_{r-1}\mid \mathcal E _{r-1}]   +  M \mathbb P(\mathcal E _{r-1}\setminus \mathcal E _{r}) \\
\le &\  \E[ \Z_{T_{r-3}}(K-r+1)\mid \mathcal{E}_{r-2} ] + \sum_{i=r-2}^{r-1} \Delta^{i}_{r}\E[\tau_{i}\mid \mathcal E _{i}]   +  M \sum_{i=r-2}^{r-1}\mathbb P(\mathcal E _{i}\setminus \mathcal E _{i+1}) \\
&\vdots \\
\le &\  \E[ \Z_{T_{0}}(K-r+1)\mid \mathcal{E}_{1} ] + \sum_{i=1}^{r-1} \Delta^i_{r} \cdot \E[\tau_{i}\mid \mathcal E _{i}]   +  M \sum_{i=1}^{r-1}\mathbb P(\mathcal E _{i}\setminus \mathcal E _{i+1}) \\
= &\   \sum_{i=1}^{r-1} \Delta^i_{r} \cdot \E[\tau_{i}\mid \mathcal E _{i}]   +  M \sum_{i=1}^{r-1}\mathbb P(\mathcal E _{i}\setminus \mathcal E _{i+1})
\end{align*}
Together with \eqref{equation_theorem_tau4} and Lemma~\ref{lemma_Error_r}, it holds that 
\begin{align}
\E[\tau_r \mid \mathcal E _r] 
&\le \tfrac{(K-r+1)   } {1-(K-r+1)f(K-r+1|[K-r+1]) } \sum_{i=1}^{r-1} \Delta^i_{r} \E[\tau_{i}\mid \mathcal E _{i}] +\tfrac{(K-r+1) M \sum_{i=2}^{r} Q_*^i} {1-(K-r+1)f(K-r+1|[K-r+1]) }. \label{equation_theorem_tau_result2}
\end{align}

\vspace{0.2 cm}
\underline{\underline{Deriving the bound.}} We will show that there exists $ q \in (0,1) $ such that 
\begin{align}\label{eq:time in each stage}
\E[\tau_r \mid \mathcal E _r ] \leq M \cdot D(f,r) + o_M(q^M) \quad \text{ for every }  r \in [K-1].
\end{align}
Let $ q_0 $ and $ Q_\ast^r = (K-r+1) q_0^M $ be taken from Lemma~\ref{lemma_Error_r} and pick $ q_0 < q < 1 $. We show \eqref{eq:time in each stage} by induction on $ r $. The $ r = 1 $ case follows directly by \eqref{equation_theorem_tau_result1}. Suppose \eqref{eq:time in each stage} holds for  $ i \in [r -1] $, then 
\begin{align*}
\E[\tau_r \mid \mathcal E _r] 
&\stackrel{(a)}{\le} \tfrac{(K-r+1)   } {1-(K-r+1)f(K-r+1|[K-r+1]) } \sum_{i=1}^{r-1} \Delta^i_{r} \E[\tau_{i}\mid \mathcal E _{i}] +\tfrac{(K-r+1)(K-1)(r-1) } {1-(K-r+1)f(K-r+1|[K-r+1]) } M  q_0^M \\
&\stackrel{(b)}{\le} \tfrac{(K-r+1)   } {1-(K-r+1)f(K-r+1|[K-r+1]) } \sum_{i=1}^{r-1} \Delta^i_{r} M D(f,i)  +\tfrac{(K-r+1)^2 (r-1) } {1-(K-r+1)f(K-r+1|[K-r+1]) } M  q_0^M+ o_M(q^M)
\\
&{=} \tfrac{(K-r+1)   } {1-(K-r+1)f(K-r+1|[K-r+1]) } \sum_{i=1}^{r-1} \Delta^i_{r} M D(f,i)  + o_M(q^M)\\
&\stackrel{(c)}{=} M \cdot D(f,r) + o_M(q^M),
\end{align*}
where part (a) follows from \eqref{equation_theorem_tau_result2} and plugging in $ \sum_{i=2}^{r} Q_*^i \leq (r-1)(K-1)q_0^M$; (b) follows from the induction hypothesis; and (c) follows from the expressions in \eqref{equation_definition_Dfr_NE}. As a result,
\begin{align*}
T^{\dagger} &= \sum_{r=1}^{K-1} \E[\tau_r \mid \mathcal E _r ]
\leq \sum_{r=1}^{K-1} D(f, r) M  + o_M(q^M) = \frac{\log(1/p)M}{I^{\mathrm N}(f)} + o_M(1),
\end{align*}
where the last equality comes from the hardness measure $ I^{\mathrm N}(f)= \log\left(\tfrac 1 p \right) \left[\sum_{r=1}^{K-1} D(f,r)\right]^{-1} $.

\vspace{0.2 cm}
\noindent\textbf{Step 3 (Bounding $T^{\ddagger}$).}  Note that $\E[\tau_1 \cdot \mathbbm{1}\{\mathcal E _1^c\} ] = 0$. Thus, we consider $\E[\tau_r \cdot \mathbbm{1}\{\mathcal E _r^c\} ]$ for arbitrary $r\in [K-1] \setminus \{ 1\}$ in the following.

Conditioned on any fixed realization of previous stages such that $\mathcal E _r$ does not occur, 
$$
\sum_{i=1}^{K-r} W_{t}\left(S_r^i \right)-(K-r)W_{t}\left(S_r^{K-r+1}\right)  - \left( \sum_{i=1}^{K-r} f(S_r^i|S_r)-(K-r)f(S_r^{K-r+1}|S_r) \right) t
$$
is a martingale for $t\ge T_{r-1}$, and hence we have 
\begin{align}
    M &\ge\E\left [ \sum_{i=1}^{K-r} W_{T_r}\left(S_r^i\right)-(K-r)W_{T_r}\left(S_r^{K-r+1}\right) \right] \notag \\
    &=\E\left [ \sum_{i=1}^{K-r} \left(W_{T_r}\left(S_r^i\right) - W_{T_{r-1}}\left(S_r^i\right) \right)-(K-r)\left(W_{T_r}\left(S_r^{K-r+1}\right) -W_{T_{r-1}}\left(S_r^{K-r+1}\right) \right)\right] \notag  \\ 
    &\phantom{=} + \sum_{i=1}^{K-r} W_{T_{r-1}}\left(S_r^i\right)-(K-r)W_{T_{r-1}}\left(S_r^{K-r+1}\right)  \notag \\
    &=  \left( \sum_{i=1}^{K-r} f(S_r^i|S_r)-(K-r)f(S_r^{K-r+1}|S_r) \right) \E[\tau_r]+ \sum_{i=1}^{K-r} W_{T_{r-1}}\left(S_r^i\right)-(K-r)W_{T_{r-1}}\left(S_r^{K-r+1}\right).
\label{equation_theorem_tau_step3_1}
\end{align}

For all $i\in[K-r]$, it holds that 
$$ W_{T_{r-1}}\left(S_r^i\right)-W_{T_{r-1}}\left(S_r^{K-r+1}\right)\ge -M.$$
Otherwise, the algorithm would have already been
terminated. Thus, we have 
$$
\sum_{i=1}^{K-r} W_{T_{r-1}}\left(S_r^i\right)-(K-r)W_{T_{r-1}}\left(S_r^{K-r+1}\right) \ge -(K-r)M.
$$

In addition, due to the definition of $\mathcal M_p$, we have
\begin{align*}
    \sum_{i=1}^{K-r} f(S_r^i|S_r)-(K-r)f(S_r^{K-r+1}|S_r) &= 1 - (K-r+1) f(S_r^{K-r+1}|S_r)  \\
    &\ge 1 - \frac {(K-r+1)(1-p)p^{K-r}} {1-p^{K-r+1}} \\
    &= \frac {(K-r)p^{K-r+1}-(K-r+1)p^{K-r}+1} {1-p^{K-r+1}}.
\end{align*}

Together with \eqref{equation_theorem_tau_step3_1},  we have
\begin{align*}
    \E[\tau_r] &\le \frac {(K-r+1)({1-p^{K-r+1}})M}{(K-r)p^{K-r+1}-(K-r+1)p^{K-r}+1},
\end{align*}
which is conditioned on any fixed realization of previous stages satisfying $\mathcal E _r^c$.

By taking expectation with respect to all realizations of previous stages satisfying $\mathcal E _r^c$, we can get
 \begin{align}
 \label{equation_theorem_tau_step3_2}
    \E[\tau_r\cdot \mathbbm{1}\{\mathcal E _r^c\}] 
    &\le \frac {(K-r+1)({1-p^{K-r+1}})M}{(K-r)p^{K-r+1}-(K-r+1)p^{K-r}+1} \mathbb P(\mathcal E _r^c).
\end{align}

Since $\mathcal E _r^c = \bigcup_{i=1}^{r-1} \left(\mathcal E _{i}\setminus \mathcal E _{i+1}\right) $ and $\mathcal E _{i}\setminus \mathcal E _{i+1}$ for $i \in [r-1]$ are pairwise mutually exclusive events, along with Lemma~\ref{lemma_Error_r}, 
 \begin{align}
 \label{equation_theorem_tau_step3_3}
\mathbb P(\mathcal E _r^c) = \sum_{i=1}^{r-1}\mathbb P(\mathcal E _{i}\setminus \mathcal E _{i+1}) \le \sum_{i=2}^{r} Q_*^i.
\end{align}

Finally, by combining \eqref{equation_theorem_tau_step3_2} and \eqref{equation_theorem_tau_step3_3}, we can bound $T^{\ddagger}$ as
\begin{align*}
    T^{\ddagger} &\ =\  \sum_{r=2}^{K-1} \E[\tau_r \cdot \mathbbm{1}\{\mathcal E _r^c\} ] \ \le\   \sum_{r=2}^{K-1} \sum_{i=2}^{r} \frac {(K-r+1)({1-p^{K-r+1}})MQ_*^i}{(K-r)p^{K-r+1}-(K-r+1)p^{K-r}+1}    .
\end{align*}

Notice that $T^{\ddagger} = o_{M}(1)$, as a result of the definition of $Q_*^i$ for $i \in [r]\setminus \{1\}$ in Lemma~\ref{lemma_Error_r}. Therefore, the proof of Proposition~\ref{proposition_tau_general} is completed, and we have 
\begin{equation}
\label{equation_oM1}
\E [\tau] \le \frac{\log(1/p)M}{I^{\mathrm N}(f)} + \sum_{r=2}^{K-1} \left( \hat D(f, r)  -D(f, r) \right)M  + \sum_{r=2}^{K-1} \sum_{i=2}^{r} \frac {(K-r+1)({1-p^{K-r+1}})MQ_*^i}{(K-r)p^{K-r+1}-(K-r+1)p^{K-r}+1}.
\end{equation}
However, note that the residual term
\begin{align*}
&\sum_{r=2}^{K-1} \left( \hat D(f, r)  -D(f, r) \right)M  + \sum_{r=2}^{K-1} \sum_{i=2}^{r} \frac {(K-r+1)({1-p^{K-r+1}})MQ_*^i}{(K-r)p^{K-r+1}-(K-r+1)p^{K-r}+1} \\
={} &\sum_{r=2}^{K-1} \left( \hat D(f, r)  -D(f, r) \right)M  + \sum_{r=2}^{K-1} \sum_{i=2}^{r} \frac {(K-r+1)({1-p^{K-r+1}})}{(K-r)p^{K-r+1}-(K-r+1)p^{K-r}+1} M\Big[\sum_{\ell=1}^{K-i+1}\left( {q_*^{i,\ell} }\right)^M\Big] \\
={}&  o_M(1).
\end{align*}
That finishes the proof.
\Halmos
\endproof

\vspace{0.3 cm}
\proof{Proof of Lemma~\ref{lemma_Error_r}. }
Fix an arbitrary stage $ r \in [K-1] $ throughout the proof. Recall that the item that is eliminated in stage $r-1$ is denoted as $a_{r-1}$.  
Therefore, we can decompose the  probability of interest as follows:
\begin{align*}
    \mathbb P(\mathcal E _{r-1}\setminus \mathcal E _{r}) &= \mathbb P (\mathcal E _{r-1} \wedge \{a_{r-1} \neq K-r+2\})  
    = \sum_{i=1}^{K-r+1} \mathbb P (\mathcal E _{r-1}\wedge \{a_{r-1} = i\}).
\end{align*}

Next, fix an item $i\in[K-r+1]$. We will analyze $ \mathbb P (\mathcal E _{r-1}\wedge \{a_{r-1} = i\})$, 
which is the probability that \textsc{NE}  eliminates the worst item correctly in each of the first $r-2$ stages and eliminates item $i$ in stage $r-1$. 

\noindent\textbf{Step 1. }  At certain time step $t$ in any stage $\bar r \in [r-1]$, suppose that $\mathcal E _{\bar r}$ holds, which ensures that the current active set is $[K-\bar r + 1]$. For every $\alpha > 0$, we consider the process 
$$
 Y^{r, i}_t(\alpha) := \sum_{j=1}^{K-r+2} W_t(j) - (K-r+2)W_t(i)+\alpha(W_t(K-r+2)-W_t(i)).
$$
We claim that there exist $ \alpha > 0 $ and $0<q<1$ (both potentially dependent on $ f $ but independent of $ M $) such that $\left(  1 / q\right)^{Y^{r, i}_t(\alpha)}$ is a supermartingale, i.e.,
\begin{equation}
\label{equation_condition2}
\E\left [ \left( \frac 1 q\right)^{Y^{r, i}_{t+1}(\alpha)} \,\middle\vert\,  \left( \frac 1 q\right)^{Y^{r, i}_t(\alpha)} \right] \le \left( \frac 1 q\right)^{Y^{r, i}_t(\alpha)}.
\end{equation}

To see why, first pick a sufficiently large $\alpha$ so that
\begin{equation}
\label{equation_condition1}
\E[ Y^{r, i}_{t+1}(\alpha)\mid Y^{r, i}_t(\alpha)]<Y^{r, i}_t(\alpha),
\end{equation}
which is equivalent to
\begin{equation*}
    \sum_{j=1}^{K-r+2} \! f(j|[K-\bar r + 1]) - (K-r+2)f(i\mid[K-\bar r + 1])+\alpha\Big(f(K-r+2|[K-\bar r + 1])\\-f(i|[K-\bar r + 1])\Big) < 0,
\end{equation*}
That is always possible because
$f(K-r+2|[K-\bar r + 1])-f(i|[K-\bar r + 1])<0.$ It suffices to show that as long as \eqref{equation_condition1} holds, there exist $0<q<1$ such that $\left(  1 / q\right)^{Y^{r, i}_t(\alpha)}$ is a supermartingale, i.e., \eqref{equation_condition2} holds. In fact, that is equivalent to
\begin{align}
      \frac 1 q\left( \sum_{j=1}^{K-r+1} f(j|[K-\bar r + 1]) -f(i|[K-\bar r + 1]) \right)  + q^{K-r+1+\alpha} \cdot f(i|[K-\bar r + 1])  \notag \\ + \frac 1 {q^{1+\alpha}} \cdot f(K-r+2|[K-\bar r + 1]) + (1 - f([K-r+2]|[K-\bar r + 1]))\le 1 . \label{equation_condition3}
\end{align}
For ease of reference, let
\begin{align*}
    g\left( q \right) &:= \frac 1 q\left( \sum_{j=1}^{K-r+1} f(j|[K-\bar r + 1]) -f(i|[K-\bar r + 1]) \right)  + q^{K-r+1+\alpha} \cdot f(i|[K-\bar r + 1])   \\ &\phantom{:=} + \frac 1 {q^{1+\alpha}} \cdot f(K-r+2|[K-\bar r + 1]) + (1 - f([K-r+2]|[K-\bar r + 1])).
\end{align*}
Notice that
\begin{align*}
    g\left( 1 \right) &= \left( \sum_{j=1}^{K-r+1} f(j|[K-\bar r + 1]) -f(i|[K-\bar r + 1]) \right)  + f(i|[K-\bar r + 1])   \\ &\phantom{== } + f(K-r+2|[K-\bar r + 1]) + (1 - f([K-r+2]|[K-\bar r + 1])) \\
    &=1 .
\end{align*}
Besides,
\begin{align*}
    &\phantom{=\ } g'(1) \\
    &= -\left( \sum_{j=1}^{K-r+1} f(j|[K-\bar r + 1]) -f(i|[K-\bar r + 1]) \right)  + ({K-r+1+\alpha}) \cdot f(i|[K-\bar r + 1])   \\ &\phantom{==} -({1+\alpha}) \cdot f(K-r+2|[K-\bar r + 1]) \\
    &= -\left( \sum_{j=1}^{K-r+2} f(j|[K-\bar r + 1]) - (K-r+2)f(i|[K-\bar r + 1])+\alpha(f(K-r+2|[K-\bar r + 1]) -f(i|[K-\bar r + 1]))\right) \\
    &> 0.
\end{align*}
Hence, there must exist $0<q<1$ such that \eqref{equation_condition3} holds, which is also independent of $ M $. Now, for every $ r,i $, pick $q_*^{r,i}, \alpha_*^{r,i} \in (0,1)$ -- independent of $ M $ -- so that \eqref{equation_condition1} and \eqref{equation_condition3} hold for all $ \bar r \in[r-1] $.  With such choices of $q_*^{r,i}$ and $\alpha_*^{r,i}$, for any stage $\bar r \in [r-1]$, given that $\mathcal E _{\bar r}$ occurs, $\left(  \frac 1  {q_*^{r,i}}\right)^{Y^{r, i}_t(\alpha_*^{r,i})}$ is a supermartingale until the end of the current stage.

\vspace{0.2 cm}
\noindent\textbf{Step 2. }   Following the above result, we can sequentially apply the optional stopping theorem in the first $r-1$ stages. Since ${Y^{r, i}_{T_0}(\alpha_*^{r,i})} = 0$, we have 
\begin{align*}
    1 &= \E \left[ \left( \frac 1  {q_*^{r,i} }\right)^{Y^{r, i}_{T_0}(\alpha_*^{r,i})} \right]\ge \E \left[ \left( \frac 1  {q_*^{r,i} }\right)^{Y^{r, i}_{T_1}(\alpha_*^{r,i})} \right]  \\
    &= \E \left[ \left( \frac 1  {q_*^{r,i} }\right)^{Y^{r, i}_{T_1}(\alpha_*^{r,i})} \cdot \mathbbm{1}\left\{ \mathcal E _{1} \right\} \right] \ge \E \left[ \left( \frac 1  {q_*^{r,i} }\right)^{Y^{r, i}_{T_1}(\alpha_*^{r,i})} \cdot \mathbbm{1}\left\{ \mathcal E _{2} \right\} \right] \\
     &\ge \E \left[ \left( \frac 1  {q_*^{r,i} }\right)^{Y^{r, i}_{T_2}(\alpha_*^{r,i})} \cdot \mathbbm{1}\left\{ \mathcal E _{2} \right\} \right] \ge \E \left[ \left( \frac 1  {q_*^{r,i} }\right)^{Y^{r, i}_{T_2}(\alpha_*^{r,i})} \cdot \mathbbm{1}\left\{ \mathcal E _{3} \right\} \right] \\
    &\ge \cdots \\
    &\ge \E \left[ \left( \frac 1  {q_*^{r,i} }\right)^{Y^{r, i}_{T_{r-2}}(\alpha_*^{r,i})} \cdot \mathbbm{1}\left\{ \mathcal E _{r-2} \right\} \right] \ge \E \left[ \left( \frac 1  {q_*^{r,i} }\right)^{Y^{r, i}_{T_{r-2}}(\alpha_*^{r,i})} \cdot \mathbbm{1}\left\{ \mathcal E _{r-1} \right\} \right] \\
    &\ge \E \left[ \left( \frac 1  {q_*^{r,i} }\right)^{Y^{r, i}_{T_{r-1}}(\alpha_*^{r,i})} \cdot \mathbbm{1}\left\{ \mathcal E _{r-1} \right\} \right] 
\end{align*}
where we also utilize the fact that  $\mathcal E _{1}\supseteq \mathcal E _{2}\supseteq \cdots\supseteq \mathcal E _{K-1}$ and the nonnegativity of $\left(  \frac 1  {q_*^{r,i}}\right)^{Y^{r, i}_t(\alpha_*^{r,i})}$.

Note that $\mathcal E_{r-1}$ implies $S_{r-1} = [K-r+2]$. Furthermore, if $a_{r-1} = i$, then it must be the case that  $W_{T_{r-1}}(i) \le W_{T_{r-1}}(K-r+2)$ and $\sum_{j=1}^{K-r+2} W_{T_{r-1}}(j) - (K-r+2)W_{T_{r-1}}(i) = M $,
which gives ${Y^{r, i}_{T_{r-1}}(\alpha_*^{r,i})}\ge M$ since $\alpha_*^{r,i}>0$.

Hence, we can get 
\begin{align*}
    1 &\ge \E \left[ \left( \frac 1  {q_*^{r,i} }\right)^{Y^{r, i}_{T_{r-1}}(\alpha_*^{r,i})} \cdot \mathbbm{1}\left\{ \mathcal E _{r-1} \right\} \right] \\ &\ge \E \left[ \left( \frac 1  {q_*^{r,i} }\right)^{Y^{r, i}_{T_{r-1}}(\alpha_*^{r,i})} \cdot \mathbbm{1}\left\{ \mathcal E _{r-1}\wedge \{a_{r-1} = i\} \right\} \right] \ \ge\  \mathbb P (\mathcal E _{r-1}\wedge \{a_{r-1} = i\}) \cdot \left( \frac 1  {q_*^{r,i} }\right)^M,
\end{align*}
which is equivalent to 
\begin{align*}
    \mathbb P (\mathcal E _{r-1}\wedge \{a_{r-1} = i\}) \le \left( {q_*^{r,i} }\right)^M.
\end{align*}
Therefore, 
$
\mathbb P(\mathcal E _{r-1}\setminus \mathcal E _{r}) \le Q_*^r = \sum_{i=1}^{K-r+1}\left( {q_*^{r,i} }\right)^M
$, which finishes the proof.\Halmos
\endproof

\subsection{Proof of Proposition~\ref{proposition_error}}
\label{appendix_proposition_error}
\proof{Proof of Proposition~\ref{proposition_error}. }
Here we also adopt the notation introduced in the proof of Proposition~\ref{proposition_tau_general}. Recall that for any stage $r \in [K-1]$, the active set of size $K-r+1$ is referred to as  $S_r$. Therefore, we are interested in bounding
\begin{align}
 \mathbb P (i_{\mathrm{out}} \neq 1 ) &= \mathbb P (1\not\in S_K)  
= \sum_{r=1}^{K-1 }\mathbb P (\{1\not\in S_{r+1}\}\wedge \{1\in S_r\}). \label{proposition_error_result1}
\end{align}

In the following, we will analyze $\mathbb P (\{1\not\in S_{r+1}\}\wedge \{1\in S_r\})$, which represents the probability of eliminating the best item in the given item set $[K]$ (i.e., item $1$) in stage $r$.

\vspace{0.2 cm}
\noindent\textbf{Step 1. } For any stage $r\in[K-1]$, condition on any fixed realization of previous stages such that the best item is not eliminated prior to stage $r$, i.e., $ 1\in S_r$. 

Then we claim that for this fixed realization of $S_r = \{ S_r^1, S_r^2, \ldots, S_r^{K-r+1}\}$ with $S_r^1 = 1$, 
$$
{\left(\frac 1 p\right)}^{\sum_{i=2}^{K-r+1}W_{t}(S_r^{i}) -(K-r) W_{t}(S_r^1)}
$$
is a supermartingale for $ t \ge T_{r-1}$.  To verify this favorable property, it suffices to show 
\begin{align*}
    \frac 1 p  \sum_{i=2}^{K-r+1}f(S_r^{i}|S_r) + p^{K-r} f(S_r^{1}|S_r) \le 1.
\end{align*}
Actually, the above inequality is equivalent to 
\begin{align*}
    f(S_r^{1}|S_r)\ge \frac {1-p} {1-p^{K-r+1}},
\end{align*}
which holds trivially due to the definition of $\mathcal M_p$. Therefore, by the optional stopping theorem, it holds that
\begin{align*}
    &\phantom{\ge\ } {\left(\frac 1 p\right)}^{\sum_{i=2}^{K-r+1}W_{T_{r-1}}(S_r^{i}) -(K-r) W_{T_{r-1}}(S_r^1)}\\
    &\ge \E \left[ {\left(\frac 1 p\right)}^{\sum_{i=2}^{K-r+1}W_{T_{r}}(S_r^{i}) -(K-r) W_{T_{r}}(S_r^1)}  \right] \\
    &\ge \mathbb P (1\not \in S_{r+1}) \cdot {\left(\frac 1 p\right)}^M
\end{align*}
where the last inequality follows from the fact that if $1\in S_r$ but $1\not \in S_{r+1}$, then  $${\sum_{i=2}^{K-r+1}W_{T_{r}}(S_r^{i}) -(K-r) W_{T_{r}}(S_r^1)}=M.$$

Again by taking expectation with respect to all realizations of previous stages satisfying $1 \in S_r $, we can derive
\begin{align}
\mathbb P  (\{1\not\in S_{r+1}\}\wedge \{1\in S_r\}) \le  \E \left[ {\left(\frac 1 p\right)}^{\sum_{i=2}^{K-r+1}W_{T_{r-1}}(S_r^{i}) -(K-r) W_{T_{r-1}}(S_r^1)}  \cdot \mathbbm{1}\{ 1\in S_r\} \right]  \cdot p^M. \label{equation_error1}
\end{align}

\vspace{0.2 cm}
\noindent\textbf{Step 2.} Consider any stage $r\in[K-1]$. For ease of presentation, we define $\hat S_r := S_r \setminus \{S_r^1 \} =\{ S_r^2, S_r^3, \ldots, S_r^{K-r+1}\}$. Notice that $\hat S_r$ is also a random variable. In addition, we use lower-case
$\hat s_r$ and $\{ s_r^i\}_{i=2}^{K-r+1}$ to denote the indicators of the specific realizations of $\hat S_r$ and $\{ S_r^i\}_{i=2}^{K-r+1}$, respectively.  Then we have 
\begin{align}
    &\phantom{\ge\ } \E \left[ {\left(\frac 1 p\right)}^{\sum_{i=2}^{K-r+1}W_{T_{r-1}}(S_r^{i}) -(K-r) W_{T_{r-1}}(S_r^1)}  \cdot \mathbbm{1}\{ 1\in S_r\} \right] \notag \\
    &= \sum_{\hat s_r \subseteq [K]\setminus \{1\}} \E \left[ {\left(\frac 1 p\right)}^{\sum_{i=2}^{K-r+1}W_{T_{r-1}}(s_r^{i}) -(K-r) W_{T_{r-1}}(1)}  \cdot \mathbbm{1}\left\{ \{1\}\cup \hat s_r = S_r \right\} \right] \notag  \\
    &= \sum_{\hat s_r \subseteq [K]\setminus \{1\}} \E \left[ {\left(\frac 1 p\right)}^{\sum_{i=2}^{K-r+1}W_{T_{r-1}}(s_r^{i}) -(K-r) W_{T_{r-1}}(1)}  \cdot \mathbbm{1}\left\{ \{1\}\cup \hat s_r \subseteq S_r \right\} \right]. \label{equation_error2}
\end{align}
Next, for any ${\hat s_r \subseteq [K]\setminus \{1\}}$, we will show via mathematical induction that 
\begin{align}
\label{equation_error3}
\E \left[ {\left(\frac 1 p\right)}^{\sum_{i=2}^{K-r+1}W_{T_{r-1}}(s_r^{i}) -(K-r) W_{T_{r-1}}(1)}  \cdot \mathbbm{1}\left\{ \{1\}\cup \hat s_r \subseteq S_r \right\} \right] \le 1.
\end{align}
Consider the first stage, where the active set $S_1$ is $ [K]$. Since 
$$
{\left(\frac 1 p\right)}^{\sum_{i=2}^{K-r+1}W_{t}(s_r^{i}) -(K-r) W_{t}(1)}
$$
is a supermartingale in the first stage, by the optional stopping theorem, it holds that 
\begin{align*}
    1 &= \E \left[ {\left(\frac 1 p\right)}^{\sum_{i=2}^{K-r+1}W_{T_0}(s_r^{i}) -(K-r) W_{T_0}(1)}  \right]  \\
    &\ge \E \left[ {\left(\frac 1 p\right)}^{\sum_{i=2}^{K-r+1}W_{T_1}(s_r^{i}) -(K-r) W_{T_1}(1)}   \right] \\
    &\ge \E \left[ {\left(\frac 1 p\right)}^{\sum_{i=2}^{K-r+1}W_{T_1}(s_r^{i}) -(K-r) W_{T_1}(1)}  \cdot \mathbbm{1}\left\{ \{1\}\cup \hat s_r \subseteq S_2 \right\} \right].
\end{align*}

Suppose that for all $j $ in $\{2,3,\dots,\bar r-1\}$, 
\begin{align*}
\E \left[ {\left(\frac 1 p\right)}^{\sum_{i=2}^{K-r+1}W_{T_{j -1}}(s_r^{i}) -(K-r) W_{T_{j -1}}(1)}  \cdot \mathbbm{1}\left\{ \{1\}\cup \hat s_r \subseteq S_j  \right\} \right] \le 1
\end{align*}
is correct. Then consider the ${\bar r} $-th stage. Since for any fixed realization of $S_{\bar r} $ such that $\{1\}\cup \hat s_r \subseteq S_{\bar r} $, 
$$
{\left(\frac 1 p\right)}^{\sum_{i=2}^{K-r+1}W_{t}(s_r^{i}) -(K-r) W_{t}(1)}
$$
is a supermartingale for $t \ge T_{{\bar r} -1}$, again by the optional stopping theorem, it holds that 
\begin{align*}
    1 &\ge \E \left[ {\left(\frac 1 p\right)}^{\sum_{i=2}^{K-r+1}W_{T_{{\bar r} -1}}(s_r^{i}) -(K-r) W_{T_{{\bar r} -1}}(1)}  \cdot \mathbbm{1}\left\{ \{1\}\cup \hat s_r \subseteq S_{\bar r}  \right\} \right] \\
    &\ge \E \left[ {\left(\frac 1 p\right)}^{\sum_{i=2}^{K-r+1}W_{T_{{\bar r} }}(s_r^{i}) -(K-r) W_{T_{{\bar r} }}(1)}  \cdot \mathbbm{1}\left\{ \{1\}\cup \hat s_r \subseteq S_{\bar r}  \right\} \right] \\
    &\ge \E \left[ {\left(\frac 1 p\right)}^{\sum_{i=2}^{K-r+1}W_{T_{{\bar r} }}(s_r^{i}) -(K-r) W_{T_{{\bar r} }}(1)}  \cdot \mathbbm{1}\left\{ \{1\}\cup \hat s_r \subseteq S_{\bar r+1}  \right\} \right]
\end{align*}
which establishes the induction step and further proves \eqref{equation_error3}.  Note that the cardinality of $\hat s_r$ is $K-r$.
Therefore, combining \eqref{equation_error2} and \eqref{equation_error3} gives 
\begin{align*}
    \E \left[ {\left(\frac 1 p\right)}^{\sum_{i=2}^{K-r+1}W_{T_{r-1}}(S_r^{i}) -(K-r) W_{T_{r-1}}(S_r^1)}  \cdot \mathbbm{1}\{ 1\in S_r\} \right] \le \binom{K-1} {K-r}.
\end{align*}
Together with \eqref{equation_error1},  we can get
\begin{align}
\label{proposition_error_result2}
    \mathbb P  (\{1\not\in S_{r+1}\}\wedge \{1\in S_r\}) \le \binom{K-1} {K-r}p^M.
\end{align}

\vspace{0.2 cm}
\noindent\textbf{Step 3. } Finally, by plugging \eqref{proposition_error_result2} into \eqref{proposition_error_result1}, we have 
\begin{align*}
 \mathbb P (i_{\mathrm{out}} \neq 1 )
&\ =\  \sum_{r=1}^{K-1 }\mathbb P (\{1\not\in S_{r+1}\}\wedge \{1\in S_r\}) \ \le\   \sum_{r=1}^{K-1 }\binom {K-1} {K-r}p^M \ =\  (2^{K-1}-1)p^M
\end{align*}
which completes the proof of Proposition~\ref{proposition_error}.\Halmos
\endproof

\subsection{Proofs for Section~\ref{sec:understanding-NE-beyond-OA}}\label{sec:understanding IN}

\tbl{
\proof{Proof of Lemma~\ref{lma:MNL-small-top-gap}.}
Invoking \eqref{equation_prop_OA_1} for a general $ p $-separable preference $ f $, 
$D(f, r) = \frac {1 - \sum_{i=1}^{r-1}  \left( f([K-r]|[K-i+1])-(K-r)f(K-r+1|[K-i+1])\right)D(f,i)} {1-(K-r+1)f(K-r+1|[K-r+1])}$ for all $ r \in [K-1]\setminus \{1\} $. Substituting $ r = K-1 $, we have
\begin{align*}
D(f, K-1) = \frac {1 - \sum_{i=1}^{K-2}  \left( f(1|[K-i+1])-f(2|[K-i+1])\right)D(f,i)} {1-2f(2|[2])}.
\end{align*}
Now, suppose 
$f \in \mathcal{M}_p^\mathrm{MNL}$ with $v_1 = 1$ and $v_2 = p$. By arranging terms, we have
\begin{align*}
1 = &\big( 1-2f(2|[2]) \big) D(f, K-1) + \sum_{i=1}^{K-2}  \left( f(1|[K-i+1])-f(2|[K-i+1])\right)D(f,i) \\
= &\big( 1-2 \cdot \tfrac{p}{1+p} \big) D(f, K-1) + \sum_{i=1}^{K-2}  \left( \tfrac{1-p}{1+p+v_3 + \cdots + v_{K-i+1}}\right)D(f,i)  \\
\leq  &\tfrac{1-p}{1+p}  D(f, K-1) + \sum_{i=1}^{K-2}  \tfrac{1-p}{1+p}D(f,i)  = \tfrac{1-p}{1+p}  \sum_{i=1}^{K-1} D(f,i).
\end{align*}
Therefore, $ \sum_{i=1}^{K-1} D(f,i) \geq \tfrac{1+p}{1-p} $. As a result, $ I^{\mathrm N}(f):= \log\left(\tfrac 1 p \right) \left[\sum\nolimits_{r=1}^{K-1} D(f,r)\right]^{-1} \leq \log\left(\tfrac{1}{p}\right) \cdot \frac{1 - p}{1 + p}  $.\Halmos 
\endproof	
}

\vskip 0.2 cm

\tbl{
\proof{Proof of Lemma~\ref{lma:dominating-item}.} Since the items are relabeled according to the ground-truth ranking and $ \IN(f) $ only depends on $ f $ through the nested collection of display sets $ \mathcal{N} $, it holds that  $ \frac{f(1\mid S)}{f(i\mid S)} \ge R $ for every item $i>1$ and  display set $ S \in \mathcal{N} $. Therefore, $ f(i\mid S) \le \frac{f(1\mid S)}{R} \le \frac{1}{R}. $ In particular, for every admissible pair $(r,i)$, \[ \Delta_r^i = f(K-r+1\mid [K-i+1]) - f(K-r+2\mid [K-i+1]) \le f(K-r+1\mid [K-i+1]) \le \tfrac{1}{R}. \] Similarly, $ f(K-r+1\mid [K-r+1])\le \tfrac{1}{R}. $ 

Define the partial sum $ S_r:=\sum_{k=1}^r D(f,k),$ for $ r=1,\ldots,K-1 $. Let us first bound $S_1$.  Note that, under the full display set $[K]$, $ f(\ell\mid [K])\ge f(K\mid [K]), $ for every $ \ell=2,\ldots,K. $ Hence 
\begin{align*} f(K\mid [K]) &= \tfrac{f(K\mid [K])} {\sum_{\ell=1}^K f(\ell\mid [K])} \le \tfrac{f(K\mid [K])} {f(1\mid [K])+(K-1)f(K\mid [K])} = \tfrac{1} {f(1\mid [K])/f(K\mid [K])+(K-1)} \le \tfrac{1}{R+K-1}. 
\end{align*} 
Therefore, $ S_1 = D(f,1) = \frac{1}{1-Kf(K\mid [K])} \le \frac{1}{1-\frac{K}{R+K-1}} = \frac{R+K-1}{R-1}. $ 

Next, consider any $r=2,\ldots,K-1$. By the recursive definition of $D(f,r)$, 
\begin{align*} D(f,r) &= \tfrac{(K-r+1)\sum_{i=1}^{r-1}\Delta_r^i D(f,i)} {1-(K-r+1)f(K-r+1\mid [K-r+1])}\le \tfrac{(K-r+1)S_{r-1}/R} {1-(K-r+1)/R}. 
\end{align*} 
Since $R\ge K$, the denominator is positive, and thus $ D(f,r) \le \frac{(K-r+1)S_{r-1}}{R-K+r-1}. $ It follows that 
\begin{align*} 
S_r &= S_{r-1}+D(f,r)\le S_{r-1} \left( 1+\tfrac{K-r+1}{R-K+r-1} \right) = S_{r-1} \tfrac{R}{R-K+r-1}. 
\end{align*} 
Iterating this inequality from $r=2$ to $r=K-1$ yields 
\begin{align*} S_{K-1} &\le \tfrac{R+K-1}{R-1} \prod_{r=2}^{K-1} \tfrac{R}{R-K+r-1} = \tfrac{R+K-1}{R-1} \prod_{m=2}^{K-1} \tfrac{R}{R-m}. \end{align*} 
Recall that $ \IN(f) = \log\left(\frac{1}{p}\right) \left[ \sum_{k=1}^{K-1}D(f,k) \right]^{-1} $. We thus obtain \[ \IN(f) \ge \log\left(\frac{1}{p}\right) \frac{R-1}{R+K-1} \prod_{m=2}^{K-1} \left(1-\frac{m}{R}\right). \] 

Finally, note that $ \frac{R-1}{R+K-1} \prod_{m=2}^{K-1} \left(1-\frac{m}{R}\right) \rightarrow 1 $ as $ R$ grows. Recall from \eqref{eq:approximating IOA} that $ I_*^{\mathrm{OA}} \leq \frac{1-p}{1+p} \log\left(\frac{1}{p}\right) $. Therefore,  for every $\varepsilon>0$ there exists a sufficiently large threshold on $R$ such that \[ \IN(f) \ge (1-\varepsilon)\log\left(\tfrac{1}{p}\right) \ge \tfrac{(1+p)(1-\varepsilon)}{1-p} \,I_*^{\mathrm{OA}}. 
\] This proves the claim. \Halmos 
\endproof

}

\subsection{Proof of Proposition~\ref{prop_OA}}
\label{appendix_prop_OA}

Let us provide a roadmap of the proof. First, we demonstrate in Lemma~\ref{lemma_prop_OA_1} that the class $\mathcal M_p^{\mathrm{OA}}$ minimizes $\{I^{\mathrm N}(f): {f\in \mathcal M_p}\}$. Next in Lemma~\ref{lemma_prop_OA_2}, we prove that for any preference $f\in \mathcal M_p^{\mathrm{OA}}$, $I^{\mathrm N}(f) = I_*^{\mathrm{OA}}$.
Finally, Proposition~\ref{prop_OA} follows directly from Lemma~\ref{lemma_prop_OA_1} and Lemma~\ref{lemma_prop_OA_2}.

\begin{lemma}
\label{lemma_prop_OA_1}
It holds that
$
\mathcal M_p^{\mathrm{OA}} \subseteq \argmin _{f\in \mathcal M_p} {I^{\mathrm N}(f)}.
$
\end{lemma}

\begin{lemma}
\label{lemma_prop_OA_2}
For any preference $f\in \mathcal M_p^{\mathrm{OA}}$, it holds that
$
    I^{\mathrm N}(f) =   {I_*^{\mathrm{OA}}} .
$
\end{lemma}

\subsubsection{Proof of Lemma~\ref{lemma_prop_OA_1}}
 \proof{Proof of Lemma~\ref{lemma_prop_OA_1}. }
We will prove the desired result in the following steps.

\noindent\textbf{Step 1. } Recall that for any general preference $f\in\mathcal M_p$ and $r\in [K-1] \setminus \{ 1\}$, we define $D(f, r)$ in~\eqref{equation_definition_Dfr_NE}. In the following, we will prove via induction that for all $r\in [K-1] \setminus \{ 1\}$, it holds that
\begin{align}
\label{equation_prop_OA_induction}
\sum_{i=1}^{r-1}  (f([K-r+1]|[K-i+1])-(K-r+1)f(K-r+2|[K-i+1]))D(f,i)= 1.
\end{align}

We only need to consider $K\ge 3$ as the case that $K =2$ is vacuous. For $r=2$, the claim of \eqref{equation_prop_OA_induction} is equivalent to 
\begin{align*}
    \big(f([K-1]|[K])-(K-1)f(K|[K])\big)D(f,1)= 1,
\end{align*}
which holds trivially due to the definition of  $D(f,1)$. Now suppose that \eqref{equation_prop_OA_induction} is true for $r= \bar r$ with  $2\le \bar r \le K-2$. Then we can derive
\begin{align*}
    &\phantom{=\ } \sum_{i=1}^{\bar r }  (f([K-\bar r]|[K-i+1])-(K-\bar r)f(K-\bar r+1|[K-i+1]))D(f,i) - 1 \\
    &= \sum_{i=1}^{\bar r }  (f([K-\bar r]|[K-i+1])-(K-\bar r)f(K-\bar r+1|[K-i+1]))D(f,i)  \\
    &\phantom{=\ } - \sum_{i=1}^{\bar r-1}  (f([K-\bar r+1]|[K-i+1])-(K-\bar r+1)f(K-\bar r+2|[K-i+1]))D(f,i) \\
    &= \sum_{i=1}^{\bar r-1} (K-\bar r+1) (f(K-\bar r+2|[K-i+1]))-f(K-\bar r+1|[K-i+1])))D(f,i) \\
    &\phantom{=\ } + (f([K-\bar r]|[K-\bar r +1])-(K-\bar r)f(K-\bar r+1|[K-\bar r +1]))D(f,\bar r) \\
    &= \sum_{i=1}^{\bar r-1} (K-\bar r+1) (f(K-\bar r+2|[K-i+1]))-f(K-\bar r+1|[K-i+1])))D(f,i) \\
    &\phantom{=\ } + (1-(K-\bar r+1)f(K-\bar r+1|[K-\bar r +1]))D(f,\bar r) = 0,
\end{align*}
where the last equality results from the definition of  $D(f,\bar r) $. Therefore, \eqref{equation_prop_OA_induction} is also true for  $r= \bar r +1$ and  the induction step is completed. By mathematical induction, we can conclude that 
our claim \eqref{equation_prop_OA_induction} holds for all $r\in [K-1] \setminus \{ 1\}$.

Next, combining \eqref{equation_definition_Dfr_NE} and \eqref{equation_prop_OA_induction} results in another helpful expression of $D(f, r)$, i.e., 
\begin{align}
\label{equation_prop_OA_1}
D(f, r) = \frac {1 - \sum_{i=1}^{r-1}  \left( f([K-r]|[K-i+1])-(K-r)f(K-r+1|[K-i+1])\right)D(f,i)} {1-(K-r+1)f(K-r+1|[K-r+1])}
,
\end{align}
for all $r\in [K-1] \setminus \{ 1\}$.

\vspace{0.2 cm}
\noindent\textbf{Step 2. } For any general preference $f\in\mathcal M_p$, due to the definition of $D(f, 1)$ and Lemma~\ref{lemma_fOA}, we have
\begin{align*}
    D(f, 1) &= \frac 1 {1- Kf(K|[K])} = \frac 1 {f([K-1]|[K])- (K-1)f(K|[K])} \le   \frac {1-p^K} {(K-1)p^K-Kp^{K-1}+1},
\end{align*}
which further gives 
\begin{align}
\label{equation_prop_OA_add_1}
\frac {(1-p)p^{K-1}} {1-p^{K}}D(f, 1) \le  \frac {(1-p)p^{K-1}} {(K-1)p^K-Kp^{K-1}+1}.
\end{align}
For any $r\in [K-1] \setminus \{ 1\}$, again by Lemma~\ref{lemma_fOA}, we have
$$
f([K-r]|[K-i+1])-(K-r)f(K-r+1|[K-i+1]) \ge \frac {(K-r)p^{K-r+1}-(K-r+1)p^{K-r}+1} {1-p^{K-i+1}}
$$
for any  $i\in[r-1]$, and 
$$
{1-(K-r+1)f(K-r+1|[K-r+1])} \ge  \frac {({K-r})p^{K-r+1}-{(K-r+1)}p^{{K-r}}+1} {1-p^{K-r+1}} .
$$
Plug the above two inequalities into the expression of $D(f, r)$ in \eqref{equation_prop_OA_1}, then we have 
\begin{align}
\label{equation_prop_OA_add_20}
D(f, r) +  \sum_{i=1}^{r-1}  \frac {1-p^{K-r+1}}  {1-p^{K-i+1}} D(f,i)\le  \frac {1-p^{K-r+1}} {({K-r})p^{K-r+1}-{(K-r+1)}p^{{K-r}}+1} .
\end{align}
Multiplying \eqref{equation_prop_OA_add_20} by different coefficients for all $r \in [K-1]\setminus \{1\}$, we can get
\begin{equation}
\label{equation_prop_OA_add_2}
\frac {(1-p)p^{K-r}} {1-p^{K-r+1}} D(f, r) +\sum_{i=1}^{r-1} \frac {(1-p)p^{K-r}}  {1-p^{K-i+1}} D(f, i)  \le \frac  {(1-p)p^{K-r}} {({K-r})p^{K-r+1}-{(K-r+1)}p^{{K-r}}+1} 
\end{equation}
for all $r\in [K-2]\setminus \{1\}$, and
\begin{equation}
\label{equation_prop_OA_add_3}
D(f, K-1 )+\sum_{i=1}^{K-2} \frac {1-p^{2}} {1-p^{K-i+1}} D(f, i) \le\frac  {1+p}{1-p}.
\end{equation}

\vspace{0.2 cm}
\noindent\textbf{Step 3. } Finally, adding up \eqref{equation_prop_OA_add_1}, \eqref{equation_prop_OA_add_2} and \eqref{equation_prop_OA_add_3} leads to 
\begin{align*}
    & \phantom{\le{}}\sum_{r=1}^{K-2}\frac {(1-p)p^{K-r}} {1-p^{K-r+1}} D(f, r) +\sum_{r=2}^{K-2} \sum_{i=1}^{r-1} \frac {(1-p)p^{K-r}}  {1-p^{K-i+1}} D(f, i)  + D(f, K-1 )+\sum_{i=1}^{K-2} \frac {1-p^{2}} {1-p^{K-i+1}} D(f, i)\\
    &\le  \frac {(1-p)p^{K-1}} {(K-1)p^K-Kp^{K-1}+1} + \sum_{r=2}^{K-2}\frac  {(1-p)p^{K-r}} {({K-r})p^{K-r+1}-{(K-r+1)}p^{{K-r}}+1} +\frac  {1+p}{1-p} \\
    \Longleftrightarrow &\phantom{\le{}} \sum_{r=1}^{K-2}\frac {(1-p)p^{K-r}} {1-p^{K-r+1}} D(f, r) +\sum_{r=1}^{K-3} \sum_{i=r+1}^{K-2} \frac {(1-p)p^{K-i}}  {1-p^{K-r+1}} D(f, r)  +\sum_{r=1}^{K-2} \frac {1-p^{2}} {1-p^{K-r+1}} D(f, r)+ D(f, K-1 )\\
    &\le  \sum_{r=1}^{K-2}\frac  {(1-p)p^{K-r}} {({K-r})p^{K-r+1}-{(K-r+1)}p^{{K-r}}+1} +\frac  {1+p}{1-p} \\
    \Longleftrightarrow &\phantom{\le{}} \sum_{r=1}^{K-2}\frac {p^{K-r}-p^{K-r+1}} {1-p^{K-r+1}} D(f, r) +\sum_{r=1}^{K-3}  \frac {p^2-p^{K-r}}  {1-p^{K-r+1}} D(f, r)  +\sum_{r=1}^{K-2} \frac {1-p^{2}} {1-p^{K-r+1}} D(f, r)+ D(f, K-1 )\\
    &\le  \sum_{r=1}^{K-2}\frac  {(1-p)p^{K-r}} {({K-r})p^{K-r+1}-{(K-r+1)}p^{{K-r}}+1} +\frac  {1+p}{1-p} \\
    \Longleftrightarrow &\phantom{\le{}} \sum_{r=1}^{K-1}D(f,r) \le  \sum_{r=1}^{K-2}\frac  {(1-p)p^{K-r}} {({K-r})p^{K-r+1}-{(K-r+1)}p^{{K-r}}+1} +\frac  {1+p}{1-p}.
\end{align*}

Therefore, we obtain that for any general preference $f\in\mathcal M_p$, 
\begin{align}
    \frac {\log(1/p)} {\IN(f)}= \sum_{r=1}^{K-1} D(f,r )\le  \sum_{r=1}^{K-2}\frac  {(1-p)p^{K-r}} {({K-r})p^{K-r+1}-{(K-r+1)}p^{{K-r}}+1} +\frac  {1+p}{1-p}.
    \label{equation_prop_OA_result}
\end{align}

Note that the above upper bound of $\frac {\log(1/p)} {\IN(f)}$ does not depend on the particular choice of $f$. Furthermore, in view of Lemma~\ref{lemma_fOA}, exact equality in 
\eqref{equation_prop_OA_result} can be achieved if $f\in \mathcal M_p^{\mathrm{OA}}$. As a result, we conclude that 
$
\mathcal M_p^{\mathrm{OA}} \subseteq \argmin _{f\in\mathcal  M_p} {\IN(f)}.
$
as desired.\Halmos
\endproof

\begin{lemma}
\label{lemma_fOA}
For any \(r\in\{2,\ldots,K\}\) and \(i\in[r-1]\),
\begin{equation}
\label{equation_lemma_fOA_problem}
\min_{f \in \mathcal M_p} f([i]|[r])-if(i+1|[r]) = \frac {ip^{i+1}-(i+1)p^i+1} {1-p^r}.
\end{equation}

Furthermore, the minimum is attained if and only if
$
f(j|[r]) = \frac{1-p}{1-p^r}p^{j-1}
$
for all $j\in[r]$.

\end{lemma}

\proof{Proof of Lemma~\ref{lemma_fOA}. }
Notice that only the preference on $[r]$ (i.e., $f(j|[r])$ for $j\in[r]$) matters in terms of the minimization problem~\eqref{equation_lemma_fOA_problem}. For ease of notation, we denote $x_j :=f(j|[r])$ for all $j\in[r]$. Then the problem \eqref{equation_lemma_fOA_problem} of interest can be reformulated as the following optimization problem:
\begin{equation*}
\begin{aligned}
    & \min_x\, \sum_{j=1}^i x_j -ix_{i+1} \\
    &\text{ s.t. } \, px_j -x_{j+1}\ge 0, \,\forall j\in[r-1] \\
    &\phantom{\text{ s.t. }}\, \sum_{j=1}^rx_j = 1\\
    &\phantom{\text{ s.t. }}\, x_j\ge 0,\, \forall j\in[r] .
\end{aligned}
\end{equation*}

We let $h(x) := \sum_{j=1}^i x_j -ix_{i+1}$. Due to the constraints on $x$, for all $1\le j_1< j_2\le r-1$, it holds that 
\begin{equation}
    x_{j_1} \ge p^{j_1-j_2} x _{j_2},
    \label{equation_lemma_fOA_0}
\end{equation}
where the exact equality is achieved if and only if $px_j=x_{j+1}$ for all $j_1\le j<j_2$. Therefore, by \eqref{equation_lemma_fOA_0}, we can get 
\begin{align*}
\sum_{j=1}^i x_j \ge \left( \sum_{j=1}^i p^{-j}\right) x_{i+1},
\end{align*}
which is equivalent to 
\begin{align*}
-ix_{i+1} \ge \frac {-ip^i(1-p)} {1-p^i} \sum_{j=1}^i x_j .
\end{align*}
Then we can bound $h(x)$ as follows:
\begin{align}
   h(x) &\ge \sum_{j=1}^i x_j - \frac {ip^i(1-p)} {1-p^i} \sum_{j=1}^i x_j =  \frac {ip^{i+1}-(i+1)p^i+1} {1-p^i} \sum_{j=1}^i x_j. \label{equation_lemma_fOA_result1}
\end{align}
For any $j\in[i]$, again by \eqref{equation_lemma_fOA_0}, we have 
\begin{align}
    x_j &= \frac{1-p^i} {1-p^r} x_j+  \frac{p^i-p^r} {1-p^r} x_j\notag \\
    &\ge \frac{1-p^i} {1-p^r} x_j+  \frac{p^i-p^r} {1-p^r}  \cdot \frac 1 {\sum_{j'=i+1}^r p^{j'-j}} \sum_{j'=i+1}^r x_{j'} \notag \\
    &= \frac{1-p^i} {1-p^r} x_j+  \frac{p^{j-1}(1-p)} {1-p^r}   \sum_{j'=i+1}^r x_{j'}. \label{equation_lemma_fOA_result2}
\end{align}
Adding up \eqref{equation_lemma_fOA_result2} for all $j\in[i]$ implies that
\begin{align*}
    \sum_{j=1}^i x_j &\ge \sum_{j=1}^i\left(\frac{1-p^i} {1-p^r} x_j+  \frac{p^{j-1}(1-p)} {1-p^r}   \sum_{j'=i+1}^r x_{j'} \right) \\
    &= \frac{1-p^i} {1-p^r}\sum_{j=1}^ix_j+ \sum_{j'=i+1}^r x_{j'} \sum_{j=1}^i \frac{p^{j-1}(1-p)} {1-p^r} \\
    &=  \frac{1-p^i} {1-p^r}\sum_{j=1}^ix_j+ \frac{1-p^i} {1-p^r}\sum_{j'=i+1}^r x_{j'} = \frac{1-p^i} {1-p^r}\sum_{j=1}^r x_j.
\end{align*}
Together with \eqref{equation_lemma_fOA_result1}, we conclude that 
\begin{align}
h(x)\ge \frac {ip^{i+1}-(i+1)p^i+1} {1-p^r}.
\label{equation_lemma_fOA_result3}
\end{align}
It is straightforward to check the lower bound in \eqref{equation_lemma_fOA_result3} can be binding if and only if 
$$
x_j = \frac{1-p}{1-p^r}p^{j-1} 
$$
for all $j\in[r]$. Thus, the proof of Lemma~\ref{lemma_fOA} is finished.\Halmos
\endproof

\subsubsection{Proof of Lemma~\ref{lemma_prop_OA_2}}

\proof{Proof of Lemma~\ref{lemma_prop_OA_2}. }
It suffices to show for any preference  $f\in\mathcal M_p^{\mathrm{OA}}$, 
\begin{align}
\label{equation_corollary_oa}
    \frac{\log(1/p)}{\IN(f)}  = \frac{\log(1/p)}{I_{*}^{\mathrm{OA}}},
\end{align}
where 
$$
{I_{*}^{\mathrm{OA}}} = (1-p)\log \left( \frac 1 p \right)\left( 1 + \sum_{j=2}^{K} \frac{p^{j-1}} {1+2p+\cdots+(j-1)p^{j-2}} \right)^{-1}
$$
as shown in \citet{feng2022robust}. In fact, according to \eqref{equation_prop_OA_result} in the proof of Lemma~\ref{lemma_prop_OA_1}, we have 
\begin{align*}
    \frac {\log(1/p)} {\IN(f)}=  \sum_{r=1}^{K-2}\frac  {(1-p)p^{K-r}} {({K-r})p^{K-r+1}-{(K-r+1)}p^{{K-r}}+1} +\frac  {1+p}{1-p}
\end{align*}
for any preference  $f\in\mathcal M_p^{\mathrm{OA}}$. For the right-hand side of \eqref{equation_corollary_oa}, by the combinatorial identity
$$
1+2p+\cdots+(j-1)p^{j-2} = \frac{(j-1) p^{j}-j p^{j-1}+1}{(1-p)^{2}},
$$
it holds that
\begin{align*}
    \frac{\log(1/p)}{I_{*}^{\mathrm{OA}}} &= \frac 1 {1-p} \left( 1 + \sum_{j=2}^{K} \frac{p^{j-1}} {1+2p+\cdots+(j-1)p^{j-2}} \right) \\
    &= \frac 1 {1-p} \left( 1 + \sum_{j=2}^{K} \frac{(1-p)^2 p^{j-1}} {(j-1) p^{j}-j p^{j-1}+1} \right) \\
    &= \frac 1 {1-p}  + \frac {(1-p)p} {p^2-2p+1} + \sum_{j=3}^{K}  \frac{(1-p) p^{j-1}} {(j-1) p^{j}-j p^{j-1}+1}  \\
    &= \frac  {1+p}{1-p} + \sum_{r=1}^{K-2} \frac{(1-p)p^{K-r}} {(K-r)p^{K-r+1}-(K-r+1)p^{K-r}+1}  = \frac{\log(1/p)}{\IN(f)}
\end{align*}
which leads to the desired result of \eqref{equation_corollary_oa}. Therefore, Lemma~\ref{lemma_prop_OA_2} is proved.\Halmos
\endproof

\section{Interpretation from the SPRT Perspective}
\label{section_Interpretation}
In this appendix, we provide a comprehensive explanation of how the elimination criterion of \textsc{NE} and the partition criterion of \textsc{NP} are developed, from the perspective of 
sequential probability ratio tests (SPRT). We consider any time step $t$ and assume that the underlying preference $f$ belongs to $\mathcal M_p^{\mathrm{OA}}$. For ease of notation, we assume that the active item set is $[K]$ and relabel the $K$ items in descending order based on their voting scores (within this appendix only). Specifically, after relabeling, we have $W_t(1)\ge W_t(2)\ge\cdots\ge W_t(K)$. In what follows, we demonstrate that the maximum likelihood estimation (MLE) of the unknown global ranking $\sigma_f$ is exactly the identity ranking $(1,2,\ldots,K)$, and our  criteria can be derived from the associated generalized likelihood ratios.

\subsection{Maximum Likelihood Estimation} 
\label{subsection_MLE}
Consider a fixed history of display sets and customers' choices up to and including time $t$ (i.e., $S_1,X_1,\ldots,S_t,X_t$). For any hypothesis preference  $\hat f \in \mathcal M_p^{\mathrm{OA}}$, its log-likelihood function is referred to as $$\mathcal{L}(\hat f):= \sum_{\ell=1}^t \log \hat f(X_\ell|S_\ell).$$
According to \citet{feng2022robust}, there exists a constant $\phi$ such that 
$$
\mathcal{L}(\hat f) =  \phi +\log(p)\left(\sum_{(i, j): i \neq j} \mathbbm{1} \left\{\sigma_{\hat f}(j)<\sigma_{\hat f}(i)\right\} w_{i j}^t  \right) 
$$
where
$
w_{i j}^t := \sum_{\ell=1}^t \mathbbm{1}\left\{\{i, j\} \subseteq S_{\ell} \text { and } X_{\ell}=i\right\}
$
for all distinct $i, j \in[K]$.

Note that every preference in $\mathcal M_p^{\mathrm{OA}}$ can be uniquely determined by its global ranking. Therefore, we can obtain the MLE of $\sigma_{\hat f}$ by solving the following integer linear programming problem:
\begin{equation}
\label{equation_MLE_integer}
\begin{aligned}
    & \min_x\, \sum_{(i, j): i \neq j} x_{j i} w_{i j}^t \\
    &\text{ s.t. } \, x_{i j}+x_{j k}+x_{k i} \geq 1, \quad &\forall \text { distinct } i, j, k \in[K]  \\
    &\phantom{\text{ s.t. }}\, x_{i j}+x_{j i}=1, \quad &\forall \text { distinct } i, j \in[K]\\
    &\phantom{\text{ s.t. }}\, x_{i j} \in\{0,1\} . \quad &\forall \text { distinct } i, j \in[K] 
\end{aligned}
\end{equation}

To solve \eqref{equation_MLE_integer}, it is useful to take note of the property highlighted in Lemma~\ref{lemma_MLE_relabel}, which is a straightforward result of the nested structures of our proposed algorithms.

\begin{lemma}
\label{lemma_MLE_relabel}
For any $i<j$, it holds that $w_{ij}^t\ge w_{ji}^t$.
\end{lemma}

For the objective function of \eqref{equation_MLE_integer}, Lemma~\ref{lemma_MLE_relabel} allows us to obtain a lower bound as: 
\begin{align*}
    \sum_{(i, j): i \neq j} x_{j i} w_{i j}^t = \sum_{(i, j): i < j} \left( x_{j i} w_{i j}^t + x_{ij} w_{ji}^t \right) 
    \ge \sum_{(i, j): i > j} w_{i j}^t ,
\end{align*}
which does not depend on $x$. Furthermore, the lower bound can be achieved by a feasible solution that corresponds to  the identity ranking:
$$
x_{ij} = \begin{cases} 1 & \text{ if } i < j, \\ 0 & \text{ if }  i >j. \end{cases}
$$
Thus, $\sum_{(i, j): i > j} w_{i j}^t $ is indeed the optimal value of the optimization problem \eqref{equation_MLE_integer}. Altogether, with respect to our algorithms \textsc{NE} and \textsc{NP}, the MLE of the unknown global ranking can be found easily by sorting the voting scores, and the maximum log-likelihood is equal to 
\begin{equation}
 \max_{\hat f \in \mathcal M_p^{\mathrm{OA}}} \mathcal{L}(\hat f) = \phi + \log(p)  \sum_{(i, j): i > j} w_{i j}^t.
\label{equation_MLE_result}   
\end{equation}

\subsection{Nested Elimination}
Suppose that $k+1$ is still active at time $t$, we consider two hypotheses:
\begin{align*}
&H_0: k+1 \text{ is not the best item} \quad \text{ vs. } \quad
H_1: k+1 \text{ is the best item}.
\end{align*}

Let ${\mathcal{GL}_0}$ and ${\mathcal{GL}_1}$ denote the generalized log-likelihood function of $H_0$ and $H_1$, respectively. Then according to \eqref{equation_MLE_result}, we have
$$
\frac {\mathcal{GL}_0-\phi} {\log(p)} = \sum_{(i, j): i > j} w_{i j}^t = \sum_{i=1}^K (i-1)W_t(i).
$$

On the other hand, using the similar idea as that in Section~\ref{subsection_MLE}, we can calculate $\mathcal{GL}_1$ by solving another integer linear programming problem:
\begin{equation}
\label{equation_H1_integer}
\begin{aligned}
    & \min_x\, \sum_{(i, j): i \neq j} x_{j i} w_{i j}^t \\
    &\text{ s.t. } \, x_{i j}+x_{j n}+x_{n i} \geq 1, \quad &\forall \text { distinct } i, j, n \in[K]  \\
    &\phantom{\text{ s.t. }}\, x_{i j}+x_{j i}=1, \quad &\forall \text { distinct } i, j \in[K]\\
    &\phantom{\text{ s.t. }}\, x_{i j} \in\{0,1\} , \quad &\forall \text { distinct } i, j \in[K]  \\
    &\phantom{\text{ s.t. }}\, x_{(k+1)i} = 1, \quad &\forall i\neq k+1
\end{aligned}
\end{equation}
whose optimal solution is given by
$$
x_{ij} = \begin{cases}
1 & \text{ if } i < j, i \neq k+1, j \neq k+1,
\\ 0 & \text{ if }  i >j, i \neq k+1, j \neq k+1, 
\\ 1 & \text{ if } i = k+1 , 
\\ 0 & \text{ if } j = k+1 .
\end{cases}
$$

Therefore, we have 
$
\frac {\mathcal{GL}_1-\phi} {\log(p)} = \sum_{(i, j): i > j} w_{i j}^t + \sum_{i=1}^k w_{i(k+1)}^t-\sum_{i=1}^k w_{(k+1)i}^t = \sum_{i=1}^K (i-1)W_t(i) + \sum_{i=1}^k W_t(i)-kW_t(k+1).$ As a result, the generalized likelihood ratio for testing $H_0$ against $H_1$ can be written as 
$$
\mathcal{GL}_0 - \mathcal{GL}_1 = -\log\left(p \right) \cdot  \left( \sum_{i=1}^k W_t(i)-kW_t(k+1)\right),
$$
which is proportional to the quantity of interest in the elimination criterion of \textsc{NE}.

\subsection{Nested Partition}\label{sub:likelihood test for NP}
Similarly, in the partition criterion, we evaluate the following pair of hypotheses:
\begin{align*}
H_0: \text{ $S_{\mathrm{high}}$ and $S_{\mathrm{low}}$ are correctly partitioned,} \quad \text{ vs. } \quad H_1:  \text{ $S_{\mathrm{high}}$ and $S_{\mathrm{low}}$ are incorrectly partitioned}.
\end{align*}

Without loss of generality, we assume that $S_{\mathrm{high}}=\{ 1,\ldots,n\}$ and $S_{\mathrm{low}}=\{n+1, \ldots, K\}$ for some $n\in[K-1]$.
Let ${\mathcal{GL}_0}$ and ${\mathcal{GL}_1}$ denote the generalized log-likelihood function of $H_0$ and $H_1$, respectively. According to \eqref{equation_MLE_result}, we have
$$
\frac {\mathcal{GL}_0-\phi} {\log(p)} = \sum_{(i, j): i > j} w_{i j}^t = \sum_{i=1}^K (i-1)W_t(i).
$$

On the other hand, following the similar approach, we can compute $\mathcal{GL}_1$ by solving the following integer linear programming problem:
\begin{equation}
\begin{aligned}
    & \min_x\, \sum_{(i, j): i \neq j} x_{j i} w_{i j}^t \\
    &\text{ s.t. } \, x_{i j}+x_{j k}+x_{k i} \geq 1, \quad &\forall \text { distinct } i, j, k \in[K]  \\
    &\phantom{\text{ s.t. }}\, x_{i j}+x_{j i}=1, \quad &\forall \text { distinct } i, j \in[K]\\
    &\phantom{\text{ s.t. }}\, x_{i j} \in\{0,1\} , \quad &\forall \text { distinct } i, j \in[K]  \\
    &\phantom{\text{ s.t. }}\, \sum_{i=1}^n \sum_{j=n+1}^K x_{ji} \ge 1  
\end{aligned}
\end{equation}
whose optimal value is 
$$
\frac {\mathcal{GL}_1-\phi} {\log(p)} = \sum_{(i, j): i > j} w_{i j}^t + \min_{i\in[n],\, j>n}  \left( w_{ij}^t - w_{ji}^t \right) = \sum_{i=1}^K (i-1)W_t(i) + W_t(n)-W_t(n+1).
$$

Consequently, the generalized likelihood ratio for testing $H_0$ against $H_1$ can be expressed as
$$
\mathcal{GL}_0 - \mathcal{GL}_1 = -\log\left(p \right) \cdot  \left( W_t(n)-W_t(n+1) \right),
$$
which is proportional to the quantity of interest in the partition criterion of our algorithm \textsc{NP}.

\section{Analysis of Nested Partition}
Before presenting the analysis of \textsc{Nested Partition}, we note that certain symbols utilized in the analysis of \textsc{Nested Elimination} (Appendix~\ref{appendix_analysis_NE}) are recurrently employed in this appendix as well, albeit with a slight abuse of notation. Although these symbols maintain similar definitions, they are tailored to address distinct identification problems and are thus defined accordingly. Readers are advised to interpret the notation in both contexts independently. For example, $\tau_r, T_r, S_r$ and $Q_*^r$ for $r\in[K]$.

\subsection{Preliminaries}
\label{subsection_ranking_Preliminaries}

For the overarching analysis spanning multiple executions of the subroutine, the concept of binary tree representation plays a crucial role. Specifically, we will introduce two types of binary trees to facilitate the analysis and presentation of our approach. One type signifies the expected procedure of our algorithm for a particular customer preference, while the other characterizes the stochastic behavior of our algorithm, akin to the illustrated example in Figure~\ref{figure_rankingtree_example}.

\vspace{0.2 cm}
\noindent\textbf{Deterministic Generating Binary Tree. } Fix a customer preference $f\in\mathcal{M}_p$. Recall that computing the hardness quantity $\JN(f)$ amounts to unrolling the recursion~\eqref{equation_DfSW}, and this unrolling \emph{deterministically} traces out a binary tree, which also represents the expected procedure of our algorithm \textsc{NP}. We call it the \emph{deterministic generating binary tree} of $f$ and write $\mathrm{DGBT}(f)$.

Every invocation of the recursion takes a display set $S$ as its argument, and no display set is ever passed as an argument more than once. We may therefore identify each node of $\mathrm{DGBT}(f)$ with the display set $S$ supplied to the corresponding invocation, so that the root is the initial argument $[K]$. Reading the two cases of~\eqref{equation_DfSW} as the structure of the tree: a node $S$ with $|S|=1$ issues no further call and is a leaf, while a node with $|S|>1$ descends into the sets $\bar S_{\mathrm{high}}$ and $\bar S_{\mathrm{low}}$ of the earlier description, which we take to be its left and right child, respectively. Consequently, $\mathrm{DGBT}(f)$ is a full binary tree, in which every node has either no children or exactly two; its $K$ leaves correspond to the $K$ items, and it has exactly $K-1$ internal (non-leaf) nodes.

For ease of reference, we enumerate the $K-1$ internal nodes of $\mathrm{DGBT}(f)$ as $\bar S_1,\ldots,\bar S_{K-1}$, indexed by $i$ from $1$ to $K-1$ following a depth-first and left-first search. In particular, the root is indexed as $\bar S_1=[K]$. Figure~\ref{figure_DGBT_example} illustrates a typical $\mathrm{DGBT}(f)$ for $K=7$. 
\begin{figure}[htbp]
	\begin{center}
		\begin{tikzpicture}[
		level/.style={level distance=15mm},
		level 1/.style={sibling distance=80mm},
		level 2/.style={sibling distance=40mm},
		level 3/.style={sibling distance=20mm},
		arrow/.style={->, line width=1.5pt, >=stealth}
		]
		\node {$\bar S_{1}=[7]$}
		child[arrow] {node {$\bar S_{2}=\{1,2,3\}$}
			child[arrow] {node {$\{1\}$}}
			child[arrow] {node {$\bar S_{3}=\{2,3\}$}
				child[arrow] {node {$\{2\}$}}
				child[arrow] {node {$\{3\}$}}
			}
		}
		child[arrow] {node {$\bar S_{4}=\{4,5,6,7\}$}
			child[arrow] {node {$\bar S_{5}=\{4,5\}$}
				child[arrow] {node {$\{4\}$}}
				child[arrow] {node {$\{5\}$}}
			}
			child[arrow] {node {$\bar S_{6}=\{6,7\}$}
				child[arrow] {node {$\{6\}$}}
				child[arrow] {node {$\{7\}$}}
			}
		};
		\end{tikzpicture}
		\caption{An example of deterministic generating binary tree $\mathrm{DGBT}(f)$ with $K=7$.}
		\label{figure_DGBT_example}
	\end{center}
\end{figure}
For each node $S$ of $\mathrm{DGBT}(f)$, write $w_S$ for the score vector with which the recursion~\eqref{equation_DfSW} is invoked on $S$. Since $f$ is fixed and $w_{\bar S_i}$ is thereby determined, we abbreviate, for every internal node $\bar S_i$,
$
\bar\tau(\bar S_i):=\bar\tau(f,\bar S_i,w_{\bar S_i}),$ $
i^*(\bar S_i):=i^*(f,\bar S_i,w_{\bar S_i}),$ and 
$D(\bar S_i):=D(f,\bar S_i,w_{\bar S_i}).
$
In particular, the two children of $\bar S_i=[a,b]$ are $\bar S_{\mathrm{high}}=[a,\,i^*(\bar S_i)]$ and $\bar S_{\mathrm{low}}=[i^*(\bar S_i)+1,\,b]$. We caution that the subscripts $\mathrm{high}/\mathrm{low}$ name the two children of a node \emph{relative to that node}, whereas the numeric subscripts $1,\ldots,K-1$ form a single global enumeration of all internal nodes; the two coincide only at the level of the underlying sets.  

\tbl{
For ease of reference, let us state a few basic facts about $ \mathrm{DGBT}(f) $ below.
\begin{lemma}
	\label{lemma:adjacent-gap-accounting}
	Fix \(f\in\mathcal M_p\) whose underlying ranking is the identity ranking. For the internal nodes
	\(\bar S_1,\ldots,\bar S_{K-1}\) of \(\mathrm{DGBT}(f)\), $ \bar\tau(\bar S_r)\geq0 $. In addition,
	$$
	D(f,[K],\mathbf 0)
	=
	\sum_{r=1}^{K-1}\bar\tau(\bar S_r)>0.
	$$
	Moreover, for every \(i\in[K-1]\),
	\begin{equation}
	\label{eq:DGBT-unit-separation}
	\sum_{\substack{r\in[K-1]:\\
			\{i,i+1\}\subseteq\bar S_r}}
	\bar\tau(\bar S_r)
	\bigl(
	f(i\mid\bar S_r)-f(i+1\mid\bar S_r)
	\bigr)
	=1.
	\end{equation}
\end{lemma}

\proof{Proof of Lemma~\ref{lemma:adjacent-gap-accounting}.}
At the root, every adjacent voting-score gap is zero. Suppose that,
when an internal node \(S\) is processed with score vector \(w_S\),
$
0\leq w_S(i)-w_S(i+1)\leq1
$
for every adjacent pair in \(S\). Since
\(f(i\mid S)-f(i+1\mid S)>0\), the definition of
\(\bar\tau(f,S,w_S)\) gives
\[
0\leq\bar\tau(f,S,w_S)
\leq
\frac{1-w_S(i)+w_S(i+1)}
{f(i\mid S)-f(i+1\mid S)}.
\]
Consequently, after the score update, every adjacent gap remaining
in either child still belongs to \([0,1]\). Induction over the tree
therefore proves that all node durations are nonnegative. Unrolling
the recursion in \eqref{equation_DfSW} gives
\[
D(f,[K],\mathbf 0)
=
\sum_{r=1}^{K-1}\bar\tau(\bar S_r).
\]

Now fix \(i\in[K-1]\). The internal nodes containing both \(i\) and
\(i+1\) form a path from the root to the node at which this pair is
separated. Along this path, their voting-score gap increases by
\[
\bar\tau(\bar S_r)
\bigl(
f(i\mid\bar S_r)-f(i+1\mid\bar S_r)
\bigr)
\]
at node \(\bar S_r\). The gap starts at zero and equals one immediately
after the pair is separated, because the split condition at index
\(i\) is then binding. Summing the increments along the path proves the second part of the claim.
\Halmos
\endproof
}

\noindent\textbf{Stochastic Generating Binary Tree. } 
For any customer preference $f\in\mathcal{M}_p$, the execution process of our algorithm \textsc{NP} (detailed in Algorithm~\ref{algo_ranking}) \emph{randomly} produces a generating binary tree. We denote the stochastic binary tree generated for preference $f$ as $\mathrm{SGBT}(f)$, which exhibits numerous similarities with $\mathrm{DGBT}(f)$. 
Within $\mathrm{SGBT}(f)$, each node is also uniquely identified by a display set $S$, which serves as the input for the subroutine Algorithm~\ref{algo_partition}. Similarly, $\mathrm{SGBT}(f)$ is a full binary tree consisting of $K$ leaf nodes  and $K-1$ internal nodes. For any internal node $S$, its  left child and right child are constructed by the subsets created in the subroutine, namely $S_{\mathrm{high}}$ and $S_{\mathrm{low}}$, respectively.

\begin{figure}[htbp]
\begin{center}
\begin{minipage}{1.0\columnwidth}
\centering
\begin{tikzpicture}[
  level/.style={level distance=12mm},
  level 1/.style={sibling distance=80mm},
  level 2/.style={sibling distance=40mm},
  level 3/.style={sibling distance=20mm},
  arrow/.style={->, line width=1.5pt, >=stealth},
  xscale=0.9, yscale=0.9
]
\node {$S_{1}=[7]$}
  child[arrow] {node {$S_{2}=\{1,2,3\}$}
    child[arrow] {node {$\{1\}$}}
    child[arrow] {node {$S_{3}=\{2,3\}$}
      child[arrow] {node {$\{2\}$}}
      child[arrow] {node {$\{3\}$}}
    }
  }
  child[arrow] {node {$S_{4}=\{4,5,6,7\}$}
    child[arrow] {node {$S_{5}=\{4,5\}$}
      child[arrow] {node {$\{4\}$}}
      child[arrow] {node {$\{5\}$}}
    }
    child[arrow] {node {$S_{6}=\{6,7\}$}
      child[arrow] {node {$\{6\}$}}
      child[arrow] {node {$\{7\}$}}
    }
  };
\end{tikzpicture}
\end{minipage}
\begin{minipage}{1.0\columnwidth}
\centering
\vspace{6pt}
\begin{tikzpicture}
\draw[dashed] (-7.5, 0) -- (7.5, 0);
\end{tikzpicture}
\vspace{6pt}
\end{minipage}
\begin{minipage}{1.0\columnwidth}
\centering
\centering
\begin{tikzpicture}[
  level/.style={level distance=12mm},
  level 1/.style={sibling distance=80mm},
  level 2/.style={sibling distance=40mm},
  level 3/.style={sibling distance=20mm},
  level 4/.style={sibling distance=15mm},
  arrow/.style={->, line width=1.5pt, >=stealth},
  xscale=0.9, yscale=0.9
]
\node {$S_{1}=[7]$}
  child[arrow] {node {$S_{2}=\{1,2\}$}
    child[arrow] {node {$\{1\}$}}
    child[arrow] {node {$\{2\}$}
    }
  }
  child[arrow] {node {$S_{3}=\{3,4,5,6,7\}$}
    child[arrow] {node {$S_{4}=\{3,4\}$}
      child[arrow] {node {$\{3\}$}}
      child[arrow] {node {$\{4\}$}}
    }
    child[arrow] {node {$S_{5}=\{5, 6,7\}$}
      child[arrow] {node {$\{5\}$}}
      child[arrow] {node {$S_{6}=\{6, 7\}$}
      child[arrow] {node {$\{6\}$}}
      child[arrow] {node {$\{7\}$}}
      }
    }
  };
\end{tikzpicture}
\end{minipage}
\begin{minipage}{1.0\columnwidth}
\centering
\vspace{6pt}
\begin{tikzpicture}
\draw[dashed] (-7.5, 0) -- (7.5, 0);
\end{tikzpicture}
\vspace{6pt}
\end{minipage}
\begin{minipage}{1.0\columnwidth}
\centering
\begin{tikzpicture}[
  level/.style={level distance=12mm},
  level 1/.style={sibling distance=80mm},
  level 2/.style={sibling distance=40mm},
  level 3/.style={sibling distance=20mm},
  arrow/.style={->, line width=1.5pt, >=stealth},
  xscale=0.9, yscale=0.9
]
\node {$S_{1}=[7]$}
  child[arrow] {node {$S_{2}=\{1,4,5\}$}
    child[arrow] {node {$\{1\}$}}
    child[arrow] {node {$S_{3}=\{4,5\}$}
      child[arrow] {node {$\{4\}$}}
      child[arrow] {node {$\{5\}$}}
    }
  }
  child[arrow] {node {$S_{4}=\{2,3,6,7\}$}
    child[arrow] {node {$S_{5}=\{2,3\}$}
      child[arrow] {node {$\{2\}$}}
      child[arrow] {node {$\{3\}$}}
    }
    child[arrow] {node {$S_{6}=\{6,7\}$}
      child[arrow] {node {$\{6\}$}}
      child[arrow] {node {$\{7\}$}}
    }
  };
\end{tikzpicture}
\end{minipage}
\vspace{-0pt}
\caption{Three examples of SGBTs with $K=7$. }
\label{figure_SGBT_example}
\end{center}
\end{figure}

For ease of presentation, we assign labels to all the $K-1$ internal nodes as $S_i$, where $i$ ranges from 1 to $K-1$, in the chronological order of their use as input for Algorithm~\ref{algo_partition}. As a result, the root node is always represented as $S_1 = [K]$. Figure~\ref{figure_SGBT_example} depicts three possible instances of $\mathrm{SGBT}(f)$ with $K=7$. Notably, the first example aligns with the expected behavior of our algorithm, as demonstrated in $\mathrm{DGBT}(f)$ (refer to Figure~\ref{figure_DGBT_example}). Specifically, it satisfies that $S_i = \bar S_i$ for all $i\in[K-1]$. In contrast, the second example, while deviating from $\mathrm{DGBT}(f)$, still produces a correct ranking. However, the third example yields an inaccurate ranking.

Note that in the subroutine Algorithm~\ref{algo_partition}, no interaction with the customers occurs (i.e., the timer is not updated) when the input $S_{\mathrm{active}}$ is a singleton. Therefore, to analyze the stopping time of our algorithm \textsc{NP}, the entire procedure can be divided into $K-1$ stages based on $S_{\mathrm{active}}$. For any stage $r \in [K-1]$, the active item set $S_{\mathrm{active}}$ is $S_r$. For convenience, we denote the number of time steps in stage $r$ as $\tau_r$, and its cumulative time is denoted as $T_r$. As such, the stopping time of our algorithm can be represented as $
\tau = T_{K-1} = \sum_{r=1}^{K-1} \tau_r.$

\tbl{
\subsection{Understanding $ J^N(f) $}\label{append:understanding JN}

\proof{Proof of Proposition~\ref{prop:MNL_NP}.} 
Because 
\(\fMNL\in\mathcal M_p^{\mathrm{MNL}}\), $ p_i \leq p \text{ for every }i\in \{2, \ldots, K\}. 
$
Under the MNL model, \eqref{eq:DGBT-unit-separation} in Lemma~\ref{lemma:adjacent-gap-accounting} becomes 
\[
1 = \sum_{\substack{r\in[K-1]:\\ \{i-1,i\}\subseteq\bar S_r}}\, \bar\tau(\bar S_r) \frac{v_{i-1}-v_i} {\sum_{\ell\in\bar S_r}v_\ell}. 
\]
Multiplying by $ \frac{v_{i-1}+v_i}{v_{i-1}-v_i} = \frac{1+p_i}{1-p_i}, $ summing over \(i\in[K]\setminus\{1\}\), and collecting the terms associated with each node \(\bar S_r=[a,b]\) gives 
\begin{equation} 
\label{eq:MNL_exact_decomposition} 
\sum_{i=2}^{K}\frac{1+p_i}{1-p_i} = \sum_{r=1}^{K-1} \bar\tau(\bar S_r) \left( 2-\frac{v_a+v_b}{\sum_{\ell=a}^b v_\ell} \right). 
\end{equation} 
For every internal node \([a,b]\), $ v_a+v_b\leq\sum_{\ell=a}^b v_\ell, $ so the multiplier in \eqref{eq:MNL_exact_decomposition}, $ \left( 2-\frac{v_a+v_b}{\sum_{\ell=a}^b v_\ell} \right) $, is at least one. On the other hand, 
\[
\frac{\sum_{\ell=a}^b v_\ell}{v_a} = 1+p_{a+1}+p_{a+1}p_{a+2} +\cdots+p_{a+1}p_{a+2}\cdots p_b \leq 1+p+\cdots+p^{b-a} < \frac1{1-p}. 
\]
Hence 
$
2-\tfrac{v_a+v_b}{\sum_{\ell=a}^b v_\ell} < 1+p. 
$
Given that $ D(\fMNL,[K],\mathbf0) =  \sum_{r=1}^{K-1} \bar\tau(\bar S_r)  $ by Lemma~\ref{lemma:adjacent-gap-accounting}, we conclude that 
\[
D(\fMNL,[K],\mathbf0) \leq \sum_{i=2}^{K}\frac{1+p_i}{1-p_i} \leq (1+p)D(\fMNL,[K],\mathbf0). 
\] 
The result now follows by taking reciprocals and applying $\JN(\fMNL) = \frac{\log(1/p)}{D(\fMNL,[K],\mathbf0)}. $
\Halmos 
\endproof
}

\vskip 0.3 cm

\tbl{
\vskip .2 cm
\noindent \textbf{Example 3: Preference Classes with a Hard-to-Learn Pair.}
We now partially extend our MNL analysis and consider a general $ p $-separable preference subfamily in which the gap between one pair of
items is \textit{binding}.

\begin{definition}\label{def:hard pair}
	Fix $0<\rho<p<1$ and $\ell\in [K-1]$. We say that a preference $f \in \mathcal{M}_p$ is
	\emph{$(\rho,p)$-separated with bottleneck pair $(\ell,\ell+1)$} if, for every display set $S$:
	\begin{enumerate}
		\item If $\{\ell,\ell+1\}\subseteq S$, then $ \frac{f(\ell+1\mid S)}{f(\ell\mid S)}=p. $
		\item For every pair $k<k'$ in $S$ with $(k,k')\neq(\ell,\ell+1)$,
		$ \frac{f(k'\mid S)}{f(k\mid S)}\le \rho. $
	\end{enumerate}
\end{definition}

This family includes the preference instances in which exactly one pair of items is hard to distinguish, regardless of whether that pair is top- or bottom-ranked. One example is the MNL preference whose
attraction scores satisfy $ \frac{v_{\ell+1}}{v_{\ell}} = p $ for some item $ \ell $.

The next result shows that the hardness measures of all instances in this family are essentially
governed, up to a constant, by $ \log\left(\tfrac 1 p \right) \left(K+\tfrac{1}{1-p}\right)^{-1} $.

\begin{lemma}\label{lma:hardest_pair}
	Suppose the preference $ f $ is $ (\rho, p) $-separated with a bottleneck pair. Then, for every fixed
	$ \rho $, there exist constants $ c = 1/4 $ and
	$ C = \max\!\left\{\tfrac{4}{(1-\rho)^2},2\right\} $ such that
	$
	c \left(K+\tfrac{1}{1-p}\right)  \,\leq \, D(f,[K], \mathbf 0) \, \leq \, C \left(K+\tfrac{1}{1-p}\right).
	$
	As a result,
	\begin{align*}
	C^{-1} \log\left(\tfrac 1 p \right) \left(K+\tfrac{1}{1-p}\right)^{-1}
	\,\leq \, \JN(f) \, \leq \,
	c^{-1} \log\left(\tfrac 1 p \right) \left(K+\tfrac{1}{1-p}\right)^{-1}.
	\end{align*}
\end{lemma}

\proof{Proof of Lemma~\ref{lma:hardest_pair}.}
Pick $0<\rho<p<1$, $\ell\in [K-1]$, and a preference  $f \in \mathcal{M}_p$ that is $(\rho,p)$-separated with bottleneck pair $(\ell,\ell+1)$ as required by Definition \ref{def:hard pair}.  To evaluate $ D(f, [K], w) $ using \eqref{equation_DfSW}, a recursion tree is unfolded. When the recursion process processes a display set $ S = [a,b] $, the following happens: it evaluates $\bar\tau(f,S,w)$, updates $w$ on $S$, and replaces $S$ by the two sub-display sets $[a,i^*]$ and $[i^*+1,b]$. Each processed display set $ S $ contributes $\bar\tau_S:=\bar\tau(f,S,w)$ to the total $D(f,[K], \mathbf 0)$.  For shorthand notation, we write $\Delta_i(S)=f(i\mid S)-f(i+1\mid S) > 0$ for the choice probability gap at adjacent pair $(i,i+1)$ in $ S $, $w_S$ for the score vector at the call that processes $S$, and $g_i(S):=w_S(i)-w_S(i+1)$ for the score gap.

We first claim $g_i(S)\in[0,1]$ for every processed display set $S$ and every adjacent pair $(i,i+1)$ in $S$.
This holds for the root call ($w_{[K]}=0$). For the inductive step, suppose the claim holds for a processed $S=[a,b]$. Then
\[
\tfrac{1-g_i(S)}{\Delta_i(S)} \ge \min_i\tfrac{1-g_i(S)}{\Delta_i(S)} = \bar\tau_S\ge0.
\]
Hence the updated gap $g_i(S)+\bar\tau_S\Delta_i(S)$ lies in $[0,1]$.
Every adjacent pair in a child display set $[a,i^*]$ or $[i^*+1,b]$ is also an adjacent pair in $S$ with the same updated gap, so the claim passes to both children.

\underline{Lower bound.}
Fix $i\in\{1,\ldots,K-1\}$. The gap $g_i(S)$ starts at zero in the root call and grows by $\bar \tau_S\Delta_i(S)$ each time a display set $S\supseteq\{i,i+1\}$ is processed. The only way to separate $i$ and $i+1$ is to split at index $i$; since the recursion terminates in singletons, this must eventually happen. At that call, $i$ is the minimizer defining $\bar \tau_S$, so $g_i(S)+ \bar\tau_S\Delta_i(S)=1$ exactly. After this, $i$ and $i+1$ never co-occur. Hence
\[
\sum_{S\text{ processed}:\;\{i,i+1\}\subseteq S}\bar\tau_S\Delta_i(S)=1.
\]
Summing over $i=1,\ldots,K-1$ and telescoping within each processed $S=[a,b]$,
\[
K-1
=\sum_S\bar\tau_S\sum_{i=a}^{b-1}\Delta_i(S)
=\sum_S\bar\tau_S\bigl(f(a\mid S)-f(b\mid S)\bigr)
\le \sum_S\bar\tau_S
=D(f,[K], \mathbf 0).
\]
Specializing to $i=\ell$: Definition~\ref{def:hard pair} gives
$\Delta_\ell(S)=(1-p)f(\ell\mid S)\le 1-p$ for every processed $S\supseteq\{\ell,\ell+1\}$, so the identity yields
$1\le (1-p)D(f,[K], \mathbf0)$. Combining with $D\ge K-1$ and $K\ge2$,
\[
D(f,[K], \mathbf 0)\ge \max\left\{K-1,\tfrac1{1-p}\right\}
\ge \frac14\left(K+\tfrac1{1-p}\right).
\]

\underline{Upper bound.}
We first establish a gap claim: every processed $S=[a,b]\ne[\ell,\ell+1]$ satisfies $\max_i\Delta_i(S)\ge(1-\rho)^2/4$. 

\emph{Case $a\ne\ell$.}
Then $(a,a+1)$ is not the bottleneck pair, so
$\Delta_a(S)\ge(1-\rho)f(a\mid S)$.
Consecutive ratios in $S$ are each at most $\rho$, except possibly the bottleneck ratio $p$.
Splitting $\sum_{i\in S}f(i\mid S)=1$ at the bottleneck yields two geometric series with ratio $\rho$, each with leading term at most $f(a\mid S)$, so
$1\le 2f(a\mid S)/(1-\rho)$ and
$\Delta_a(S)\ge (1-\rho)^2/2$.

\emph{Case $a=\ell$.}
Then $b\ge\ell+2$. Since $f(\ell+1\mid S)=p f(\ell\mid S)$ and subsequent ratios are at most $\rho$,
\[
1\le f(\ell\mid S)\!\left(1+\tfrac{p}{1-\rho}\right)
\le \frac{2-\rho}{1-\rho}\,f(\ell\mid S),
\quad\text{so}\quad
f(\ell\mid S)\ge \frac{1-\rho}{2-\rho}.
\]
Now $\Delta_\ell(S)=(1-p)f(\ell\mid S)$ and $\Delta_{\ell+1}(S)\ge(1-\rho)pf(\ell\mid S)$, so
\[
\max\bigl\{\Delta_\ell(S),\,\Delta_{\ell+1}(S)\bigr\}
\;\ge\;\max\{1-p,\,(1-\rho)p\}\cdot f(\ell\mid S)
\;\ge\;\tfrac{(1-\rho)^2}{(2-\rho)^2}
\;\ge\;\tfrac{(1-\rho)^2}{4},
\]
where the second inequality uses $\max\{1-p,(1-\rho)p\}\ge(1-\rho)/(2-\rho)$, attained at $p=1/(2-\rho)$.

\smallskip

Now we bound $\bar\tau_S$ for each processed display set. By the gap claim, every processed $S\ne[\ell,\ell+1]$ has some adjacent pair $(i,i+1)$ with $\Delta_i(S)\ge(1-\rho)^2/4$. Since we have shown that $1-g_i(S)\le 1$, $\bar\tau_S\le 4/(1-\rho)^2$.
For $S=[\ell,\ell+1]$, normalization and the bottleneck ratio give
$\Delta_\ell=(1-p)/(1+p)\ge(1-p)/2$, so $\bar\tau_S\le 2/(1-p)$.
The recursion processes exactly $K-1$ display sets (each step replaces one by two, going from $1$ to $K$), so
\[
D(f,[K], \mathbf 0)\le \tfrac{4}{(1-\rho)^2}(K-1)+\tfrac{2}{1-p}
\le C\!\left(K+\tfrac1{1-p}\right),
\]
with $C=\max\{4/(1-\rho)^2,\,2\}$. \Halmos
\endproof
}

\subsection{Proof of Theorem~\ref{theorem_ranking_fixedconfidence}}
\label{appendix_theorem_ranking_fixedconfidence}

In a manner akin to the proof of Theorem~\ref{theorem_fixedconfidence} presented in Appendix~\ref{appendix_theorem_fixedconfidence}, we introduce two intermediate results with the tuning parameter $M$ serving as an intermediary. Specifically, Proposition~\ref{proposition_tau_general_ranking} demonstrates that the expected stopping time of our algorithm \textsc{NP}, with input parameter $M$, is asymptotically upper bounded by ${\log(1/p)M}/{\JN(f)}$ as $M$ tends to infinity. Additionally, in Proposition~\ref{proposition_error_ranking}, we provide an upper bound on the error probability of \textsc{NP},  which does not depend on the specific preference instance $f$. The detailed proofs of Proposition~\ref{proposition_tau_general_ranking} and 
Proposition~\ref{proposition_error_ranking} can be found in Appendices~\ref{appendix_proposition_tau_general_ranking} and 
\ref{appendix_proposition_error_ranking}, respectively.

\begin{proposition}[Expected stopping time of \textsc{NP}]
\label{proposition_tau_general_ranking}
	For any customer preference $f\in\mathcal M_p$,  \textsc{NP} ensures that
	\begin{equation*}
	\E [\tau] \le \frac{\log(1/p)M} {\JN(f)} + O_M(1)
	\end{equation*}
	where the  $O_M(1)$ term is specified in \eqref{equation_oM1_ranking} in the corresponding proof.
\end{proposition}

\begin{proposition}[Error probability of \textsc{NP}]
	\label{proposition_error_ranking}
	For any customer preference $f\in\mathcal M_p$, \textsc{NP} outputs a ranking $\sigma_{\mathrm{out}} $ satisfying
	\begin{equation*}
	\mathbb P (\sigma_{\mathrm{out}} \neq \sigma_* ) \le  (K-1)\cdot p^M. 
	\end{equation*}
\end{proposition}

In accordance with Propositions~\ref{proposition_tau_general_ranking} and~\ref{proposition_error_ranking}, the expected stopping time of \textsc{NP}  grows approximately linearly with respect to the exogenous parameter $M$, while its error probability undergoes exponential decay. Using these results, we now proceed to present the proof of Theorem~\ref{theorem_ranking_fixedconfidence}.

\vspace{0.2 cm}
\proof{Proof of Theorem~\ref{theorem_ranking_fixedconfidence}. }Consider any fixed confidence level $\delta\in(0,1)$. Proposition~\ref{proposition_error_ranking} implies that the choice of $ M $ in Equation~\eqref{equation_ranking_M} guarantees the error probability is no more than $ \delta $, for any customer preference $f\in\mathcal M_p$. Moreover, since $ \mathbb{E}[\tau] < +\infty $ due to Proposition~\ref{proposition_tau_general_ranking}, $\mathbb P (\tau<\infty) = 1$. Therefore, our algorithm \textsc{NP} is $ \delta $-PAC. 
In addition, based on Proposition~\ref{proposition_tau_general_ranking} as well as Equation~\eqref{equation_oM1_ranking} in the corresponding proof, with $M = \frac{\log(1/\delta)+\log(K-1)} {\log(1/p)}$, the expected stopping time ${\E [\tau]}$ can be upper bounded as follows:
\begin{align*}
\E [\tau] &\le \frac{\log(1/p)M}{\JN(f)} + \underbrace{ \sum_{r=1}^{K-1} \left( \hat \tau_{\mathrm{U}} (\bar S_r) -\bar \tau (\bar S_r) \right)M }_{O_M(1)} + \underbrace{  \sum_{r=2}^{K-1} \frac{2(K-1)(1+p)MQ_*^r} {1-p} }_{o_M(1)}  \\
&= \frac{\log(1/\delta)}{\JN(f)} + \frac{\log(K-1)}{ \JN(f)} + \underbrace{ \sum_{r=1}^{K-1} \left( \hat \tau_{\mathrm{U}} (\bar S_r) -\bar \tau (\bar S_r) \right)M }_{O_{1/\delta}(1)} + \underbrace{  \sum_{r=2}^{K-1} \frac{2(K-1)(1+p)MQ_*^r} {1-p} }_{o_{1/\delta}(1)} 
\end{align*}

Consequently, the theorem is established by setting
\begin{equation}
\label{equation_odelta1_ranking}
C_f' := \frac{\log(K-1)}{ \JN(f)} + \sup_{\delta\in (0,1)} \left( { \sum_{r=1}^{K-1} \left( \hat \tau_{\mathrm{U}} (\bar S_r) -\bar \tau (\bar S_r) \right)M } + {  \sum_{r=2}^{K-1} \frac{2(K-1)(1+p)MQ_*^r} {1-p} }  \right) < +\infty.
\end{equation}\Halmos
\endproof

\subsection{Proof of Proposition~\ref{proposition_tau_general_ranking}}
\label{appendix_proposition_tau_general_ranking}

\proof{Proof of Proposition~\ref{proposition_tau_general_ranking}.}
The structure of the proof of Proposition~\ref{proposition_tau_general_ranking} is similar to that of Proposition~\ref{proposition_tau_general}, although the specific details differ significantly. Using the notation introduced in Section~\ref{subsection_ranking_Preliminaries}, our goal is to bound the expected stopping time $\E [\tau]  = \sum_{r=1}^{K-1} \E[\tau_r ]$.

\vspace{0.2 cm}
\noindent\textbf{Step 1 (Decomposition of the expected stopping time).}  
For any stage $r \in [K-1]$, we state that our algorithm \textsc{NP} adheres to the structure of $\mathrm{DGBT}(f)$ if the left and right children of $S_r$ match those of $\bar S_r$. Then for each $r \in [K]$, let $\mathcal{E}_r$ indicate the event  that our algorithm \textsc{NP} precisely mirrors the structure of $\mathrm{DGBT}(f)$ in each of the first $r-1$ stages. As a result, $\mathcal{E}_1$ always holds true, and it subsequently follows that $\mathcal{E}_1 \supseteq \mathcal{E}_2 \supseteq \cdots \supseteq \mathcal{E}_K$.

Using the linearity of expectation, we can decompose the expected stopping time as follows:
\begin{align*}
    \E [\tau] = \sum_{r=1}^{K-1} \E[\tau_r ] 
    =\sum_{r=1}^{K-1} \E[\tau_r \cdot \mathbbm{1}\{\mathcal E _r\} ]+\sum_{r=1}^{K-1}  \E[\tau_r \cdot \mathbbm{1}\{\mathcal E _r ^c\} ].
\end{align*}
For convenience, we also introduce two shorthand notation:
\begin{align*}
{\everymath={\displaystyle}
T^{\dagger} = \sum_{r=1}^{K-1} \E[\tau_r \cdot \mathbbm{1}\{\mathcal E _r\} ] \quad \text{ and } \quad 
T^{\ddagger} = \sum_{r=1}^{K-1} \E[\tau_r \cdot \mathbbm{1}\{\mathcal E _r^c\} ].
}
\end{align*}

In the subsequent two steps, we will bound $T^{\dagger}$ and $T^{\ddagger}$ separately. Specifically, we will show $T^{\dagger} \le  \frac{\log(1/p)M}{\JN(f)}+  O_M(1)$ and $T^{\ddagger} =  o_M(1)$.

\vspace{0.2 cm}
\noindent\textbf{Step 2 (Bounding $T^{\dagger}$).} {First, it is crucial to highlight a frequently utilized fact: for any stage $r \in [K-1]$ and items $i$ and $j$ in $S_r$, it holds that $|W_{t}(i) - W_{t}(j)| \le (K-1)M$. Otherwise, by the pigeonhole principle, the current stage would have already been terminated. }

For each stage $r\in [K-1]$, conditioned on the event $\mathcal E _r$, we have $S_{\mathrm{active}} = S_r = \bar S_r $, which thereby ensures that $ i^*(\bar S_r)\in S_{\mathrm{active}} $. Since 
$$
W_t(i^*(\bar S_r)) - W_t(i^*(\bar S_r)+1) - \left(f(i^*( \bar S_r)|\bar S_r)-f(i^*(\bar S_r)+1|\bar S_r) \right)t
$$
is a martingale for $t\ge T_{r-1}$, we can apply the optional stopping theorem to obtain 
\begin{align}
    &\phantom{\, =\, } \left(f(i^*( \bar S_r)|\bar S_r)-f(i^*(\bar S_r)+1|\bar S_r) \right)\E[\tau_r\cdot \mathbbm{1}\{\mathcal E _r\} ] \notag \\  
    &= \E \left [(W_{ T_r}(i^*(\bar S_r)) - W_{ T_r}(i^*(\bar S_r)+1))\cdot \mathbbm{1}\{\mathcal E _r\}\right] - \E\left[ \left ( W_{T_{r-1}}(i^*(\bar S_r)) - W_{T_{r-1}}(i^*(\bar S_r)+1) \right)\cdot \mathbbm{1}\{\mathcal E _r\} \right].
    \label{equation_bound_stage_gap}
\end{align}

\noindent\textbf{Step 2a (Upper-bounding \eqref{equation_bound_stage_gap}).} Consider the first term in \eqref{equation_bound_stage_gap} under two scenarios. First, if $\mathcal{E}_{r+1}$ holds true, then it implies that
$$
\min_{i\in S_r, i\le i^*(\bar S_r)} W_{T_r}(i)-\max_{i\in S_r, i> i^*(\bar S_r)} W_{T_r}(i)= M.
$$
Consequently, we can establish an upper bound for $W_{ T_r}(i^*(\bar S_r)) - W_{ T_r}(i^*(\bar S_r)+1)$ in the following manner:
\begin{align*}
    &\phantom{\, =\, }W_{ T_r}(i^*(\bar S_r)) - W_{ T_r}(i^*(\bar S_r)+1)\\
    &= M + \left( W_{ T_r}(i^*(\bar S_r)) - \min_{i\in S_r, i\le i^*(\bar S_r)} W_{ T_r}(i) \right ) + \left( \max_{i\in S_r, i> i^*(\bar S_r)} W_{ T_r}(i) -W_{ T_r}(i^*(\bar S_r)+1) \right) \\
    &\le M + \sum_{i\in S_r, i< i^*(\bar S_r)} \left ( {W_{ T_r}(i^*(\bar S_r))-W_{ T_r}(i)} \right) _+ + \sum_{i\in S_r, i> i^*(\bar S_r)+1} \left( {W_{ T_r}(i)-W_{ T_r}(i^*(\bar S_r)+1)}\right)_+ \\
    &\le M + \sum_{i\in S_r, i< i^*(\bar S_r)} \frac {1} {e\log(1/p)} \cdot {\frac 1 p} ^{W_{ T_r}(i^*(\bar S_r))-W_{ T_r}(i)} + \sum_{i\in S_r, i> i^*(\bar S_r)+1} \frac {1} {e\log(1/p)} \cdot {\frac 1 p} ^{W_{ T_r}(i)-W_{ T_r}(i^*(\bar S_r)+1)}
\end{align*}
where the last inequality follows from the numerical fact that $(x)_+ = \max(x, 0) \le \frac {1} {e\log(1/p)} \cdot {\frac 1 p} ^ x$ for any $x\in \mathbb Z $.

Secondly, if $\mathcal{E}_{r+1}$ is false, it follows that $W_{T_r}(i^* (\bar S_r)) - W_{T_r}(i^*(\bar S_r)+1) \leq (K-1)M$. By merging these two scenarios, we arrive at the inequality:
\begin{align*}
    &\phantom{=\ }\E \left [(W_{ T_r}(i^*(\bar S_r)) - W_{ T_r}(i^*(\bar S_r)+1))\cdot \mathbbm{1}\{\mathcal E _r\}\right] \\
    &\le  M+ \sum_{i\in S_r, i< i^*(\bar S_r)} \frac {1} {e\log(1/p)} \cdot \E\left[ {\frac 1 p} ^{W_{ T_r}(i^*(\bar S_r))-W_{ T_r}(i)}  \cdot \mathbbm{1}\{\mathcal E _{r+1}\}\right] \\
    & \phantom{=\ M}  + \sum_{i\in S_r, i> i^*(\bar S_r)+1} \frac {1} {e\log(1/p)} \cdot \E\left[{\frac 1 p} ^{W_{ T_r}(i)-W_{ T_r}(i^*(\bar S_r)+1)}\cdot \mathbbm{1}\{\mathcal E _{r+1}\}\right] + (K-1)M\cdot \mathbb P(\mathcal E _{r}\setminus \mathcal E _{r+1}) \\
    &\le  M+ \sum_{i\in S_r, i< i^*(\bar S_r)} \frac {1} {e\log(1/p)} \cdot \E\left[ {\frac 1 p} ^{W_{ T_r}(i^*(\bar S_r))-W_{ T_r}(i)}  \cdot \mathbbm{1}\{\mathcal E _{r}\}\right] \\
    & \phantom{=\ M}  + \sum_{i\in S_r, i> i^*(\bar S_r)+1} \frac {1} {e\log(1/p)} \cdot \E\left[{\frac 1 p} ^{W_{ T_r}(i)-W_{ T_r}(i^*(\bar S_r)+1)}\cdot \mathbbm{1}\{\mathcal E _{r}\}\right] + (K-1)M\cdot \mathbb P(\mathcal E _{r}\setminus \mathcal E _{r+1}).
\end{align*}

For any item $i\in S_r$ satisfying $i< i^*(\bar S_r)$, due to the display rule of our algorithm, items $i$ and $i^*(\bar S_r)$ are always displayed together before ${T_r}$. Together with the definition of $\mathcal M_p$,  we can deduce that
$
{\frac 1 p}^{W_t(i^*(\bar S_r))-W_t(i)}
$
is a supermartingale. Thus, it is straightforward to show via induction that 
$$
\begin{aligned}
&\phantom{\; = } \E\left[ {\frac 1 p} ^{W_{ T_r}(i^*(\bar S_r))-W_{ T_r}(i)}  \cdot \mathbbm{1}\{\mathcal E _{r}\}\right] \le \E\left[ {\frac 1 p} ^{W_{ 0}(i^*(\bar S_r))-W_{ 0}(i)}  \cdot \mathbbm{1}\{\mathcal E _{1}\}\right]  = 1 .
\end{aligned}
$$
Similarly, for any item ${i\in S_r}$ satisfying $ i> i^*(\bar S_r)+1$, it holds that 
$$
\begin{aligned}
& \E\left[{\frac 1 p} ^{W_{ T_r}(i)-W_{ T_r}(i^*(\bar S_r)+1)}\cdot \mathbbm{1}\{\mathcal E _{r}\}\right] \le 1 .
\end{aligned}
$$
Therefore, we can establish that
\begin{equation}
\label{equation_stager_0_1}
\begin{aligned}
    \E \left [(W_{ T_r}(i^*(\bar S_r)) - W_{ T_r}(i^*(\bar S_r)+1))\cdot \mathbbm{1}\{\mathcal E _r\}\right]   \le M + \frac {K-2} {e\log(1/p)}+ (K-1)M\cdot \mathbb P(\mathcal E _{r}\setminus \mathcal E _{r+1}).
\end{aligned}
\end{equation}

Consider the second term in \eqref{equation_bound_stage_gap}. Let $\bar S_{r_p}$, where $r_p \in [r-1]$, denote the parent node of $\bar S_r$. In other words, $\bar S_{r_p} = \mathrm{Parent}(\bar S_r)$. 
When the event $\mathcal E _{r_p+1}$ occurs, neither item $i^*(\bar S_r)$ nor item $i^*(\bar S_r)+1$ is displayed between time $T_{r_p}+1$ and $T_{r-1}$. This leads to the following equation:
$$
\E\left[ \left ( W_{T_{r-1}}(i^*(\bar S_r)) - W_{T_{r-1}}(i^*(\bar S_r)+1) \right)\cdot \mathbbm{1}\{\mathcal E _{r_p+1}\} \right] = \E\left[ \left ( W_{T_{r_p}}(i^*(\bar S_r)) - W_{T_{r_p}}(i^*(\bar S_r)+1) \right)\cdot \mathbbm{1}\{\mathcal E _{r_p+1}\} \right].
$$
Subsequently, we can deduce that 
\begin{equation}
\label{equation_stager_1}
\begin{aligned}
    &\phantom{=\ } \E\left[ \left ( W_{T_{r-1}}(i^*(\bar S_r)) - W_{T_{r-1}}(i^*(\bar S_r)+1) \right)\cdot \mathbbm{1}\{\mathcal E _r\} \right] \\
    & = \E\left[ \left ( W_{T_{r-1}}(i^*(\bar S_r)) - W_{T_{r-1}}(i^*(\bar S_r)+1) \right)\cdot \mathbbm{1}\{\mathcal E _{r_p+1}\} \right]  \\
    &\phantom {=\ }- \E\left[ \left ( W_{T_{r-1}}(i^*(\bar S_r)) - W_{T_{r-1}}(i^*(\bar S_r)+1) \right)\cdot \mathbbm{1}\{\mathcal E _{r_p+1} \setminus \mathcal E _{r}\} \right] \\
    & \ge \E\left[ \left ( W_{T_{r_p}}(i^*(\bar S_r)) - W_{T_{r_p}}(i^*(\bar S_r)+1) \right)\cdot \mathbbm{1}\{\mathcal E _{r_p+1}\} \right] - (K-1)M \mathbb P (\mathcal E _{r_p+1} \setminus \mathcal E _{r}) \\
    &= \E\left[ \left ( W_{T_{r_p}}(i^*(\bar S_r)) - W_{T_{r_p}}(i^*(\bar S_r)+1) \right)\cdot \mathbbm{1}\{\mathcal E _{r_p}\} \right] \\
    &\phantom{=\ } -\E\left[ \left ( W_{T_{r_p}}(i^*(\bar S_r)) - W_{T_{r_p}}(i^*(\bar S_r)+1) \right)\cdot \mathbbm{1}\{\mathcal E _{r_p} \setminus \mathcal E _{r_p+1}\} \right]- (K-1)M \mathbb P (\mathcal E _{r_p+1} \setminus \mathcal E _{r}). 
\end{aligned}
\end{equation}

In stage $r_p$, supposing that $\mathcal E _{r_p}$ occurs, the active set is $\bar S_{r_p}$ and 
$$
W_t(i^*(\bar S_r)) - W_t(i^*(\bar S_r)+1) - \left(f(i^*( \bar S_r)|\bar S_{r_p})-f(i^*(\bar S_r)+1|\bar S_{r_p}) \right) t
$$
is a martingale for $t\ge T_{r_p-1}$. Thus, by the optional stopping theorem, we can obtain
\begin{equation}
\label{equation_stager_2}
\begin{aligned}
    &\phantom{=\ } \E\left[ \left ( W_{T_{r_p}}(i^*(\bar S_r)) - W_{T_{r_p}}(i^*(\bar S_r)+1) \right)\cdot \mathbbm{1}\{\mathcal E _{r_p}\} \right] \\
    &= \E\left[ \left ( W_{T_{r_p-1}}(i^*(\bar S_r)) - W_{T_{r_p-1}}(i^*(\bar S_r)+1) \right)\cdot \mathbbm{1}\{\mathcal E _{r_p}\} \right] + \left(f(i^*( \bar S_r)|\bar S_{r_p})-f(i^*(\bar S_r)+1|\bar S_{r_p}) \right) \E[\tau_{r_p} \cdot \mathbbm{1}\{\mathcal E _{r_p}\} ].
\end{aligned}
\end{equation}
Furthermore, since $\{\mathcal E _{r_p} \setminus \mathcal E _{r_p+1}\} \cup \{ \mathcal E _{r_p+1} \setminus \mathcal E _{r}\} = \mathcal E _{r_p} \setminus \mathcal E _{r}$, we have 
\begin{equation}
\label{equation_stager_3}
\begin{aligned}
    &\phantom{=\ } \E\left[ \left ( W_{T_{r_p}}(i^*(\bar S_r)) - W_{T_{r_p}}(i^*(\bar S_r)+1) \right)\cdot \mathbbm{1}\{\mathcal E _{r_p} \setminus \mathcal E _{r_p+1}\} \right] + (K-1)M \mathbb P (\mathcal E _{r_p+1} \setminus \mathcal E _{r}) \\
    &\le  (K-1)M \mathbb P (\mathcal E _{r_p} \setminus \mathcal E _{r_p+1})  + (K-1)M \mathbb P (\mathcal E _{r_p+1} \setminus \mathcal E _{r})  \\
    &= (K-1)M \mathbb P (\mathcal E _{r_p} \setminus \mathcal E _{r}) .
\end{aligned}
\end{equation}
Combining \eqref{equation_stager_1}, \eqref{equation_stager_2} and \eqref{equation_stager_3} gives 
\begin{align*}
    &\phantom{=\ } \E\left[ \left ( W_{T_{r-1}}(i^*(\bar S_r)) - W_{T_{r-1}}(i^*(\bar S_r)+1) \right)\cdot \mathbbm{1}\{\mathcal E _r\} \right] \\
    &\ge  \E\left[ \left ( W_{T_{r_p-1}}(i^*(\bar S_r)) - W_{T_{r_p-1}}(i^*(\bar S_r)+1) \right)\cdot \mathbbm{1}\{\mathcal E _{r_p}\} \right]   \\
    &\phantom{= \ } + \left(f(i^*( \bar S_r)|\bar S_{r_p})-f(i^*(\bar S_r)+1|\bar S_{r_p}) \right) \E[\tau_{r_p} \cdot \mathbbm{1}\{\mathcal E _{r_p}\} ]- (K-1)M \mathbb P (\mathcal E _{r_p} \setminus \mathcal E _{r}).    
\end{align*}

Notice that the preceding analysis of the quantity $\E\left[ \left ( W_{T_{r-1}}(i^*(\bar S_r)) - W_{T_{r-1}}(i^*(\bar S_r)+1) \right)\cdot \mathbbm{1}\{\mathcal E _r\} \right]$ can be equivalently applied to $\E\left[ \left ( W_{T_{r_p-1}}(i^*(\bar S_r)) - W_{T_{r_p-1}}(i^*(\bar S_r)+1) \right)\cdot \mathbbm{1}\{\mathcal E _{r_p}\} \right]$, and further extended to encompass all the ancestors of $\bar S_r$. As a result, we can deduce that 
\begin{align}
    &\phantom{=\ } \E\left[ \left ( W_{T_{r-1}}(i^*(\bar S_r)) - W_{T_{r-1}}(i^*(\bar S_r)+1) \right)\cdot \mathbbm{1}\{\mathcal E _r\} \right] \notag  \\
    &\ge  \sum_{i: \bar S_i \in \mathrm{Ancestor}(\bar S_r)}  \left(f(i^*( \bar S_r)|\bar S_{i})-f(i^*(\bar S_r)+1|\bar S_{i}) \right) \E[\tau_{i} \cdot \mathbbm{1}\{\mathcal E _{i}\} ]- (K-1)M \mathbb P (\mathcal E _{1} \setminus \mathcal E _{r})  \label{equation_stager_0_2}
\end{align}
where the last inequality follows from the fact that $\bar S_1$ is the root node of $\mathrm{DGBT}(f)$.

By substituting \eqref{equation_stager_0_1} and  \eqref{equation_stager_0_2} into \eqref{equation_bound_stage_gap}, we can conclude that for all stage $r \in [K-1]$,
\begin{align*}
    &\phantom{\, =\, } \left(f(i^*( \bar S_r)|\bar S_r)-f(i^*(\bar S_r)+1|\bar S_r) \right)\E[\tau_r\cdot \mathbbm{1}\{\mathcal E _r\} ]  \\  
    &\le  M + \tfrac {K-2} {e\log(1/p)}+ (K-1)M\cdot \mathbb P(\mathcal E _{r}\setminus \mathcal E _{r+1}) \\
    &\phantom{=\ }  -\sum_{i: \bar S_i \in \mathrm{Ancestor}(\bar S_r)}  \left(f(i^*( \bar S_r)|\bar S_{i})-f(i^*(\bar S_r)+1|\bar S_{i}) \right) \E[\tau_{i} \cdot \mathbbm{1}\{\mathcal E _{i}\} ] +  (K-1)M \mathbb P (\mathcal E _{1} \setminus \mathcal E _{r}) \\
    &= M -\sum_{i: \bar S_i \in \mathrm{Ancestor}(\bar S_r)}  \left(f(i^*( \bar S_r)|\bar S_{i})-f(i^*(\bar S_r)+1|\bar S_{i}) \right) \E[\tau_{i} \cdot \mathbbm{1}\{\mathcal E _{i}\} ]  + \tfrac {K-2} {e\log(1/p)} + (K-1)M Q_*^{r+1}
\end{align*}
where the last inequality follows from Lemma~\ref{lemma_Error_r_ranking} and $\mathcal E _{1} \setminus \mathcal E _{r+1} = \mathcal E _{r+1}^c$.

\noindent\textbf{Step 2b (Lower-bounding \eqref{equation_bound_stage_gap}).} Consider the first term in \eqref{equation_bound_stage_gap}. First, if $\mathcal{E}_{r+1}$ holds true, then $W_{ T_r}(i^*(\bar S_r)) - W_{ T_r}(i^*(\bar S_r)+1)\ge M$. Conversely, if $\mathcal{E}_{r+1}$ is false, then $W_{T_r}(i^* (\bar S_r)) - W_{T_r}(i^*(\bar S_r)+1) \geq -(K-1)M$. Therefore, we can arrive at the inequality:
\begin{align}
\label{equation_stager_0_3}
    &\phantom{=\ }\E \left [(W_{ T_r}(i^*(\bar S_r)) - W_{ T_r}(i^*(\bar S_r)+1))\cdot \mathbbm{1}\{\mathcal E _r\}\right] \ge M-M\mathbb P (\mathcal E _{r+1}^c) -(K-1)M \cdot \mathbb P(\mathcal E _{r}\setminus \mathcal E _{r+1}).
\end{align}

For the second term in \eqref{equation_bound_stage_gap}, employing a method similar to the one utilized in deriving the lower bound \eqref{equation_stager_0_2}, we can establish a symmetric upper bound:
\begin{align}
    &\phantom{=\ } \E\left[ \left ( W_{T_{r-1}}(i^*(\bar S_r)) - W_{T_{r-1}}(i^*(\bar S_r)+1) \right)\cdot \mathbbm{1}\{\mathcal E _r\} \right] \notag  \\
    &\le  \sum_{i: \bar S_i \in \mathrm{Ancestor}(\bar S_r)}  \left(f(i^*( \bar S_r)|\bar S_{i})-f(i^*(\bar S_r)+1|\bar S_{i}) \right) \E[\tau_{i} \cdot \mathbbm{1}\{\mathcal E _{i}\} ]+ (K-1)M \mathbb P (\mathcal E _{1} \setminus \mathcal E _{r})  \label{equation_stager_0_4}.
\end{align}

By substituting \eqref{equation_stager_0_3} and  \eqref{equation_stager_0_4} into \eqref{equation_bound_stage_gap}, we can deduce that for all stage $r \in [K-1]$,
\begin{align*}
    &\phantom{\, =\, } \left(f(i^*( \bar S_r)|\bar S_r)-f(i^*(\bar S_r)+1|\bar S_r) \right)\E[\tau_r\cdot \mathbbm{1}\{\mathcal E _r\} ]  \\  
    &\ge M -\sum_{i: \bar S_i \in \mathrm{Ancestor}(\bar S_r)}  \left(f(i^*( \bar S_r)|\bar S_{i})-f(i^*(\bar S_r)+1|\bar S_{i}) \right) \E[\tau_{i} \cdot \mathbbm{1}\{\mathcal E _{i}\} ]  - KM Q_*^{r+1} .
\end{align*}

\noindent\textbf{Step 2c (Putting it together).} To reduce clutter and ease the reading, for all $r\in [K-1]$, we introduce 
$$
\hat \tau_{\mathrm{U}} (\bar S_r) := \frac {1-\sum_{i: \bar S_i \in \mathrm{Ancestor}(\bar S_r)}  \left(f(i^*( \bar S_r)|\bar S_{i})-f(i^*(\bar S_r)+1|\bar S_{i}) \right) \hat \tau_{\mathrm{L}} (\bar S_i)  + \frac {K-2} {eM\log(1/p)} + (K-1) Q_*^{r+1} } {\left(f(i^*( \bar S_r)|\bar S_r)-f(i^*(\bar S_r)+1|\bar S_r) \right)}
$$
and 
$$
\hat \tau_{\mathrm{L}} (\bar S_r) := \frac {1-\sum_{i: \bar S_i \in \mathrm{Ancestor}(\bar S_r)}  \left(f(i^*( \bar S_r)|\bar S_{i})-f(i^*(\bar S_r)+1|\bar S_{i}) \right) \hat \tau_{\mathrm{U}} (\bar S_i) - K Q_*^{r+1} } {\left(f(i^*( \bar S_r)|\bar S_r)-f(i^*(\bar S_r)+1|\bar S_r) \right)}.
$$
It can be straightforwardly verified, based on the definition of $Q_*^r$, that both $\hat \tau_{\mathrm{U}} (\bar S_r) - \bar \tau (\bar S_r)$ and $\hat \tau_{\mathrm{L}} (\bar S_r) - \bar \tau (\bar S_r)$ are on the order of $O_M(1/M)$. Due to the upper and lower bounds of \eqref{equation_bound_stage_gap}, we can obtain that for all $r\in [K-1]$,
$$
\hat \tau_{\mathrm{L}} (\bar S_r) M \le \E[\tau_r\cdot \mathbbm{1}\{\mathcal E _r\} ] \le \hat \tau_{\mathrm{U}} (\bar S_r) M,
$$
which leads to 
\begin{align*}
     T^{\dagger} = \sum_{r=1}^{K-1} \E[\tau_r \cdot \mathbbm{1}\{\mathcal E _r\} ] &\le \sum_{r=1}^{K-1} \hat \tau_{\mathrm{U}} (\bar S_r) M \\
     &= \sum_{r=1}^{K-1} \bar \tau (\bar S_r) M  + \sum_{r=1}^{K-1} \left( \hat \tau_{\mathrm{U}} (\bar S_r) -\bar \tau (\bar S_r) \right)M \ =\  \frac{\log(1/p)M}{\JN(f)} + \sum_{r=1}^{K-1} \left( \hat \tau_{\mathrm{U}} (\bar S_r) -\bar \tau (\bar S_r) \right)M 
\end{align*}
as desired.

\vspace{0.2 cm}
\noindent\textbf{Step 3 (Bounding $T^{\ddagger}$).} The analysis of the first stage is straightforward since $\E[\tau_1 \cdot \mathbbm{1}\{\mathcal E _1^c\} ] = 0$. For any subsequent stage $r\in [K-1]\setminus \{1\}$, we temporarily condition on any fixed realization of previous stages such that $\mathcal E _r^c$  occurs. Let $i_{\min}$ and $i_{\max}$ represent the best and worst items within the active set $S_r$, respectively. Since 
$$
W_t(i_{\min}) - W_t(i_{\max}) - \left(f(i_{\min}| S_r)-f(i_{\max}|S_r) \right)t
$$
is a martingale for $t\ge T_{r-1}$, we can invoke the optional stopping theorem to obtain 
\begin{align*}
   \big(f(i_{\min}| S_r)-f(i_{\max}|S_r) \big) \E[\tau_r] \notag = \E \left [W_{ T_r}(i_{\min}) - W_{ T_r}(i_{\max})\right] - \E\left[ W_{T_{r-1}}(i_{\min}) - W_{T_{r-1}}(i_{\max})  \right] \le 2(K-1)M.
\end{align*}
Note that $ f(\min(S)|S) - f(\max(S)|S)\ge \frac {1-p}{1+p} $ for any customer preference $f\in\mathcal M_p$ and display set $S\in \mathcal S$. Therefore,
\begin{align*}
    \phantom{\, =\, }  \E[\tau_r ]  
    &\le \frac{2(K-1)M} {f(i_{\min}| S_r)-f(i_{\max}|S_r) } \le \frac{2(K-1)(1+p)M} {1-p},
\end{align*}
which is conditioned on any fixed realization of previous stages satisfying $\mathcal E _r^c$. By taking expectation with respect to all realizations of previous stages satisfying $\mathcal E _r^c$, we can get
 \begin{align*}
    \E[\tau_r\cdot \mathbbm{1}\{\mathcal E _r^c\}] 
    &\le \frac{2(K-1)(1+p)M} {1-p} \mathbb P(\mathcal E _r^c).
\end{align*}
Therefore, by Lemma~\ref{lemma_Error_r_ranking}, we can bound $T^{\ddagger}$ as
\begin{align*}
    T^{\ddagger} &= \sum_{r=2}^{K-1} \E[\tau_r \cdot \mathbbm{1}\{\mathcal E _r^c\} ] \le  \sum_{r=2}^{K-1} \frac{2(K-1)(1+p)MQ_*^r} {1-p}  .
\end{align*}

\noindent\textbf{Putting all pieces together}, the proof of Proposition~\ref{proposition_tau_general_ranking} is completed, and we have 
\begin{equation}
\label{equation_oM1_ranking}
\E [\tau] \le \frac{\log(1/p)M}{\JN(f)} + \underbrace{ \sum_{r=1}^{K-1} \left( \hat \tau_{\mathrm{U}} (\bar S_r) -\bar \tau (\bar S_r) \right)M }_{O_M(1)} + \underbrace{  \sum_{r=2}^{K-1} \frac{2(K-1)(1+p)MQ_*^r} {1-p} }_{o_M(1)}  .
\end{equation}
That finishes the proof. \Halmos 
\endproof

\begin{lemma}
For any customer preference $f\in\mathcal M_p$ and any $r \in [K]\setminus \{1\}$,
$$
\mathbb P(\mathcal E _r^c)  \le Q_*^r := \sum_{i=1}^{r-1}2K\left( {q_*^{i} }\right)^M 
$$
where $q_*^i\in (0,1)$ is defined  in the corresponding proof and does not depend on $M$.
\label{lemma_Error_r_ranking}
\end{lemma}

\proof{Proof of Lemma~\ref{lemma_Error_r_ranking}.}

Observe that within each stage, the increment of any voting score can be modeled by a binomial distribution. Building on this understanding, our proof is grounded in the idea that when voting scores closely align with their expected values, the behavior of our algorithm closely resembles the structure of the deterministic binary tree $\mathrm{DGBT}(f)$.  To articulate this formally, let \( \varepsilon > 0 \).
For all $i\in[K-1]$, we define $n_i := \lfloor M\bar \tau(\bar S_i) \rfloor$ and the event
$$
\mathcal A_i := \left\{ \forall j\in S_i, s\in [\min (n_i, \tau_i)] : \left | W_{s+T_{i-1}}(j)- W_{T_{i-1}}(j) - s f(j|S_i)\right|< \varepsilon M \right\},
$$
which indicates that for each item $j\in S_i$, the difference between the increment of its voting score during stage $i\in[K-1]$ and its expected value is at most $\varepsilon M$.

Note that within the partition criterion of our algorithm and the definition of $\mathcal A_i$, all the quantities related to the voting scores scale with $M$. Thus, in accordance with the definition of $\mathrm{DGBT}(f)$, there exists a sufficiently small $\varepsilon$ {(not depending on $M$)} such that for  any $r \in [K]\setminus \{1\}$, if $\mathcal A_i$ holds for all  $i \in [r-1]$, then $\mathcal E _r$ is true. Therefore, using Lemma~\ref{lemma_hoeffding1}, we can establish
\begin{align*}
   \mathbb P(\mathcal E _r^c)  &\le  \mathbb P\left( \bigcup _{i=1}^{r-1} \mathcal A_i^c \right)   
   = \sum_{i=1}^{r-1} \mathbb P\left( \bigcup  _{j=1}^{i-1} \mathcal A_j \cup \mathcal A_i^c \right) \ \le\  \sum_{i=1}^{r-1} \mathbb P( \mathcal E _i \cup \mathcal A_i^c) \\
   &\le \sum_{i=1}^{r-1} \mathbb{P} \left( S_i = \bar S_i \text{ and }\exists j\in S_i, s\in [\min (n_i, \tau_i)]: \left | W_{s+T_{i-1}}(j)- W_{T_{i-1}}(j) - s f(j|S_i)\right| \geq \varepsilon M \right) \\ 
   &\le  \sum_{i=1}^{r-1}  2 K \exp \left(-\frac {2 \varepsilon^2 M^2} {n_i}\right) \ \le\   \sum_{i=1}^{r-1}  2 K \exp \left(-\frac {\varepsilon^2 M} {\bar \tau(\bar S_i)}\right) \ =\  \sum_{i=1}^{r-1}2K\left( {q_*^{i} }\right)^M 
\end{align*}
where we set $q_*^{i} : = \exp \left(-{ \varepsilon^2 / \bar \tau(\bar S_i)  }\right)$.
\Halmos 
\endproof

\begin{lemma}[Hoeffding's maximal inequality \citealp{hoeffding1963probability}]
\label{lemma_hoeffding1}
Let $X_1, \ldots, X_n$ be independent Bernoulli random variables with mean $\mu$. Then for any $\varepsilon > 0$, it holds that
$
\mathbb{P} \left( \exists s\in [n]: \left | \sum_{t=1}^s \left( X_t-\mu\right)\right|\geq \varepsilon  \right) \leq 2 \exp \left(-\frac {2 \varepsilon^2} n\right).
$
\end{lemma}

\subsection{Proof of Proposition~\ref{proposition_error_ranking}}
\label{appendix_proposition_error_ranking}

\proof{Proof of Proposition~\ref{proposition_error_ranking}.}

First, according to the procedure outlined in Algorithm~\ref{algo_ranking}, we can decompose the error probability of interest as follows.  During each execution of the subroutine Algorithm~\ref{algo_partition}, if the active item set $S_{\mathrm{active}}$ consists of multiple items, it is partitioned into two parts: $S_{\mathrm{high}}$ and $S_{\mathrm{low}}$, with items in $S_{\mathrm{high}}$ considered better than those in $S_{\mathrm{low}}$. On the other hand, if $S_{\mathrm{active}}$ contains only one item, the ranking of that item is determined directly.  For any pair of items $i$ and $i'$, we say they are \textit{separated} if they are placed into different subsets during an execution of Algorithm~\ref{algo_partition}. Recall that we assume the global ranking of $f$ is the identity ranking $\sigma_*$. Therefore, if the output $\sigma_{\mathrm{out}} $ is incorrect, there must exist $i^* \in [K-1]$ such that items $i^*$ and $i^*+1$ are mistakenly separated. Specifically, item $i^*$ is partitioned into $S_{\mathrm{low}}$ while item $i^*+1$ is partitioned into $S_{\mathrm{high}}$. For any $i^* \in [K-1]$, we denote the event that items $i^*$ and $i^*+1$ are incorrectly separated as $\mathrm{Error}(i^*)$. Thus, we have
\begin{align}
 \mathbb P (\sigma_{\mathrm{out}} \neq \sigma_*) \le \sum_{i^*=1}^{K-1} \mathbb P (\mathrm{Error}(i^*)). \label{proposition_error_ranking_result1}
\end{align}

Consider any fixed $i^* \in [K-1]$. Let $\hat T_{i^*}$ denote the time step at which items $i^*$ and $i^*+1$ are separated, which is clearly a stopping time. Due to the display rule of our algorithm, prior to time $\hat T_{i^*}$, items $i^*$ and $i^*+1$ are always displayed together. Taking into account the definition of $\mathcal M_p$,  we can deduce that
$$
{\left(\frac 1 p\right)}^{W_t(i^*+1)-W_t(i^*)}
$$
is a supermartingale for $ t \leq \hat T_{i^*}$.  Consequently, by the optional stopping theorem, we have
\begin{align}
    1 \ge \mathbb E\left[{\left(\frac 1 p\right)}^{W_{\hat T_{i^*}}(i^*+1)-W_{\hat T_{i^*}}(i^*)}\right]
    \ge \mathbb P (\mathrm{Error}(i^*)) \cdot {\left(\frac 1 p\right)}^M \label{proposition_error_ranking_result2}
\end{align}
where the last inequality follows from the fact that if items $i^*$ and $i^*+1$ are incorrectly separated, then  $W_t(i^*+1)-W_t(i^*) \ge M$. Finally, by combining \eqref{proposition_error_ranking_result1} and \eqref{proposition_error_ranking_result2}, we arrive at 
\begin{align*}
 \mathbb P (\sigma_{\mathrm{out}} \neq \sigma_*) \le \sum_{i^*=1}^{K-1} p^M = (K-1)\cdot p^M,
\end{align*}
which completes the proof of Proposition~\ref{proposition_error_ranking}.
\Halmos 
\endproof

\tbl{
\subsection{Proof of Proposition~\ref{prop_OA_ranking}}
\label{appendix_prop_OA_ranking}

\proof{Proof of Proposition~\ref{prop_OA_ranking}.}
Fix any $f\in\mathcal M_p$. Without loss of generality, suppose that
the underlying ranking is $(1,2,\ldots,K)$. Recall from Section~\ref{subsection_ranking_Preliminaries} that
$\bar S_1,\ldots,\bar S_{K-1}$ are the non-singleton display sets
generated when the recursion defining $D(f,[K],\mathbf 0)$ is
unrolled.

We first establish the upper bound on $D(f,[K],\mathbf 0)$. Invoking Lemma~\ref{lemma:adjacent-gap-accounting}, it holds that 
\begin{equation}
\sum_{S\in
	\mathrm{Ancestor}(\bar S_r)\cup\{\bar S_r\}}
\bar\tau(S)
\Bigl(
f(i^*(\bar S_r)\mid S)
-
f(i^*(\bar S_r)+1\mid S)
\Bigr)
=1.
\label{eq:DGBT-unit-separation-special}
\end{equation}
We observe that the recursion partitions each contiguous display set until only
singletons remain. Consequently, the split indices
$
i^*(\bar S_1),\ldots,i^*(\bar S_{K-1})
$
are exactly $1,\ldots,K-1$, each appearing once. We next introduce an auxiliary lemma, which is a critical step of the proof.
\begin{lemma}
	\label{lemma_weighted_DGBT}
	Fix $f\in\mathcal M_p$ and, without loss of generality, suppose that
	its underlying ranking is $(1,2,\ldots,K)$. Let
	$\mathcal T=\mathrm{DGBT}(f)$. For every $i\in[K-1]$, there is a unique internal (i.e., non-leaf) node whose
	split index is $i$. Let it be denoted by $ S^*(i) $. Let $L_i$ and $R_i$ be its left and right children. Assign the weight parameter
	\[
	\alpha_i:=
	\begin{cases}
	1+p,
	& |L_i|=|R_i|=1,\\[1mm]
	1,
	& \min\{|L_i|,|R_i|\}=1<\max\{|L_i|,|R_i|\},\\[1mm]
	1-p,
	& |L_i|\ge2\ \text{and}\ |R_i|\ge2.
	\end{cases}
	\]
	to every split index $ i $. Then, for every internal node $S=[a,b]$ of $\mathcal T$,
	\begin{equation}
	\sum_{i=a}^{b-1}\alpha_i=|S|-1+p.
	\label{eq:tree_weight_sum}
	\end{equation}
	Moreover,
	\begin{equation}
	\sum_{i=a}^{b-1}
	\alpha_i
	\bigl(f(i\mid S)-f(i+1\mid S)\bigr)
	\ge 1-p.
	\label{eq:tree_weight_gap}
	\end{equation}
\end{lemma}

Intuitively, the weights $\alpha_i$ are carefully crafted with the purpose that is revealed by \eqref{eq:DGBT-weighted-sum} below. We choose them so that, for every
non-singleton display set $S=[a,b]$ generated by $\mathrm{DGBT}(f)$, \eqref{eq:tree_weight_sum} is satisfied.
Then the contribution of each $\bar\tau(S)$ on the right-hand side of
\eqref{eq:DGBT-weighted-sum} is at least $(1-p)\bar\tau(S)$. At the
same time, we want $\sum_{i=1}^{K-1}\alpha_i$ to be as small as possible,
since this sum forms the left-hand side of that identity and hence
determines the resulting upper bound on the total duration. The weights
constructed below satisfy these two requirements with
$
\sum_{i=1}^{K-1}\alpha_i=K-1+p,
$
which yields the desired sharp bound.

Now, let $\{\alpha_i\}_{i=1}^{K-1}$ be the weights from
Lemma~\ref{lemma_weighted_DGBT}. Multiply
\eqref{eq:DGBT-unit-separation-special} by
$\alpha_{i^*(\bar S_r)}$ and sum over $r\in[K-1]$. On the right-hand
side, this gives
$
\sum_{r=1}^{K-1}\alpha_{i^*(\bar S_r)}
=
\sum_{i=1}^{K-1}\alpha_i
$ due to the observation above.
On the left-hand side, collecting all terms involving the same
$\bar\tau(\bar S_j)$ and using the preceding observation gives $ \sum_{j=1}^{K-1}
\bar\tau(\bar S_j)
\sum_{i=\min(\bar S_j)}^{\max(\bar S_j)-1}
\alpha_i
\bigl(
f(i\mid\bar S_j)-f(i+1\mid\bar S_j)
\bigr). $ As a result,
\begin{align}
\sum_{i=1}^{K-1}\alpha_i
&=
\sum_{j=1}^{K-1}
\bar\tau(\bar S_j)
\sum_{i=\min(\bar S_j)}^{\max(\bar S_j)-1}
\alpha_i
\bigl(
f(i\mid\bar S_j)-f(i+1\mid\bar S_j)
\bigr).
\label{eq:DGBT-weighted-sum}
\end{align}

Applying Lemma~\ref{lemma_weighted_DGBT} first to the root
$\bar S_1=[K]$ and then to each $\bar S_j$, we obtain
$$
\sum_{i=1}^{K-1}\alpha_i=K-1+p
$$
and
\[
\sum_{i=\min(\bar S_j)}^{\max(\bar S_j)-1}
\alpha_i
\bigl(
f(i\mid\bar S_j)-f(i+1\mid\bar S_j)
\bigr)
\ge 1-p.
\]
Since $\bar\tau(\bar S_j)\ge0$, substituting these relations into
\eqref{eq:DGBT-weighted-sum} yields
\[
K-1+p
\ge
(1-p)\sum_{j=1}^{K-1}\bar\tau(\bar S_j) = (1-p)D(f,[K],\mathbf 0),
\]
where the last equality comes from $ \sum_{j=1}^{K-1}\bar\tau(\bar S_j) = D(f,[K],\mathbf 0) $ by unrolling the recursion in \eqref{equation_DfSW}. Consequently,
\begin{equation*}
D(f,[K],\mathbf 0)
\le
\frac{K-1+p}{1-p}.
\end{equation*}
As a result, \eqref{equation_JNf_definition} implies that
$
\JN(f)
=
\frac{\log(1/p)}{D(f,[K],\mathbf 0)}
\ge
\log(1/p)\frac{1-p}{K-1+p}
=
J_*^{\mathrm{OA}}.
$

\vskip 0.3 cm
It remains to show that the bound is attained by the OA preference $ f^{\mathrm{OA}} $. We do so by characterizing the tree DGBT($ f^{\mathrm{OA}} $) when unfolding $ D(f^{\mathrm{OA}},[K],\mathbf 0) $.

Recall that $\bar \tau(S) = \min_{i\in S, i\neq \max(S)} \tfrac {1-w(i)+w(i+1)} {f(i|S)-f(i+1|S)} \text{ and } 
i^*(S) := \argmin_{i\in S, i\neq \max(S)} \tfrac {1-w(i)+w(i+1)} {f(i|S)-f(i+1|S)}$ for every node $ S $ in DGBT($ f^{\mathrm{OA}} $).

Consider the root $\bar S_1=[K]$. For every $i\in[K-1]$,
$
f^{\mathrm{OA}}(i\mid[K])
-f^{\mathrm{OA}}(i+1\mid[K])
=
\frac{p^{i-1}(1-p)}{1+p+\cdots+p^{K-1}}.
$
Since the initial score vector is zero, the candidate duration associated
with the split at $i$ is
$
\frac{1+p+\cdots+p^{K-1}}{p^{i-1}(1-p)},
$
which is uniquely minimized at $i=1$. Therefore,
\[
i^*(\bar S_1)=1,
\qquad
\bar\tau(\bar S_1)=\tfrac{1+p+\cdots+p^{K-1}}{1-p}.
\]
The root thus separates $\{1\}$ from $[2,K]$. After the score update,
\[
w_{[2,K]}(i)-w_{[2,K]}(i+1)
=
\tfrac{1+p+\cdots+p^{K-1}}{1-p}
\tfrac{p^{i-1}(1-p)}{1+p+\cdots+p^{K-1}}
=
p^{i-1}, \quad \text{ for } i=2,\ldots,K-1.
\]
After that, one can verify using standard induction that for every $r=2,\ldots,K-1$, the recursion reaches
$\bar S_r=[r,K]$ with
$
w_{\bar S_r}(i)-w_{\bar S_r}(i+1)
=
p^{i-r+1}$ for every 
$i=r,\ldots,K-1,
$
and then splits $\bar S_r$ at $r$ after duration
$
\bar\tau(\bar S_r)=1 + p + \cdots + p^{K-r} .
$ In other words, in DGBT($ f^{\mathrm{OA}} $),
the non-singleton sets and their durations are
\[
[1,K],[2,K],\ldots,[K-1,K],
\]
with $ \bar\tau([1,K])=\frac{1+p+\cdots+p^{K-1}}{1-p} $ and 
\[
\bar\tau([r,K])=1 + p + \cdots + p^{K-r}  = \tfrac{1-p^{K-r+1}}{1-p} \quad \text{ for }\quad 
r=2,\ldots,K-1.
\]
Therefore,
\begin{align*}
D(f^{\mathrm{OA}},[K],\mathbf 0) = \sum_{j=1}^{K-1}\bar\tau([j,K]) 
&=
\tfrac{1+p+\cdots+p^{K-1}}{1-p}+\sum_{m=2}^{K-1}\tfrac{1-p^{m}}{1-p} =
\tfrac{
	1+p+\cdots+p^{K-1}+\sum_{m=2}^{K-1}(1-p^m)
}{1-p}=
\tfrac{K-1+p}{1-p}.
\end{align*}
It thus follows that
$
\JN(f^{\mathrm{OA}})
=
\frac{\log(1/p)}
{D(f^{\mathrm{OA}},[K],\mathbf 0)}
=
\log(1/p)\frac{1-p}{K-1+p}
=
J_*^{\mathrm{OA}}.
$
Together with the lower bound
on $\JN(f)$ established above, this proves that
$
\min_{f\in\mathcal M_p}\JN(f)
=
\JN(f^{\mathrm{OA}})
=
J_*^{\mathrm{OA}}.
$
\Halmos
\endproof

\vskip 0.2 cm

\proof{Proof of Lemma~\ref{lemma_weighted_DGBT}.}
Because $\mathcal T$ is an ordered full binary tree, the internal nodes
in the subtree rooted at $S=[a,b]$ are in one-to-one correspondence
with the adjacent split indices $a,\ldots,b-1$. The internal node corresponding to split index $i$, $S^*(i)$, is hence well defined. Define
\[
g(m):=
\begin{cases}
0,&m=1,\\
m-1+p,&m\ge2.
\end{cases}
\]
Consider an internal node $Q$ that splits at $q=i^*(Q)$, and denote its
two children by $Q_L$ and $Q_R$. The definition of $\alpha_q$ gives
\begin{equation}
\alpha_q
=
g(|Q|)-g(|Q_L|)-g(|Q_R|).
\label{eq:alpha_node_decomposition}
\end{equation}
This is by enumerating all the possibilities of the right-hand side equalling $1+p$, $1$, or $1-p$ according as both, exactly one, or neither of the two children is a singleton. Summing \eqref{eq:alpha_node_decomposition} over all internal nodes in
the subtree rooted at $S$ produces a telescoping sum. Every nonroot
internal node appears once with a positive sign and once with a
negative sign, while each leaf contributes $g(1)=0$. Consequently,
\[
\sum_{i=a}^{b-1}\alpha_i
=
g(|S|)
=
|S|-1+p,
\]
which proves \eqref{eq:tree_weight_sum}.

To prove \eqref{eq:tree_weight_gap}, we prove the following slightly stronger claim. For every internal
node $Q=[c,d]$ and every nonnegative vector
$(x_c,\ldots,x_d)$ satisfying
$
x_{j+1}\le p x_j
$ for every $ j=c,\ldots,d-1,
$
we have
\begin{equation}
\sum_{j=c}^{d-1}\alpha_j(x_j-x_{j+1})
\ge
(1-p)\sum_{j=c}^d x_j.
\label{eq:homogeneous_tree_gap}
\end{equation}
We proceed by induction on $|Q|$. If $|Q|=2$, then $\alpha_c=1+p$, and
$
\begin{aligned}
&(1+p)(x_c-x_{c+1})
-(1-p)(x_c+x_{c+1})=2(px_c-x_{c+1})\ge0.
\end{aligned}
$
Thus \eqref{eq:homogeneous_tree_gap} holds in the base case.

Now let $Q=[c,d]$ split at $k$ with children
$
Q_L=[c,k],
$ and 
$Q_R=[k+1,d].
$
Write
\[
X_L:=\sum_{j=c}^k x_j,
\qquad
X_R:=\sum_{j=k+1}^d x_j.
\]
According to the induction hypothesis, \eqref{eq:homogeneous_tree_gap} applies to every non-leaf child by restricting the vector $ (x_c, \ldots, x_d) $ to that child. Suppose first that neither child is a singleton. Then
$\alpha_k=1-p$, and hence
\[
\begin{aligned}
\sum_{j=c}^{d-1}\alpha_j(x_j-x_{j+1})
&\ge
(1-p)X_L+(1-p)X_R
+(1-p)(x_k-x_{k+1})\ge
(1-p)(X_L+X_R),
\end{aligned}
\]
where the last inequality follows from
$x_{k+1}\le p x_k\leq x_k$. Suppose next that $Q_L$ is a singleton and $Q_R$ is not. In this case,
$X_L=x_k$ and $\alpha_k=1$. Therefore,
\[
\begin{aligned}
\sum_{j=c}^{d-1}\alpha_j(x_j-x_{j+1})
&\ge
(1-p)X_R+x_k-x_{k+1}\ge
(1-p)X_R+(1-p)x_k=
(1-p)(X_L+X_R),
\end{aligned}
\]
where the second inequality uses $x_{k+1}\le p x_k$. Finally, suppose that $Q_R$ is a singleton and $Q_L$ is not. Then
$X_R=x_{k+1}$ and $\alpha_k=1$. Moreover,
$
x_k-x_{k+1}
\ge
\left(\frac1p-1\right)x_{k+1}
\ge
(1-p)x_{k+1}.
$
It follows that
\[
\begin{aligned}
\sum_{j=c}^{d-1}\alpha_j(x_j-x_{j+1})
&\ge
(1-p)X_L+x_k-x_{k+1}\ge
(1-p)(X_L+X_R).
\end{aligned}
\]
Combining the three possibilities completes the induction and proves
\eqref{eq:homogeneous_tree_gap}.

To obtain \eqref{eq:tree_weight_gap}, apply
\eqref{eq:homogeneous_tree_gap} with
$
x_i=f(i\mid S)$ for every $ i\in S $. Because $f$ is $ p $-separable and the underlying ranking is the identity ranking,
$
f(i+1\mid S)\le p f(i\mid S).
$
Moreover, $\sum_{i\in S}f(i\mid S)=1$. Hence
$
\sum_{i=a}^{b-1}
\alpha_i
\bigl(f(i\mid S)-f(i+1\mid S)\bigr)
\ge1-p,
$
as desired.
\Halmos
\endproof
}

\section{Lower Bound of Full-Ranking Identification}
\label{appendix_lower_bound_ranking}
\subsection{Proof of Theorem~\ref{theorem_lower_bound_ranking}}
\label{appendix_theorem_lower_bound_ranking}
The proof of Theorem~\ref{theorem_lower_bound_ranking} closely resembles that of Fact~\ref{fact_lower_bound} for the best-item identification problem, leveraging the change-of-measure argument. By employing this technique, the probabilities of a given event under different probability measures are related through the Kullback--Leibler (KL) divergence between the two measures. In our context, one can directly apply the change-of-measure argument for general hypothesis testing as outlined in \citet[Appendix B]{feng2022robust}, which can capture our setting as a special case.

\subsection{Proof of Proposition \ref{prop_OA_ranking_bound}}
\label{appendix_prop_OA_ranking_bound}
Recall that the OA preferences are uniquely defined only up to permutation. Let $\Sigma$ represent the collection of all  permutations. To enhance clarity and minimize confusion, for any permutation $\sigma \in \Sigma$, we refer to the corresponding OA preference as $f^{\mathrm{OA}}_\sigma$. In particular, $f^{\mathrm{OA}}_{\sigma_*}$  corresponds to the identity ranking $\sigma_* = (1,2,\ldots, K)$. Furthermore, we define $d_{S}(\sigma) :=  D_S(f^{\mathrm{OA}}_{\sigma_*} \| f^{\mathrm{OA}}_{\sigma})$ for any permutation $\sigma \in \Sigma$. Then the max-min optimization problem \eqref{eq_OA_J_maxmin} is equivalent to the following linear programming problem:
\begin{equation}
\begin{aligned}
    & \max _{\lambda, u} \quad u \\
    &\text{ s.t. } \, \sum_{S \in \mathcal{S}} d_S(\bar{\sigma}) \cdot \lambda(S) \geq u, \, \forall \bar{\sigma} \neq \sigma_* \\
    &\phantom{\text{ s.t. }}\, \sum_{S \in \mathcal{S}} \lambda(S)=1\\
    &\phantom{\text{ s.t. }}\, \lambda(S) \geq 0, \, \forall S \in \mathcal{S}.
\end{aligned}
\tag{LP-P}
\label{LP-P}
\end{equation}

To further analyze the problem \eqref{LP-P}, we can express its dual problem as follows:
\begin{equation}
\begin{aligned}
    & \min _{\mu, l} \quad l \\
    &\text{ s.t. } \, \sum_{\bar{\sigma} \neq \sigma_*} d_S(\bar{\sigma}) \cdot \mu(\bar{\sigma}) \leq l, \, \forall S \in \mathcal{S} \\
    &\phantom{\text{ s.t. }}\, \sum_{\bar{\sigma} \neq \sigma_*} \mu(\bar{\sigma})=1 \\
    &\phantom{\text{ s.t. }}\, \mu(\bar{\sigma}) \geq 0, \, \forall {\bar{\sigma} \neq \sigma_*} .
\end{aligned}
\tag{LP-D}
\label{LP-D}
\end{equation}

We now present several technical lemmas for these two LPs, with proofs presented later. First, we introduce an important subclass of rankings that is useful in our analysis. For each $m\in[K-1]$, let $\hat{\sigma}_m$ denote the ranking obtained from the identity ranking $\sigma_*$ by transposing the two consecutive items $m$ and $m+1$, so that $\hat\sigma_m(m)=m+1$, $\hat\sigma_m(m+1)=m$, and $\hat\sigma_m(j)=j$ for every $j\notin\{m,m+1\}$. In particular, $\hat\sigma_1=(2,1,3,\ldots,K)$ and $\hat\sigma_{K-1}=(1,\ldots,K-2,K,K-1)$. Lemma~\ref{lemma_counterpart_lemma6} offers a characterization of the value of $d_{S}(\cdot)$ for these rankings.

\begin{lemma}
\label{lemma_counterpart_lemma6}
Given any display set $S \in \mathcal{S}$ and $m\in[K-1]$,
$$
d_S\left(\hat{\sigma}_m\right)= \begin{cases} \log\left( \frac 1 p \right)\cdot  \frac {p^{i-1}(1-p)^2}{1-p^n} & \text { if } m \in S \text{ and } m+1 \in S \\ 0 & \text { otherwise }\end{cases} 
$$
where $i=\sigma_*(m|S)$ and $n=|S|$.
\end{lemma}

Lemmas~\ref{lemma_counterpart_lemma7} and \ref{lemma_counterpart_lemma8} below provide feasible solutions for \eqref{LP-P} and \eqref{LP-D}, respectively.

\begin{lemma}
\label{lemma_counterpart_lemma7}
Let $ \lambda^* $ be given by \eqref{eq:optimal OAM allocation for ranking} and 
$
u^* =   \log \left(\tfrac{1}{p}\right) \cdot \tfrac {1-p} {K-1+p}
$.  Then $(\lambda^*, u^*)$ is feasible in \eqref{LP-P}. 
\end{lemma}

\begin{lemma}
\label{lemma_counterpart_lemma8}
Let $ l^* =   \log \left(\tfrac{1}{p}\right) \cdot \tfrac {1-p} {K-1+p} $ and
$$
\mu^*(\bar{\sigma})= 
\begin{cases}\tfrac {1} {K-1+p} & \text { if } \bar{\sigma}=\hat{\sigma}_m \text{ for some } m \in [K-2]  \\ 
\tfrac {p+1} {K-1+p} & \text { if } \bar{\sigma}=\hat{\sigma}_{K-1}  \\
0 & \text { otherwise }
\end{cases}
$$
Then $(\mu^*, l^*)$ is feasible in \eqref{LP-D}. 
\end{lemma}

\vspace{0.3 cm}
\proof{Proof of Proposition~\ref{prop_OA_ranking_bound}.}
Since $u^* = l^* = \log \left(\frac{1}{p}\right) \cdot \frac {1-p} {K-1+p}$, by the weak duality theorem of linear programming, we can deduce that the optimal value of \eqref{LP-P} is precisely $\log \left(\frac{1}{p}\right) \cdot \frac {1-p} {K-1+p}$.
\Halmos 
\endproof
\vspace{0.3 cm}

\subsubsection*{Proofs of Auxiliary Lemmas}
\label{subsection_auxiliary_lemma} \proof{Proof of Lemma~\ref{lemma_counterpart_lemma6}.}
Note that the closed-form expression for $ f_\sigma^{\mathrm{OA}} $ is given by 
$$ f_\sigma^{\mathrm{OA}}(j|S) = \frac {1-p} {1-p^{|S|}} p^{\sigma(j|S)-1} \quad \text{ for all } S\in\mathcal S \text{ and } j\in S.$$

If either $m \not \in S$ or $m+1 \not \in S$, then for all $j\in S$, it holds that $\sigma_*(j|S) = \hat{\sigma}_m(j|S)$, which leads to $f_{\sigma_*}^{\mathrm{OA}}(j|S) = f_{\hat{\sigma}_m}^{\mathrm{OA}}(j|S)$. Therefore, this case is trivial since 
$$
\begin{aligned}
    d_S\left(\hat{\sigma}_m\right) & = D_S(f^{\mathrm{OA}}_{\sigma_*} \| f^{\mathrm{OA}}_{\hat{\sigma}_m}) = \sum_{j\in S}  f^{\mathrm{OA}}_{\sigma_*}(j|S) \log \frac {f^{\mathrm{OA}}_{\sigma_*}(j|S)} {f^{\mathrm{OA}}_{\hat{\sigma}_m}(j|S)} = 0.
\end{aligned}
$$

If $m  \in S$ and $m+1 \in S$, it suffices to consider items $m$ and $m+1$ since  $f_{\sigma_*}^{\mathrm{OA}}(j|S) = f_{\hat{\sigma}_m}^{\mathrm{OA}}(j|S)$ for all $j \in S\setminus \{m, m+1\}$. Using the notation $i=\sigma_*(m|S)$ and $n=|S|$, we have 
$
f_{\sigma_*}^{\mathrm{OA}}(m|S) = \tfrac {1-p} {1-p^{n}} p^{i-1}, \  f_{\sigma_*}^{\mathrm{OA}}(m+1|S) = \tfrac {1-p} {1-p^{n}} p^{i}, \ 
f_{\hat{\sigma}_m}^{\mathrm{OA}}(m|S) = \tfrac {1-p} {1-p^{n}} p^{i}, \ \text{ and } \  f_{\hat{\sigma}_m}^{\mathrm{OA}}(m+1|S) = \tfrac {1-p} {1-p^{n}} p^{i-1}.
$
Thus, we can obtain
$$
\begin{aligned}
    d_S\left(\hat{\sigma}_m\right) 
    &= \sum_{j\in S}  f^{\mathrm{OA}}_{\sigma_*}(j|S) \log \tfrac {f^{\mathrm{OA}}_{\sigma_*}(j|S)} {f^{\mathrm{OA}}_{\hat{\sigma}_m}(j|S)} = f_{\sigma_*}^{\mathrm{OA}}(m|S) \log \tfrac {f_{\sigma_*}^{\mathrm{OA}}(m|S)} {f_{\hat{\sigma}_m}^{\mathrm{OA}}(m|S)} + f_{\sigma_*}^{\mathrm{OA}}(m+1|S) \log \tfrac {f_{\sigma_*}^{\mathrm{OA}}(m+1|S)} {f_{\hat{\sigma}_m}^{\mathrm{OA}}(m+1|S)} = \log\left( \tfrac 1 p \right)\cdot  \tfrac {p^{i-1}(1-p)^2}{1-p^n} .
\end{aligned}
$$
This completes the proof of Lemma~\ref{lemma_counterpart_lemma6}.
\Halmos 
\endproof

\vspace{30pt}
\proof{Proof of Lemma~\ref{lemma_counterpart_lemma7}.}
First, it is straightforward to see for all $S \in \mathcal{S}$, $\lambda^*(S) \geq 0$. Second, by summing up $\lambda^*(S)$ for all $S \in \mathcal{S}$, we have 
$$
\begin{aligned}
\sum_{S \in \mathcal{S}} \lambda^*(S) &= \sum_{n=2}^{K-1} \frac {1-p^{K-n+1}} {K-1+p} + \frac {1-p^K} {(1-p)(K-1+p)} =  \sum_{n=2}^{K-1} \frac {1-p^{K-n+1}} {K-1+p} + \frac {1+p+\cdots+p^{K-1}} {K-1+p} 
= 1.
\end{aligned}
$$
Next, we will verify the first line of inequality constraints in \eqref{LP-P}. Utilizing the dominance result in Lemma~\ref{lemma_counterpart_lemma4}, it suffices to consider $\bar{\sigma} = \hat{\sigma}_m$ for all $m\in[K-1]$.  For any fixed $m\in[K-1]$, by Lemma~\ref{lemma_counterpart_lemma6}, we can verify that 
\begin{align*}
    &\phantom{\ =}\sum_{S \in \mathcal{S}} d_S(\hat{\sigma}_m) \cdot \lambda^*(S) \\&= \sum_{n=2}^{m} \log\left( \frac 1 p \right)\cdot  \frac {p^{m-n}(1-p)^2}{1-p^{K-n+1}} \cdot \frac {1-p^{K-n+1}} {K-1+p} + \log\left( \frac 1 p \right)\cdot  \frac {p^{m-1}(1-p)^2}{1-p^{K}} \cdot \frac {1-p^K} {(1-p)(K-1+p)}    \\
    &= \log\left( \frac 1 p \right)\cdot  \frac{(1-p)(1-p^{m-1})}{K-1+p} + \log\left( \frac 1 p \right)\cdot  \frac{(1-p)p^{m-1}}{K-1+p} \\
    &= \log \left(\frac{1}{p}\right) \cdot \frac {1-p} {K-1+p} 
    = J_*^{\mathrm{OA}},
\end{align*}
which concludes the proof of Lemma~\ref{lemma_counterpart_lemma7}.
\Halmos 
\endproof

\vspace{30pt}

\begin{lemma}
\label{lemma_counterpart_lemma4}
Consider any permutation $\sigma \neq \sigma_*$. There must exist $\hat{\sigma}_m$ with $m\in[K-1]$ such that 
$
d_S(\sigma) \ge d_S(\hat{\sigma}_m) 
$
for all $S\in \mathcal S$.
\end{lemma}

\proof{Proof of Lemma~\ref{lemma_counterpart_lemma4}.}
Since $\sigma \neq \sigma_*$, there must exist two items $i_1$ and $i_2$ such that $i_1 < i_2$ and $\sigma(i_1) = \sigma(i_2)+1$. For example, for $\sigma = (4, 1, 3, 2)$, we have $\sigma(1) = \sigma(3)+1$, which is not consistent with the identity ranking $\sigma_*$. Now we introduce another permutation $\bar \sigma$ such that
$$
\bar \sigma(i) =
\begin{cases}
    \sigma(i) & \text{if } i \neq i_1 \text{ and } i\neq i_2\\
    \sigma(i_2) & \text{if } i = i_1 \\
    \sigma(i_1) & \text{if } i = i_2. 
\end{cases}
$$
In other words, the rankings of items $i_1$ and $i_2$ are reversed. Consider any $S\in \mathcal S$. We claim that $d_S(\sigma) \ge d_S(\bar{\sigma}) $. 

If both $i_1 \not \in S$ and $i_2 \not \in S$, it is  trivially true that $d_S(\sigma) = d_S(\bar{\sigma})$.

If exactly one of $i_1$ and $i_2$ belongs to $S$, then the local ranking of $i_1$ or $i_2$ in $S$ is the same for $\bar \sigma$ and $\sigma$. Hence, we have $f_{\sigma}^{\mathrm{OA}}(j|S) = f_{\bar{\sigma}}^{\mathrm{OA}}(j|S)$ for all $j\in S$, which also implies $d_S(\sigma) = d_S(\bar{\sigma}) $.

If both $i_1 \in S$ and $i_2 \in S$, the desired result $d_S(\sigma) \ge d_S(\bar{\sigma}) $ is equivalent to 
$$
\begin{aligned}
&\phantom{\ =} f^{\mathrm{OA}}_{\sigma_*}(i_1|S) \log \tfrac {f^{\mathrm{OA}}_{\sigma_*}(i_1|S)} {f^{\mathrm{OA}}_{{\sigma}}(i_1|S)} + f^{\mathrm{OA}}_{\sigma_*}(i_2|S) \log \tfrac {f^{\mathrm{OA}}_{\sigma_*}(i_2|S)} {f^{\mathrm{OA}}_{{\sigma}}(i_2|S)} \\
&\ge f^{\mathrm{OA}}_{\sigma_*}(i_1|S) \log \tfrac {f^{\mathrm{OA}}_{\sigma_*}(i_1|S)} {f^{\mathrm{OA}}_{\bar{\sigma}}(i_1|S)} + f^{\mathrm{OA}}_{\sigma_*}(i_2|S) \log \tfrac {f^{\mathrm{OA}}_{\sigma_*}(i_2|S)} {f^{\mathrm{OA}}_{\bar{\sigma}}(i_2|S)} = f^{\mathrm{OA}}_{\sigma_*}(i_1|S) \log \tfrac {f^{\mathrm{OA}}_{\sigma_*}(i_1|S)} {f^{\mathrm{OA}}_{{\sigma}}(i_2|S)} + f^{\mathrm{OA}}_{\sigma_*}(i_2|S) \log \tfrac {f^{\mathrm{OA}}_{\sigma_*}(i_2|S)} {f^{\mathrm{OA}}_{{\sigma}}(i_1|S)}.
\end{aligned}
$$
Since $f^{\mathrm{OA}}_{\sigma_*}(i_1|S) \ge f^{\mathrm{OA}}_{\sigma_*}(i_2|S)$ and ${f^{\mathrm{OA}}_{{\sigma}}(i_1|S)} \le {f^{\mathrm{OA}}_{{\sigma}}(i_2|S)}$, by the rearrangement inequality, the above inequality holds.  Therefore, our claim that $d_S(\sigma) \ge d_S(\bar{\sigma}) $ for all $S\in \mathcal S$ is true.

Intuitively, the manipulation above resembles a single step in sorting. We can continue this \emph{sorting} process until arriving at the identity ranking $\sigma_*$. Consequently, a sequence of permutations can be obtained with the values of $d_S(\cdot)$ being non-increasing. For our previous example, we can derive the sequence as follows:
$$
(4, 1, 3, 2) \rightarrow (3, 1, 4, 2)  \rightarrow (2, 1, 4, 3)  \rightarrow (1, 2, 4, 3) \rightarrow (1, 2, 3, 4).
$$
By construction, the penultimate permutation is equal to $\hat{\sigma}_m$ with a certain $m\in[K-1]$. 
\Halmos 
\endproof

\vspace{.3 cm}

\proof{Proof of Lemma~\ref{lemma_counterpart_lemma8}.}
The nonnegativity and sum-to-one constraints are immediate from the
definition of $\mu^*$. It remains to verify the first constraint in
\eqref{LP-D}. Fix an arbitrary display set
$S=\{s_1<s_2<\cdots<s_n\}\in\mathcal S$, where $n=|S|$.
By Lemma~\ref{lemma_counterpart_lemma6},
$d_S(\hat\sigma_m)$ is nonzero only if
$\{m,m+1\}\subseteq S$. If $m=s_j$ and $m+1=s_{j+1}$,
then $\sigma_*(m\mid S)=j$. Therefore,
\begin{align*}
&\sum_{\bar\sigma\ne\sigma_*}
d_S(\bar\sigma)\mu^*(\bar\sigma)=
\tfrac{\log(1/p)(1-p)^2}
{(K-1+p)(1-p^n)}
\Bigg[
\sum_{\substack{j\in[n-1]:\\
		s_{j+1}=s_j+1,\;s_j\le K-2}}
p^{j-1}
+
(1+p)p^{n-2}
\mathbbm{1}\{\{K-1,K\}\subseteq S\}
\Bigg].
\end{align*}
If $\{K-1,K\}\not\subseteq S$, the expression in brackets is at
most $\sum_{j=0}^{n-2}p^j$. If $\{K-1,K\}\subseteq S$, this pair
occupies positions $n-1$ and $n$, and the expression in brackets is
at most
$
\sum_{j=0}^{n-3}p^j+(1+p)p^{n-2}
=
\sum_{j=0}^{n-1}p^j.
$
Thus, in either case, the expression in brackets is at most
$
\sum_{j=0}^{n-1}p^j=\frac{1-p^n}{1-p}.
$
Consequently,
\begin{align*}
\sum_{\bar\sigma\ne\sigma_*}
d_S(\bar\sigma)\mu^*(\bar\sigma)
&\le
\tfrac{\log(1/p)(1-p)^2}
{(K-1+p)(1-p^n)}
\tfrac{1-p^n}{1-p}=
\log(1/p)\frac{1-p}{K-1+p}
=
l^*.
\end{align*}
Hence $(\mu^*,l^*)$ is feasible in \eqref{LP-D}.
\Halmos 
\endproof

\vspace{0.3 cm}

\subsection{Proof of Proposition~\ref{prop_worstcase_ranking}}
\label{appendix_prop_worstcase_ranking}

\proof{Proof of Proposition~\ref{prop_worstcase_ranking}.}

Consider any OA preference $   f^{\mathrm{OA}} \in  \mathcal M_p^{\mathrm{OA}}$. By definition, we have $\min_{f\in \mathcal M_p} J_*(f) \le J_*(f^{\mathrm{OA}})$.  On the other hand, note that ${\widetilde{\mathcal{M}}}_p^{\mathrm{OA}}(f^{\mathrm{OA}}) \subseteq {\widetilde{\mathcal{M}}}_p(f^{\mathrm{OA}}) $. Therefore, 
$$
\begin{aligned}
\min_{f\in \mathcal M_p} J_*(f) \le  J_*(f^{\mathrm{OA}}) &= \sup_{\lambda\in \mathcal P (\mathcal S)} \inf_{f'\in {\widetilde{\mathcal{M}}}_p(f^{\mathrm{OA}})}  D_\lambda(f^{\mathrm{OA}} \| f')  \le \sup_{\lambda\in \mathcal P (\mathcal S)} \inf_{f'\in {\widetilde{\mathcal{M}}}_p^{\mathrm{OA}}(f^{\mathrm{OA}})}  D_\lambda(f^{\mathrm{OA}} \| f')  = J_*^{\mathrm{OA}},
\end{aligned}
$$
where the last equality follows from Proposition~\ref{prop_OA_ranking_bound}.
\Halmos 
\endproof

\section{Nested Elimination for Full Ranking}
\label{appendix_NE_ranking}
The pseudocode for \textsc{NE-Ranking} is presented in Algorithm~\ref{algo_NE_ranking}, which is also parameterized by a tuning parameter $M>0$.  From the perspective of sequential probability ratio tests (SPRT), at each time step $t$, \textsc{NE-Ranking} examines the following two hypotheses under OA instances:
\begin{align*}
&H_0: \hat i^* _t \text{ is the best item among $S_{\mathrm{active}}$,} \\
&H_1: \hat i^* _t \text{ is not the best item among $S_{\mathrm{active}}$.}
\end{align*}

We illustrate the system dynamics under \textsc{NE-Ranking} in Figure~\ref{process_NE-ranking}. As can be seen, \textsc{NE-Ranking} actually leads to more complex system dynamics than \textsc{NP}.  Specifically, the partition criteria of \textsc{NP} lead to closed polytopes while those of \textsc{NE-Ranking} lead to open boundaries (i.e., disjoint union of cones), which makes \textsc{NP} more amenable to theoretical analysis. The more refined boundary of \textsc{NP} also explains why it can demonstrate better non-asymptotic empirical performances.

\begin{algorithm}[h]
	\caption{Nested Elimination for Full Ranking (\textsc{NE-Ranking})}
	\label{algo_NE_ranking}
	\hspace*{0.02in} {\bf Input:} Tuning parameter $M>0.$

    \hspace*{0.02in} {\bf Output:}  The candidate full ranking $\sigma_{\mathrm{out}}$.
	\begin{algorithmic}[1]
		\STATE Initialize voting score $W_0(i)\leftarrow 0$ for all $i\in[K]$, active item set $S_{\mathrm{active}}\leftarrow [K]$, $t \leftarrow 0$.
		\WHILE{$|S_{\mathrm{active}}|>1$}
		\STATE Update the timer: $t \leftarrow t+1$.
		\STATE Display the active set $S_{\mathrm{active}}$,  and observe the choice $X_t \in S_{\mathrm{active}}$.
		\STATE Update voting scores based on $ X_t $:
		\begin{equation*}
		W_t(i) \leftarrow \begin{cases}
		W_{t-1}(i) + 1 & \text{if } i = X_t \\
		W_{t-1}(i) & \text{if }  i \neq X_t.
		\end{cases}
		\end{equation*}
            \STATE{Update the active set:}
            \\ (i) Find the most voted item $\hat i^* _t = \argmax_{i\in S_{\mathrm{active}}} W_t(i)$.
            \\ (ii) If {$W_t(\hat i^* _t)-\max_{i\neq \hat i^* _t}W_t(i)\ge M$}, then:
            \\ $ \phantom{\text{(ii)}}$ (a) Set $S_{\mathrm{active}}\leftarrow S_{\mathrm{active}}\setminus \{ \hat i^* _t \}$  \\ $ \phantom{\text{(ii)}}$ (b)  Assign the ranking of $\hat i^* _t$ to be $K-|S_{\mathrm{active}}|$.
            \ENDWHILE
            \STATE  Assign the ranking of the only element of $S_{\mathrm{active}}$ to be $K$.
	\end{algorithmic}
\end{algorithm}

\begin{figure}[htbp]
	\centering
	\hspace{-2cm}
	\includegraphics[width=.7\textwidth]{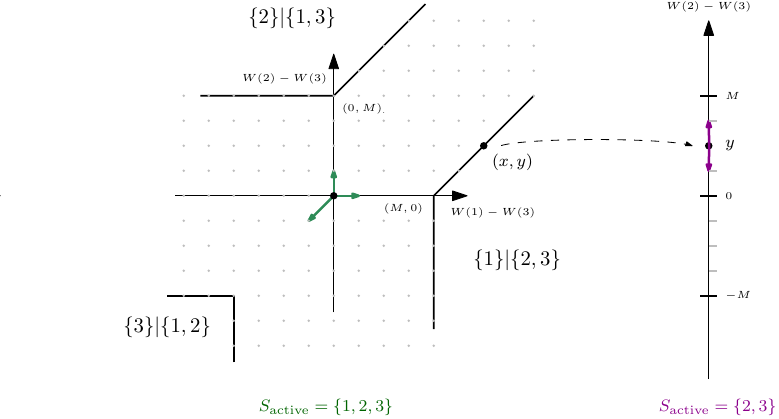}
	\caption{\textbf{A visualization of the system dynamics under \textsc{NE-Ranking}.} 
		Let \( K = 3 \) and consider the first stage where the full display set \( [3] = \{1, 2, 3\} \) is used. The projected state variables \( \big(W(1) - W(3),\, W(2) - W(3)\big) \) and the random walk dynamics are the same as \textsc{NE} illustrated in Figure \ref{fig:process_NE}. Under \textsc{NE-Ranking}, the first stage finishes when the random walk hits the boundary of any of the three cones in the plot, where each cone corresponds to a possibility of one item to be eliminated. For example, $ \{1\}|\{2,3\}$ means that item 1 is identified as the top-ranked one and eliminated from the active set. In the illustrated path, item~1 is eliminated and the active set is updated to $ \{2,3\}$, which is an event with high probability under any OA preference instance. In the next stage, the system dynamic is the same as that under \textsc{NP} in Figure~\ref{process_NP}. Depending on which endpoint the random walk hits, the resulting ranking is either $ (1,2,3) $ or $ (1,3,2) $.
	} 
\label{process_NE-ranking}
\end{figure}

\section{Additional Implementation Details and Numerical Results}
\label{appendix_experiments}
\subsection{Additional Implementation Details of \textsc{MTP}}

\noindent\textbf{Initialization.} At the initial time step, i.e., $t=1$, we randomly assign a ranking on the item set $[K]$ as the estimated global ranking, and use this ranking to determine the first display set. In fact, through extensive tests, we find that the initialization step has minimal influence on the overall performance.

\noindent\textbf{Stopping Rule. } For the threshold function used in the stopping rule of \textsc{MTP}, we follow the one indicated in the experimental parts of \citet{feng2022robust}, i.e., 
$
\log ((K-1)(K-1) !) + \log(1/\delta).
$

\noindent\textbf{Optimization Solver.} As we noted in Section~\ref{subsection_best_comparisons}, both the display rule and the stopping rule of \textsc{MTP} require solving some combinatorial optimization problems. Throughout the experiments, we follow the exact integer linear programming formulation in \citet{feng2022robust}, and utilize Gurobi 9.5.2 as the optimization solver.

\subsection{Construction of General Preferences}

Both the Netflix Prize and Debian Logo datasets are provided by PrefLib \citep{MaWa13a}. The Netflix Prize dataset \citep{BeLa07} consists of $823$ preference rankings over $4$ movies, while the Debian Logo dataset consists of $143$ preference rankings over $8$ candidates for the Debian logo. To generate one general (not worst-case) preference from each raw dataset, we consider each preference ranking as an interaction between the company and the customers, and hence the top-ranked item is treated as the choice of the customer. Next, we fit an MNL model (defined in Definition~\ref{definition_lucetype}), using maximum likelihood estimation. Finally, note that both  final outputs, $f_1$ and $f_2$, belong to $\mathcal M_p$ with $p=0.9$.

\begin{definition}[\citet{luce1959individual}]
\label{definition_lucetype}
    Under the multinomial logit (MNL) model, a preference $f$ is characterized by a non-negative vector of attraction scores $\{v_1,v_2,\ldots,v_K \}$, and the probability that item $i$ is chosen from the display set $S\in \mathcal S$ is
    $
    f(i|S) = \frac {v_i} {\sum_{j\in S} v_j}.
    $
\end{definition}

\tbl{\section{Comparison with the Full-Display Policy }\label{appendix:comparison with full display}
	\label{appendix_full_display} 
	
	A natural baseline for our problem is the \emph{full-display policy}, which offers the assortment $[K]$ to every customer. Because the feedback is i.i.d., a standard concentration argument can, in principle, also certify low error probability with $O(\log(1/\delta))$ samples. The purpose of this appendix is to make precise the sense in which \textsc{NE} and \textsc{NP} improve over this simple baseline, and to quantify the improvement. Following the tradition in the fixed-confidence literature (e.g., \citealp{garivier2016optimal,feng2022robust}) and consistent with the rest of this paper, the object of comparison is $ \lim_{\delta \downarrow 0} \mathbb{E}_f[\tau]/\log(1/\delta) $, the leading coefficient of $\log(1/\delta)$ as $\delta\downarrow0$, in the worst case over $\mathcal M_p$.  We compare against \emph{two} versions of the baseline, which isolate different sources of the advantage of \textsc{NE}/\textsc{NP}:
	\begin{itemize}
		\item[(a)] the full-display policy where the stopping rule and sample-complexity certificate come from Chernoff/Hoeffding concentration bounds on the empirical choice frequencies; and
		\item[(b)] the \emph{asymptotically optimal} full-display policy with the best possible stopping and decision rules.
	\end{itemize}
	In this way, we can cleanly separate the benefit of a tighter analysis or stopping rule from that of the \emph{display rule} itself. Table~\ref{tbl:benchmark} collects the resulting worst-case leading coefficients; the rest of this appendix derives its entries and reads off the separations.
	
	\begin{table}[t]
		\centering
		\small
		\setlength{\tabcolsep}{5pt}
		\renewcommand{\arraystretch}{1.9}
		\caption{Worst-case Leading Coefficients of $\log(1/\delta)$ over $\mathcal M_p$.}
		\label{tbl:benchmark}
		\begin{tabular}{lcc}
			\toprule
			Leading coeff.\ of $\log(1/\delta)$ & Best-Item Identification & $ \quad  $ Full-Ranking Identification \\
			\midrule
			(a)~Full display (concentration bound) & $\tfrac{2\,(1+p+\cdots+p^{K-1})^2}{(1-p)^2}$ & $+\infty$ \\
			(b)~Full display (optimal policy) & $\tfrac{1+p+\cdots+p^{K-1}}{\log(1/p)\,(1-p)}$ & $+\infty$ \\
			(c)~\textsc{NE}\,/\,\textsc{NP} (this paper) & $ \tfrac{1}{I_*^{\mathrm{OA}}} \left(\le \tfrac{1+2p}{\log(1/p)\,(1-p)}\right)$ & $\frac{1}{J_*^{\mathrm{OA}}} \left(=\tfrac{K-1+p}{\log(1/p)(1-p)}\right)$  \\
			(d)~Information-theoretic lower bound & $\tfrac{1}{I_*^{\mathrm{OA}}}$ & $\frac{1}{J_*^{\mathrm{OA}}}$ \\
			\bottomrule
		\end{tabular}
	\end{table}

	\subsection{Characterizing the Performance of Full-Display}
	
	\vskip 0.2cm
	\noindent  \textbf{The full-display policy implied by concentration inequalities.}  Under full display, at every round the learner offers $S=[K]$ and observes a choice $X_t\in[K]$ drawn i.i.d.\ from the categorical distribution with parameters $p_i:=f(i\mid[K])$, $i\in[K]$. Let $\widehat p_i(T):=\tfrac1T\sum_{t=1}^T\mathbf 1\{X_t=i\}$ denote the empirical frequencies after $T$ rounds, and let $p_{(1)}\ge p_{(2)}\ge\cdots\ge p_{(K)}$ be the decreasing rearrangement of $(p_i)$. The following concentration certificate thus follows.\footnote{We thank an anonymous referee for providing this concentration-based benchmark.}
	
	\begin{lemma}[Concentration-based full-display guarantee]
		\label{lemma:benchmark_concentration}
		Fix $f\in\mathcal M_p$ and let $\Delta(f)$ be the smallest choice-probability gap that the task must resolve:
		\[
		\Delta(f) := p_{(1)}-p_{(2)}\quad\text{(best-item identification)},
		\qquad
		\Delta(f) := \min_{i\neq j}\bigl|p_i-p_j\bigr|\quad\text{(full-ranking identification)}.
		\]
		If $\Delta(f)>0$, then the full-display policy that stops at
		\begin{equation}\label{eq:benchmark_T}
		T \;=\; \Big\lceil\, \tfrac{2}{\Delta(f)^2}\log\!\big(\tfrac1\delta\big) + \tfrac{2}{\Delta(f)^2}\log(2K)\,\Big\rceil
		\end{equation}
		and reports the order of $\bigl(\widehat p_i(T)\bigr)_{i\in[K]}$ is $\delta$-PAC. As a result, its leading coefficient of $\log(1/\delta)$ equals $2/\Delta(f)^2$.
	\end{lemma}
	
	\proof{Proof of Lemma~\ref{lemma:benchmark_concentration}.}
	By Hoeffding's inequality, $\mathbb P\bigl(|\widehat p_i(T)-p_i|\ge \Delta(f)/2\bigr)\le 2\exp(-T\Delta(f)^2/2)$ for each fixed $i$. A union bound over the $K$ coordinates shows that, with probability at least $1-2K\exp(-T\Delta(f)^2/2)$, every empirical frequency lies within $\Delta(f)/2$ of its mean. On that event, any two items whose true means differ by at least $\Delta(f)$ are ranked correctly by $(\widehat p_i(T))$: for best-item identification the top item's frequency exceeds every other by construction, and for full-ranking identification the same holds for every adjacent pair in the true order. Requiring $2K\exp(-T\Delta(f)^2/2)\le\delta$ yields \eqref{eq:benchmark_T}, whose leading term in $\log(1/\delta)$ has coefficient $2/\Delta(f)^2$.
	\Halmos
	\endproof
	
	\vskip 0.2 cm
	Since the learner does not know $ f $, and hence $ \Delta(f) $ \textit{a priori}, the tightest feasible $ \Delta $ is then governed by the smallest gap $\Delta$ that $\mathcal M_p$ can impose, i.e., 
	$$
	\Delta = \inf_{f \in \mathcal{M}_p} \Delta (f),
	$$
	which we analyze below.
	
	\begin{lemma}[Best-item-selection gap]\label{lemma:gap_selection}
		For every $f\in\mathcal M_p$, the top gap satisfies
		\[
		p_{(1)}-p_{(2)}\;\ge\;\frac{1-p}{1+p+\cdots+p^{K-1}},
		\]
		with equality at the ordinal-attraction (OA) instance $p_i^\star=\frac{p^{\,i-1}}{1+p+\cdots+p^{K-1}}$. Consequently, the worst-case concentration coefficient for best-item identification is
		\[
		\frac{2}{\Delta^2} = \frac{2}{\bigl(p^\star_{(1)}-p^\star_{(2)}\bigr)^2}
		\;=\;\frac{2\,(1+p+\cdots+p^{K-1})^2}{(1-p)^2}.
		\]
	\end{lemma}
	
	\proof{Proof of Lemma~\ref{lemma:gap_selection}.}
	Relabel items so that $p_1\ge p_2\ge\cdots\ge p_K$. By $p$-separability, $p_{i+1}\le p\,p_i$, hence $p_i\le p^{\,i-1}p_1$ and
	$1=\sum_{i=1}^K p_i\le p_1\sum_{i=0}^{K-1}p^{\,i}$, so $p_1\ge (1+p+\cdots+p^{K-1})^{-1}$. Therefore
	$p_1-p_2\ge (1-p)\,p_1\ge \tfrac{1-p}{1+p+\cdots+p^{K-1}}$, and both inequalities are tight exactly at $p_i^\star$.
	\Halmos
	\endproof
	
	\vskip 0.2cm 
	\begin{lemma}[Full-ranking-identification gap]\label{lemma:gap_ranking}
		Pick $ K \geq 3 $. For every $p\in(0,1)$, $$\Delta = \inf_{f\in\mathcal M_p}\ \min_{i\neq j}\bigl|p_i-p_j\bigr|=0.$$ Hence the concentration coefficient is \emph{vacuous} for full-ranking identification.
	\end{lemma}
	
	\proof{Proof of Lemma~\ref{lemma:gap_ranking}.}
	Fix an arbitrarily small $\varepsilon$, and set
	\[
	p_1=1-(1+p+\cdots+p^{K-2})\,\varepsilon,
	\qquad
	p_i=\varepsilon\,p^{\,i-2}\ \ (i=2,\ldots,K).
	\]
	Then $\sum_{i=1}^K p_i=1$; the ratios satisfy $p_{i+1}/p_i=p$ for $i\ge 2$, and $p_2/p_1\le p$ for $\varepsilon$ sufficiently small. Hence $f\in\mathcal M_p$ with strict ranking $(1,2,\ldots,K)$. Yet the gap $p_2-p_3=\varepsilon(1-p)\to0$ as $\varepsilon\downarrow0$, so $\min_{i\neq j}|p_i-p_j|$ can be made arbitrarily small.
	\Halmos
	\endproof
	\vskip 0.2 cm
	
	In other words, Lemma~\ref{lemma:gap_ranking} says that $p$-separability, which describes the pairwise choice probability \emph{ratios}, is a  strictly milder condition than separability in the pairwise choice-probability \emph{gaps}. As a consequence, no stopping rule of the concentration form \eqref{eq:benchmark_T} is $\delta$-PAC over all of $\mathcal M_p$, which is why row~(a) of Table~\ref{tbl:benchmark} is vacuous in the second column. 
	
	\vskip 0.2cm
	\noindent  \textbf{The asymptotically optimal full-display policy.} The concentration certificate is loose on two counts: Chernoff/Hoeffding bounds are not exponent-optimal in $\delta$, and threshold stopping can be conservative. To this end, \cite{feng2022robust} characterizes the worst-case asymptotically optimal full display policy (Proposition 5), which leads to a worst-case leading coefficient of
	\begin{align*}
	\frac{1}{I^F}, \quad \text{ where } \quad  I^F = (1-p)\log \left(\tfrac{1}{p}\right) \frac{1}{1+p+\cdots + p^{K-1}}.
	\end{align*}

	\begin{lemma}[Performance of Full-Display in the Ranking-Identification Problem]
		\label{lemma:full_display_ranking_infinite}
		Suppose $K\geq 3$. Any PAC family of full-display policies for full-ranking identification satisfies
		\[
		\sup_{f\in\mathcal M_p}
		\liminf_{\delta\downarrow0}
		\frac{\mathbb E_f[\tau]}{\log(1/\delta)}
		=+\infty.
		\]
	\end{lemma}
	
	\proof{Proof of Lemma~\ref{lemma:full_display_ranking_infinite}.}
	Following the construction in Lemma~\ref{lemma:gap_ranking}, pick any preference instance $ f^\varepsilon \in \mathcal{M}_p $ with full-display choice probabilities
	$
	p_1^\varepsilon
	=
	1-(1+p+\cdots+p^{K-2})\varepsilon,
	$ and 
	$p_i^\varepsilon
	=
	\varepsilon p^{\,i-2},
	$ for $ i=2,\ldots,K
	$, where $ \varepsilon $ is sufficiently small. For example, one could pick $ f^\varepsilon $ to be an MNL instance with attraction scores $ (p_1^\varepsilon, \ldots, p_K^\varepsilon) $. Now, consider an alternative preference $ \widetilde{f^\varepsilon} $ with full-display choice probabilities $\widetilde p^\varepsilon$ obtained by transposing the last two
	coordinates of $p^\varepsilon$:
	\[
	\widetilde p_i^\varepsilon
	=
	\begin{cases}
	p_K^\varepsilon, & i=K-1;\\
	p_{K-1}^\varepsilon, & i=K;\\
	p_i^\varepsilon, & \text{otherwise}.
	\end{cases}
	\]
	Then $\widetilde f^\varepsilon\in\mathcal M_p$ and has global ranking
	$(1,\ldots,K-2,K,K-1)$. Under full display, the choice distributions of
	$f^\varepsilon$ and $\widetilde f^\varepsilon$ are respectively
	$p^\varepsilon$ and $\widetilde p^\varepsilon$. Their KL divergence therefore
	satisfies
	\begin{align*}
	D_{[K]}\bigl(f^\varepsilon\Vert\widetilde f^\varepsilon\bigr)
	&=
	p_{K-1}^\varepsilon
	\log\left(\tfrac{p_{K-1}^\varepsilon}{p_K^\varepsilon}\right)
	+
	p_K^\varepsilon
	\log\left(\tfrac{p_K^\varepsilon}{p_{K-1}^\varepsilon}\right) =
	\varepsilon p^{K-3}(1-p)\log\left(\tfrac1p\right).
	\end{align*}
	
	Now consider any PAC family of full-display policies and introduce the event 
	$
	A:=\bigl\{\sigma_{\mathrm{out}}=\sigma_{f^\varepsilon}\bigr\}.
	$
	Since the full-display policy is $\delta$-PAC, we have
	$
	\mathbb P_{f^\varepsilon}(A)\geq1-\delta,
	$ and $
	\mathbb P_{\widetilde f^\varepsilon}(A)\leq\delta .
	$
	Applying the standard change-of-measure argument, for $\delta<1/2$,
	\[
	\mathbb E_{f^\varepsilon}[\tau]\,
	D_{[K]}\bigl(f^\varepsilon\Vert\widetilde f^\varepsilon\bigr)
	\geq
	\operatorname{kl}(1-\delta,\delta),
	\]
	where $\operatorname{kl}(\cdot,\cdot)$ denotes the binary relative entropy and satisfies $
	\lim_{\delta\downarrow0}
	\frac{\operatorname{kl}(1-\delta,\delta)}
	{\log(1/\delta)}
	=1. 
	$ Hence
	it follows that
	\[
	\liminf_{\delta\downarrow0}
	\frac{\mathbb E_{f^\varepsilon}[\tau]}
	{\log(1/\delta)}
	\geq
	\frac{1}
	{\varepsilon p^{K-3}(1-p)\log(1/p)}.
	\]
	The right-hand side diverges as $\varepsilon\downarrow0$. Since
	$f^\varepsilon\in\mathcal M_p$ for every sufficiently small $\varepsilon$,
	taking the supremum over $f\in\mathcal M_p$ proves the result.
	\Halmos
	\endproof

	\subsection{Magnitude of the Improvement}\label{append:magnitude of the Improvement}
	
	Let us now turn to how \textsc{NE} and \textsc{NP} improve upon these two versions of full-display policies, as reflected in Rows~(c) and (d) of Table~\ref{tbl:benchmark}. For \textsc{NE}, Theorem~\ref{theorem_fixedconfidence} combined with Proposition~\ref{prop_OA} and the bounds \eqref{eq:approximating IOA} on $I_*^{\mathrm{OA}}$ give worst-case coefficient $\frac{1}{I_*^{\mathrm{OA}}}\le \tfrac{1+2p}{\log(1/p)(1-p)}$, and the matching lower bound $\frac{1}{I_*^{\mathrm{OA}}}$ is in row~(d). For \textsc{NP}, Theorem~\ref{theorem_ranking_fixedconfidence} with Proposition~\ref{prop_OA_ranking} gives a worst-case leading coefficient of $ \frac{1}{J_*^{\mathrm{OA}}} = \tfrac{K-1+p}{\log(1/p)(1-p)}$, which matches the worst-case lower bound.
	
	\medskip
	\noindent\textbf{Best-item identification.} Dividing the first-column entries, we obtain
	\[
	\frac{\text{(a)}}{\text{(c)}}\;\ge\;\frac{2\,(1+p+\cdots+p^{K-1})^2\,\log(1/p)}{(1-p)(1+2p)}
	\quad \text{ and } \quad 
	\frac{\text{(b)}}{\text{(c)}}\;\ge\;\frac{1+p+\cdots+p^{K-1}}{1+2p}.
	\]
	Both ratios are therefore \emph{arbitrarily large} when $p$ is close to one and $K$ is large, at rates of $\Omega(\min\{K^2,1/(1-p)^2\})$ and $\Omega(\min\{K,1/(1-p)\})$, respectively. Intuitively, the disadvantage of full display happens because it dilutes choices across non-top items and renders each observation less informative about the top-ranked item. In comparison, \textsc{NE} progressively focuses its comparisons as the horizon proceeds. The benefit is most pronounced when $p$ is close to one, where comparison noise is most severe.

	\medskip
	\noindent\textbf{Full-ranking identification.} Here $\text{(a)}/\text{(c)}=+\infty$ by Lemma~\ref{lemma:gap_ranking} and $\text{(b)}/\text{(c)}=+\infty$ by Lemma~\ref{lemma:full_display_ranking_infinite}. In other words, the worst-case performance of  full-display policies is essentially vacuous, yet \textsc{NP} provides strong performance guarantees.  Intuitively, that is because \textsc{NP} concentrates comparisons on small contiguous subsets in which adjacent items compete directly.

	\medskip  \noindent\textbf{Numerical Experiments.}
	To complement the theoretical comparison, we also include Figures~\ref{figure_worst_case_fulldisplay-R1} and~\ref{figure_worst_case_ranking_fulldisplay-R1}, which
	compare the empirical stopping times of \textsc{NE} and \textsc{NP} against the full-display benchmarks
	at OA instances. For best-item identification we include both benchmarks; for full-ranking
	identification we report only the asymptotically optimal full-display policy, since the concentration-based full-display policy is much worse and off the plotted scale.

	\begin{figure}[tb]
		\begin{minipage}{1.0\columnwidth}
			\centering
			\hspace{-.7 cm}
			\subfigure[$K=5, p=0.9$.]{
				\begin{minipage}[b]{0.24\textwidth}
					\includegraphics[width=1\textwidth]{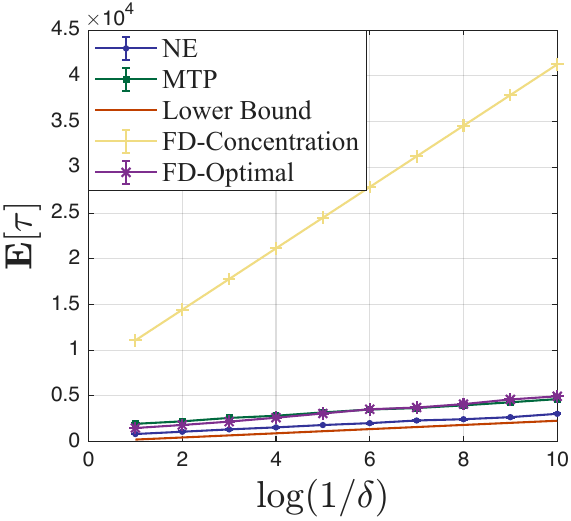} 
				\end{minipage}
			}
			\hspace{-.2 cm}
			\subfigure[$K=15, p=0.9$.]{
				\begin{minipage}[b]{0.24\textwidth}
					\includegraphics[width=1\textwidth]{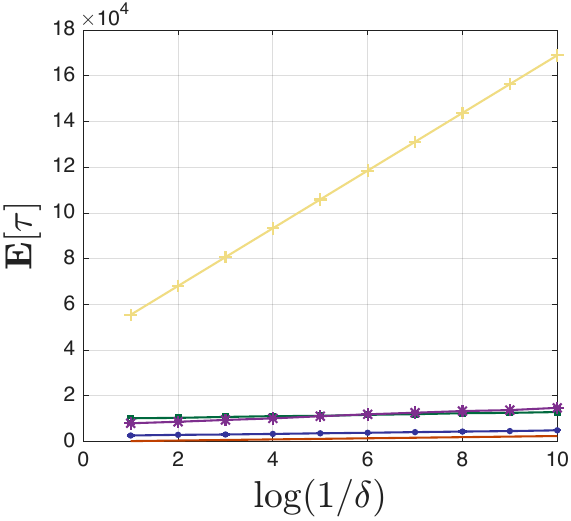} 
				\end{minipage}
			}
			\hspace{-.2 cm}
			\subfigure[$K=5, p=0.6$.]{
				\begin{minipage}[b]{0.24\textwidth}
					\includegraphics[width=1\textwidth]{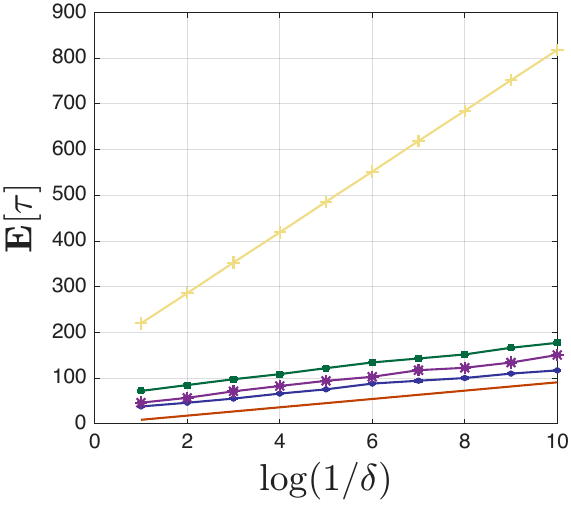} 
				\end{minipage}
			}
			\hspace{-.2 cm}
			\subfigure[$K=15, p=0.6$.]{
				\begin{minipage}[b]{0.24\textwidth}
					\includegraphics[width=1\textwidth]{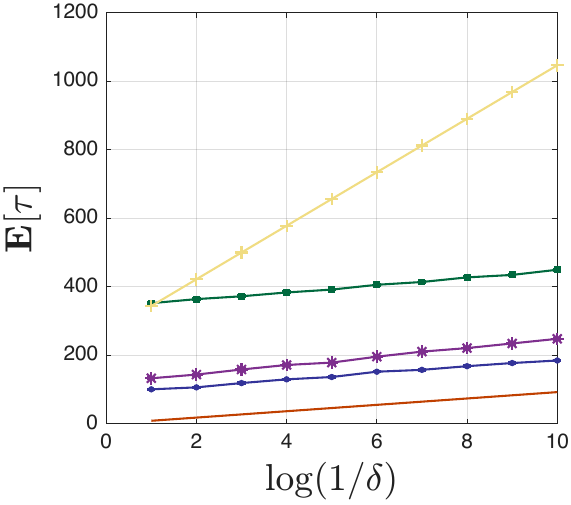} 
				\end{minipage}    
			}
		\end{minipage}
		\vspace{3pt}
		\caption{\textbf{Best-Item Identification: Empirical stopping times of full-display benchmarks under OA preference instances.} Different values of $\delta$, $ K $, $ p $ are considered.} \label{figure_worst_case_fulldisplay-R1}
	\end{figure}
	\begin{figure}[tb]
		\begin{minipage}{1.0\columnwidth}
			\centering
			\hspace{-.7 cm}
			\subfigure[$K=5, p=0.9$.]{
				\begin{minipage}[b]{0.24\textwidth}
					\includegraphics[width=1\textwidth]{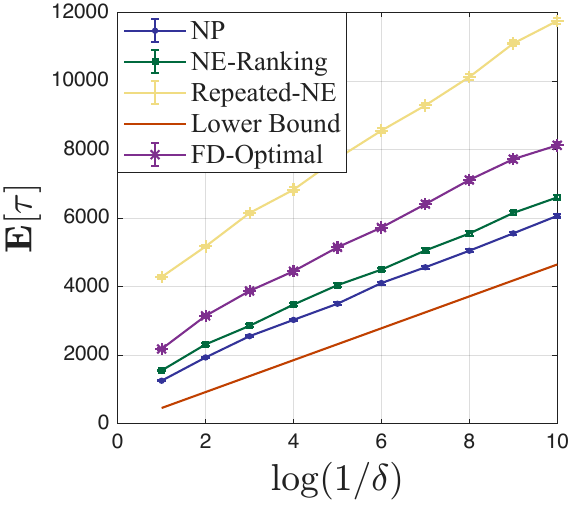} 
				\end{minipage}
			}
			\hspace{-.2 cm}
			\subfigure[$K=15, p=0.9$.]{
				\begin{minipage}[b]{0.24\textwidth}
					\includegraphics[width=1\textwidth]{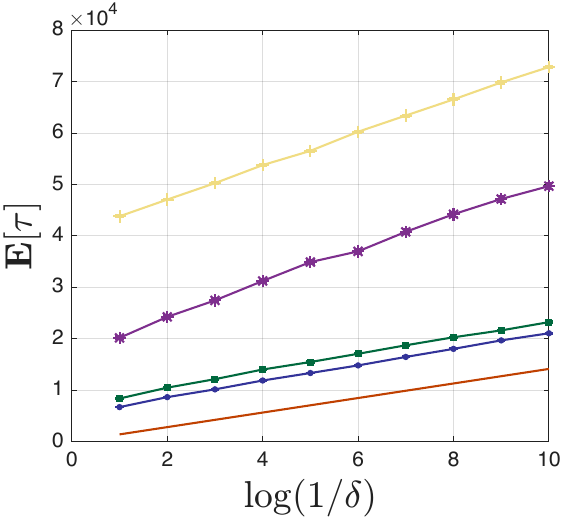} 
				\end{minipage}
			}
			\hspace{-.2 cm}
			\subfigure[$K=5, p=0.6$.]{
				\begin{minipage}[b]{0.24\textwidth}
					\includegraphics[width=1\textwidth]{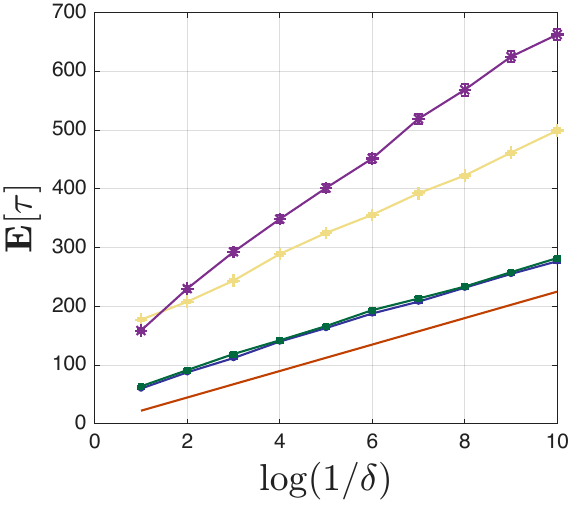} 
				\end{minipage}
			}
			\hspace{-.2 cm}
			\subfigure[$K=15, p=0.6$.]{
				\begin{minipage}[b]{0.24\textwidth}
					\includegraphics[width=1\textwidth]{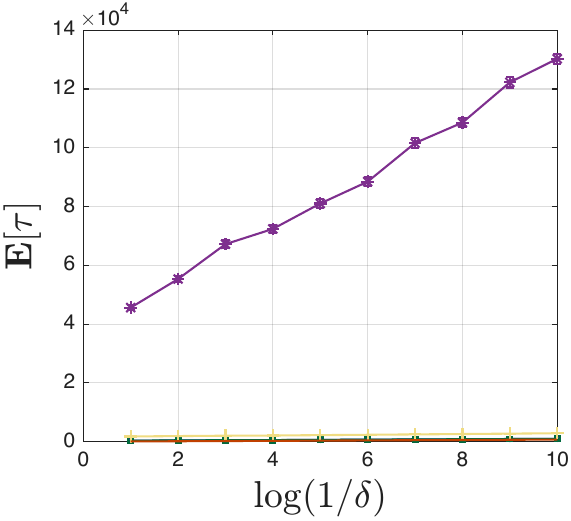} 
				\end{minipage}    
			}
		\end{minipage}
		\vspace{3pt}
		\caption{\textbf{Full-Ranking Identification: Empirical stopping times of full-display benchmarks under OA preference instances.} Different values of $\delta$, $ K $, $ p $ are considered.} \label{figure_worst_case_ranking_fulldisplay-R1}
	\end{figure}
	The numerical results are consistent with the theoretical comparison above. In the best-item identification problem, the concentration-based full-display policy requires substantially more samples than \textsc{NE}, reflecting the conservativeness of using a uniform concentration certificate and a deterministic sample size. The optimal full-display policy improves upon this concentration-based benchmark, but its slope as a function of $\log(1/\delta)$ still tends to be larger than that of \textsc{NE}. The same pattern is even more visible in the full-ranking identification problem: the optimal full-display benchmark has a noticeably larger slope than \textsc{NP}, especially as $K$ grows or as $p$ decreases.
}

\tbl{
\section{Proofs for Section~\ref{sec_extension_capacity}}

\subsection{Proofs for Best-Item Identification}
\label{appendix_theorem_fixedconfidence_capacity}

\proof{Proof of Proposition~\ref{prop:constant optimality}.} Recall that $ \log\left(\tfrac{1}{p}\right)(1-p)\tfrac{1}{1+2p}
\leq
I_{*, \kappa}^{\mathrm{OA}} $. As a result, $ \IHNE = I_{*, \kappa}^{\mathrm{OA}} /N_{\mathrm{call}} \geq \log\left(\tfrac{1}{p}\right)(1-p)\tfrac{1}{1+ 2p} \lceil \frac {K-1} {\kappa - 1} \rceil^{-1}  $. As a result, 
\begin{align*}
\frac{\IHNE}{I_{\kappa}^{\mathrm{worst}}}
&\geq
\frac{
	\log\left(\tfrac{1}{p}\right)(1-p)\tfrac{1}{1+2p}
	\left\lceil \tfrac{K-1}{\kappa-1}\right\rceil^{-1}
}{
	\frac{
		\log(1/p)\left(
		(\kappa-1)(1+p^\kappa)
		-\frac{2(p-p^\kappa)}{1-p}
		\right)
	}{
		(K-1)(1-p^\kappa)
	}
}\ =\ 
\frac{
	(K-1)\sum_{j=0}^{\kappa-1}p^j
}{
	(1+2p)
	\left\lceil \frac{K-1}{\kappa-1}\right\rceil
	\sum_{n=1}^{\kappa-1}n(\kappa-n)p^{n-1}
},
\end{align*}
where we use
\[
(\kappa-1)(1+p^\kappa)-\frac{2(p-p^\kappa)}{1-p}
=
(1-p)^2\sum_{n=1}^{\kappa-1}n(\kappa-n)p^{n-1}.
\]
Since \(\lceil x\rceil\le x+1\) and \(n(\kappa-n)\le \kappa^2/4\), we have
$
\frac{K-1}{
	\left\lceil \frac{K-1}{\kappa-1}\right\rceil
}
\ge
\frac{(K-1)(\kappa-1)}{K+\kappa-2}
$
and
$
\sum_{n=1}^{\kappa-1}n(\kappa-n)p^{n-1}
\le
\frac{\kappa^2}{4}\sum_{j=0}^{\kappa-1}p^j.
$
Therefore, the ratio can be further lower bounded by
\[
\frac{\IHNE}{I_{\kappa}^{\mathrm{worst}}}
\ge
\frac{4(K-1)(\kappa-1)}
{(1+2p)\kappa^2(K+\kappa-2)}
\ge
\frac{2(\kappa-1)}
{(1+2p)\kappa^2},
\]
where the last inequality follows from \(\kappa\le K\). Since \(p<1\), we have \(1+2p<3\), and hence \(\IHNE/I_{\kappa}^{\mathrm{worst}}\ge 2(\kappa-1)/(3\kappa^2)\), as claimed. \Halmos
\endproof

\subsubsection*{Proof of Theorem~\ref{theorem_fixedconfidence_capacity}}
Similar to the analysis of \textsc{NE} and \textsc{NP}, we proceed by establishing the bounds on the expected stopping time and the error probability with respect to the tuning parameter $M$. These intermediate results are formally stated in Proposition~\ref{proposition_tau_general_capacity} and Proposition~\ref{proposition_error_capacity} below. The proof of Theorem~\ref{theorem_fixedconfidence_capacity} follows directly from combining these two results.

\vspace{0.2 cm}
\begin{proposition}[Expected stopping time of \textsc{HNE}]
	\label{proposition_tau_general_capacity}
	For any customer preference $f\in\mathcal M_p$,  \textsc{HNE} ensures that
	$
	\E [\tau] \le \frac{\log(1/p)N_{\mathrm{call}} M} {{I_{*, \kappa}^{\mathrm{OA}}}} + o_M(1).
	$
\end{proposition}
\vspace{0.2 cm}

\vspace{0.2 cm}
\begin{proposition}[Error probability of \textsc{HNE}]
	\label{proposition_error_capacity}
	For any customer preference $f\in\mathcal M_p$, \textsc{HNE} outputs an item $i_{\mathrm{out}} $ satisfying $
	\mathbb P(i_{\mathrm{out}}\neq 1)\le N_{\mathrm{call}}\beta_{\kappa}p^M.
	$
\end{proposition}
\vspace{0.2 cm}

\vspace{0.2 cm}
\proof{Proof of Proposition~\ref{proposition_tau_general_capacity}.}
Recall that \textsc{HNE} organizes the item set $[K]$ into a tournament structure, and the total number of calls to the subroutine is exactly $N_{\mathrm{call}} =\lceil \frac {K-1} {\kappa - 1} \rceil$. For each subroutine call, the batch size is at most $\kappa$.
According to Proposition~\ref{prop_OA} and Proposition~\ref{proposition_tau_general}, the expected sample complexity for identifying the winner within any specific batch is upper-bounded by the complexity of the worst-case OA instance of size $\kappa$, denoted by $I_{*, \kappa}^{\mathrm{OA}}$.
Specifically, for any batch, the expected sample complexity is bounded by $\frac{\log(1/p)M} {{I_{*, \kappa}^{\mathrm{OA}}}} + o_M(1)$. 
This remains valid for the final call, as $I_{*, \kappa}^{\mathrm{OA}}$ is monotone non-increasing in $\kappa$.
By the linearity of expectation, summing over all subroutine calls yields the desired bound.
\Halmos
\endproof

\vspace{0.4 cm}

\proof{Proof of Proposition~\ref{proposition_error_capacity}.} Applying Proposition~\ref{proposition_error}, the probability that \textsc{NE} eliminates the best item from a set of size $\kappa$ is bounded by $\beta_{\kappa} p^M$.  Applying the union bound over the $ N_{\mathrm{call}} $ calls of \textsc{NE}, we conclude that 
$
\mathbb P (i_{\mathrm{out}} \neq 1 ) \leq  N_{\mathrm{call}} \, \beta_{\kappa} p^M.$\Halmos
\endproof

\vskip 0.3cm 
\subsubsection*{Proof of Theorem~\ref{thm_worstcase_capacity}}
\label{appendix_thm_worstcase_capacity}

We begin by stating the information-theoretic lower bound adapted for the capacity-constrained setting, which parallels Fact~\ref{fact_lower_bound} in the unconstrained case. Let $\mathcal P (\mathcal S^{\kappa})$ denote the collection of all probability distributions on the collection of display sets $\mathcal S^{\kappa}$ subject to the capacity constraint. That is, the support of any $\lambda \in \mathcal P (\mathcal S^{\kappa})$ is restricted to display sets with size at most $\kappa$.

\vspace{0.2 cm}
\begin{theorem}[Information-theoretic lower bound with capacity constraints]
\label{theorem_lower_bound_capacity}
For any preference $f\in\mathcal M_p$, let 
\begin{align}\label{eq:max-min_capacity}
I_{*,\kappa}(f):= \sup_{\lambda\in \mathcal P (\mathcal S^{\kappa})} \ \inf_{f'\in \overline{\mathcal{M}}_p(f)}  D_\lambda(f \| f').
\end{align}
Then any $\delta$-PAC policy with capacity constraint $\kappa$ satisfies
$
\E [\tau] \ge  \frac {\log(1/\delta) - \log 2.4} {I_{*,\kappa}(f)}.
$
\end{theorem}
\vspace{0.2 cm}

The proof of Theorem~\ref{theorem_lower_bound_capacity} follows the standard change-of-measure argument akin to the proof of Theorem~\ref{theorem_lower_bound_ranking}, and is omitted here for brevity. 

\vskip 0.2 cm
To establish the worst-case sample complexity stated in Theorem~\ref{thm_worstcase_capacity}, it suffices to show that there exists a preference instance $f \in \mathcal M_p$ such that the hardness quantity $I_{*,\kappa}(f)$ is upper bounded by $I_{\kappa}^{\mathrm{worst}}$. To this end, we focus on the class of OA preferences. Specifically, let ${\overline{\mathcal{M}}}_p^{\mathrm{OA}}(f) := \{f'\in \mathcal M_p^{\mathrm{OA}}: \sigma_{f'}(1)\neq \sigma_f(1)\}$ be the collection of alternative preferences with distinct best items within the class of OA preferences. Then, the max-min optimization problem defining $I_{*,\kappa}(f^{\mathrm{OA}})$ can be relaxed to the following upper bound:
\begin{align}
I_{*,\kappa}(f^{\mathrm{OA}}) \le \sup_{\lambda\in \mathcal P (\mathcal S^{\kappa})} \ \inf_{f'\in {\overline{\mathcal{M}}}_p^{\mathrm{OA}}(f^{\mathrm{OA}})}  \  D_\lambda(f^{\mathrm{OA}} \| f') 
\label{equation_maxmin_capacity}
\end{align}

Drawing inspiration from the proof of Proposition~\ref{prop_OA_ranking_bound}, the max-min optimization problem~\eqref{equation_maxmin_capacity} is equivalent to the following linear programming problem:
\begin{equation}
\begin{aligned}
    & \max _{\lambda, u} \quad u \\
    &\text{ s.t. } \, \sum_{S \in \mathcal{S}^{\kappa}} d_S(\bar{\sigma}) \cdot \lambda(S) \geq u, \, \forall \bar{\sigma} \in \overline{\Sigma}\left(\sigma_*\right) \\
    &\phantom{\text{ s.t. }}\, \sum_{S \in \mathcal{S}^{\kappa}} \lambda(S)=1\\
    &\phantom{\text{ s.t. }}\, \lambda(S) \geq 0, \, \forall S \in \mathcal{S}^{\kappa},
\end{aligned}
\tag{LP-P-{$\kappa$}}
\label{LP-P-kappa}
\end{equation}
where $\overline{\Sigma}(\sigma_*) = \{\sigma : \sigma^{-1}(1) \neq \sigma_*^{-1}(1)\}$ is the set of rankings whose top-ranked item differs from that of $\sigma_*$, and $d_S(\bar{\sigma}) = D_S(f^{\mathrm{OA}}_{\sigma_*} \| f^{\mathrm{OA}}_{\bar{\sigma}})$. 

The dual problem of \eqref{LP-P-kappa} can be expressed as
\begin{equation}
\begin{aligned}
    & \min _{\mu, l} \quad l \\
    &\text{ s.t. } \, \sum_{\bar{\sigma} \in \overline{\Sigma}\left(\sigma_*\right)} d_S(\bar{\sigma}) \cdot \mu(\bar{\sigma}) \leq l, \, \forall S \in \mathcal{S}^{\kappa} \\
    &\phantom{\text{ s.t. }}\, \sum_{\bar{\sigma} \in \overline{\Sigma}\left(\sigma_*\right)} \mu(\bar{\sigma})=1 \\
    &\phantom{\text{ s.t. }}\, \mu(\bar{\sigma}) \geq 0, \, \forall \bar{\sigma} \in \overline{\Sigma}\left(\sigma_*\right).
\end{aligned}
\tag{LP-D-{$\kappa$}}
\label{LP-D-kappa}
\end{equation}

By the weak duality theorem, any feasible solution $(\mu, l)$ to the dual problem \eqref{LP-D-kappa} provides an upper bound on the optimal value of the primal problem, and consequently an upper bound on $I_{*,\kappa}(f^{\mathrm{OA}})$. We construct such a feasible solution in Lemma~\ref{lemma_feasible} below.
\vspace{0.2 cm}
\begin{lemma}
\label{lemma_feasible}
Let $ l^* =  I_{\kappa}^{\mathrm{worst}} $ and define $\mu^*$ as
$$
\mu^*(\bar{\sigma})= \begin{cases}\frac 1 {K-1}, & \text { if } \bar{\sigma}=\tilde{\sigma}_m, \text{ for } m=2, \ldots, K \\ 0, & \text { otherwise }\end{cases}
$$
where $\tilde{\sigma}_m:=(2,3, \ldots, m, 1, m+1, \ldots, K)$ for $m=2, \ldots, K$. Then $(\mu^*, l^*)$ is feasible in \eqref{LP-D-kappa}. 
\end{lemma}
\vspace{0.2 cm}

Therefore, Theorem~\ref{thm_worstcase_capacity} follows immediately from weak duality.
Specifically, we have established that
$$
\min_{f \in \mathcal{M}_p} I_{*,\kappa}(f) \le I_{*,\kappa}(f^{\mathrm{OA}}) \le l^* = I_{\kappa}^{\mathrm{worst}}.
$$
Substituting this upper bound on the hardness quantity into the lower bound result of Theorem~\ref{theorem_lower_bound_capacity} yields the desired conclusion.
\Halmos

\vspace{0.6 cm}
\proof{Proof of Lemma~\ref{lemma_feasible}.}
We first introduce some necessary notation from \citet{feng2022robust}. For any $n \in[K]$, let $s_n=(1-p) p^{n-1}$. We define:
$$
\begin{aligned}
\mathfrak{a}_n &:= \sum_{k=1}^{n-1}\left(s_k-s_n\right) \log \frac{s_k}{s_{k+1}} =\log \left(\frac{1}{p}\right)(1-p)\left[1+p+\cdots+p^{n-2}-(n-1) p^{n-1}\right] \\
\mathfrak{b}_n &:= \sum_{k=1}^n s_k =(1-p)\left(1+p+\cdots+p^{n-1}\right)=1-p^n.
\end{aligned}
$$
Based on these definitions, we utilize the following fact from \citet{feng2022robust} regarding the value of $d_{S}(\cdot)$ for the rankings $\tilde{\sigma}_m$. Given any set $S \in \mathcal{S}$ and $m \in [K-1]$,
$$
d_S\left(\tilde{\sigma}_m\right)=\begin{cases}
\mathfrak{a}_i / \mathfrak{b}_n & \text { if } m \in S \\
0 & \text { if } m \notin S
\end{cases}, \quad \text { where } \quad i=\sigma_*(m \mid S) \text { and } n=|S|.
$$

We proceed to verify that the proposed solution satisfies the constraints of the dual problem \eqref{LP-D-kappa}. The non-negativity condition and the requirement that $\mu^*$ sums to unity are trivially satisfied.
For the inequality constraint, consider any display set $S \in \mathcal{S}^\kappa$. We have 
\begin{align*}
\sum_{\bar{\sigma} \in \overline{\Sigma}\left(\sigma_*\right)} d_S(\bar{\sigma}) \cdot \mu^*(\bar{\sigma}) &= \sum_{m=2}^K d_S(\tilde{\sigma}_m) \cdot \frac{1}{K-1} = \frac{1}{K-1} \sum_{m \in S, m \neq 1} \frac{\mathfrak{a}_{\sigma_*(m|S)}}{\mathfrak{b}_{|S|}}.
\end{align*}

Note that the sum $\sum_{m \in S, m \neq 1}  \mathfrak{a}_{\sigma_*(m|S)}$ is bounded by $\sum_{k=1}^{|S|} \mathfrak a_k = {\mathfrak a_1 + \mathfrak a_2 + \cdots + \mathfrak a_{|S|}} $. Since $|S| \le \kappa$, we obtain that \begin{align*}
\sum_{\bar{\sigma} \in \overline{\Sigma}\left(\sigma_*\right)} d_S(\bar{\sigma}) \cdot \mu^*(\bar{\sigma})  &\le  \max_{2\le n \le \kappa} \frac {\mathfrak a_1 + \mathfrak a_2 + \cdots + \mathfrak a_n} {(K-1)\mathfrak b_n} .
\end{align*}

Substituting the explicit expressions for $\mathfrak{a}_n$ and $\mathfrak{b}_n$, the objective function becomes
\begin{align*}
\max_{2\le n \le \kappa} \frac {\sum_{k=1}^n \mathfrak a_k} {(K-1)\mathfrak b_{n}} &= \max_{2\le n \le \kappa}  \frac{ \log(1/p) \left( (n-1)(1+p^n) - \frac{2(p-p^{n})}{1-p} \right) } { (K-1)(1-p^n) }.
\end{align*}
Through straightforward algebra, one can verify that the maximum is attained when $n = \kappa$. 

Altogether, we have 
\begin{align*}
\sum_{\bar{\sigma} \in \overline{\Sigma}\left(\sigma_*\right)} d_S(\bar{\sigma}) \cdot \mu^*(\bar{\sigma}) \le \frac{ \log(1/p) \left( (\kappa-1)(1+p^\kappa) - \frac{2(p-p^{\kappa})}{1-p} \right) } { (K-1)(1-p^\kappa) }  = l^*.
\end{align*}
This confirms that the inequality constraint of \eqref{LP-D-kappa} holds for all $S \in \mathcal{S}^\kappa$, thereby completing the proof.
\Halmos
\endproof
}

\subsection{Proofs for Full-Ranking Identification}

\tbl{
The following lemma provides uniform guarantees for \textsc{NE} when $ K = 2 $.
	\begin{lemma}[Two-item nested comparison]
		\label{lemma_pairwise_ne}
		For every $f\in\mathcal M_p$, every pair of distinct items $i,j$, and
		every integer $m\ge 1$, let $T_{ij}(m)$ be the number of samples used
		by $\textsc{NE}_2(i,j;m)$. Then
		\begin{align*}
		\mathbb P_f\!\left(
		\textsc{NE}_2(i,j;m)
		\text{{ returns the lower-ranked item}}
		\right)
		\le p^m \quad \text{ and } \quad 
		\mathbb E_f[T_{ij}(m)]
		\le \frac{(1+p)m}{1-p}.
		\end{align*}
	\end{lemma}
	
	\proof{Proof of Lemma~\ref{lemma_pairwise_ne}.}
	
	Let the pair $ \{i,j\} $ be given and suppose $ i $ is the preferred item. The score difference $ \{W_t(i)-W_t(j)\}  $ under \textsc{NE} forms a one-dimensional random walk, which increases by one with probability \( f(i \mid \{i,j\}) \geq \tfrac{1}{1 + p} \) and decreases by one with probability \( f(j \mid \{i,j\}) \leq \tfrac{p}{1 + p} \) at each step, and stops once it hits the boundary $ \{-m,m\} $. Since there is no overshoot once hitting, the hitting probability and expected hitting time are thus standard to obtain; see Proposition~EC.1 of \cite{zhu2025power} for a more refined and slightly more general analysis.
	\Halmos
	\endproof

	\vskip 0.3 cm
	\proof{Proof of
		Theorem~\ref{theorem_fixedconfidence_capacity_ranking}.}
	
	We divide the proof into three steps.
	
	\smallskip
	\noindent
	\underline{Step 1. Analysis within one sort-verification cycle.} Let us start from the \textit{sort} step and invoke a well-known fact: the standard binary \textsc{MergeSort} uses at most $ K\log_2K $
	comparisons, regardless of the path of comparison outcomes (\citealp{parberry1998mergesort,sedgewick2011algorithms}). If all the \textsc{NE} outputs are correct, \textsc{MergeSort} outputs the correct ranking $ \sigma_f^{-1} $. Therefore, Lemma~\ref{lemma_pairwise_ne} combined with the union bound
	therefore implies that the probability that the candidate ranking $ \widehat\pi $ is incorrect satisfies
	\begin{align*}
	\mathbb P_f(\widehat\pi\neq\sigma_f^{-1})
	&\le
	K\log_2Kp^{m_\delta}
	\le
	\eta_\delta.
	\end{align*}
	The expected number of samples used during candidate generation satisfies
	\begin{equation}
	\label{eq:candidate-time-proof}
	\mathbb E_f[\tau_{\mathrm{cand}}]
	\le
	\frac{K\log_2K(1+p)m_\delta}{1-p}.
	\end{equation}
	
	We now proceed to the \textit{verification} step. 
	Suppose first that $\widehat\pi=\sigma_f^{-1}$. By Lemma~\ref{lemma_pairwise_ne} and a
	union bound, the probability that at least one of the $K-1$
	verification calls fails is at most
	$(K-1)p^{M_\delta}$. Consequently, for every single cycle, the probability of generating the correct ranking and accepting it, denoted by $s_\delta$,
	satisfies
	\begin{equation}
	\label{eq:correct-acceptance-cycle}
	s_\delta
	\ge
	(1-\eta_\delta)
	\bigl(1-(K-1)p^{M_\delta}\bigr)
	\ge
	(1-\eta_\delta)(1-\delta/4).
	\end{equation}
	
	Suppose instead that $\widehat\pi\neq\sigma_f^{-1}$. Then there is at least one adjacent inversion contained in the permutation $ \widehat\pi $. Lemma~\ref{lemma_pairwise_ne} implies that the \textit{verification} step will pass with probability at most $p^{M_\delta}$. Consequently, for every single cycle, the probability of generating an incorrect ranking and accepting it, denoted by $w_\delta$,
	satisfies
	\begin{equation*}
	w_\delta
	\le
	\eta_\delta p^{M_\delta}
	\le
	\frac{\eta_\delta\delta}{4(K-1)}.
	\end{equation*}
	Invoking Lemma~\ref{lemma_pairwise_ne}, the expected sample complexity of the \textit{verification} step in one cycle is at most
	$
	\frac{(K-1)(1+p)M_\delta}{1-p}.
	$ Together with \eqref{eq:candidate-time-proof}, the expected cost
	$B_\delta$ of one cycle satisfies
	\begin{equation}
	\label{eq:one-cycle-time}
	B_\delta
	\le
	\frac{K\log_2K(1+p)m_\delta}{1-p} + \frac{(K-1)(1+p)M_\delta}{1-p}.
	\end{equation}

	\smallskip
	\noindent
	\underline{Step 2. Overall error probability and sample complexity.}
	The cycles use fresh observations and hence are independent and
	identically distributed, and are repeated until the verification step is passed. Each cycle has acceptance probability 
	$s_\delta+w_\delta>0$, so the algorithm terminates (with acceptance) almost surely. In addition, the probability of incorrect ranking conditional on acceptance is at most $\frac{w_\delta}{s_\delta+w_\delta}$. Therefore, it holds that
	\begin{equation*}
	\mathbb P_f(\sigma_{\mathrm{out}}\neq\sigma_f)
	=
	\frac{w_\delta}{s_\delta+w_\delta}
	\le
	\frac{w_\delta}{s_\delta} \le\frac{\delta}{9}
	\le\delta,
	\end{equation*}
	where the second inequality from the last is because $\eta_\delta\le1/4$, which implies that
	$
	s_\delta
	\ge
	\frac34\left(1-\frac{\delta}{4}\right)
	\ge\frac{9}{16},
	$
	and
	$
	w_\delta
	\le
	\frac{\delta}{16(K-1)}
	\le\frac{\delta}{16}.
	$
	
	To derive the total sample complexity, let $ \tau_c^i $ be the length of cycle $ i $ and $ N_c $ be the number of cycles. It holds that
	\begin{align*}
	\mathbb E_f[\tau]
	= \mathbb E_f[\sum_{i=1}^{\infty} \tau_c^i \mathbb{I}\{N_c \geq i\}  ] =\sum_{i=1}^{\infty}  \mathbb E_f[ \tau_c^i]\cdot  \mathbb P_f (N_c \geq i ) = \sum_{i=1}^{\infty}  B_\delta (1-s_\delta-w_\delta)^{i-1} = \tfrac{B_\delta}{s_\delta+w_\delta},
	\end{align*}
	where the second equality is because cycle $ i $ is independent of the event $ \{N_c \geq i\} $, which only depends on up to cycle $ i-1 $. Drawing on \eqref{eq:one-cycle-time} and \eqref{eq:correct-acceptance-cycle}, we thus have
	\[
	\mathbb E_f[\tau]
	\le
	\frac{B_\delta}{s_\delta} \le
	\frac{
		K\log_2K(1+p)m_\delta
	}{
		(1-p)(1-\eta_\delta)(1-\delta/4)
	} + \frac{
		(K-1)(1+p)M_\delta
	}{
		(1-p)(1-\eta_\delta)(1-\delta/4)
	}
	\]
	as claimed, which is notably a quantity independent of $ f $.
	
	\smallskip
	\noindent
	\underline{Step 3. Putting things together.} Finally, as $\delta\downarrow0$,
	$
	\eta_\delta\to 0,
	$
	$
	m_\delta
	=
	O\!\left(\log\log(1/\delta)\right),
	$
	and
	$
	M_\delta
	=
	\frac{\log(1/\delta)}{\log(1/p)}
	+O(1).
	$
	Dividing the preceding sample-complexity bound by
	$\log(1/\delta)$ therefore yields
	\[
	\limsup_{\delta\downarrow0}
	\sup_{f\in\mathcal M_p}
	\frac{\mathbb E_f[\tau]}{\log(1/\delta)}
	\le \frac{(K-1)(1+p)}{(1-p)\log(1/p)}.
	\]
	This completes the proof.
	\Halmos
	\endproof
}

\vspace{0.3 cm}
\proof{Proof of Theorem~\ref{thm_worstcase_capacity_ranking}.}

The proof structure for Theorem~\ref{thm_worstcase_capacity_ranking} closely parallels that of Theorem~\ref{thm_worstcase_capacity} presented in Appendix~\ref{appendix_thm_worstcase_capacity}, though certain details exhibit significant variations due to the differing problem setting. To avoid redundancy, we streamline the shared components and focus primarily on the distinctions.

First, we formulate the dual of the linear programming problem associated with the information-theoretic lower bound for the capacity-constrained ranking problem:
\begin{equation}
\begin{aligned}
& \min _{\mu, l} \quad l \\
&\text{ s.t. } \, \sum_{\bar{\sigma} \neq \sigma_*} d_S(\bar{\sigma}) \cdot \mu(\bar{\sigma}) \leq l, \quad \forall S \in \mathcal{S}^{\kappa} \\
&\phantom{\text{ s.t. }}\, \sum_{\bar{\sigma} \neq \sigma_*} \mu(\bar{\sigma})=1 \\
&\phantom{\text{ s.t. }}\, \mu(\bar{\sigma}) \geq 0, \quad \forall {\bar{\sigma} \neq \sigma_*} .
\end{aligned} \tag{LP-D-{$\kappa$}-ranking} \label{LP-D-kappa-ranking}
\end{equation}

Recall from the proof of Proposition~\ref{prop_OA_ranking_bound} in Appendix~\ref{appendix_prop_OA_ranking_bound} the rankings $\hat{\sigma}_m$, $m \in [K-1]$, each obtained from $\sigma_*$ by transposing items $m$ and $m+1$. Lemma~\ref{lemma_counterpart_lemma6} characterizes the value of $d_{S}(\cdot)$ under these specific rankings.

By the weak duality theorem, any feasible solution to the dual problem~\eqref{LP-D-kappa-ranking} yields an upper bound on the primal optimal value. Thus, to obtain the desired worst-case bound, it suffices to construct a feasible solution $(\mu^*, l^*)$. To this end, we set $l^* = J_{\kappa}^{\mathrm{worst}}$ and define $\mu^*$ as follows:
$$
\mu^*(\bar{\sigma})= \begin{cases}\frac{1}{K-1}, & \text { if } \bar{\sigma}=\hat{\sigma}_m \text{ for } m\in [K-1]; \\ 0, & \text { otherwise. }\end{cases}
$$

We now verify the feasibility of $(\mu^*, l^*)$ in \eqref{LP-D-kappa-ranking}. The non-negativity and sum-to-one constraints on $\mu^*$ are trivially satisfied. For the inequality constraint, consider any display set $S \in \mathcal{S}^{\kappa}$. We have:
\begin{align*}
\sum_{\bar{\sigma} \neq \sigma_*} d_S(\bar{\sigma}) \cdot \mu^*(\bar{\sigma}) 
& = \sum_{m=1}^{K-1} d_S(\hat{\sigma}_m) \cdot \frac{1}{K-1} \\
&\le \frac{\log(1/p) (1-p)^2}{K-1} \cdot \frac{1+p+\cdots+p^{|S|-2}}{1-p^{|S|}} \\
&\le \frac{\log(1/p) (1-p)^2}{K-1} \cdot \frac{1+p+\cdots+p^{\kappa-2}}{1-p^{\kappa}} \\
&= \frac{\log(1/p) (1-p)^2}{K-1} \cdot \frac{\frac{1-p^{\kappa-1}}{1-p}}{1-p^{\kappa}} \\
&= \frac{\log(1/p) (1-p)(1-p^{\kappa-1})}{(K-1)(1-p^\kappa)} \\
&= J_{\kappa}^{\mathrm{worst}} = l^*.
\end{align*}
This confirms that the inequality constraint is satisfied for all $S \in \mathcal{S}^{\kappa}$. Therefore, $(\mu^*, l^*)$ is a valid feasible solution, which completes the proof.
\Halmos
\endproof

\tbl{
\section{Empirical Error Probabilities}\label{append:empirical error probability}

In this appendix, we report additional numerical results on the empirical error probabilities of the different methods; see Figures~\ref{figure_worst_case_error_probability}--\ref{figure_general_ranking_error_probability}. These figures complement the stopping-time comparisons in Section~\ref{section_experiment}. For ease of comparison, we also plot the corresponding guaranteed error probabilities induced by the choices of the tuning parameters used in the experiments. Recall that all the reported quantities are averaged over $512$ independent trials.\footnote{With this finite simulation budget, very small error probabilities cannot be estimated with high precision. Therefore, these figures are mainly intended to provide a qualitative check on the conservativeness of the theoretical guarantees.} The following observations emerge from Figures~\ref{figure_worst_case_error_probability}--\ref{figure_general_ranking_error_probability}:
\begin{enumerate}[label = (\roman*)]
	\item Across all the settings we consider, the empirical error probabilities of all methods stay below their corresponding guaranteed error probabilities. This is consistent with the fixed-confidence theoretical guarantees.
	
	\item For the best-item identification problem, the empirical error probabilities of \textsc{NE} are closer to their theoretical guarantees than those of \textsc{MTP}; similarly, for the full-ranking identification problem, the empirical error probabilities of \textsc{NP} are closer to their theoretical guarantees than those of \textsc{NE-Ranking} and \textsc{Repeated-NE}. This suggests that \textsc{NE} and \textsc{NP} are less conservative in their use of samples.
	
	\item The gap between the empirical error probabilities of \textsc{NE} and its theoretical guarantee in the selection problem is noticeably larger than the corresponding gap for \textsc{NP} in the ranking problem. We attribute this difference to the choice of the parameter $M$: in the ranking problem, the choice of $M$ for \textsc{NP} is tighter, which leads to a less conservative error-probability guarantee.
\end{enumerate}

\begin{figure}[tb]
	\begin{minipage}{1.0\columnwidth}
		\centering
		\hspace{-.7 cm}
		\subfigure[$K=5, p=0.9$.]{
			\begin{minipage}[b]{0.24\textwidth}
				\includegraphics[width=1\textwidth]{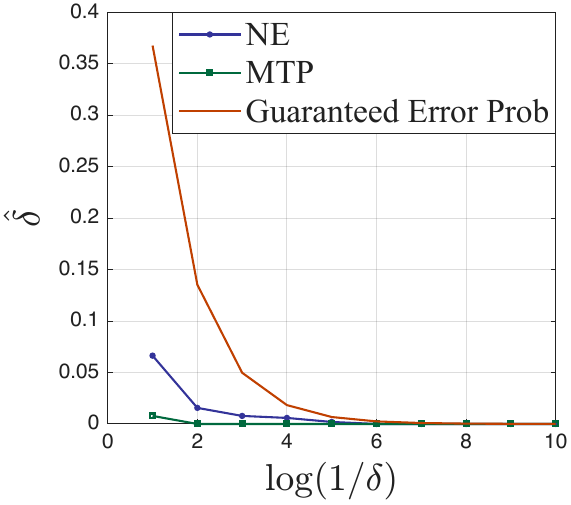}
			\end{minipage}
		}
		\hspace{-6pt}
            \hspace{-6pt}
		\subfigure[$K=15, p=0.9$.]{
			\begin{minipage}[b]{0.24\textwidth}
				\includegraphics[width=1\textwidth]{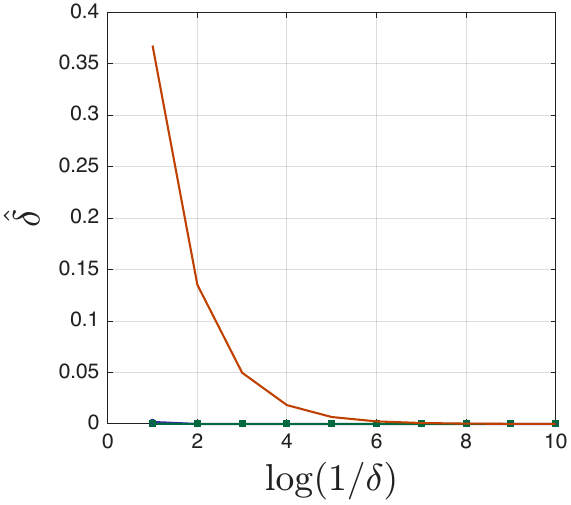} 
			\end{minipage}
		}
		\hspace{-6pt}
		\subfigure[$K=5, p=0.6$.]{
			\begin{minipage}[b]{0.24\textwidth}
				\includegraphics[width=1\textwidth]{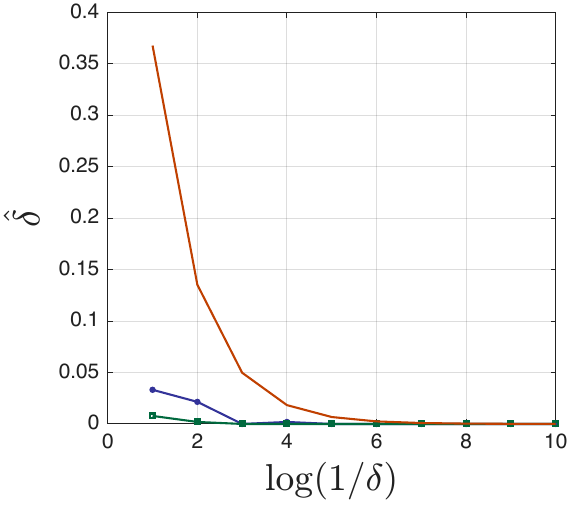} 
			\end{minipage}
		}
		\hspace{-6pt}
		\subfigure[$K=15, p=0.6$.]{
			\begin{minipage}[b]{0.24\textwidth}
				\includegraphics[width=1\textwidth]{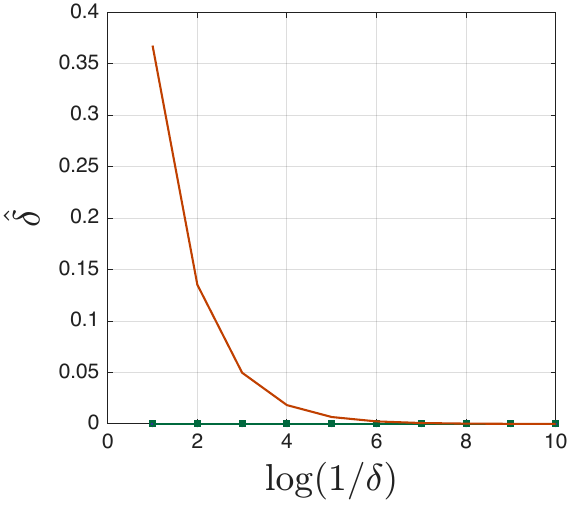} 
			\end{minipage}    
		}
	\end{minipage}
        \vspace{3pt}
	\caption{\textbf{Empirical error probabilities of \textsc{NE} and \textsc{MTP} under the hardest-to-learn (i.e., OA) preference instances.}} \label{figure_worst_case_error_probability}
\end{figure}
\begin{figure}[htbp]
	\centering
	\vspace{2pt}
	\begin{minipage}{1.0\columnwidth}
		\centering
		\hspace{-1pt}
		\subfigure[Preference $f_1$]{
			\begin{minipage}[b]{0.25\textwidth}
				\includegraphics[width=1\textwidth]{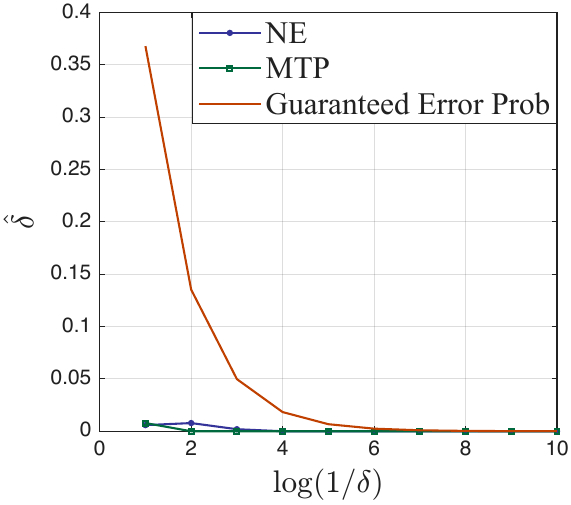} 
			\end{minipage}
		}
		\hspace{-1pt}
		\subfigure[Preference $f_2$]{
			\begin{minipage}[b]{0.25\textwidth}
				\includegraphics[width=1\textwidth]{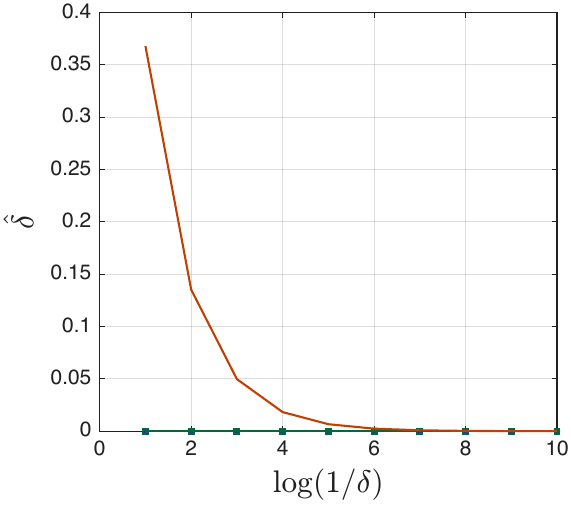}  
			\end{minipage}
		}
	\end{minipage}
	\caption{\textbf{Empirical error probabilities of \textsc{NE} and \textsc{MTP} under the real data.}} 
	\label{figure_general_error_probability}
	\vspace{4pt}
\end{figure}
\begin{figure}
	\centering
	\hspace{-15pt}
	\subfigure[$K=5, p=0.9$ ]{\includegraphics[width=0.24 \textwidth]{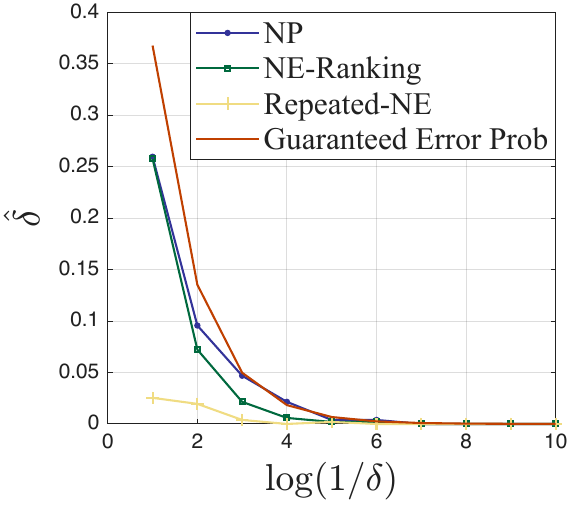}}
	\subfigure[$K=15, p=0.9$ ]{\includegraphics[width=0.24 \textwidth]{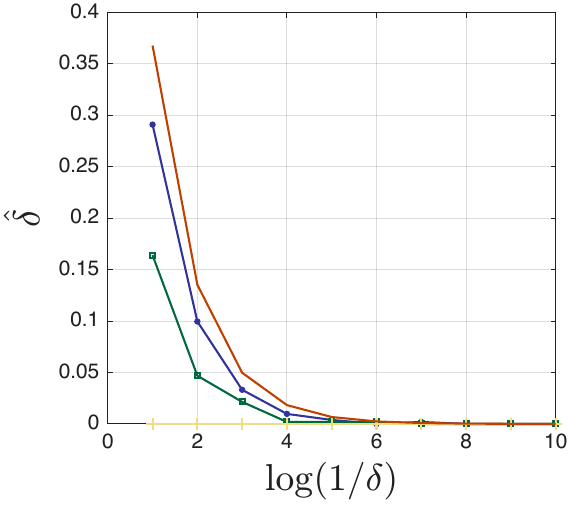}}
	\subfigure[$K=5, p=0.6$ ]{\includegraphics[width=0.24 \textwidth]{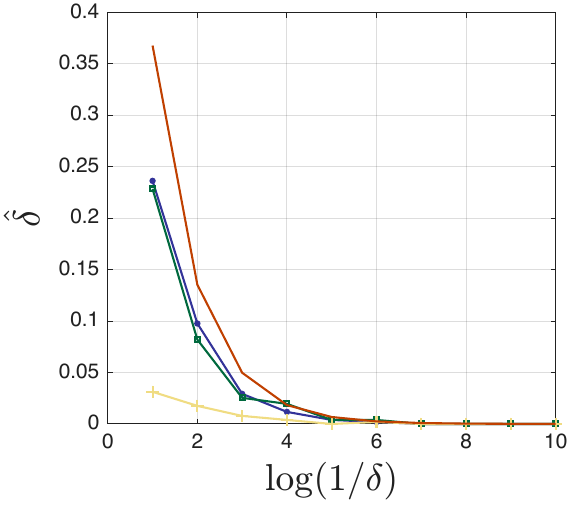}}
	\subfigure[$K=15, p=0.6$ ]{\includegraphics[width=0.24 \textwidth]{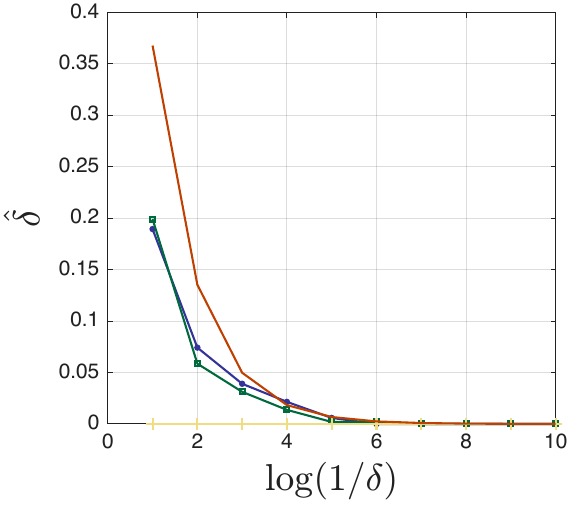}}
	\vspace{3pt}
	\caption{\textbf{Empirical error probabilities of \textsc{NP}, \textsc{NE-Ranking}, and \textsc{Repeated-NE} under the hardest-to-learn (i.e., OA) preference instances.} } \label{figure_worst_case_ranking_error_probability}
\end{figure}
\begin{figure}
	\centering
	\hspace{-1pt}
	\subfigure[Preference $f_1$]{
		\begin{minipage}[b]{0.25\textwidth}
			\includegraphics[width=1\textwidth]{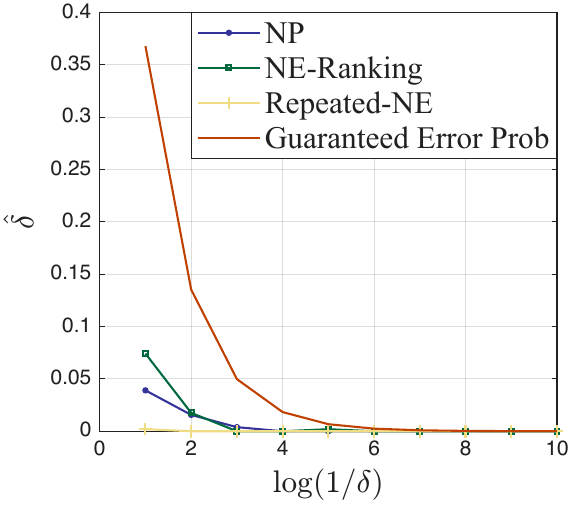} 
		\end{minipage}
	}
	\hspace{-1pt}
	\subfigure[Preference $f_2$]{
		\begin{minipage}[b]{0.25\textwidth}
			\includegraphics[width=1\textwidth]{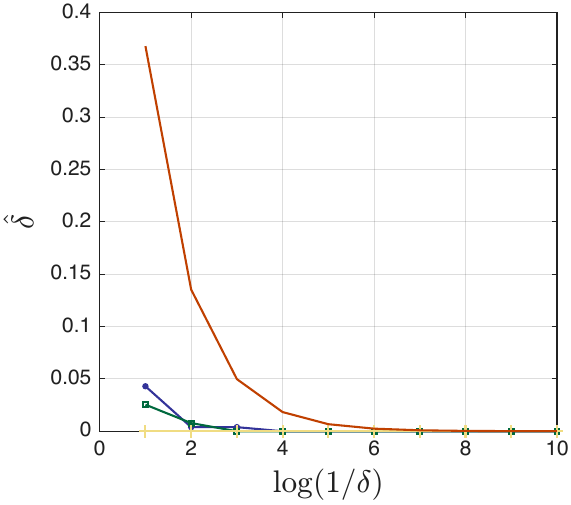}  
		\end{minipage}
	}
	\caption{\textbf{Empirical error probabilities of \textsc{NP}, \textsc{NE-Ranking}, and \textsc{Repeated-NE} under real data.}} 
	\label{figure_general_ranking_error_probability}
\end{figure}
}

\tbl{
\section{Non-identifiability of Mixture Models}
\label{append:mixture-nonidentifiability}

\begin{proposition}\label{prop:nonidentifiability}
	Let $K=4$. There exist a preference $f$ and four preferences $g_1,g_2,h_1,h_2$ such that
	$f=\tfrac12 g_1+\tfrac12 g_2=\tfrac12 h_1+\tfrac12 h_2$, with
	\[
	g_1\in\mathcal M_{20/27},\quad g_2\in\mathcal M_{22/27},\quad h_1\in\mathcal M_{22/27},\quad h_2\in\mathcal M_{4/5},\quad f\in\mathcal M_{27/29},
	\]
	whose associated rankings are $1\!\succ\!3\!\succ\!4\!\succ\!2$, $2\!\succ\!4\!\succ\!3\!\succ\!1$, $3\!\succ\!1\!\succ\!2\!\succ\!4$, $4\!\succ\!2\!\succ\!1\!\succ\!3$ and $1\!\succ\!2\!\succ\!3\!\succ\!4$ respectively. In particular, the sets of top-ranked items under the two decompositions are $\{1,2\}$ and $\{3,4\}$, which are disjoint.
\end{proposition}

\proof{Proof of Proposition \ref{prop:nonidentifiability}.}
Table~\ref{table_nonid} specifies the five choice probabilities on all eleven display sets. The identity $g_1+g_2=h_1+h_2=2f$ is verified entrywise, and the stated membership in each $\mathcal M_p$ follows by checking every ratio $x(i\mid S)/x(j\mid S)$ whenever $j$ is preferred to $i$, the largest of which equals the quoted parameter in each case.
\hfill\Halmos
\endproof

\begin{table}
	
	\begin{center}
		\small
		\begin{tabular}{lccccc}
			\toprule
			& \multicolumn{2}{c}{Decomposition A} & \multicolumn{2}{c}{Decomposition B} & Aggregate \\
			\cmidrule(r){2-3} \cmidrule(r){4-5} \cmidrule(r){6-6}
			& $g_1$ & $g_2$ & $h_1$ & $h_2$ & $f$ \\
			\textit{ranking} & $1\succ3\succ4\succ2$ & $2\succ4\succ3\succ1$ & $3\succ1\succ2\succ4$ & $4\succ2\succ1\succ3$ & $1\succ2\succ3\succ4$ \\
			\textit{dispersion} & $20/27$ & $22/27$ & $22/27$ & $4/5$ & $27/29$ \\
			\midrule
			$\{1,2\}$     & $(.98,.02)$          & $(.06,.94)$          & $(.98,.02)$          & $(.06,.94)$          & $(.52,.48)$          \\
			$\{1,3\}$     & $(.98,.02)$          & $(.06,.94)$          & $(.06,.94)$          & $(.98,.02)$          & $(.52,.48)$          \\
			$\{1,4\}$     & $(.98,.02)$          & $(.06,.94)$          & $(.98,.02)$          & $(.06,.94)$          & $(.52,.48)$          \\
			$\{2,3\}$     & $(.06,.94)$          & $(.98,.02)$          & $(.06,.94)$          & $(.98,.02)$          & $(.52,.48)$          \\
			$\{2,4\}$     & $(.06,.94)$          & $(.98,.02)$          & $(.98,.02)$          & $(.06,.94)$          & $(.52,.48)$          \\
			$\{3,4\}$     & $(.98,.02)$          & $(.06,.94)$          & $(.98,.02)$          & $(.06,.94)$          & $(.52,.48)$          \\
			\addlinespace[2pt]
			$\{1,2,3\}$   & $(.72,.02,.26)$      & $(.02,.66,.32)$      & $(.44,.02,.54)$      & $(.30,.66,.04)$      & $(.37,.34,.29)$      \\
			$\{1,2,4\}$   & $(.72,.02,.26)$      & $(.02,.66,.32)$      & $(.72,.26,.02)$      & $(.02,.42,.56)$      & $(.37,.34,.29)$      \\
			$\{1,3,4\}$   & $(.78,.20,.02)$      & $(.02,.44,.54)$      & $(.38,.60,.02)$      & $(.42,.04,.54)$      & $(.40,.32,.28)$      \\
			$\{2,3,4\}$   & $(.02,.66,.32)$      & $(.72,.02,.26)$      & $(.32,.66,.02)$      & $(.42,.02,.56)$      & $(.37,.34,.29)$      \\
			\addlinespace[2pt]
			$\{1,2,3,4\}$ & $(.54,.02,.40,.04)$  & $(.04,.52,.06,.38)$  & $(.34,.22,.42,.02)$  & $(.24,.32,.04,.40)$  & $(.29,.27,.23,.21)$  \\
			\bottomrule
		\end{tabular}
	\end{center}
	\vskip 0.2 cm
	
	\caption{\textbf{Two decompositions of the same preference.}}
	\label{table_nonid}
\end{table}
}

\tbl{
\section{Summary of Incremental Contributions Relative To The Conference Version}\label{appendix:conference}

While the conference version focused exclusively on the learning-to-select problem, this version extends the framework to address the more complex and previously unexplored learning-to-rank problem, as detailed in the newly added Section~\ref{section_fullranking} as well as the expanded numerical experiments in Section~\ref{section_experiment2}. This extension is made possible through further analysis and development of our earlier methodology, providing new insights into the solution structure and offering a deeper understanding of the algorithms. For example, the nested structure in the \textit{ranking identification} problem is a surprising discovery for us. These advancements broaden the applicability of our approach to a wider range of online learning problems. This version further adds the capacity-constrained treatment of both best-item and full-ranking identification problems, together with new algorithms and corresponding upper and lower bounds, in Section~\ref{sec_extension_capacity}, as well as characterizes the hardness quantities $\IN(f)$ and $\JN(f)$ beyond the worst case for structured preference families in Sections~\ref{sec:understanding-NE-beyond-OA} and~\ref{sec:understanding-NP-beyond-OA}. The appendices contain three further
	additions, including (i) a quantification of the advantage of \textsc{NE} and \textsc{NP} over
	the simple full display policy for both learning
	objectives (Appendix~\ref{appendix:comparison with full display}), (ii) a
	non-identifiability result for the mixture preference model
	(Appendix~\ref{append:mixture-nonidentifiability}), and (iii) an expanded numerical
	study (Appendix~\ref{append:empirical error probability}).
}
\end{APPENDICES}



\end{document}